\documentclass[pdflatex,sn-mathphys-num]{sn-jnl}% Math and Physical Sciences Numbered Reference Style 
\usepackage{graphicx}%
\usepackage{multirow}%
\usepackage{amsmath,amssymb,amsfonts}%
\usepackage{amsthm}%
\usepackage{mathrsfs}%
\usepackage[title]{appendix}%
\usepackage{xcolor}%
\usepackage{textcomp}%
\usepackage{manyfoot}%
\usepackage{booktabs}%
\usepackage{algorithm}%
\usepackage{algorithmicx}%
\usepackage{listings}%
\usepackage{graphicx}
\usepackage{subcaption}

\definecolor{p0}{HTML}{227600}
\definecolor{p1}{HTML}{222f7b}

\usepackage[noend]{algpseudocode}
\algnewcommand{\LineComment}[1]{\Statex \(\triangleright\) #1}
\algnewcommand\algorithmicinput{\textbf{Input:}}
\algnewcommand\algorithmicoutput{\textbf{Output:}}
\algnewcommand\Input{\item[\algorithmicinput]}
\algnewcommand\Output{\item[\algorithmicoutput]}
\newcommand{\Break}{\State \textbf{break} }
\def\algbackskip{\hskip-\ALG@thistlm}

\usepackage{comment}

\usepackage{graphicx}
\usepackage{rotating}
\usepackage{booktabs}    % Optional, for better table rules
\usepackage{adjustbox}   % Optional, better resizing control

\theoremstyle{thmstyleone}%
\theoremstyle{thmstyletwo}%

\theoremstyle{thmstylethree}%

\begin{document}

\title[Expanding Data-Agnostic Pivotal Instances Selection Models with Proximity Trees and Ensemble Learning]{%
    \begin{minipage}{\textwidth}
        \centering
        \textcolor{blue}{%
            \normalsize
            This manuscript was accepted for publication in
            \textit{Machine Learning} on May 2, 2026.
        }

        \vspace{0.5em}
        \rule{0.85\textwidth}{0.4pt}
        \vspace{0.8em}

        Expanding Data-Agnostic Pivotal Instances Selection Models with
        Proximity Trees and Ensemble Learning
    \end{minipage}%
}


\author*[1]{\fnm{Alessio} \sur{Cascione}}\email{alessio.cascione@phd.unipi.it}

\author[1]{\fnm{Mattia} \sur{Setzu}}\email{mattia.setzu@unipi.it}

\author[1,2]{\fnm{Cristiano} \sur{Landi}}\email{cristiano.landi@phd.unipi.it}

\author[1]{\fnm{Paolo Maria} \sur{Mancarella}}\email{paolo.mancarella@unipi.it}

\author[1,2]{\fnm{Riccardo} \sur{Guidotti}}\email{riccardo.guidotti@unipi.it}

\affil*[1]{\orgdiv{Department of Computer Science}, \orgname{University of Pisa}, \orgaddress{\street{Largo B. Pontecorvo}, \city{Pisa}, \postcode{56127}, \state{PI}, \country{Italy}}}

\affil[2]{\orgname{ISTI-CNR}, \orgaddress{\street{Via G. Moruzzi}, \city{Pisa}, \postcode{56127}, \state{PI}, \country{Italy}}}


\abstract{

As decision-making processes grow more complex, machine learning tools have become essential for tackling business and societal challenges. 
However, many existing methods rely on decision-making procedures that are difficult to interpret. 
Since humans naturally make decisions by comparing new cases with a few representative examples, we aim to design an approach that selects such \textit{pivots} to construct an interpretable predictive model. Inspired by decision trees, we propose a hierarchical, interpretable-by-design pivot selection model based on the similarity between pivots and input instances. 
Our method functions both as a pivot selection technique and a standalone predictive model. 
Extending beyond single pivots, we incorporate pairs of pivots that are used by proximity and oblique trees, as well as ensembles, which enhance the versatility and effectiveness of our proposal.
Additionally, our approach is data modality-agnostic, leveraging pre-trained networks for data transformation.
Experiments across diverse datasets, including tabular data, text, images, and time series, demonstrate the effectiveness of our approach, outperforming alternative instance selection strategies and achieving competitive results against state-of-the-art interpretable models while maintaining a minimal number of pivots.

}

%%================================%%
%% Sample for structured abstract %%
%%================================%%




\maketitle

\section{Introduction}
\label{sec:introduction}
Machine Learning (ML) models have become indispensable in aiding human decision-making across diverse domains, including healthcare, online threat detection, and consumer behavior analysis~\citep{chui2022state,chatzakou2019cyber,DeFauw2018ex1,guidotti2019market}.
Despite their remarkable performance, these models often rely on complex and opaque architectures, making it challenging for both experts and end-users to interpret their decision-making processes.
While ML models can equal or even surpass human performance in specific tasks, their underlying reasoning mechanisms significantly diverge from human cognitive processes~\cite{yang2022unbox}.
Given the growing reliance on ML-driven decisions, enhancing model interpretability and demystifying the inner workings of these ``black-box'' systems is essential~\cite{DBLP:conf/aies/KasirzadehC21}.
This need for transparency is the primary objective of Explainable AI (XAI)~\cite{bodria2023benchmarking}.

A promising approach to fostering interpretability in ML models is leveraging similarity-based reasoning grounded in \textit{discriminative} and \textit{descriptive} elements.
We hypothesize that models capable of reasoning through exemplary instances provide a more intuitive and interpretable framework for decision-makers, analysts, and general users~\cite{waa2021evaluating}.
Human cognition naturally employs \emph{case-based reasoning}~\cite{schank2014knowledge}, wherein past experiences are stored and retrieved to solve new problems.
Although the retrieval mechanism itself may not always be explicit, reasoning through analogous cases remains inherently interpretable.
This cognitive strategy is deeply ingrained in human perception, enabling even young children to recognize and interact with novel objects based on their resemblance to familiar ones~\cite{spelke2022babies}.
Furthermore, this reasoning process spans multiple modalities, including visual recognition of familiar faces, auditory comparison of musical genres, and gustatory identification of culinary styles based on past experiences~\cite{johnson2010mental}.
At its core, similarity-based reasoning constitutes a universal cognitive framework that extends across diverse data modalities and domains~\cite{golding1995review}.

Case-based reasoning offers substantial advantages in promoting interpretability across various fields, including medical diagnostics~\cite{healthcase2006}, financial risk assessment~\cite{li2022data}, text analysis~\cite{DBLP:conf/acl/0001GKLL22,hong2023protorynet}, and time-series and image processing~\cite{adebayo2018sanity}.
Recent research~\cite{nguyen2021effectiveness,jeyakumar2020howcan} highlights the effectiveness of this approach, demonstrating that human users often prefer instance-based reasoning over more abstract feature-based methods.
Building on these insights, we underscore the importance of high-quality training data to ensure meaningful similarity between pivots and instances for predictive modeling.
Poor data diversity or bias can result in unrepresentative cases, thereby reducing model interpretability and reliability.
Conversely, feature-based approaches may offer enhanced robustness in such scenarios by emphasizing the role of specific attributes in shaping predictions.

Therefore, our goal is to develop a suite of interpretable case-based models that identify both descriptive and discriminative cases to support decision-making tasks.
To this end, we introduce \textsc{PivotTree}, a hierarchical and interpretable case-based model inspired by decision trees~\cite{breiman1984classification}.
At its core, \textsc{PivotTree} serves as a \textit{selection} model, capable of extracting a representative subset of training instances, referred to as \emph{pivots}. 
However, beyond its role in instance selection, \textsc{PivotTree} can also function as a \textit{predictive} model. 
By employing a similarity-based decision tree structure, it classifies new instances by routing them through its hierarchy, ultimately providing both a prediction and an interpretable explanation.
Unlike traditional decision trees, where explanations consist of explicit rule sets, \textsc{PivotTree} provides explanations based on similarities with pivots. 
This approach aligns with instance-based models, as \textsc{PivotTree} functions as both a selection method and a predictive model, encoding instances within a similarity space to facilitate case-based reasoning.  

A key challenge of \textsc{PivotTree} lies in the interpretability of the distance metric used to compare instances with pivots, requiring users to understand how the metric influences the model’s decision-making process. 
To address this and further enhance both interpretability and generalization, we draw inspiration from research on oblique trees~\cite{wickramarachchi2016hhcart,murthy1994system} and proximity forest methods in time series classification~\cite{DBLP:journals/datamine/LucasSPOZGPW19,tan2025proximity,zhang2021proximity}. 
Specifically, we introduce within \textsc{PivotTree} \textit{oblique splits} and \textit{proximity splits}, which refine the decision process by comparing a test instance against two pivots and routing predictions based on the closest match.  
To further improve predictive performance, we integrate \textsc{PivotTree} and its variants into ensemble models. 
By leveraging Random Forests~\cite{breiman2001randomforest} composed of multiple \textsc{PivotTree} classifiers, we enhance the model robustness while preserving interpretability. 
Additionally, to mitigate the complexity of ensemble models, we employ the splitting stump forests approach~\cite{alkhouryW2024splitting}, which extracts compact ensembles of weak learners from a Random Forest~\cite{breiman2001randomforest}, striking a balance between accuracy and efficiency.  
Lastly, \textsc{PivotTree} and its variants are inherently \emph{data-agnostic}, making them adaptable across various data modalities, further broadening their applicability in diverse decision-making scenarios.

\begin{figure}[t]
    \centering
    \begin{subfigure}[b]{\textwidth}
        \centering
        \includegraphics[width=0.4\textwidth]{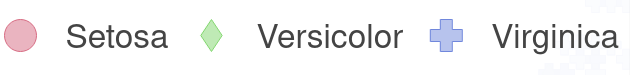}
    \end{subfigure}
    \hfill
    \begin{subfigure}[b]{0.27\textwidth}
        \centering
        \includegraphics[width=\textwidth]{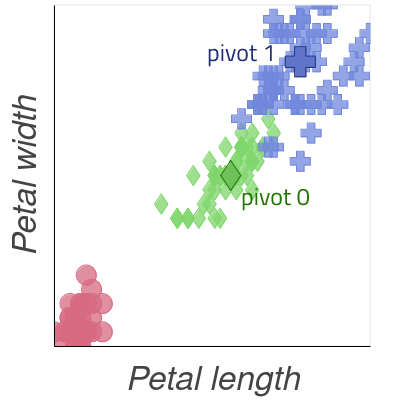}
        \caption{Select}
        \label{fig:pivots}
    \end{subfigure}
    \hfill
    \begin{subfigure}[b]{0.27\textwidth}
        \centering
        \includegraphics[width=\textwidth]{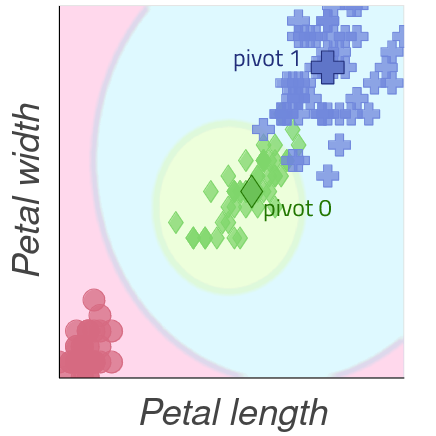}
        \caption{Predict}
        \label{fig:example:similarities}
    \end{subfigure}
    \hfill
    \begin{subfigure}[b]{0.42\textwidth}
        \centering
        \includegraphics[width=\textwidth]{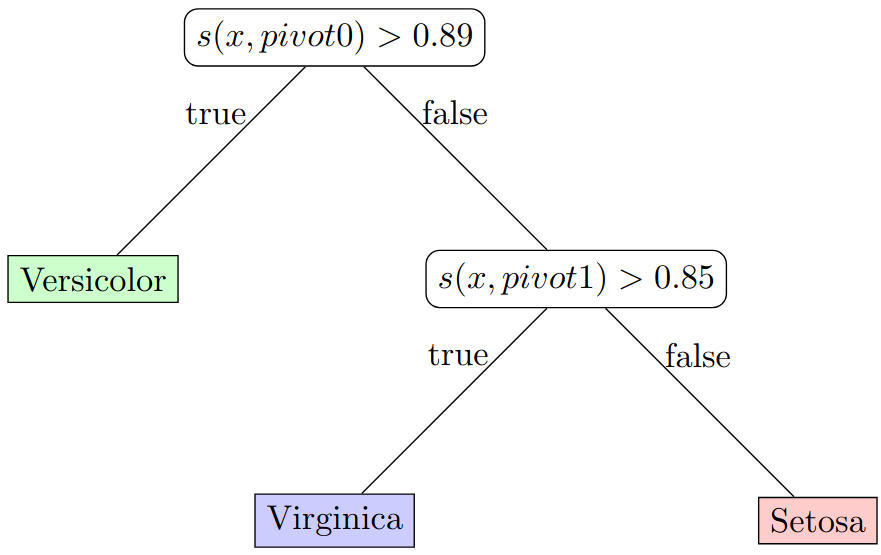}
        \caption{Explain}
        \label{fig:example:tree}
    \end{subfigure}
    \hfill    
    \caption{\textsc{PivotTree} as \textit{(a)} selector, \textit{(b)} interpretable model, \textit{(c)} decision tree w.r.t. the two \textcolor{p0}{\textit{pivot 0}} and \textcolor{p1}{\textit{pivot 1}} highlighted with larger markers.}
    \label{fig:example}
\end{figure}
\begin{figure}[t]
    \centering
    \begin{subfigure}[b]{\textwidth}
        \centering
        \includegraphics[width=0.4\textwidth]{img_legend.png}
    \end{subfigure}
    \hfill
    \begin{subfigure}[b]{0.32\textwidth}
        \centering
        \includegraphics[width=0.9\textwidth]{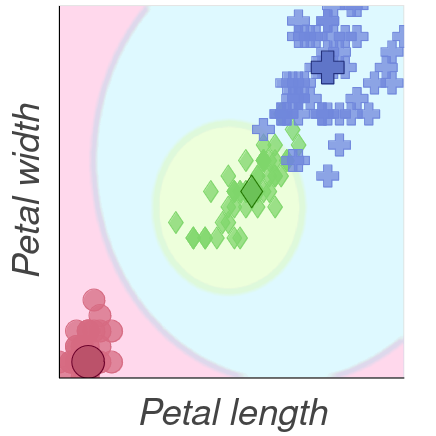}
        \caption{Univariate}
        \label{fig:boundaries:univariate}
    \end{subfigure}
    \hfill
    \begin{subfigure}[b]{0.32\textwidth}
        \centering
        \includegraphics[width=0.9\textwidth]{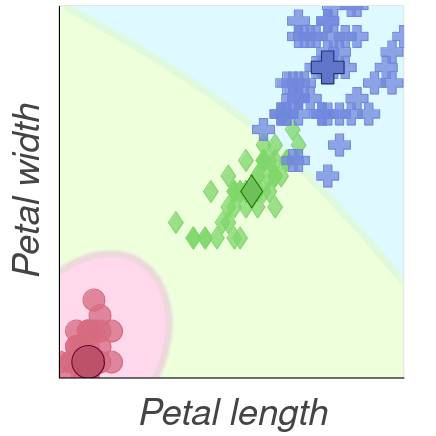}
        \caption{Multivariate}
        \label{fig:boundaries:multivariate}
    \end{subfigure}
    \hfill
    \begin{subfigure}[b]{0.32\textwidth}
        \centering
        \includegraphics[width=0.9\textwidth]{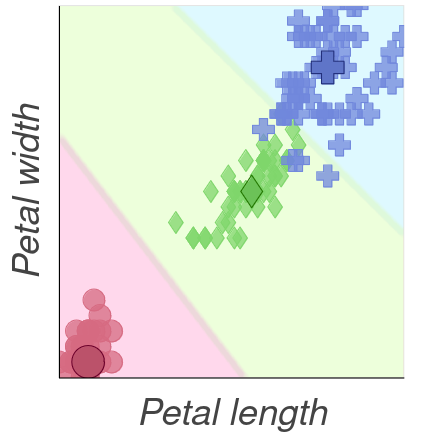}
        \caption{Proximity}
        \label{fig:boundaries:proximity}
    \end{subfigure}
    \hfill    
    \caption{\textsc{PivotTree} decision boundaries induced by \textit{(a)} univariate, \textit{(b)} multivariate, and \textit{(c)} proximity splits w.r.t. the same three pivots highlighted with larger markers.}
    \label{fig:boundaries}
\end{figure}

Figure~\ref{fig:example} provides an example of \textsc{PivotTree} using the well-known \texttt{iris} dataset, where flowers are classified based on their petal characteristics.
The process begins with \textsc{PivotTree} selecting a subset of representative instances, referred to as \textit{pivots} (Figure~\ref{fig:example} \textit{(a)}), which serve as exemplars for classification. 
These pivots form the foundation of a case-based model, allowing new instances to be represented in terms of their similarity to the selected pivots (Figure~\ref{fig:example} \textit{(b)}). 
Building on this pivot selection, \textsc{PivotTree} constructs a hierarchical structure that classifies instances based on their proximity to different pivots. 
This hierarchy takes the form of a decision tree (Figure~\ref{fig:example} \textit{(c)}), where test instances traverse the tree, progressively aligning with the most similar pivots until they reach a classification leaf.
For example, given a test instance $x$: if its similarity to \textit{pivot 0} exceeds 0.89 (following the left branch), $x$ is classified as a \textit{Versicolor} flower. 
Otherwise (following the right branch), if $x$'s similarity to \textit{pivot 1} is greater than 0.85 (left branch), $x$ is classified as a \textit{Virginica}.
If neither condition is met, $x$ is classified as a \textit{Setosa}.
In contrast, a traditional decision tree would establish decision boundaries using feature-based rules, such as:
``If petal length $<$ 2.4, then classify as \textit{Setosa}; else if petal width $<$ 1.7, then classify as \textit{Versicolor}; otherwise, classify as \textit{Virginica}''.
However, this conventional approach has two key limitations: \emph{(i)} it can only model axis-parallel splits, restricting its flexibility,  \emph{(ii)} it is ineffective for data types where features lack clear semantic meaning.
\textsc{PivotTree} overcomes these limitations by enabling interpretability in complex domains such as images, text, and time series -- areas where conventional interpretable models often struggle both in performance and clarity.
Moreover, besides the univariate splitting strategy described above, \textsc{PivotTree} is designed to support alternative splitting approaches. 
These include a multivariate strategy based on oblique splits~\cite{wickramarachchi2016hhcart,DBLP:conf/kdd/KairgeldinC24}, and a proximity-based strategy, which refines decision choices by selecting the closest pivot between two candidates~\cite{DBLP:journals/datamine/LucasSPOZGPW19}.
Each strategy provides further expressivity to the model as shown in Figure~\ref{fig:boundaries}.
Univariate strategies are the most stable and provide the fastest induction times.
Multivariate strategies are more powerful, and leverage similarity to \textit{multiple} pivots.
Finally, proximity strategies provide comparison-based splits: each node holds a set of pivots, one per class, each associated to a different subtree, and determining the instances' routing.
Furthermore, unlike state-of-the-art distance-based predictive models such as \textsc{knn}~\cite{fix1985discriminatory}, our method introduces a hierarchical structure, enhancing the interpretability and efficiency of similarity-based classification.

Experiments conducted on 45 datasets spanning diverse modalities, including tabular data, time series, images, and text, demonstrate that \textsc{PivotTree} and its variants produce interpretable predictive models that match the effectiveness of state-of-the-art approaches while significantly reducing complexity, as measured by the number of pivots and decision stumps.  
A qualitative analysis and a real case study on oral lesion diagnosis highlight the model’s strong performance across different data types.

The rest of the paper is organized as follows.
In Section~\ref{sec:related} we illustrate the related works Section~\ref{sec:method} illustrates our proposals, and Section~\ref{sec:experiments} reports the experimental results. 
Finally, Section~\ref{sec:conclusion} summarizes our contributions, details its limitations and discusses future research directions.

\section{Related Work}
\label{sec:related}
We can categorize case-based methods in two main families: \textit{selection} methods, aiming to, given a fixed data representation, learn the proper pivots through similarity on said representation; and \textit{representation} methods, which instead fix a similarity function, and aim to learn a proper instance representation. 
Such representations in turn are used to select the pivots.
We underline that in this section we adopt the term \textit{pivot} to refer to the instances selected by different proposals in the state-of-the-art which do not necessarily adopt this term.

\smallskip
\textbf{Selection.} Underlying selection methods is the assumption of a fixed data representation.
Among them, we can distinguish three subclasses of methods: \textit{covering}, \textit{clustering}, and \textit{partitioning} methods.
\emph{Covering} methods aim to group records around pivots.
\textsc{$\varepsilon$-ball}~\cite{bien2011epsilon} jointly learns a set of distance-based neighboring \textit{coverages} centered on a set of pivots.
Pivots are optimized to be as few as possible, while coverages to be as class-pure as possible.
The resulting pivots are thus laid on a ``flat'' structure where no structure defines the relationship \emph{among} pivots.
Rather than selected, pivots may also be learned: \textsc{M-Peer}~\cite{DBLP:journals/telo/FilhoLP22} learns a set of synthetic pivots by evolutionary optimization, searching for pivots with high coverage and purity.

\emph{Clustering} algorithms can provide more nuanced pivot-to-pivot relationships.
The \textsc{MiniMax} algorithm~\cite{bien2011hierarchical} builds on agglomerative clustering by identifying cluster representatives and aggregating them hierarchically, resulting in a hierarchy of prototypes.
\textsc{PivotTree} improves on \textsc{MiniMax} by greatly improving on its complexity, and by leveraging pivots to perform prediction.
\emph{Partitioning} algorithms segment the feature space, assigning a pivot to each segment.
\textsc{ProximityForest}~\cite{DBLP:journals/datamine/LucasSPOZGPW19} induces a forest of similarity-based decision trees routing instances according to two pivot similarities.
Notably, pivots are selected randomly, and so is the similarity function, thus yielding highly randomized trees.
Its extension, \textsc{ProximityForest 2.0}~\cite{DBLP:journals/datamine/TanHSW25} enriches the approach by considering a large set of similarities.
Similarity scores may be adaptive~\cite{DBLP:conf/kdd/SatheA17}, or comparative~\cite{DBLP:conf/icml/HaghiriGL18,DBLP:conf/pakdd/ShiWYL18}.
In the latter case, each node is assigned \textit{two} pivots, and higher similarity to either of them routes instances in either subtree.
This family of approaches has found large success on time series data, where shapelets~\cite{DBLP:journals/datamine/KarlssonPB16} or other transformations, e.g., frequency transformations, are often first applied.

With respect to covering algorithms, \textsc{PivotTree} constructs hierarchies of pivots, thus improving model interpretability.
Following partitioning algorithms, \textsc{PivotTree} partitions the feature space, but unlike \textsc{ProximityForest}, it adopts a pivot selection strategy and a fixed similarity function, greatly improving the robustness and variance of its results.
Moreover, pivots are carefully chosen, rather than randomly selected.

\smallskip
\textbf{Representation.}
Unlike selection methods, \textit{representation methods} fix a similarity function and rely on learning a proper representation of the data to find pivots.
Unsurprisingly, these models are often neural, hence lacking interpretability.
Still, they often retain some interpretable components.
Among the first proposals,~\cite{frosst2017distilling} and~\cite{nauta2021prototree} leverage decision tree-like structures, and introduce soft decision trees, wherein nodes hold pivots, and instances are routed probabilistically towards multiple paths in the tree, thus creating fuzzy chains of pivots.
A similar approach is implemented in \textsc{CNN-Trees}~\cite{DBLP:conf/cvpr/ZhangYMW19}, where the hierarchy of learned pivots also holds specificity properties, each layer of the tree identifying pivots more specific than the previous.
Moreover, each pivot also provides a score indicating its contribution to the final prediction.
Notably, unlike soft trees, routing paths are crisp, thus significantly improving the interpretability of the model inference.
In tree-based representation models, interpretability is provided by both the pivots themselves and their hierarchical structure.

Integrating linear models instead, \textsc{ProtoPNet}~\cite{DBLP:conf/nips/ChenLTBRS19} lays pivots on a flat structure.
Here, interpretability lies in similarity scores assigned to each pivot, but no intra-pivot structure is considered.
A neural model first embeds pivots and instances, then computes similarity between them on a penultimate layer, yielding similarity scores that are fed to a linear scoring model.
The same approach is also leveraged to induce trees of pivots~\cite{DBLP:conf/hcomp/HaseCLR19}, which also integrates a novel class detection mechanism: whenever an instance is sufficiently different from all pivots, or is similar to many different pivots, then the prediction is rejected.
Initially developed for computer vision data, this approach has since been applied to text~\cite{DBLP:conf/acl/0001GKLL22,DBLP:journals/jmlr/HongWB23}, sequential data~\cite{DBLP:conf/kdd/MingXQR19}, and audio~\cite{DBLP:conf/dis/FedeleGP24}. 
The interpretability of prototype-based networks has also been evaluated through adversarial analyses, highlighting the need for more robust architectures~\cite{baniecki2025birds}.
Further improving the expressivity of scoring linear models, Generalized Additive Models (GAMs)~\cite{DBLP:journals/siamrev/McCaffrey92} have also seen application in pivot-based prediction.
\textsc{ProtoNAM}~\cite{DBLP:journals/corr/abs-2410-04723} introduces a neural model wherein representations for instances and pivots are learned jointly, and then provided to a downstream linear model for prediction.
The representations are learned on a feature-basis, thus, unlike other models~\cite{DBLP:conf/nips/ChenLTBRS19}, the instance-pivot relationship is learned at an extremely fine-grained level.
Yet, the model introduces a set of non-neural components that negatively impact interpretability.
Overall, \textsc{PivotTree} shares the tree structure with some models, but improves on them by providing either proper pivot selection, or fast induction.

\section{Methodology}
\label{sec:method}
In this section, we present \textsc{PivotTree} (\textsc{pt}), an interpretable case-based hierarchical pivot selection model inspired by decision trees~\cite{breiman1984classification}, its variants considering multiple pivots for each split, and different ensembles of \textsc{PivotTree}s.
First, we formalize the problem setting of solving a classification task through case-based reasoning. 
Then, we present different strategies for selecting pivots and performing node splitting during the tree-growing process. 
Finally, we discuss the integration of such methods into ensemble approaches inspired by Random Forest classifiers~\cite{breiman2001randomforest}.

\subsection{Problem Setting}
Given a set of instances represented as real-valued $m$-dimensional feature vectors\footnote{We consistently treat data instances as real-valued vectors. Data transformation employed in the experimental section to maintain coherence with this assumption will be specified when needed.} in $\mathbb{R}^{m}$, and given a set of class labels $C = \{1, \ldots, c\}$, we assume the existence of an unknown ground-truth function $g:\mathbb{R}^m \to C$ mapping each vector in $\mathbb{R}^{m}$ to one of the $c$ classes in $C$.
The objective of case-based reasoning is to learn a function $f: \mathbb{R}^m \rightarrow C$ approximating $g$, with $f$ being defined as a function of $k$ exemplary cases named \textit{pivots}.
As explained in Section~\ref{sec:related}, similarity-based case-based models define $f$ on a similarity space $\mathcal{S}$, often inversely denoted as ``distance space'', induced by a similarity function $s: \mathbb{R}^m \times \mathbb{R}^m \rightarrow \mathbb{R}$ quantifying the similarity of instances~\cite{DBLP:series/smpai/PekalskaD05}.

Given a training set $\langle X, Y \rangle$, with a set $X = \{x_1, \dots, x_n\}$ of $n$ instances, and $Y = \{y_1, \dots, y_n\}$ associated class labels, a similarity function $s$, our objective is to learn a function $\pi: \mathbb{R}^{n\times m} \rightarrow \mathbb{R}^{k \times m}$ that takes as input $X$  and returns a set $P \subseteq X$, i.e., $\pi_s(X) = P$, of $k$ pivots such that the performance of $f$ are maximized.

In practical terms, given a training set $\langle X, Y \rangle$ and a similarity function $s$, the selection method $\pi$ selects $k$ pivots $P$ from $X$.
Through the similarity function $s$ and the pivots $P$, the dataset $X \in \mathbb{R}^{m}$ is mapped into the \textit{similarity space} $\mathcal{S}$ w.r.t. the selected pivots, and thus encoded into a representation $Z \in \mathbb{R}^{n \times k}$ where $Z_{i,j}$ is the similarity between the $i$-th instance in $X$ and the $j$-th pivot in $P$.
Hence, the predictive model $f$ is trained on $\langle Z, Y \rangle$.
Then, given a test instance $x \in \mathbb{R}^m$, $x$ is first mapped to a similarity vector $z = \langle s(x, p_1), \dots, s(x, p_k) \rangle$ yielding its similarity to the set $P$ of pivots.
Then, $z$ is provided to $f$, which performs the prediction.

\subsection{Usage of PivotTree}
Aiming for transparency of the case-based predictive model $f$, our objective is to use as an interpretable model $f$, i.e., a decision tree~\cite{breiman1984classification} or k-Nearest Neighbors (kNN)~\cite{guidotti2019survey}.

When $f$ is implemented as a decision tree (\textsc{dt}), we explore three forms of splits:
\begin{itemize}
    \item \textbf{Univariate Split.} Is a ``traditional'' \textit{axis-parallel} split condition of the form $s(x, p_i) \geq \beta$, i.e., ``if the similarity between instance $x$ and pivot $p_i$ is greater or equal than $\beta$, then ...'', that allows to easily understand the logic adopted by inspecting $x$ and $p_i$ for every condition of the rule.
    
    \item \textbf{Multivariate Split.} Is an \textit{oblique} split condition~\cite{wickramarachchi2016hhcart,murthy1994system} of the form 
    $\omega_{i} s(x, p_i) + \omega_{j} s(x, p_j) \geq \beta$, i.e., ``if the weighted sum of the similarities between instance $x$ and pivots $p_i$ and $p_j$, with weights $\omega_{i}$ and $\omega_{j}$, is greater or equal then $\beta$, then ...'', that allows to easily understand the logic adopted by inspecting $x$ and $p_i$ against $x$ and $p_j$ for every condition of the rule. 
    A multivariate split allows to leverage multiple pivots (limited to two to grant interpretability), thus increasing the expressivity of the model. 
    Moreover, similarity to each pivot is weighted by a learnable weight $\omega$.
    \item \textbf{Proximity Split.} Is a \textit{proximity} split condition inspired by the Proximity Forest developed for Time Series Classification~\cite{DBLP:journals/datamine/LucasSPOZGPW19,tan2025proximity,zhang2021proximity} of the form 
    $s(x, p_i) > s(x, p_j)$, i.e. ``if instance $x$ is closer to pivot $p_i$ than to pivot $p_j$, then ...'' that allows to understand in a much easier way the logic adopted by comparing $x$ against $p_i$ and $p_j$ for every condition of the rule.
    Also, this type of split allows to reason in terms of relative, rather than absolute similarity, providing an additional and different layer of interpretability for the user.

\end{itemize}

On the other hand, when $f$ is implemented as a \textsc{knn}, every decision is based on the similarity with a few neighbors, typically between one and five, in the 
similarity space $\mathcal{S}$ obtained by computing the similarity between each instance with respect to the selected pivots.
As an alternative, we consider a variant where \textsc{knn} uses only the set of selected pivots as a training set, performing the prediction w.r.t. the original feature space.
A human user just needs to inspect $x$ and the similarities with the pivots $P$ and the instances in the neighborhood. 
When the number of pivots is kept small, the interpretability of both methods increases, limiting the expressiveness.
Vice versa, using a selection model $\pi$ that returns a large number $k$ of pivots can increase the performance at the cost of interpretability.
Our proposal aims to balance these two aspects by allowing the selection of a small number of pivots that still guarantee comparable performance to interpretable predictive models.

\subsection{Pivot Tree Algorithm}
We present here \textsc{PivotTree} implementing the selection function $\pi$.
Much like decision tree induction algorithms~\cite{breiman1984classification}, \textsc{PivotTree} greedily learns a hierarchy of nodes, each node splitting instances towards one of its two children, ultimately reaching terminal leaf nodes, which are associated with a classification label.
The splitting is based on \textit{discriminative} pivots and \textit{descriptive} pivots.
Let $X_t$ be the instances constrained by the decision path at iteration $t$ in the tree construction, and $Y_t$ the associated class labels. 
We call \textit{discriminative pivots} the instances of class $c$, i.e., $p^- \in X^{(c)}_t = \{ x_i | x_i \in X_t \wedge y_i = c \}$ that maximize the impurity gain when partitioning $X_t$ w.r.t. the similarity to such instances. 
We discuss details of the different approaches for selecting such instances in the next section.
Furthermore, besides discriminative pivots, for each iteration $v$, \textsc{PivotTree} also identifies \textit{descriptive pivots}.
A descriptive pivot is an instance of class $c$, i.e., $p^+ \in \{ x_i | x_i \in X_t \wedge y_i = c \}$ that maximizes the similarity with all the other instances described by the same node and belonging to the same class, i.e., $p^+ = \arg\max_{p' \in X^{(c)}_t}\sum_{x_i \neq p' \in X^{(c)}_t} s(x_i, p')$.

\begin{algorithm}[t]
\caption{\textsc{PivotTree}$(X, Y)$}
\label{alg:pivottree_glance}
\begin{algorithmic}[1]
    \Input $\langle X, Y \rangle$ data and labels, $s$ similarity function, $\mathit{maxdepth}$ maximum tree depth
    \Output $P$ set of pivots, $T$ learned tree
    \State $T \gets \emptyset$; $P \gets \emptyset$; 
    \Comment{\texttt{\scriptsize Variables initialization}}
    \State $S \gets \langle s(x_i, x_j) \rangle \; \forall x_i, x_j \in X \times X$ \Comment{\texttt{\scriptsize Calculate similarity matrix}}
    \State $P, T \gets \Call{ptr}{X, Y, T, P, S}$; \Comment{\texttt{\scriptsize Start of recursive procedure}}
    \State \Return $P, T$
    \newline
    
    \Function{ptr}{$X, Y, T, P, S$}
        \If{$\Call{depth}{T} \leq \mathit{maxdepth}$}
            \For{$c \in C$}
                \State $P^- \gets \Call{GetDisc}{X, Y, S}$; \Comment{\texttt{\scriptsize Get discriminative pivots}}
                \State $P^+ \gets \{arg\max_{x \in X} \sum_{x \neq x' \in X^{(l)}} s(x, x')$\}; \Comment{\texttt{\scriptsize Get descriptive pivot}}
                \State $P \gets P \cup P^- \cup P^+$; \Comment{\texttt{\scriptsize Add pivots to result set}}
            \EndFor
            \State $X_l, X_r, Y_l, Y_r \gets \Call{SplitData}{X, Y, P}$; \Comment{\texttt{\scriptsize Split data w.r.t. $P$}}
            \State $P_l, T_l \gets \Call{ptr}{X_l, Y_l, T, P, S}$ 
            \Comment{\texttt{\scriptsize Recourse on left child}}
            \State $P_r, T_r \gets \Call{ptr}{X_r, Y_r, T, P, S}$
            \Comment{\texttt{\scriptsize Recourse on right child}}
            \State $T \gets \Call{addSplitToTree}{T, T_l, T_r}$; \Comment{\texttt{\scriptsize Add split to tree}}
            
            \State \Return $P, T$; \Comment{\texttt{\scriptsize Return current pivots and tree}}
        \Else
            \State $P^+ \gets \{arg\max_{x \in X} \sum_{x \neq x' \in X^{(l)}} s(x, x')$\}; \Comment{\texttt{\scriptsize Get descriptive pivot}}
            \State $P \gets P \cup P^+$; \Comment{\texttt{\scriptsize Add pivots to result set}}
            \State \Return $P, \Call{makeLeaf}{T}$; \Comment{\texttt{\scriptsize Return current pivots and leaf}}
        \EndIf
    \EndFunction
    
\end{algorithmic}
\end{algorithm}

In Algorithm~\ref{alg:pivottree_glance}, we illustrate the pseudo-code for training a \textsc{PivotTree}.
Given the dataset and labels $\langle X, Y \rangle$, the similarity function $s$, the maximum tree depth $\mathit{maxdepth}$, \textsc{PivotTree} returns the set $P$ of selected pivots, and the trained decision tree $T$ (line 4).
After initializing the tree and pivots (line 1),
\textsc{PivotTree} calculates a similarity matrix $S$ between all pairs of instances in $X$ (line 2).
Then, the recursive procedure \textsc{ptr} is started (line 3).
The \textsc{ptr} procedure is illustrated in lines 5--19 in Algorithm~\ref{alg:pivottree_glance}.
If the current depth of the tree \textsc{depth}$(T)$ is lower than the maximum tree depth $\mathit{maxdepth}$ (line 6), then for each class, the most discriminative and most descriptive pivots are selected and added to the result set $P$ (lines 7--10)\footnote{To ease the computational burden, and similarly to other implementations, e.g., \texttt{scikit-learn}, only a subset of candidate splits is tested.}.
We notice that, since the similarity matrix $S$ is calculated at the beginning, the pairwise similarities to select the most discriminative and descriptive pivots are available without performing any calculus.
The set $P$ of discriminative and descriptive pivots is then used to select the best split to partition the data with the \textsc{SplitData} function, again maximizing the Information Gain w.r.t. the similarities w.r.t. the pivots in $P$ (line 11).
After that, \textsc{PivotTree} recourses on the left and right subsets $X_l, Y_l$ and $X_r, Y_r$ and composes the tree returned (lines 12--14).
On the other hand, if the maximum depth (line 16) or other stopping conditions are met, then the current pivots, augmented with the descriptive pivots of the records in the leaf, and a leaf itself (lines 17--19), are returned.
Thus, the complexity of the \textsc{PivotTree} is theoretically bounded by the calculus of the similarity matrix $S$.

Besides being used as a pivot selector method ($\pi$), we underline that \textsc{PivotTree} can be employed as a standalone predictive model by combining the encoding in the similarity space and the tree induction $f$.
In this case, we do not need to train additional interpretable models, as both pivot selection and case-based prediction are already integrated into the model.
The variability in how a \textsc{PivotTree} can be learned lies in the different ways the \textsc{GetDisc} and \textsc{SplitData} functions can be implemented. 
We discuss proposed strategies for implementing both functions in the next section.

\subsection{Pivot Tree Variants}
\label{sec:flavors}
We present here \textsc{PivotTree} variants according to the different strategies adopted.

\smallskip
\textbf{Univariate Pivot Tree (}\textsc{pt}\textbf{).}
The ``standard'' implementation of both \textsc{GetDisc} and \textsc{SplitData} employs univariate decision splitting function.
Let $X_t$ be the instances constrained by the decision path at iteration $t$, and $Y_t$ the associated class labels. 
For each $c$ occurring in $Y_t$, we consider the similarity of each instance $x \in X_t $ w.r.t. each one of the $c$ class instances in $X_t$, collecting similarity values for each instance in a real-valued vector.
This is a way to represent the instances of $X_{t}$ in a feature space where each dimension stands for the similarity an instance has to another $c$-class instance. 
On such data representation, a univariate linear splitting function $f_{i}$ can be learned not differently from how it is learned in the original feature space. 
Since each feature in this space corresponds to a $c$-class instance, finding the optimal feature-threshold pair to partition the data for impurity gain in this context corresponds to identifying the $c$-class instance that best separates $X_t$ based on similarity. 
This is the discriminative $c$-class pivot of $X_t$. 
Therefore, \textsc{GetDisc} is implemented by performing a univariate split on the newly defined similarity feature-space, utilizing standard approaches such as CART, ID3, or C4.5~\cite{DBLP:conf/ACISicis/ThaiphanP18}.
After finding the optimal split condition, the corresponding discriminative pivot $p^{-}$ is extracted as the pivot whose similarity feature was chosen to define the optimal split.
This process is repeated for each class $c \in Y_t$, thereby extracting a distinct {discriminative pivot} $p^{-}$ for every class.
We further specify that the set of discriminative pivots ${P^-}$ is collected along with the set of {descriptive pivots} ${P^+}$. This combined set, ${P} = {P^-} \cup {P^+}$, constitutes the {pivot collection} for the node, which is accumulated during the tree growing process. The final collection of candidate pivots extracted from each internal node after the training process can be used at the end to define the $k$ exemplar instances to be utilized by an external case-based classifier.
To finally complete the splitting procedure in this standard context, the \textsc{SplitData} function is also implemented as a univariate split. More specifically, it identifies the unique best univariate split obtained by considering the similarity feature-space defined for each $x \in X_t$ with respect to the whole set of candidate pivots $P$ previously extracted. The split is ultimately chosen as the best univariate split in terms of impurity gain within this similarity feature-space. A single pivot $p$ is then selected from the candidate set $P$ to separate $X_t$, providing the univariate split condition used by the \textsc{PivotTree} model when trained as a standalone classifier. This enables the training process to continue by routing the instances in $X_t$ to the corresponding child nodes accordingly.

The described training process, characterized by first extracting candidate pivots to be collected and potentially used for external classifiers, and at the same time learning a standalone case-based classification model, can be naturally extended to support a variety of different splits, as we describe in the following.

\textbf{Oblique Pivot Tree (}\textsc{opt}\textbf{).}
We introduce \textit{multivariate splits}, often referred to as \textit{oblique} splits, for learning \textsc{PivotTree}s.
Indeed, instead of selecting a single feature-threshold pair in the similarity feature space, we can employ a linear combination of two feature values with non-zero coefficients to determine the splits. 
The same splitting strategies used to define multivariate splits for feature-based trees, such as HHCART or bivariate CART~\cite{wickramarachchi2016hhcart,DBLP:conf/kdd/KairgeldinC24}, can be applied in this context. 
For the \textsc{GetDisc} procedure, rather than selecting a single discriminative pivot for each class, two discriminative pivots of different classes are chosen, finding the linear combination of their similarity values which better partitions the data in terms of impurity gain. 
Similarly, for the \textsc{SplitData} procedure, two pivots are selected from the candidate pool to perform the split, with each pivot contributing according to its assigned weight in the combination, choosing the best multivariate split in terms of impurity gain. 

\smallskip
\textbf{Proximity Pivot Tree (}\textsc{ppt}\textbf{).}
To implement the \textit{proximity} variant, we adopt the same strategy used by the univariate \textsc{pt} to select the discriminative pivots, but then we apply a different strategy to implement the \textsc{SplitData} procedure. 
Inspired by Proximity Forests~\cite{DBLP:journals/datamine/LucasSPOZGPW19}, we implement \textsc{SplitData} as follows. From the set of candidate pivots $P$ we select a pair of pivots $p_i, p_j$ with different class labels, and we partition $X_{t}$ in $X_{t, l} = \{x_h \in X_t | s(x_h, p_i) > s(x_h, p_j)\}$ and $X_{t, r} = \{x_h \in X_t | s(x_h, p_i) > s(x_h, p_j)\}$, such that it does not exist another pair $\hat{p_i}, \hat{p_j}$ leading to a better partitioning in terms of impurity gain. 
Intuitively, each branch (left or right) is associated with a pivot selected from the candidate pool. 
Instances in $X_t$ are then routed based on their relative similarity to the pivots of the left and right branches. 
This approach compares the similarity of an instance to the respective pivots, rather than splitting based on a single similarity value or a linear combination of similarity values.
Differently from the standard Proximity Tree~\cite{DBLP:journals/datamine/LucasSPOZGPW19}, we do not randomly sample candidates or proximity measure, but we use \textsc{pt} criterion to identify the best instances instead.

\smallskip
\textbf{Oblique Proximity Pivot Tree (}\textsc{oppt}\textbf{).}
This approach combines the \textsc{GetDisc} strategy implemented in \textsc{opt} with the \textsc{SplitData} strategy of \textsc{ppt}. 
Specifically, a multivariate splitting approach is applied w.r.t. each class to identify multiple discriminative pivots per class. 
The choice of descriptive pivots remains the same as the approaches above. 
Then, for each possible pair of candidate pivots $p_i, p_j$, the information gain of partitioning the instances routed at that node based on similarity to these pivots is evaluated. 
The pair that maximizes the information gain is selected as the optimal split condition.
To enhance the interpretability of the resulting tree structures, we restrict our search to pairs of pivots belonging to different classes.

\subsection{Random Pivot Forest}
Similarly to traditional decision trees, \textsc{PivotTree} directly used as classifier, can be integrated into an ensemble model such as Random Forest~\cite{breiman2001randomforest}. 
An ensemble of \textsc{PivotTrees}, which we name \textsc{RandomPivotForest}, can be constructed by instantiating multiple \textsc{PivotTree} estimators, each trained on a randomly selected subset of features and/or samples, and performing predictions through majority voting, aggregating the prediction of each base tree. 
We underline that this approach can be applied to any of the previously introduced \textsc{PivotTree} variants\footnote{Although using ensemble approaches with \textsc{PivotTree} as weak estimators may hinder transparency, interpretability can still be recovered by examining feature importances or by adapting methods that extract standalone interpretations from forests~\cite{bonsignori2021deriving,pensa2025explaining}. We leave this investigation to future work, focusing on a straight comparison between standard tree ensembles and \textsc{RandomPivotForest}s.}.
Furthermore, the ensemble approach can be extended by using a splitting stump forest compression strategy presented in~\cite{alkhouryW2024splitting}. 

Following their method description, given a trained forest $\Phi$, the splitting stump forest approach extracts a compact set of simple trees named, \emph{splitting stumps} whose split conditions are well balanced according to a balance score.
Let us consider a non-terminal node of one of the trees comprising $\Phi$, and let $X$ be the set of training samples reaching that node. In~\cite{alkhouryW2024splitting}, the authors define the \emph{balance score} of the split associated with that node as  $\frac{\min(|X^{\top}|, |X^{\bot}|)}{|X^{\top}| + |X^{\bot}|}$, where $X^{\top}$ denotes the set of training instances reaching that node that evaluate to \textit{true} with respect to the split condition, and $X^{\bot}$ the training instances reaching that node that evaluate to \textit{false}.
For example, if a split sends $6$ samples to the true child and $9$ samples to the false child (i.e., $6$ out of $15$ evaluate to \textsc{True}), then the balance score of that split is $0.4$. A split is considered \emph{balanced} according to~\cite{alkhouryW2024splitting} if its balance score meets or exceeds a predetermined threshold $\rho$, which determines how tolerant we are w.r.t. the presence of imbalance in the split.
If the split associated with a node is balanced, then a \emph{splitting stump} is created by treating that split as a standalone decision-tree classifier performing a single split. The stump is added to a set data structure to avoid duplicate entries. This evaluation is performed for all nodes across all trees in the forest.
At the end of this process, collecting all obtained stumps yields a new ensemble model $\Phi'$, which can be directly used for prediction or, as suggested in~\cite{alkhouryW2024splitting}, leveraged to learn a new data representation that maps each data point to a set of leaves.
More specifically, for each obtained splitting stump $\phi \in \Phi'$, a function $b_{\phi}:\mathbb{R}^m \to \{0,1\}$ can be defined where $b_{\phi}(x) = 1$ if and only if the split condition associated to $\phi$ evaluates to \textit{true} on $x$.
We can then define a new representation of the feature vector $x$ as the concatenation of the output $b_{\phi}(x)$ for all $\phi \in \Phi'$.
A new function $f$ can be learned w.r.t. this new binary vector representation of the instances.
As suggested in~\cite{alkhouryW2024splitting}, a logistic regression model can be employed due to its resource efficiency and interpretability, learning to predict the target variable based on the assignment of leaves in the stumps.
In our analysis, we experiment with the use of \textsc{RandomPivotForest} combined with the splitting stump strategy using a logistic regression classifier at the end. 
Furthermore, we also build upon the analysis conducted in~\cite{alkhouryW2024splitting} by additionally exploring an approach that, to foster interpretability, stops at the stump selection step, and treats the resulting stump forest as an ensemble of estimators which are then used directly for prediction through majority voting, without any further data transformation.

\subsection{Data Agnosticism}
By design, \textsc{PivotTree} with the different types of splits and with the ensemble is a data-agnostic model that leverages the concept of similarity to conduct both selection and prediction tasks simultaneously.
While some data types, e.g., relational data, are more amenable than others, e.g., images or text, to similarity computation, with our contribution, we aim to address all data types as one.
By decoupling similarity computation and object representation, \textsc{PivotTree} can be applied to any data type supporting a mapping to $\mathbb{R}^m$, i.e., text through language models, images through vision models, graphs through graph models, etc.
In our experimentation, besides tabular data, we focus on time series, images, and text.

\section{Experiments}
\label{sec:experiments}
We evaluate here the performance of \textsc{PivotTree} in its variants on different datasets of different types\footnote{We employ a Python implementation of \textsc{PivotTree} as part of the \textsc{RuleTree} framework available at~\url{https://github.com/fismimosa/PivotTree}}.
We aim to demonstrate that \textsc{PivotTree} is competitive w.r.t. standard classifiers while at the same time being a simple, interpretable, and effective prototype selection method.
We adopt the nomenclature introduced in Section~\ref{sec:method} to describe the splitting strategies utilized. 
Specifically, we use \textsc{pt} and \textsc{opt} to denote the univariate and oblique variants, respectively. 
Additionally, we refer to the proximity-based strategy as \textsc{ppt} and its oblique variant as \textsc{oppt}. 
When \textsc{PivotTree} models are employed as standalone \textsc{c}lassification models, we denote them as \textsc{ptc}, \textsc{pptc}, and so on. 
Conversely, when they are used solely as pivot \textsc{s}electors for a different classifier, we refer to them as \textsc{pts}, \textsc{ppts}, and so forth.
When we refer to the ensemble \textsc{RandomPivotForest} variants of such methods, we use \textsc{rpt} for the univariate case, \textsc{rppt} for the proximity-based split case, and \textsc{ropt} and \textsc{roppt} for their oblique counterparts.

\subsection{Experimental Setting}
In this section, we illustrate the datasets adopted, \textsc{PivotTree} setting and hyperparameters,
competitors and baselines, and evaluation measures.

\smallskip
\textbf{Datasets.}
In order to show the effectiveness of our proposal for different data types, we experimented with 20 tabular datasets, 10 time series datasets, 9 image datasets, and 6 text datasets, for a total of 45 different datasets. 
Table~\ref{tab:datasets} reports dataset details\footnote{Detailed preprocessing steps for the different datasets are available on the project repository.}.
The tabular datasets are split into 70\% training and 30\% testing, along with the \texttt{oral} and \texttt{pol} datasets. The remaining datasets come with predefined training and test splits. 
For the image datasets, we adhere to the splits defined in~\cite{pugnana2024benchmark}, using their calibration set as our training set and adopting their test set as ours. 
This approach prevents the risk of data leakage, as we utilize their pre-trained models to extract latent image representations\footnote{ \url{https://github.com/andrepugni/ESC}.}.
For tabular datasets, in order to perform a direct distance comparison between instances, we leave unvaried numeric and ordinal features, while we one-hot encode categorical ones. 
For image datasets, we extract latent feature representations using as trained models either a standard \textsc{ResNet} or \textsc{VGG} architecture, as employed for each distinct dataset in~\cite{pugnana2024benchmark}. 
These models are trained with standard Cross Entropy Loss, and we consider the representations obtained from both architectures as latent embeddings before passing the transformed input into the final classification layer. 
More details on the extraction process are available in our paper repository.
For textual datasets, we embed the input text with the \texttt{all-mpnet-base-v2} sentence transformer model\footnote{ \url{https://huggingface.co/sentence-transformers/all-mpnet-base-v2}}, which yields $L2$-normalized 768-dimensional dense vector with magnitude 1. 
On the basis of these encodings, the similarity $s$ is based on the Euclidean distance for each data type.
While text embeddings usually rely on cosine similarity, in~\cite{korenius2007cosine}, it is shown that under unit normalization, the two are directly proportional and thus order-preserving, justifying our uniform usage of Euclidean distance.
The datasets are normalized using z-score normalization, where the mean is subtracted, and the values are scaled to unit variance. Specifically, each train-test set within every fold is independently normalized to ensure consistency during cross-validation. 
For model assessment, the train and test sets are normalized jointly, using the statistical properties calculated from the training data.

\begin{table}[t]
\centering
\caption{Datasets used for model selection and assessment.}
\label{tab:datasets}
\setlength{\tabcolsep}{7.2mm}
\footnotesize
\begin{tabular}{l@{\hspace{4mm}}c|c|c|c|c}
\toprule
\multirow{2}{*}{} & \multicolumn{5}{c}{dataset details} \\
 & dataset & training size & test size & features & labels \\
\midrule
\multirow{16}{*}{\rotatebox{90}{\texttt{tabular}}}
    & $\texttt{ion}$ & 245 & 105 & 35 & 2 \\
    & $\texttt{fire}$ & 170 & 73 & 13 & 2 \\
    & $\texttt{yeast}$ & 1038 & 446 & 8 & 10 \\
    & $\texttt{magic}$ & 13314 & 5706 & 10 & 2 \\
    & $\texttt{sonar}$ & 145 & 63 & 60 & 2 \\
    & $\texttt{compas}$ & 5049 & 2165 & 17 & 3 \\
    & $\texttt{house}$ & 15948 & 6836 & 16 & 2 \\
    & $\texttt{german}$ & 700 & 300 & 61 & 2 \\
    & $\texttt{spamb}$ & 3220 & 1381 & 57 & 2 \\
    & $\texttt{norm}$ & 5180 & 2220 & 20 & 2 \\
    & $\texttt{lrs}$ & 371 & 160 & 100 & 9 \\
    & $\texttt{vert}$ & 217 & 93 & 6 & 2 \\
    & $\texttt{iris}$ & 105 & 45 & 4 & 3 \\
    & $\texttt{wine}$ & 4547 & 1950 & 12 & 2 \\
    & $\texttt{diva}$ & 8561 & 3669 & 330 & 2 \\
    & $\texttt{breast}$ & 398 & 171 & 30 & 2 \\
    & $\texttt{steel}$ & 1358 & 583 & 27 & 7 \\
    & $\texttt{ecoli}$ & 235 & 101 & 7 & 8 \\
    & $\texttt{heloc}$ & 7321 & 3138 & 23 & 2 \\
    & $\texttt{page}$ & 3831 & 1642 & 9 & 2 \\
\midrule
\multirow{8}{*}{\rotatebox{90}{\texttt{images}}} 
    & $\texttt{oral}$ & 454 & 89 & 256 & 3 \\
    & $\texttt{mnist}$ & 7000 & 14000 & 512 & 10 \\
    & $\texttt{cifar10}$ & 6000 & 12000 & 512 & 10 \\
    & $\texttt{catsdogs}$ & 2500 & 5000 & 512 & 2 \\
    & $\texttt{birds}$ & 1179 & 2358 & 512 & 2 \\
    & $\texttt{pets}$ & 735 & 1470 & 512 & 2 \\
    & $\texttt{organa}$ & 5883 & 11766 & 512 & 11 \\
    & $\texttt{blood}$ & 1710 & 3419 & 512 & 8 \\
    & $\texttt{svhn}$ & 9929 & 19858 & 512 & 10 \\
\midrule
\multirow{8}{*}{\rotatebox{90}{\texttt{time-series}}} 
    & $\texttt{yoga}$ & 300 & 3000 & 426 & 2 \\
    & $\texttt{star}$ & 1000 & 8236 & 1024 & 3 \\
    & $\texttt{chlorine}$ & 467 & 3840 & 166 & 3 \\
    & $\texttt{kitchen}$ & 375 & 375 & 720 & 3 \\
    & $\texttt{share}$ & 965 & 966 & 60 & 2 \\
    & $\texttt{devices}$ & 8799 & 7494 & 96 & 7 \\
    & $\texttt{gun}$ & 50 & 150 & 150 & 2 \\
    & $\texttt{worms}$ & 158 & 77 & 900 & 2 \\
    & $\texttt{ecg}$ & 500 & 4500 & 140 & 5 \\
    & $\texttt{wafer}$ & 1000 & 6164 & 152 & 2 \\
\midrule
\multirow{5}{*}{\rotatebox{90}{\texttt{text}}} 
     & $\texttt{medabs}$ & 11550 & 2888 & 768 & 5 \\
    & $\texttt{vicuna}$ & 5984 & 1511 & 768 & 2 \\
    & $\texttt{pted}$ & 5027 & 1257 & 768 & 2 \\
    & $\texttt{tgpt}$ & 3548 & 888 & 768 & 2 \\
    & $\texttt{pol}$ & 1400 & 600 & 768 & 2 \\
    & $\texttt{liar}$ & 11508 & 1267 & 768 & 6 \\
\bottomrule
\end{tabular}
\end{table}

\smallskip
\textbf{Pivot Tree Settings.}
We evaluate \textsc{PivotTree} both as a pivot selection function ($\pi$) and as a standalone interpretable classifier ($f$), considering all variants introduced in Section~\ref{sec:flavors}.
For all variants, whether used as a \textsc{s}elector or \textsc{c}lassifier, the optimal $\mathit{maxdepth}$ is searched in $\{2,3,4\}$. 
Furthermore, whenever a multivariate split is employed, we use the HHCART strategy with 2 components~\cite{wickramarachchi2016hhcart}.

When employing \textsc{PivotTree} as a selector, model selection also involves optimizing the type of pivots extracted and used for the task. 
Specifically, we evaluate four configurations to choose the final pivot set: 
\emph{(i)} using only discriminative pivots identified during training, 
\emph{(ii)} using only descriptive pivots, 
\emph{(iii)} employing both discriminative and descriptive pivots, 
\emph{(iv)} or selecting pivots exclusively from those chosen as split conditions for each internal node.  
In our results, we report only the best performance achieved among these configurations for each of the four \textsc{PivotTree} variants.

When \textsc{PivotTree} is used as \textsc{s}elector $\pi$, we evaluate as interpretable predictive models $f$ both univariate decision trees (\textsc{dt}) and k-Nearest Neighbor (\textsc{knn}) classifiers. 
The \textsc{knn} implementation is sourced from the \texttt{scikit-learn} library, while \textsc{dt} is implemented using the \texttt{RuleTree} Python library.
For \textsc{dt}, we test w.r.t a $\mathit{maxdepth}$ interval of $\{2,3,4\}$. 
For \textsc{knn}, we determine the optimal configuration by searching for $\kappa$ in $\{1,3,5\}$, where $\kappa$ represents the number of nearest neighbors considered during classification.
We explore two distinct training strategies for \textsc{knn} when combined with \textsc{PivotTree} \textsc{s}elector. 
In the first approach, \textsc{knn} is trained in the similarity feature space $Z$, where each instance is represented by its similarity to the extracted pivots. 
In this case, the entire original training set is retained as a reference but represented in the similarity space. 
Instead, in the second approach, \textsc{knn} is trained in the original feature space $X$, using as its reference set the $k$ pivots $P$ selected, where $k \leq |X|$.
We further include, as additional baselines, two rule-set models, namely \textsc{ripper}~\cite{cohen1995fast} and \textsc{irep}~\cite{furnkranz1994incremental}, as well as a model that uses CART to fit a list of univariate decision rules rather than a full tree (\textsc{grule})\footnote{Implementations are available at 
\url{https://github.com/imoscovitz/wittgenstein} and 
\url{https://github.com/csinva/imodels}. For both \textsc{irep} and \textsc{ripper}, we use the default discretization method provided by the respective implementations, which applies binning into 10 intervals.}. All these further rule-based approaches are trained on the original feature space, exploring a maximum number of rules in $\{4,8,16,32\}$.
%}
%
To distinguish between all these strategies, we denote methods trained in the similarity space with the subscript $Z$, while those trained in the original feature space considering only the pivots with the subscript $P$, e.g., \textsc{pts}$_{Z}$-\textsc{knn} and \textsc{pts}$_{P}$-\textsc{knn}.
Finally, we underline that for both \textsc{PivotTree} \textsc{c}lassifier and \textsc{PivotTree} \textsc{s}elector the interpretable predictive models $f$ is implemented through a \textsc{dt} and therefore the domain used is always the similarity space indicated with subscript $Z$, e.g., \textsc{ptc}$_{Z}$-\textsc{dt} and \textsc{pts}$_{Z}$-\textsc{dt}.

Regarding model selection, we performed stratified $5$-fold cross validation on the training set to select the best hyperparameter configuration through grid searches over the hyperparameter space, selecting the best-performing model.
For each best identified configuration, we performed model assessment on the test set.
For a sensitivity analysis of the proposed methods, examining how performance varies with different maximum numbers of pivots $k$, we refer the reader to~\cite{cascione2024pivottree}. 
Indeed, the results obtained for the various introduced variants are consistent with those already observed in~\cite{cascione2024pivottree}.

\smallskip
\textbf{Random Pivot Forest Settings.}
To evaluate the performance of \textsc{PivotTree} when integrated into an ensemble learning strategy, we follow the same model selection and evaluation strategy as described for the single estimator case, testing ensemble with each \textsc{PivotTree} variant.
We name our ensemble \textsc{RandomPivotForest} (\textsc{rpt}).
Such name is adapted according to the variant of \textsc{PivotTree} used in the ensemble, e.g., \textsc{rppt} if a proximity split is used, \textsc{ropt} if an oblique split is used, etc.
Furthermore, when experimenting with the balanced variants of \textsc{RandomPivotForest}, we consider two strategies\footnote{In \textsc{RandomPivotForest}, as we exclusively use \textsc{PivotTree}s as standalone classifiers as base estimators, we omit the subscript $Z$ for simplicity.}: 
\emph{(i)} using the stumps directly as base estimators, which we denote with the subscript \textsc{b}, e.g., \textsc{rpt}$_{\textsc{b}}$ for prediction;  
\emph{(ii)} combining the stumps with logistic regression, following the original splitting stump forest as described in~\cite{alkhouryW2024splitting}. 
In this case, we indicate this variant with the subscript \textsc{s}. 
For model selection, we explore different threshold values of $\rho$, selecting from the range $\{0.1, 0.15 \dots, 0.45\}$ in increments of $0.05$.
In our experiments, each \textsc{RandomPivotForest} comprises $100$ estimators, and the optimal $\mathit{maxdepth}$ for the base estimators is selected from $\{2,3,4\}$. 
During training, each estimator samples $\lfloor \sqrt{m} \rfloor$ features, where $m$ represents the total number of features in the original feature space. 
Additionally, in order to speed-up the training of the ensemble, we investigate a sampling strategy in which only $10\%$ of the training set is selected without replacement for each base estimator, and we compare its impact on performance against using the entire training set for each estimator.

\smallskip
\textbf{Competitors and Baselines.}
We compare \textsc{PivotTree} with the following baselines and state-of-the-art similarity-based approaches for pivot selection ($\pi$):
\begin{itemize}
    \item \textsc{kms}: runs kMeans~\cite{tan2006data} and adopts the centroids as pivots;
    \item \textsc{kmd}: runs kMedoids~\cite{tan2006data} and adopts the medoids as pivots;
    \item \textsc{ebl}: selects pivots according to the \textsc{$\varepsilon$-ball} algorithm\footnote{ \url{https://docs.seldon.io/projects/alibi/en/latest/methods/ProtoSelect.html}.}~\cite{bien2011epsilon}.
\end{itemize}
For \textsc{kms} and \textsc{kmd}, the number of pivots $k$ is selected within a grid on $k \in [2, 50]$, while for \textsc{ebl}, the grid search for $\varepsilon$ is performed on an interval between the $2^{nd}$ and the $52^{th}$ quantile of the empirical similarity distribution of the training set, as suggested in~\cite{bien2011epsilon}. 
Regarding the interpretable predictive models $f$, we use \textsc{knn} and \textsc{dt}, maintaining the same hyperparameter space $Z$ as in for \textsc{PivotTree} selectors.

As additional baselines that do not use pivots, we compare \textsc{PivotTree} against standard k-Nearest Neighbor (\textsc{knn}), decision trees (\textsc{dt}), and oblique decision trees (\textsc{odt}), all trained directly on the original feature space $X$. 
To distinguish these methods from our pivot-based approaches and other competitors using pivots, we denote them as \textsc{knn}$_X$, \textsc{dt}$_X$, and \textsc{odt}$_X$. 
For all models, we apply the same hyperparameter settings and model selection strategy as described for pivot-based approaches.  
Notably, for datasets consisting of \texttt{images} and \texttt{text}, \textsc{dt}$_X$ and \textsc{odt}$_X$ do not offer interpretability, as they rely on feature-based splits within an opaque latent embedding space. 
In contrast, \textsc{knn}, other competitors, and our approaches operate on individual cases, allowing end users to compare examples directly when making a final decision.
Similarly, as competing ensemble methods, we compare \textsc{RandomPivotForest} with traditional Random Forests trained on the original feature space, employing either univariate splits (\textsc{rdt}) or oblique splits (\textsc{rodt}). 
For the sake of completeness, we also include in our experimental evaluation Categorical Boosting (\textsc{catb})~\cite{prokhorenkova2018catboost}, Extreme Gradient Boosting (\textsc{xgb})~\cite{gohiya2018survey}, and Light Gradient Boosting Machine (\textsc{lgbm})~\cite{ke2017lgbm}.\footnote{Implementations are available at: \url{https://catboost.ai/docs/en/}, \url{https://xgboost.readthedocs.io/en/stable/}, and \url{https://lightgbm.readthedocs.io/en/stable/}. For further implementation details, we refer the reader to the repository of this paper.}
%} 
%%
All boosting models use univariate decision trees as weak learners and adopt the same values of $\mathit{maxdepth}$ and number of estimators explored for \textsc{RandomPivotForest}. For each method, we test learning rates of $0.01$, $0.1$, and $0.3$, and evaluate both full-sampling and $10\%$ subsampling strategies.
%}
%
In all cases, the comparison follows the same model-selection procedure and relies on identical hyperparameter configurations.

\smallskip
\textbf{Evaluation Measures.}
We evaluate the effectiveness of the selected pivots by measuring the weighted F1-score of the predictive models relying on the different sets of pivots\footnote{For all reported tables, we also provide results in terms of Balanced Accuracy in Appendix~\ref{sec:appendix}, which are consistent with those reported in terms of weighted F1-score.}. 
In line with the literature~\cite{bodria2023benchmarking}, we adopt model complexity, measured in terms of $k$, the number of selected pivots, as a proxy for interpretability. The underlying rationale is that a smaller number of representative instances leads to a more concise and cognitively manageable explanation of the model’s decision process~\cite{miller1956magical}.
This choice is consistent with prior work in exemplar and instance-selection approaches, where classifier or regressor performance is evaluated w.r.t. respect to subsets of selected instances,  analyzing the trade-off between predictive performance and the number of selected examples~\cite{bien2011epsilon,DBLP:journals/telo/FilhoLP22,kim2016examples}.
Note that $k$ can either be user-given, or optimized w.r.t. a given validation set.
We experiment in both settings.
As a further specification, although we cannot theoretically guarantee that the selected pivots will cover all classes, the greedy construction of the tree naturally promotes diversity, as exemplified in the qualitative examples discussed in Section~\ref{subsec:qual_results}. Instances that are discriminative at higher levels are unlikely to be selected again for subsequent splits, as the earlier splits have already partitioned those instances. This mechanism inherently encourages the tree to capture a wide variety of class-representative instances and justify the fact that we focus on the number of pivots extracted to estimate interpretability.

\begin{table}[t]
    \caption{Average weighted F1-score $\pm$ std. dev. for \textsc{PivotTree} classifiers and selectors combined with \textsc{dt} (limited to at most 20 pivots), baselines, and competitors. Subscripts indicate the average number of pivots $\pm$ std. dev.
    Best results in \textbf{bold}, second best in \textit{italics}.
    }
    \centering
    \setlength{\tabcolsep}{2.8mm}
    \label{ref:tab_dt_model_max20}
    \begin{tabular}{cccccc}
    \toprule
model & $\texttt{tabular}$ & $\texttt{images}$ & $\texttt{time-series}$ & $\texttt{text}$ & \texttt{all} \\
\midrule
$\textsc{grule}_{X}$ & $.65_{} \pm {.29}_{}$ & $.38_{} \pm {.42}_{}$ & $.40_{} \pm {.32}_{}$ & $.45_{} \pm {.32}_{}$ & $.51_{} \pm {.34}_{}$ \\
$\textsc{irep}_{X}$ & $.76_{} \pm .14_{}$ & $\textit{.90}_{} \pm .09_{}$ & $.55_{} \pm .27_{}$ & $.52_{} \pm {.25}_{}$ & $\textit{.71}_{} \pm {.23}_{}$ \\
$\textsc{ripper}_{X}$ & $.76_{} \pm .15_{}$ & $\textbf{.91}_{} \pm .08_{}$ & $.61_{} \pm .25_{}$ & $.52_{} \pm {.25}_{}$ & $\textbf{.72}_{} \pm .22_{}$ \\
\midrule
$\textsc{dt}_{X}$ & $\textbf{.81}_{} \pm .14_{}$ & $.70_{} \pm .19_{}$ & $\textbf{.64}_{} \pm .26_{}$ & $.58_{} \pm .22_{}$ & $\textbf{.72}_{} \pm .21_{}$ \\
$\textsc{odt}_{X}$ & $\textbf{.81}_{} \pm .13_{}$ & $.70_{} \pm .19_{}$ & $\textbf{.64}_{} \pm .27_{}$ & $.58_{} \pm .22_{}$ & $\textbf{.72}_{} \pm .21_{}$ \\
\midrule
$\textsc{kms}_{Z}$ & $.74_{14} \pm .16_{6}$ & $.70_{12} \pm .20_{6}$ & $.61_{12} \pm .26_{6}$ & $.54_{15} \pm .20_{4}$ & $.68_{13} \pm .21_{6}$ \\
$\textsc{kmd}_{Z}$ & $.76_{15} \pm .15_{5}$ & $.69_{12} \pm .19_{6}$ & $\textbf{.64}_{14} \pm .26_{6}$ & $.56_{17} \pm .20_{4}$ & $.69_{14} \pm .20_{6}$ \\
$\textsc{ebl}_{Z}$ & $.79_{18} \pm .15_{5}$ & $.70_{16} \pm .21_{6}$ & $.62_{16} \pm .27_{7}$ & $\textit{.59}_{20} \pm .22_{2}$ & $\textit{.71}_{18} \pm .21_{6}$ \\
\midrule
$\textsc{ptc}_{Z}$ & $.79_{\textbf{10}} \pm .14_{4}$ & $.69_{\textbf{7}} \pm .20_{4}$ & $\textit{.63}_{\textbf{9}} \pm .27_{4}$ & $\textbf{.60}_{\textbf{10}} \pm .21_{4}$ & $\textit{.71}_{\textbf{9}} \pm .21_{\textit{4}}$ \\
$\textsc{pptc}_{Z}$ & $\textit{.80}_{13} \pm .14_{6}$ & $.73_{10} \pm .17_{6}$ & $.62_{12} \pm .25_{6}$ & $.58_{\textit{11}} \pm .21_{4}$ & $\textit{.71}_{12} \pm .20_{6}$ \\
$\textsc{optc}_{Z}$ & $.76_{15} \pm .15_{7}$ & $.67_{\textit{9}} \pm .20_{7}$ & $.62_{11} \pm .27_{5}$ & $.53_{14} \pm .19_{5}$ & $.68_{13} \pm .20_{7}$ \\
$\textsc{opptc}_{Z}$ & $\textit{.80}_{12} \pm .13_{5}$ & $.73_{11} \pm .21_{6}$ & $.58_{\textit{10}} \pm {.28}_{6}$ & $.55_{13} \pm .20_{6}$ & $.70_{\textit{11}} \pm .22_{6}$ \\
\midrule
$\textsc{pts}_{Z}$ & $\textit{.80}_{\textit{11}} \pm .14_{5}$ & $.69_{\textit{9}} \pm .21_{5}$ & $.62_{11} \pm {.28}_{5}$ & $\textbf{.60}_{12} \pm .22_{5}$ & $\textit{.71}_{\textit{11}} \pm .21_{5}$ \\
$\textsc{ppts}_{Z}$ & $.78_{15} \pm .15_{6}$ & $.69_{11} \pm .21_{7}$ & $\textbf{.64}_{14} \pm {.28}_{7}$ & $\textit{.59}_{15} \pm .21_{8}$ & $.70_{14} \pm .21_{7}$ \\
$\textsc{opts}_{Z}$ & $.77_{14} \pm .14_{7}$ & $.67_{13} \pm .20_{5}$ & $.62_{12} \pm .27_{6}$ & $.54_{\textit{11}} \pm .21_{7}$ & $.68_{13} \pm .21_{6}$ \\
$\textsc{oppts}_{Z}$ & $.77_{14} \pm .16_{7}$ & $.66_{13} \pm {.23}_{7}$ & $.62_{12} \pm .26_{5}$ & $.54_{\textit{11}} \pm .20_{6}$ & $.68_{13} \pm .22_{6}$ \\
\bottomrule
\end{tabular}
\end{table}

\begin{table}[t]
    \caption{Average weighted F1-score $\pm$ std. dev. for \textsc{PivotTree} selectors combined with \textsc{knn} (limited to at most 20 pivots), baselines, and competitors. 
    Subscripts indicate the average number of pivots $\pm$ std. dev.
    Best results in \textbf{bold}, second best in \textit{italics}.
    }
    \centering
    \setlength{\tabcolsep}{3mm} %
    \label{ref:tab_knn_model_max20}
    \begin{tabular}{cccccc}
    \toprule
    model & $\texttt{tabular}$ & $\texttt{images}$ & $\texttt{time-series}$ & $\texttt{text}$ & \texttt{all} \\
\midrule
$\textsc{knn}_{{X}}$ & $\textbf{.83}_{} \pm .13_{}$ & $\textbf{.94}_{} \pm .05_{}$ & $\textit{.68}_{} \pm .27_{}$ & $\textbf{.61} \pm .22_{}$ & $\textbf{.79}_{} \pm .20_{}$ \\
\midrule
$\textsc{kms}_{{Z}}$ & $.80_{13} \pm .14_{6}$ & $\textit{.93}_{13} \pm .06_{5}$ & $\textit{.68}_{14} \pm .24_{6}$ & $.58_{15} \pm .22_{4}$ & $.77_{14} \pm .20_{5}$ \\
$\textsc{kmd}_{{Z}}$ & $.81_{15} \pm .14_{5}$ & $\textit{.93}_{14} \pm .06_{6}$ & $\textit{.68}_{\textbf{12}} \pm .24_{5}$ & $.60_{19} \pm .22_{3}$ & $\textit{.78}_{15} \pm .19_{5}$ \\
$\textsc{ebl}_{{Z}}$ & $\textit{.82}_{20} \pm .14_{3}$ & $\textit{.93}_{19} \pm .07_{4}$ & $\textit{.68}_{17} \pm .24_{6}$ & $\textbf{.62}_{20} \pm .23_{2}$ & $\textbf{.79}_{19} \pm .20_{4}$ \\
\midrule
$\textsc{pts}_{{Z}}$ & $\textit{.82}_{\textit{12}} \pm .14_{5}$ & $\textit{.93}_{15} \pm .05_{3}$ & $\textbf{.69}_{\textbf{12}} \pm .25_{7}$ & $\textbf{.62}_{\textit{13}} \pm .23_{4}$ & $\textbf{.79}_{\textit{13}} \pm .19_{5}$ \\
$\textsc{ppts}_{{Z}}$ & $\textit{.82}_{\textit{12}} \pm .14_{6}$ & $\textit{.93}_{16} \pm .05_{7}$ & $\textbf{.69}_{15} \pm .25_{5}$ & $\textit{.61}_{16} \pm .22_{7}$ & $\textit{.78}_{14} \pm .20_{6}$ \\
$\textsc{opts}_{{Z}}$ & $.81_{13} \pm .14_{5}$ & $\textit{.93}_{15} \pm .07_{7}$ & $.67_{\textit{13}} \pm .25_{5}$ & $.60_{17} \pm .22_{8}$ & $\textit{.78}_{14} \pm .20_{6}$ \\
$\textsc{oppts}_{{Z}}$ & $.81_{15} \pm .14_{5}$ & $.92_{14} \pm .07_{5}$ & $.67_{\textbf{12}} \pm .24_{7}$ & $.58_{14} \pm .22_{4}$ & $.77_{14} \pm .20_{5}$ \\
\midrule
$\textsc{kms}_{{P}}$ & $.72_{16} \pm {.20}_{5}$ & $.92_{16} \pm .07_{5}$ & $.58_{16} \pm {.28}_{5}$ & $.50_{17} \pm .20_{6}$ & $.70_{16} \pm .24_{5}$ \\
$\textsc{kmd}_{{P}}$ & $.71_{17} \pm {.20}_{4}$ & $.91_{\textit{11}} \pm .12_{6}$ & $.57_{16} \pm {.29}_{5}$ & $.53_{15} \pm .20_{5}$ & $.70_{15} \pm .24_{5}$ \\
$\textsc{ebl}_{{P}}$ & $.80_{20} \pm .14_{3}$ & $\textit{.93}_{17} \pm .06_{6}$ & $.61_{20} \pm {.29}_{1}$ & $.58_{19} \pm .22_{5}$ & $.76_{19} \pm .22_{\textit{4}}$ \\
\midrule
$\textsc{pts}_{{P}}$ & $.71_{\textit{12}} \pm .17_{4}$ & $.80_{\textbf{10}} \pm .22_{5}$ & $.56_{\textit{13}} \pm .26_{5}$ & $.54_{\textit{13}} \pm .21_{5}$ & $.67_{\textbf{12}} \pm .22_{5}$ \\
$\textsc{ppts}_{{P}}$ & $.72_{\textbf{10}} \pm {.19}_{7}$ & $.65_{14} \pm {.32}_{7}$ & $.57_{\textbf{12}} \pm {.28}_{6}$ & $.50_{\textit{13}} \pm .23_{9}$ & $.64_{\textbf{12}} \pm {.25}_{7}$ \\
$\textsc{opts}_{{P}}$ & $.73_{\textit{12}} \pm .18_{6}$ & $.73_{14} \pm .25_{6}$ & $.54_{\textit{13}} \pm {.28}_{5}$ & $.54_{15} \pm .23_{9}$ & $.66_{\textit{13}} \pm .24_{6}$ \\
$\textsc{oppts}_{{P}}$ & $.75_{\textit{12}} \pm .18_{9}$ & $.60_{12} \pm {.35}_{7}$ & $.55_{\textbf{12}} \pm {.29}_{7}$ & $.47_{\textbf{10}} \pm {.29}_{5}$ & $.64_{\textbf{12}} \pm {.27}_{7}$ \\
\bottomrule
\end{tabular}
\end{table}

\subsection{Quantitative Results}
In this section, we present a quantitative analysis of the results obtained from our experiments.

\smallskip
\textbf{Pivot Tree Classifier and Selector Evaluation.}
In Tables~\ref{ref:tab_dt_model_max20} and~\ref{ref:tab_knn_model_max20}, we report the overall performance of each model across all datasets, measured in terms of average weighted F1-score\footnote{For the sake of clarity, we do not include the standard deviation of the weighted F1-score, as it remains relatively stable across different models for each data type, typically ranging between 0.2 and 0.3.}. 
We also assess model complexity, expressed as the average number of pivots $\pm$ standard deviation.
To provide a more detailed analysis, we first examine performance separately for each data type, followed by an evaluation of overall average performance. 
In both cases, models are assessed in a \textit{constrained} setting, where the number of pivots used for classification is limited to a maximum of $20$ during model selection. 
This constraint is introduced to enhance interpretability, as excessive complexity would hinder this aspect.
Notably, all \textsc{PivotTree}-based models and baselines, except for \textsc{ebl}, naturally select approximately $20$ pivots even when no such constraint is imposed. 
For completeness, we report results for the unconstrained setting in the Appendix~\ref{sec:appendix}, along with detailed performance metrics for each model on individual datasets.

When a decision tree is adopted as interpretable predictive models $f$, as presented in Table~\ref{ref:tab_dt_model_max20}, \textsc{PivotTree} and its variants emerge as the most effective pivot selection methods in terms of both performance and complexity.  
Specifically, for \texttt{tabular} data, \textsc{ptc}$_Z$, \textsc{pts}$_Z$, and \textsc{opptc}$_Z$ rank as the second-best performing models, positioned just behind standard decision trees \textsc{dt}$_X$ and with comparable performance to \textsc{ripper}$_X$ and \textsc{irep}$_X$, which are particularly effective when applied on \texttt{images}.
These models outperform their competitors while requiring fewer pivots to achieve strong predictive performance.  
For \texttt{images}, \textsc{pptc}$_Z$ and \textsc{opptc}$_Z$ not only exceed the performance of the standard counterparts but also enhance interpretability due to their case-based reasoning structure.  
Regarding \texttt{time-series} data, \textsc{ppts}$_Z$ performs comparably to \textsc{kmd}, \textsc{dt}, and \textsc{odt}. 
In the case of \texttt{text} data, \textsc{ptc}$_Z$ and \textsc{pts}$_Z$, achieves the best performance, surpassing competitors while requiring fewer pivots.  
When considering performance across all data modalities, the \textsc{ptc}$_Z$ and \textsc{pts}$_Z$ variants consistently emerge as the second-best models, ranking immediately after standard decision trees \textsc{dt}$_X$.

As shown in Table~\ref{ref:tab_knn_model_max20}, similar observations can be made when \textsc{knn} is used as interpretable predictive models $f$. 
In general, when employing pivot selection methods, we observe that training \textsc{knn} on the entire training set using the similarity feature representation $Z$ yields better results than operating in the original feature space only considering the selected pivots $P$. 
For \texttt{tabular} data, \textsc{pts}$_{Z}$ and \textsc{ppts}$_{Z}$ perform on par with \textsc{ebl}$_{Z}$ as the second-best models, trailing the standard \textsc{knn} model by a single unit of difference while requiring fewer pivots for prediction. 
A similar trend is observed for \texttt{images}, although in this case, all models tend to achieve comparable performance when using similarity-based representations. Notably, \textsc{ebl}$_P$ achieves remarkable performance also in the original feature space.  
For \texttt{time-series} data, \textsc{pts}$_Z$ and \textsc{ppts}$_Z$ surpass standard \textsc{knn}$_X$, while for \texttt{text} data, they outperform \textsc{knn}$_X$ and achieve the same performance as \textsc{ebl}$_Z$, again requiring less pivots.  
When considering all data types, \textsc{pts}$_Z$ demonstrates performance comparable to \textsc{knn}$_X$ and \textsc{ebl}$_{Z}$ while utilizing fewer pivots than competing approaches and therefore being more interpretable.

\begin{figure}[t]
    \includegraphics[width=0.48\linewidth]{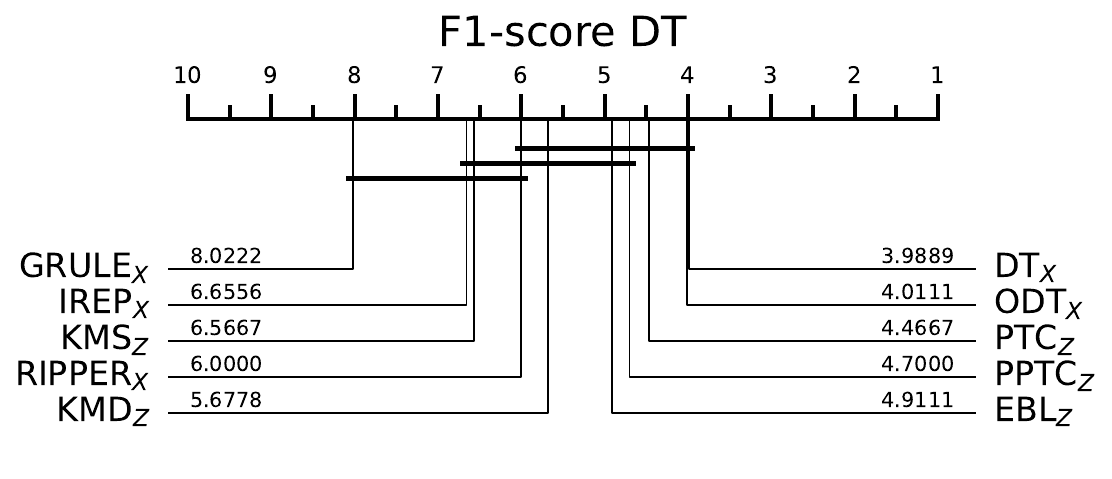}
    \includegraphics[width=0.48\linewidth]{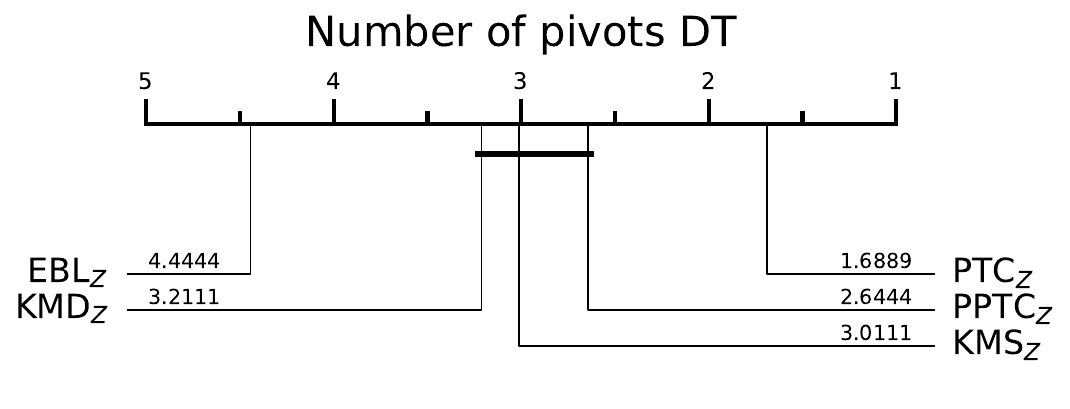}
    \includegraphics[width=0.5\linewidth]{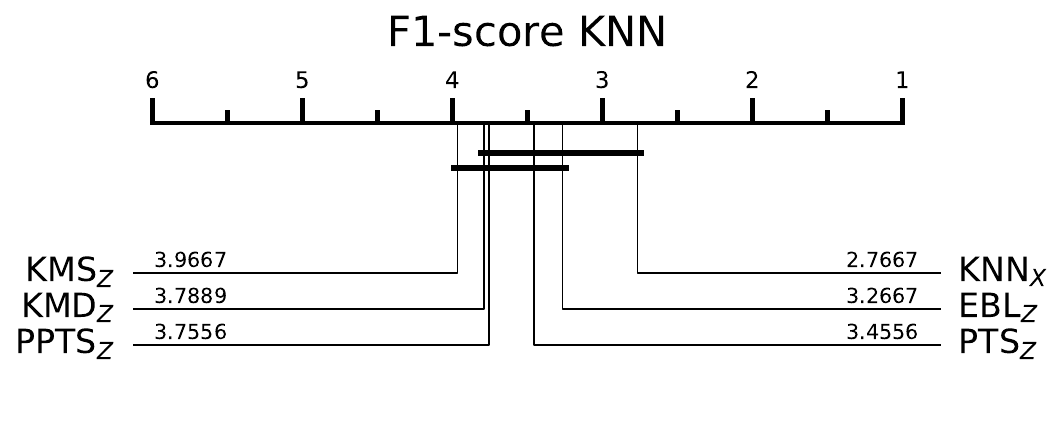}
    \includegraphics[width=0.5\linewidth]{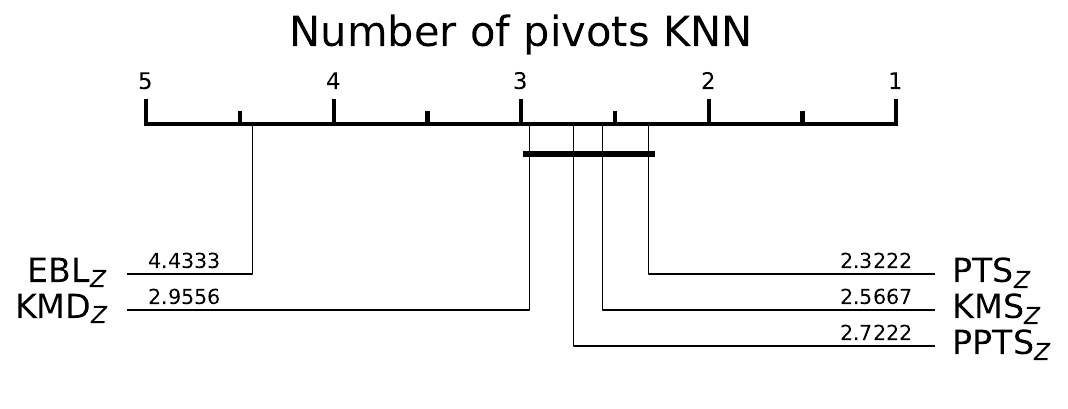}
    \caption{Critical difference plot of model's rank in terms of weighted F1-score and number of pivots against each other with Nemenyi test w.r.t all datasets, considering standard \textsc{PivotTree} variants. Models that are not significantly different at $95$\% significance level are connected. Best models on the right.}
    \label{fig:cd_plots}
\end{figure}

To further support our findings, we perform a statistical significance comparison among models across all datasets. 
We visualize the ranking of all methods relative to each other using the critical difference plots presented in Figure~\ref{fig:cd_plots}. 
In these plots, lower rank values indicate better-performing models, with the best ranks positioned on the right (refer to~\cite{demvsar2006statistical} for details). 
In Figure~\ref{fig:cd_plots}, methods that are statistically equivalent according to a post-hoc Nemenyi test are connected by black lines, indicating no significant difference in their performance.
To ensure a fair comparison without increasing the number of hypotheses (i.e., models considered), which would risk widening the critical difference threshold, we restrict the set of methods included in the statistical analysis.
For tree-based classifiers, we report only \textsc{ptc}$_{Z}$ and \textsc{pptc}$_{Z}$, corresponding to the standard standalone classification variants of \textsc{PivotTree}. The oblique versions are excluded, as they do not yield a meaningful performance improvement. Furthermore, employing these approaches either as standalone classifiers or as selectors within tree-based methods leads to comparable results, hence our decisions to only keep standalone classifiers in the comparison.
When comparing against \textsc{knn}, center-based clustering selection methods, and \textsc{ebl}, we focus exclusively on \textsc{pts}$_{Z}$ and \textsc{ppts}$_{Z}$, which represent the best-performing variants in this context. Variants trained w.r.t. selected instances in the original feature space are also excluded from the critical difference analysis, as they consistently demonstrate lower overall performance.\footnote{For completeness, Appendix~\ref{sec:appendix} reports the critical difference plots including all methods not directly presented in this section.}
We observe that for tree-based models (top), the univariate strategies of \textsc{pt} and \textsc{ppt}, rank among the top-performing models. 
These models show no statistically significant difference in performance compared to standard baselines and \textsc{ebl}, even surpassing the latter for both variants, and surpassing other rule-based baseline methods.
Moreover, in terms of complexity measured as number of pivots adopted, \textsc{PivotTree} variants rank among the least complex models while exhibiting statistically significant differences from competitors for \textsc{ptc}$_{Z}$.
For \textsc{knn} models (bottom), \textsc{pts}$_{Z}$ and \textsc{ppts}$_{Z}$ achieve strong rankings. In terms of performance, they are surpassed only by \textsc{knn} and \textsc{ebl}$_{Z}$, from which they remain statistically indistinguishable.
Additionally, \textsc{pts}$_{Z}$ outperforms all competitors in terms of low complexity, further reinforcing its interpretability.

Although time complexity is not the primary focus of this paper, we report for completeness the training runtime (in seconds) of \textsc{PivotTree} variants on representative datasets of varying sizes and dimensionalities. In Table~\ref{tab:rep_ds_training_times}, we report training time in seconds of variants of \textsc{ptc} in comparison with feature-based rule-based methods.

Although time complexity is not the primary focus of this paper, we report for completeness the training runtime (in seconds) of \textsc{PivotTree} variants on representative datasets of varying sizes and dimensionalities.

On small datasets such as \texttt{breast} and \texttt{worms}, \textsc{ptc}$_{Z}$ and \textsc{pptc}$_{Z}$ require $0.68s$ and $1.56s$ on \texttt{breast}, and $0.52s$ and $1.62s$ on \texttt{worms}, respectively. The oblique variants \textsc{ropt}$_{Z}$ and \textsc{roppt}$_{Z}$ take $3.02s$ and $4.48s$ on \texttt{breast}, and $1.03s$ and $3.98s$ on \texttt{worms}. 
On the other hand, the standard univariate \textsc{dt} exhibits very low training times, requiring $0.05s$ and $0.07s$ on the \texttt{breast} and \texttt{worms} datasets, respectively. 
In comparison, \textsc{grule}, \textsc{ripper}, and \textsc{irep} require $0.09s$, $0.74s$, and $0.59s$ on \texttt{breast}, and $0.33s$, $424s$, and $421s$ on \texttt{worms}, respectively. 
The significantly longer training times observed on \texttt{worms} are mainly due to the high dimensionality of the dataset.

For higher-dimensional datasets such as \texttt{cifar10} and \texttt{tgpt}, training times increase substantially. On \texttt{cifar10}, \textsc{ptc}$_{Z}$ and \textsc{pptc}$_{Z}$ require $38.48s$ and $176.38s$, while \textsc{ropt}$_{Z}$ and \textsc{roppt}$_{Z}$ take $93.33s$ and $400.38s$. On \texttt{tgpt}, the respective times are $14.14s$ and $17.67s$ for \textsc{ptc}$_{Z}$ and \textsc{pptc}$_{Z}$, and $24.16s$ and $31.99s$ for \textsc{ropt}$_{Z}$ and \textsc{roppt}$_{Z}$.
On \texttt{cifar10}, the standard univariate \textsc{dt} and \textsc{grule}$_{Z}$ exhibit fast training times ($1.64s$ and $1.40s$, respectively), whereas rule-based methods such as \textsc{ripper}$_{Z}$ and \textsc{irep}$_{Z}$ require substantially longer training times ($1468s$ and $1403s$). 
A similar trend is observed on \texttt{tgpt}, where \textsc{dt}$_{Z}$ and \textsc{grule}$_{Z}$ train in $25.22s$ and $4.61s$, while \textsc{ripper}$_{Z}$ and \textsc{irep}$_{Z}$ require $288s$ and $531s$, respectively.

\begin{table}[t]
\centering
\setlength{\tabcolsep}{4.5mm} %
\caption{Training time (in seconds) across representative datasets of varying size and dimensionality. 
For each dataset, best result in \textbf{bold}, second best in \textit{italics}.}
\label{tab:rep_ds_training_times}
\begin{tabular}{lrrrrrr}
\toprule
model & \texttt{breast} & \texttt{worms} & \texttt{cifar10} & \texttt{tgpt} & \texttt{magic} & \texttt{svhn} \\
\midrule
\textsc{ptc}$_Z$    
& 0.68 & 0.52 & 38.48 & \textit{14.14} & 71.86 & 64.38 \\

\textsc{pptc}$_Z$   
& 1.56 & 1.62 & 176.38 & {17.67} & 70.10 & 294.74 \\

\textsc{optc}$_Z$   
& 3.02 & 1.03 & 93.33 & 24.16 & 214.30 & 143.36 \\

\textsc{opptc}$_Z$  
& 4.48 & 3.98 & 400.38 & 31.99 & 177.48 & 515.77 \\

\midrule
\textsc{dt}$_Z$     
& \textbf{0.05} & \textit{0.07} & \textit{1.64} & 25.22 & \textit{2.84} & \textit{41.44} \\

\textsc{grule}$_Z$  
& \textit{0.09} & \textbf{0.33} & \textbf{1.40} & \textbf{4.61} & \textbf{0.05} & \textbf{0.62} \\

\textsc{ripper}$_Z$ 
& 0.74 & 424.00 & 1468.00 & 288.00 & 8.88 & 993.00 \\

\textsc{irep}$_Z$   
& 0.59 & 421.00 & 1403.00 & 531.00 & 1.26 & 1614.00 \\
\bottomrule
\end{tabular}
\end{table}

Analogous results can be observed for larger datasets. On \texttt{magic}, \textsc{ptc}$_{Z}$ and \textsc{pptc}$_{Z}$ require $71.86s$ and $70.10s$, while the oblique variants \textsc{ropt}$_{Z}$ and \textsc{roppt}$_{Z}$ take $214.30s$ and $177.48s$, respectively. On the large-scale \texttt{svhn} dataset, \textsc{ptc}$_{Z}$ and \textsc{pptc}$_{Z}$ require $64.38s$ and $294.74s$, whereas \textsc{ropt}$_{Z}$ and \textsc{roppt}$_{Z}$ need $143.36s$ and $515.77s$, respectively.
On \texttt{magic}, \textsc{grule}$_{Z}$ achieves the shortest training time ($0.05s$), followed by \textsc{dt}$_{Z}$ ($2.84s$), whereas \textsc{ripper}$_{Z}$ and \textsc{irep}$_{Z}$ require $8.88s$ and $1.26s$. 
A markedly different behavior is observed on \texttt{svhn}$_{Z}$, where \textsc{dt}$_{Z}$ and \textsc{grule}$_{Z}$ takes respectively $41.44s$ and $0.62s$, but \textsc{ripper}$_{Z}$ and \textsc{irep}$_{Z}$ scale poorly, requiring $993s$ and $1614s$, respectively.
On the other hand, prediction times remain relatively fast across all methods. 
For example, on \texttt{breast}, \textsc{ptc}$_{Z}$ requires only $0.03s$ for inference, while \textsc{ropt}$_{Z}$ takes $0.129s$. 
These values are comparable to those of \textsc{irep} and \textsc{ripper}, which require $0.06s$ and $0.07s$, respectively, and are only moderately higher than that of \textsc{dt}$_{Z}$, which requires just $0.002s$ for prediction.
On \texttt{svhn}, prediction times range from a minimum of $4.51s$ for \textsc{ptc}$_{Z}$ to a maximum of $7.41s$ for \textsc{roppt}$_{Z}$, while \textsc{dt}$_{Z}$ requires $0.06s$, but \textsc{irep}$_{Z}$ and \textsc{ripper}$_{Z}$ much more with $2291s$ and $4282s$.

As a final remark, if we were to suggest a \textsc{PivotTree} variant, our choice would likely be \textsc{ppts}$_{Z}$. 
While its performance is comparable to that of \textsc{pts}$_{Z}$ and \textsc{ptc}$_{Z}$ when combined with \textsc{knn}, it offers a higher level of interpretability when paired with \textsc{dt}. 
This is because it enables decision-making through a hierarchical process in which a test instance is compared against two pivots, guiding the traversal along two distinct directions.
Such hierarchical splits are particularly advantageous for modalities like \texttt{images} or \texttt{text}, where relative similarity to exemplar instances is often more intuitive than absolute distance thresholds. 
Conversely, in settings where a precise global distance threshold is desirable and where the original feature space has transparent semantics, variants such as \textsc{pts}$_{Z}$ may be preferable, since examining the contribution of individual feature components and comparing them to an absolute distance threshold could offer additional insight into why instances are considered similar. 
More generally, the choice of the downstream classifier also depends on whether the task prioritizes local or global explanations: pairing \textsc{PivotTree} with \textsc{knn} emphasizes local neighbourhood reasoning, whereas pairing it with \textsc{dt}s emphasizes a global, structured, and fully interpretable decision process.

\begin{table}[t]
    \caption{Average weighted F1-score $\pm$ std. dev. for \textsc{RandomPivotForest} with $100$ \textsc{PivotTree} \textsc{classifiers} as estimators, and no sampling, against baselines. 
    When the splitting stump forest is adopted, subscripts indicate the average number of stumps $\pm$ std. dev.
    Best results in \textbf{bold}, second best in \textit{italics}.}
    \label{tab:ensemble_full}
    \setlength{\tabcolsep}{1.4mm}
    \begin{tabular}{cccccc}
    \toprule

{model} & $\texttt{tabular}$ & $\texttt{images}$ & $\texttt{time-series}$ & $\texttt{text}$ & \texttt{all} \\
\midrule
$\textsc{xgb}$ & $\textbf{.87}_{\textbf{}} \pm .12_{}$ & $\textit{.94}_{\textbf{}} \pm .04_{}$ & $\textbf{.72}_{\textbf{}} \pm {.27}_{}$ & $\textbf{.70}_{\textbf{}} \pm .25_{}$ & $\textbf{.83}_{} \pm .19_{}$ \\
$\textsc{lgbm}$ & $\textbf{.87}_{\textbf{}} \pm .12_{}$ & $\textbf{.95}_{\textbf{}} \pm .04_{}$ & $\textit{.69}_{\textbf{}} \pm .26_{}$ & $\textbf{.70}_{\textbf{}} \pm .25_{}$ & $\textit{.82}_{} \pm .19_{}$ \\
$\textsc{catb}$ & $\textbf{.87}_{\textbf{}} \pm .12_{}$ & $\textit{.94}_{\textbf{}} \pm .05_{}$ & $\textbf{.72}_{\textbf{}} \pm .26_{}$ & $\textit{.69}_{\textbf{}} \pm .25_{}$ & $\textbf{.83}_{} \pm .19_{}$ \\
\midrule
$\textsc{rdt}$ & $\textit{.80}_{\textbf{}} \pm .16_{}$ & $\textit{.94}_{\textbf{}} \pm .04_{}$ & $.68_{\textbf{}} \pm {.28}_{}$ & $.55_{\textbf{}} \pm .25_{}$ & $.77_{} \pm .22_{}$ \\
$\textsc{rodt}$ & $\textit{.80}_{\textbf{}} \pm .16_{}$ & $\textit{.94}_{\textbf{}} \pm .05_{}$ & $.67_{\textbf{}} \pm {.28}_{}$ & $.55_{\textbf{}} \pm .25_{}$ & $.76_{} \pm .22_{}$ \\
\midrule
$\textsc{rpt}$ & $\textit{.80}_{\textbf{}} \pm .16_{}$ & $.76_{\textbf{}} \pm .16_{}$ & $.68_{\textbf{}} \pm {.28}_{}$ & $.58_{\textbf{}} \pm .25_{}$ & $.74_{} \pm .21_{}$ \\
$\textsc{ropt}$ & $.78_{\textbf{}} \pm .16_{}$ & $.77_{\textbf{}} \pm .15_{}$ & $.68_{\textbf{}} \pm {.28}_{}$ & $.52_{\textbf{}} \pm .25_{}$ & $.72_{} \pm .22_{}$ \\
$\textsc{rppt}$ & $.78_{\textbf{}} \pm .17_{}$ & $.93_{\textbf{}} \pm .04_{}$ & $.66_{\textbf{}} \pm {.27}_{}$ & $.54_{\textbf{}} \pm {.26}_{}$ & $.75_{} \pm .22_{}$ \\
$\textsc{roppt}$ & $.78_{\textbf{}} \pm .16_{}$ & $.93_{\textbf{}} \pm .03_{}$ & $.66_{\textbf{}} \pm {.27}_{}$ & $.53_{\textbf{}} \pm .25_{}$ & $.75_{} \pm .22_{}$ \\
\midrule
$\textsc{rdt}_{\textsc{s}}$ & $.70_{72} \pm .17_{84}$ & $.91_{282} \pm .10_{375}$ & $.62_{206} \pm .24_{205}$ & $.57_{185} \pm .20_{97}$ & $.71_{159} \pm .21_{213}$ \\
$\textsc{rodt}_{\textsc{s}}$ & $.72_{90} \pm .16_{74}$ & $.92_{280} \pm .09_{330}$ & $.63_{207} \pm .23_{212}$ & $.56_{200} \pm .19_{101}$ & $.72_{169} \pm .20_{196}$ \\
\midrule
$\textsc{rpt}_{\textsc{s}}$ & $.71_{80} \pm .16_{61}$ & $.81_{189} \pm .27_{127}$ & $.63_{163} \pm .22_{236}$ & $.56_{175} \pm .20_{103}$ & $.69_{133} \pm .21_{139}$ \\
$\textsc{ropt}_{\textsc{s}}$ & $.71_{91} \pm .16_{74}$ & $.74_{122} \pm .32_{123}$ & $.62_{165} \pm .24_{233}$ & $.57_{196} \pm .20_{95}$ & $.68_{128} \pm .22_{137}$ \\
$\textsc{rppt}_{\textsc{s}}$ & $.71_{149} \pm .16_{{120}}$ & $.93_{168} \pm .06_{159}$ & $.64_{277} \pm .22_{{290}}$ & $.57_{147} \pm .21_{54}$ & $.72_{181} \pm .20_{176}$ \\
$\textsc{roppt}_{\textsc{s}}$ & $.71_{100} \pm .17_{91}$ & $.92_{175} \pm .09_{186}$ & $.64_{268} \pm .22_{{253}}$ & $.56_{149} \pm .20_{83}$ & $.72_{159} \pm .20_{168}$ \\
\midrule
$\textsc{rdt}_{\textsc{b}}$ & $.59_{\textbf{53}} \pm {.22}_{60}$ & $.34_{368} \pm {.36}_{455}$ & $.51_{136} \pm .21_{116}$ & $.38_{98} \pm .20_{87}$ & $.50_{140} \pm {.26}_{239}$ \\
$\textsc{rodt}_{\textsc{b}}$ & $.60_{102} \pm {.22}_{{113}}$ & $.35_{190} \pm .34_{{404}}$ & $.47_{101} \pm .19_{82}$ & $.40_{108} \pm .24_{89}$ & $.49_{120} \pm {.26}_{197}$ \\
\midrule
$\textsc{rpt}_{\textsc{b}}$ & $.63_{\textit{56}} \pm {.21}_{41}$ & $.37_{134} \pm .35_{130}$ & $.49_{82} \pm .23_{107}$ & $.38_{157} \pm .20_{{143}}$ & $.52_{91} \pm {.26}_{100}$ \\
$\textsc{ropt}_{\textsc{b}}$ & $.61_{65} \pm .19_{51}$ & $.37_{112} \pm .34_{124}$ & $.48_{\textbf{63}} \pm .23_{55}$ & $.36_{97} \pm .18_{93}$ & $.50_{\textit{78}} \pm .25_{{77}}$ \\
$\textsc{rppt}_{\textsc{b}}$ & $.59_{63} \pm {.21}_{45}$ & $.36_{89} \pm {.39}_{134}$ & $.50_{111} \pm .22_{139}$ & $.41_{\textbf{46}} \pm {.27}_{31}$ & $.50_{\textbf{76}} \pm {.27}_{{93}}$ \\
$\textsc{roppt}_{\textsc{b}}$ & $.61_{62} \pm .19_{56}$ & $.33_{\textbf{59}} \pm {.36}_{60}$ & $.46_{\textit{90}} \pm .21_{87}$ & $.33_{144} \pm .17_{{203}}$ & $.48_{79} \pm {.26}_{95}$ \\
\bottomrule
\end{tabular}
\end{table}

\smallskip
\textbf{Random Pivot Forest Evaluation.}
Table~\ref{tab:ensemble_full} compares the performance of ensemble methods using the entire training set for each estimator. 
We recall to the reader that in \textsc{RandomPivotForest} we exclusively use \textsc{PivotTree}s \textsc{c}lassifiers as base estimators, i.e., \textsc{ptc}, \textsc{pptc}, \textsc{optc}, \textsc{opptc} for the different types of forests which are named accordingly as \textsc{rpt}, \textsc{rppt}, \textsc{ropt}, \textsc{roppt}, with the \textsc{c} omitted as the \textsc{s}elector option is not considered with ensemble.
We also remind the reader that we use the subscripts \textsc{s} and \textsc{b} to indicate different approaches for utilizing the splitting stump forest to reduce forest complexity. 
Specifically, \textsc{s} denotes the use of logistic regression for classification, as proposed in~\cite{alkhouryW2024splitting}, while \textsc{b} refers to classification based on majority voting.
When the splitting stump forest is not used, for \texttt{tabular} dataset, \textsc{rpt} achieves the same performance as the standard Random Forest (\textsc{rdt}) and oblique Random Forest (\textsc{rodt}). 
Overall, boosting-based approaches outperform both feature-based and case-based random forest models in our experimental evaluation.
Notably, for the \texttt{images} modality, the proximity-based variant \textsc{rppt} substantially outperforms its univariate counterpart, achieving performance levels comparable to those of \textsc{rdt} and \textsc{rodt}, which in turn approach the performance attained by the boosting-based methods.

Moreover, \textsc{RandomPivotForest}s retain a degree of interpretability, as transparency within the ensemble can, in principle, be recovered through feature importance analysis, capturing the contribution of training instances to predictions, or by employing methods that extract standalone explanations from tree ensembles~\cite{bonsignori2021deriving,pensa2025explaining}. A comprehensive investigation of these aspects is left for future work. Nevertheless, this potential provides a level of interpretability that is not available in standard models based on CNN-derived feature representations.
For the \texttt{time-series} datasets, the best performance is achieved by boosting-based approaches. Nonetheless, \textsc{rpt} and \textsc{ropt} obtain results comparable to the standard \textsc{rdt}, highlighting their competitiveness within the class of randomized tree-based methods, even though they do not outperform the boosting models.
For the \texttt{text} datasets, \textsc{rpt} achieves the best performance among the randomized tree-based methods, outperforming the standard \textsc{rdt} and \textsc{rodt}. However, its performance is still surpassed by the boosting-based approaches.

Finally, when considering all data types together, comparable results are obtained across all random forests.
We also highlight that, consistently with~\cite{Setzu2023tree}, the theoretically enhanced expressiveness of multivariate splits compared to univariate splits is not empirically confirmed. 
This holds true both for individual trees and for forests of trees, with univariate trees consistently outperforming their oblique counterparts.
When the splitting stump forest is employed, we observe a notable decrease in performance for the \texttt{tabular} and \texttt{time-series} datasets. In contrast, for \texttt{images} and \texttt{text}, the performance remains comparatively stable, indicating that the alternative representation is still effective for these data types.
We notice that in this setting, proximity-based models tend to outperform other \textsc{PivotTree} variants.  
Additionally, we find that relying solely on the resulting stumps as predictive ensembles, i.e., methods denoted with the subscript \textsc{b}, leads to generally unsatisfactory results across all approaches. 
This reinforces the advantage of using logistic regression, i.e., subscript \textsc{s}, for prediction over simple majority voting.
Finally, given the relatively high number of stumps required to achieve competitive performance, we conclude that, at least within the \textsc{RandomPivotForest} framework, the intended complexity reduction is not effectively realized for many approaches and data types. 
This suggests the need for further investigation in this direction.

\begin{table}[t]
    \caption{Average weighted F1-score $\pm$ std. dev. for \textsc{RandomPivotForest} with $100$ \textsc{PivotTree} \textsc{classifiers} as estimators, and 10\% of the training set is used to train each estimator, against baselines. 
    When the splitting stump forest is adopted, subscripts indicate the average number of stumps $\pm$ std. dev.
    Best results in \textbf{bold}, second best in \textit{italics}.}
    \label{tab:ensemble_10_perc}
    \setlength{\tabcolsep}{1.4mm}
    \begin{tabular}{cccccc}
\toprule
{model} & $\texttt{tabular}$ & $\texttt{images}$ & $\texttt{time-series}$ & $\texttt{text}$ & \texttt{all} \\
\midrule
$\textsc{xgb}$ & $\textit{.86}_{} \pm .12_{}$ & $\textit{.94}_{} \pm .04_{}$ & $.66_{} \pm .25_{}$ & $\textit{.68}_{} \pm .25_{}$ & $.81_{} \pm .19_{}$ \\
$\textsc{lgbm}$ & $\textbf{.87}_{} \pm .12_{}$ & $\textbf{.95}_{} \pm .04_{}$ & $\textit{.69}_{} \pm .26_{}$ & $\textbf{.70}_{} \pm .25_{}$ & $\textit{.82}_{} \pm .19_{}$ \\
$\textsc{catb}$ & $\textit{.86}_{} \pm .12_{}$ & $\textbf{.95}_{} \pm .04_{}$ & $\textbf{.72}_{} \pm .27_{}$ & $\textit{.68}_{} \pm .25_{}$ & $\textbf{.83}_{} \pm .19_{}$ \\
\midrule
$\textsc{rdt}$ & $.78_{} \pm .17_{}$ & $.93_{} \pm .04_{}$ & $.61_{} \pm .26_{}$ & $.51_{} \pm .27_{}$ & $.74_{} \pm .23_{}$ \\
$\textsc{rodt}$ & $.77_{} \pm .17_{}$ & $.93_{} \pm .05_{}$ & $.62_{} \pm .25_{}$ & $.51_{} \pm .27_{}$ & $.74_{} \pm .23_{}$ \\
\midrule
$\textsc{rpt}$ & $.78_{} \pm .16_{}$ & $.86_{} \pm .10_{}$ & $.64_{} \pm .26_{}$ & $.57_{} \pm .24_{}$ & $.73_{} \pm .21_{}$ \\
$\textsc{ropt}$ & $.76_{} \pm .17_{}$ & $.86_{} \pm .11_{}$ & $.64_{} \pm .26_{}$ & $.52_{} \pm .24_{}$ & $.72_{} \pm .21_{}$ \\
$\textsc{rppt}$ & $.77_{} \pm .16_{}$ & $.93_{} \pm .05_{}$ & $.63_{} \pm .26_{}$ & $.53_{} \pm .25_{}$ & $.74_{} \pm .22_{}$ \\
$\textsc{roppt}$ & $.77_{} \pm .16_{}$ & $.93_{} \pm .05_{}$ & $.63_{} \pm .26_{}$ & $.52_{} \pm .25_{}$ & $.74_{} \pm .22_{}$ \\
\midrule
$\textsc{rdt}_{\textsc{s}}$ & $.70_{74} \pm .16_{72}$ & $.92_{236} \pm .09_{320}$ & $.63_{191} \pm .23_{190}$ & $.56_{150} \pm .19_{63}$ & $.71_{142} \pm .20_{182}$ \\
$\textsc{rodt}_{\textsc{s}}$ & $.71_{89} \pm .16_{59}$ & $.92_{199} \pm .09_{185}$ & $.63_{189} \pm .22_{194}$ & $.56_{197} \pm .19_{109}$ & $.71_{148} \pm .20_{140}$ \\
\midrule
$\textsc{rpt}_{\textsc{s}}$ & $.70_{86} \pm .16_{44}$ & $.65_{69} \pm .22_{51}$ & $.62_{112} \pm .23_{82}$ & $.56_{158} \pm .20_{72}$ & $.66_{98} \pm .19_{64}$ \\
$\textsc{ropt}_{\textsc{s}}$ & $.71_{75} \pm .16_{50}$ & $.68_{\textit{49}} \pm .25_{46}$ & $.62_{107} \pm .23_{84}$ & $.56_{{144}} \pm .19_{69}$ & $.66_{{86}} \pm .20_{66}$ \\
$\textsc{rppt}_{\textsc{s}}$ & $.71_{127} \pm .17_{99}$ & $.92_{138} \pm .07_{116}$ & $.63_{224} \pm .23_{208}$ & $.57_{169} \pm .20_{80}$ & $.72_{157} \pm .20_{133}$ \\
$\textsc{roppt}_{\textsc{s}}$ & $.71_{106} \pm .16_{99}$ & $.90_{145} \pm .09_{112}$ & $.64_{193} \pm .22_{149}$ & $.57_{162} \pm .20_{88}$ & $.71_{141} \pm .20_{115}$ \\
\midrule
$\textsc{rdt}_{\textsc{b}}$ & $.61_{82} \pm .21_{73}$ & $.28_{141} \pm .28_{152}$ & $.52_{\textit{61}} \pm .24_{68}$ & $.38_{98} \pm .22_{59}$ & $.49_{91} \pm .26_{93}$ \\
$\textsc{rodt}_{\textsc{b}}$ & $.59_{{71}} \pm .21_{53}$ & $.22_{200} \pm .26_{191}$ & $.51_{76} \pm .23_{82}$ & $.37_{86} \pm .18_{65}$ & $.47_{100} \pm .26_{111}$ \\
\midrule
$\textsc{rpt}_{\textsc{b}}$ & $.63_{72} \pm .21_{47}$ & $.36_{54} \pm .36_{69}$ & $.48_{68} \pm .21_{41}$ & $.36_{89} \pm .20_{107}$ & $.51_{\textit{70}} \pm .27_{60}$ \\
$\textsc{ropt}_{\textsc{b}}$ & $.62_{\textbf{69}} \pm .20_{47}$ & $.33_{\textbf{48}} \pm .33_{70}$ & $.50_{\textbf{59}} \pm .22_{54}$ & $.36_{\textbf{41}} \pm .18_{46}$ & $.50_{\textbf{59}} \pm .26_{53}$ \\
$\textsc{rppt}_{\textsc{b}}$ & $.58_{73} \pm .22_{51}$ & $.34_{92} \pm .36_{72}$ & $.47_{81} \pm .23_{49}$ & $.35_{47} \pm .15_{44}$ & $.48_{{75}} \pm .26_{55}$ \\
$\textsc{roppt}_{\textsc{b}}$ & $.60_{\textit{70}} \pm .21_{45}$ & $.33_{124} \pm .32_{157}$ & $.47_{103} \pm .25_{100}$ & $.38_{\textbf{41}} \pm .18_{18}$ & $.49_{84} \pm .26_{91}$ \\
\bottomrule
\end{tabular}
\end{table}

Table~\ref{tab:ensemble_10_perc} reports the same methods as those in Table~\ref{tab:ensemble_full}, with the only difference being that only $10\%$ of the training set is used to train each estimator. 
This study aims to evaluate whether a significant reduction in the training set size, which can speed up the training of the individual \textsc{PivotTree}s, still guarantees acceptable performance.
By comparing the two tables, we observe that most approaches tend to slightly suffer from the reduction in the training set size of the base estimator. 
Boosting-based approaches, including \textsc{catb}, \textsc{xgb}, and \textsc{lgbm}, exhibit consistently strong and stable performance across the evaluated datasets even when sampling is introduced, confirming their effectiveness.
For \texttt{tabular} data, there is a very small reduction in performance, with the \textsc{RandomPivotForest} approaches and the \textsc{RandomPivotForest} with splitting stump forests achieving almost the same performance. 
In contrast, for \texttt{images}, we notice an opposite trend, with performance improving under the reduced sampling context for \textsc{rpt} and \textsc{ropt}, though there remains a significant gap when compared to the best model variants, which tend to achieve the same results w.r.t the ones obtained using the full training set. 
On the other hand, using a splitting stump strategy degrades their performance, in contrast to the full dataset scenario.
Notably, for \texttt{time-series} data, \textsc{rpt} and \textsc{ropt} emerge as the best models in this scenario, outperforming \textsc{rdt} and \textsc{rodt} in the standard ensemble context, while at the same time matching or surpassing their performance in the splitting stump context. 
For \texttt{text}, \textsc{rpt} maintains analogous performance to the no sampling scenario, with only a slight degradation compared to the full dataset case, both in the standard ensemble setup and when using splitting stumps. 
Finally, we notice that when generally evaluated on all data types, in the sampling context \textsc{rppt} and \textsc{roppt} reach the same average performance of \textsc{rdt}, matching its performance also in the splitting stump scenario.

\begin{table}[t]
    \centering
    \setlength{\tabcolsep}{1.2mm}
    \caption{Training time (seconds) using the full dataset (100\%) and a sampled subset (10\%). 
    Best results per dataset and sample subset size are in \textbf{bold}, second best in \textit{italics}. 
    All models use base estimators with $\mathit{maxdepth}=4$.}
    \label{tab:rpf_training_combined}
    \scriptsize
    \begin{tabular}{lcc|cc|cc|cc|cc|cc}
    \toprule
    & \multicolumn{2}{c}{\texttt{breast}}
    & \multicolumn{2}{c}{\texttt{worms}}
    & \multicolumn{2}{c}{\texttt{cifar10}}
    & \multicolumn{2}{c}{\texttt{tgpt}}
    & \multicolumn{2}{c}{\texttt{magic}}
    & \multicolumn{2}{c}{\texttt{svhn}} \\
    \cmidrule(lr){2-3}\cmidrule(lr){4-5}\cmidrule(lr){6-7}
    \cmidrule(lr){8-9}\cmidrule(lr){10-11}\cmidrule(lr){12-13}
    model & 100\% & 10\% & 100\% & 10\% & 100\% & 10\% & 100\% & 10\% & 100\% & 10\% & 100\% & 10\% \\
    \midrule
    \textsc{rpt}   
    & \textit{0.49} & \textit{0.13}
    & \textit{0.47} & \textit{0.12}
    & \textit{2.76} & \textit{0.39}
    & 3.51 & \textit{1.18}
    & 2.14 & 1.74
    & 18.63 & 0.69 \\
    
    \textsc{rppt}  
    & 1.13 & 0.32
    & 1.81 & 0.51
    & 4.21 & 1.60
    & \textit{2.02} & 1.65
    & 3.75 & 1.65
    & \textit{17.09} & \textit{0.68} \\
    
    \midrule
    \textsc{rdt}   
    & 0.65 & \textbf{0.08}
    & \textbf{0.26} & \textbf{0.09}
    & \textbf{1.09} & \textbf{0.23}
    & \textbf{1.08} & \textbf{0.40}
    & \textit{1.34} & \textbf{0.64}
    & \textbf{0.92} & \textbf{0.32} \\
    
    \textsc{lgbm}  
    & \textbf{0.30} & 0.30
    & 0.92 & 0.92
    & 45.37 & 41.97
    & 7.21 & 7.16
    & \textbf{0.89} & \textit{0.87}
    & 55.19 & 50.12 \\
    \bottomrule
    \end{tabular}
\end{table}

In Table~\ref{tab:rpf_training_combined}, we report the training times (in seconds) of \textsc{RandomPivotForest} variants on representative datasets of varying size and dimensionality. 
In particular, we compare training times obtained using the full training dataset ($100\%$) with those obtained from a $10\%$ sampled subset. 
We report the training times (in seconds) of \textsc{RandomPivotForest} variants on example datasets of varying sizes and dimensionalities. 
All results shown concern base estimators with $\mathit{maxdepth} = 4$, and we compare training times using the full dataset versus $10\%$ sampled subsets. 
We focus on univariate variants of \textsc{rpt} and \textsc{rppt}, as oblique ones did not show relevant deviations.

On small-sized datasets such as \texttt{breast} and \texttt{worms}, training times for \textsc{rpt} and \textsc{rppt} using the full training set are $0.49s$ and $1.13s$ on \texttt{breast}, and $0.47s$ and $1.81s$ on \texttt{worms}, respectively. 
These training times are comparable to those of our standard \textsc{rdt} implementation, which requires $0.65s$ on \texttt{breast} and $0.26s$ on \texttt{worms}, and are also in line with \textsc{lgbm}, which requires $0.30s$ on \texttt{breast} and $0.92s$ on \texttt{worms}.
When using only $10\%$ of the training data, training times for \textsc{rpt} and \textsc{rppt} further decrease to $0.13s$ and $0.32s$ on \texttt{breast}, and to $0.12s$ and $0.51s$ on \texttt{worms}, respectively. 
Under the same setting, \textsc{rdt} requires $0.08s$ on \texttt{breast} and $0.09s$ on \texttt{worms}, while \textsc{lgbm} requires $0.30s$ and $0.92s$ on \texttt{breast} and \texttt{worms}, respectively.

A similar trend is observed on medium-sized datasets. 
For \texttt{cifar10}, full-data training times are ${2.76s}$ for \textsc{rpt} and ${4.21s}$ for \textsc{rppt}, whereas sampled training reduces these to ${0.39s}$ and ${1.60s}$, respectively. 
For \texttt{tgpt}, training on the full dataset requires ${3.51s}$ for \textsc{rpt} and ${2.02s}$ for \textsc{rppt}, while sampled training lowers this to ${1.18s}$ and ${1.65s}$.
On the full training set, \textsc{rdt} requires $1.09s$ and $1.08s$ on \texttt{cifar10} and \texttt{tgpt}, respectively, while \textsc{lgbm} requires $45.37s$ on \texttt{cifar10} and $7.21s$ on \texttt{tgpt}. 
When using only $10\%$ of the training data, the training time of \textsc{rdt} decreases to $0.23s$ on \texttt{cifar10} and $0.40s$ on \texttt{tgpt}, whereas \textsc{lgbm} remains significantly more expensive, requiring $41.97s$ and $7.16s$ on \texttt{cifar10} and \texttt{tgpt}, respectively.

\begin{figure}[t]
    \centering
    
    \includegraphics[width=0.48\linewidth]{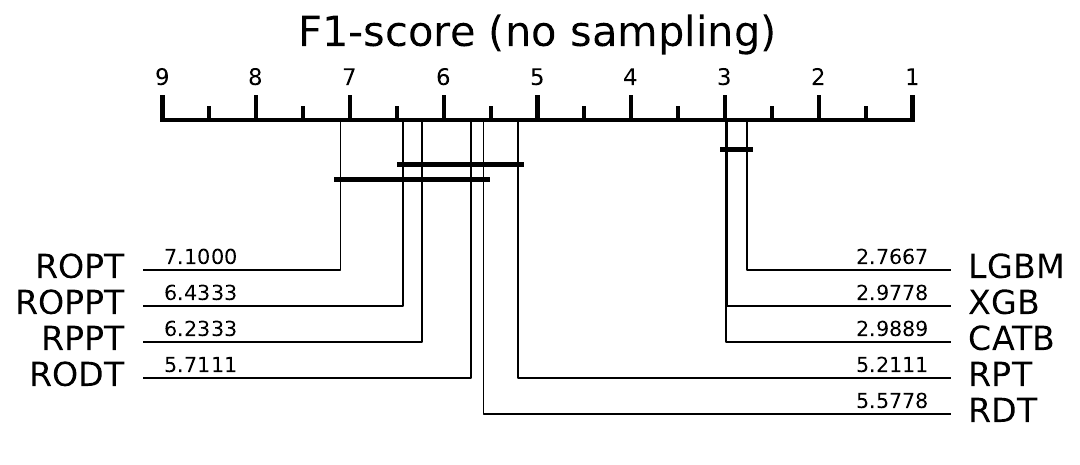}
    \includegraphics[width=0.48\linewidth]
    {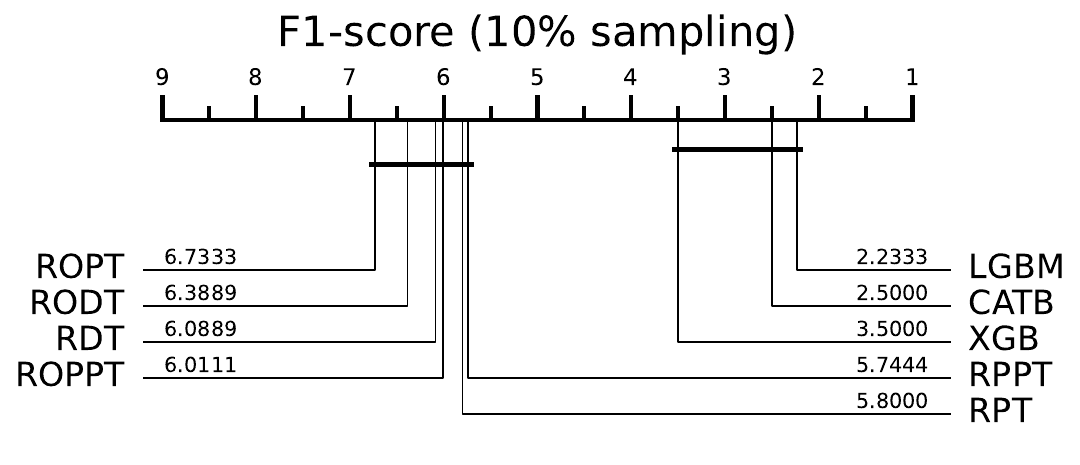}
    \caption{Critical difference plots of ensemble models' rankings in terms of weighted F1-score, evaluated using the Nemenyi test across all datasets for best performing \textsc{RandomPivotForest} variants in comparison with baselines. Models that are not significantly different at the $95\%$ significance level are connected. The best models are positioned on the right.}
    \label{fig:cd_plots_ensemble}
\end{figure}

For larger datasets, the performance gains are more substantial. 
Training times for \textsc{rdt} and \textsc{lgbm} vary significantly across datasets and data sampling. 
For \textsc{rdt}, training on \texttt{magic} requires $1.34s$, which decreases to $0.64s$ when only $10\%$ of the data is used, while on \texttt{svhn}, training requires $0.92s$, reducing to $0.32s$ with $10\%$ sampling. 
In comparison, \textsc{lgbm} is generally faster on \texttt{magic}, requiring $0.89s$ for the full dataset and $0.87s$ for $10\%$ sampling, but considerably slower on \texttt{svhn}, requiring $55.19s$ for the full dataset and $50.12s$ with $10\%$ sampling. 
On \texttt{magic}, training on the full dataset requires ${2.14s}$ for \textsc{rpt} and ${3.75s}$ for \textsc{rppt}; using the sampled context, training times drop to ${1.74s}$ and ${1.65s}$, respectively.
On the large-scale \texttt{svhn} dataset, full training times are ${18.63s}$ for \textsc{rpt} and ${17.09s}$ for \textsc{rppt}, while with sampling, they are significantly reduced to ${0.69s}$ and ${0.68s}$. 
These aspects highlight that even when using a small percentage of the training set, satisfactory predictive performance can still be achieved, while significantly speeding up training time. 
In terms of prediction times, all \textsc{RandomPivotForest} variants exhibit similar performance across both the full training and sampled settings, with times generally around $1s$ for all datasets considered. 
The minimum prediction time is $0.167s$ for \textsc{rpt} on \texttt{breast}, while the maximum is $1.23s$ for \textsc{rppt} on \texttt{tgpt}.

\begin{table}[t]
\caption{Average weighted F1-score $\pm$ std. dev. for \textsc{PivotTree} classifiers and selectors combined with \textsc{dt} in the unconstrained setting, alongside baselines and competing methods, with respect to different distance measures.
Subscripts denote the average number of pivots $\pm$ standard deviation (rounded up to nearest integer).
Best results are shown in \textbf{bold}, and second-best results in \textit{italics}.}
\centering
\setlength{\tabcolsep}{2.2mm}
\label{ref:tab_dist_comp_standalone}
\begin{tabular}{c c c c c c c}
\toprule
 & model & $\texttt{tabular}$ & $\texttt{images}$ & $\texttt{time-series}$ & $\texttt{text}$ & \texttt{all} \\
\midrule

\multirow{8}{*}{\rotatebox{90}{\texttt{euclidean}}} 
& $\textsc{ptc}^{}_{Z}$ & $.79_{\textbf{11}} \pm .14_{3}$ & $.69_{\textit{11}} \pm .20_{4}$ & $.64_{\textbf{9}} \pm .27_{3}$ & $\textit{.59}_{13} \pm .21_{2}$ & $.71_{\textbf{11}} \pm .21_{3}$ \\
& $\textsc{pptc}^{}_{Z}$ & $.80_{14} \pm .14_{4}$ & $.73_{\textit{11}} \pm .17_{5}$ & $.63_{12} \pm .25_{6}$ & $.58_{\textit{12}} \pm .21_{3}$ & $.72_{13} \pm .20_{4}$ \\
& $\textsc{optc}^{}_{Z}$ & $.76_{17} \pm .15_{6}$ & $.68_{15} \pm .20_{6}$ & $.63_{13} \pm .26_{6}$ & $.55_{23} \pm .21_{4}$ & $.68_{16} \pm .20_{6}$ \\
& $\textsc{opptc}^{}_{Z}$ & $.79_{13} \pm .14_{5}$ & $.73_{12} \pm .21_{6}$ & $.59_{11} \pm .28_{7}$ & $.55_{14} \pm .20_{4}$ & $.70_{\textbf{12}} \pm .21_{5}$ \\
\cmidrule(lr){2-7}
& $\textsc{pts}^{}_{Z}$ & $.79_{39} \pm .14_{19}$ & $.69_{50} \pm .20_{21}$ & $.63_{35} \pm .28_{31}$ & ${.60}_{61} \pm .21_{32}$ & $.71_{43} \pm .21_{25}$ \\
& $\textsc{ppts}^{}_{Z}$ & $.78_{37} \pm .14_{24}$ & $.69_{55} \pm .21_{39}$ & $.64_{37} \pm .27_{36}$ & ${.59}_{52} \pm .21_{45}$ & $.71_{43} \pm .20_{33}$ \\
& $\textsc{opts}^{}_{Z}$ & $.78_{50} \pm .14_{30}$ & $.68_{66} \pm .20_{36}$ & $.63_{47} \pm .26_{48}$ & $.58_{96} \pm .21_{68}$ & $.70_{59} \pm .20_{43}$ \\
& $\textsc{oppts}^{}_{Z}$ & $.79_{46} \pm .15_{31}$ & $.69_{69} \pm .21_{49}$ & $.61_{40} \pm .26_{34}$ & $.56_{76} \pm .20_{60}$ & $.70_{53} \pm .21_{41}$ \\
\midrule
\multirow{8}{*}{\rotatebox{90}{\texttt{cosine}}} 
& $\textsc{ptc}^{}_{Z}$ & $\textit{.81}_{\textbf{11}} \pm .13_{3}$ & $.69_{\textit{11}} \pm .20_{4}$ & $\textit{.65}_{\textit{10}} \pm .26_{4}$ & $\textbf{.60}_{12} \pm .22_{2}$ & $.72_{\textbf{11}} \pm .20_{3}$ \\
& $\textsc{pptc}^{}_{Z}$ & $.80_{\textit{12}} \pm .14_{5}$ & $\textbf{.87}_{15} \pm .10_{4}$ & $\textit{.65}_{17} \pm .24_{7}$ & $\textit{.59}_{14} \pm .22_{5}$ & $\textbf{.75}_{14} \pm .19_{6}$ \\
& $\textsc{optc}^{}_{Z}$ & $.78_{17} \pm .15_{5}$ & $.68_{16} \pm .20_{5}$ & $\textit{.65}_{13} \pm .27_{5}$ & $.58_{21} \pm .22_{5}$ & $.70_{16} \pm .21_{5}$ \\
& $\textsc{opptc}^{}_{Z}$ & $.80_{13} \pm .13_{5}$ & $\textit{.80}_{15} \pm .13_{4}$ & $.63_{15} \pm .25_{7}$ & $\textit{.59}_{\textbf{11}} \pm .22_{4}$ & $\textit{.73}_{14} \pm .19_{5}$ \\
\cmidrule(lr){2-7}
& $\textsc{pts}^{}_{Z}$ & $\textbf{.82}_{42} \pm .13_{23}$ & $.69_{51} \pm .20_{24}$ & $\textbf{.66}_{36} \pm .25_{31}$ & $\textbf{.60}_{62} \pm .21_{34}$ & $\textit{.73}_{45} \pm .20_{27}$ \\
& $\textsc{ppts}^{}_{Z}$ & $\textit{.81}_{34} \pm .14_{31}$ & $.68_{77} \pm .20_{55}$ & $.64_{46} \pm .26_{36}$ & $\textbf{.60}_{54} \pm .23_{45}$ & $.72_{48} \pm .20_{42}$ \\
& $\textsc{opts}^{}_{Z}$ & $.80_{50} \pm .14_{26}$ & $.69_{69} \pm .20_{36}$ & $.63_{43} \pm .26_{33}$ & $\textit{.59}_{87} \pm .22_{63}$ & $.71_{57} \pm .21_{38}$ \\
& $\textsc{oppts}^{}_{Z}$ & $.79_{{51}} \pm .13_{42}$ & $.68_{111} \pm .21_{72}$ & $.64_{61} \pm .25_{56}$ & $\textit{.59}_{67} \pm .21_{52}$ & $.71_{67} \pm .20_{56}$ \\
\midrule
\multirow{8}{*}{\rotatebox{90}{\texttt{manhattan}}} 
& $\textsc{ptc}^{}_{Z}$ & $.78_{\textit{12}} \pm .14_{3}$ & $.68_{\textbf{10}} \pm .21_{3}$ & $.63_{\textbf{9}} \pm .26_{4}$ & $.58_{13} \pm .20_{2}$ & $.70_{\textbf{11}} \pm .20_{3}$ \\
& $\textsc{pptc}^{}_{Z}$ & $.80_{13} \pm .14_{5}$ & $.74_{12} \pm .19_{6}$ & $\textit{.65}_{14} \pm .25_{6}$ & $.58_{\textit{12}} \pm .22_{3}$ & ${.73}_{13} \pm .20_{5}$ \\
& $\textsc{optc}^{}_{Z}$ & $.77_{16} \pm .16_{5}$ & $.68_{16} \pm .20_{6}$ & $.64_{15} \pm .26_{6}$ & $.54_{22} \pm .19_{4}$ & $.69_{17} \pm .21_{6}$ \\
& $\textsc{opptc}^{}_{Z}$ & $.78_{51} \pm .14_{31}$ & $.69_{74} \pm .21_{39}$ & $\textit{.65}_{46} \pm .25_{36}$ & $.56_{66} \pm .21_{43}$ & $.71_{56} \pm .20_{36}$ \\
\cmidrule(lr){2-7}
& $\textsc{pts}^{}_{Z}$ & $.79_{41} \pm .14_{20}$ & $.69_{48} \pm .21_{25}$ & $.63_{36} \pm .26_{31}$ & $.58_{63} \pm .20_{38}$ & $.71_{44} \pm .21_{27}$ \\
& $\textsc{ppts}^{}_{Z}$ & $.79_{36} \pm .15_{24}$ & $.69_{54} \pm .21_{38}$ & $.64_{37} \pm .25_{27}$ & $\textit{.59}_{50} \pm .21_{35}$ & $.71_{42} \pm .21_{29}$ \\
& $\textsc{opts}^{}_{Z}$ & $.80_{49} \pm .14_{21}$ & $.69_{70} \pm .21_{37}$ & ${.65}_{48} \pm .26_{37}$ & $.56_{88} \pm .20_{66}$ & $.71_{58} \pm .21_{38}$ \\
& $\textsc{oppts}^{}_{Z}$ & $.78_{50} \pm .14_{30}$ & $.69_{73} \pm .21_{39}$ & ${.65}_{45} \pm .25_{35}$ & $.56_{66} \pm .21_{42}$ & $.71_{56} \pm .20_{35}$ \\
\bottomrule
\end{tabular}
\end{table}

We complement our analysis with statistical significance tests in the ensemble context reported in Figure~\ref{fig:cd_plots_ensemble}. 
Also in this case, to ensure a fair comparison, we restrict the analysis to the standard \textsc{RandomPivotForest} variants without the splitting-stump strategy and evaluate them against the corresponding baselines\footnote{For completeness, in Appendix~\ref{sec:appendix} we also report the critical difference plots including the splitting-stump forest variants.}.
Notably, while Tables~\ref{tab:ensemble_full} and~\ref{tab:ensemble_10_perc} indicate that, although \textsc{RandomPivotForests} models do not always achieve the highest average mean performance, they attain the best rankings when compared to feature-based Random Forests approaches when a relative comparison per dataset is conducted using the Nemenyi test in both the no-sampling and $10\%$ sampling scenarios.
In the no-sampling scenario, \textsc{rpt} ranks as the best model among the Random Forests, followed by \textsc{rdt}, \textsc{rodt}, and \textsc{rppt}, with no statistically significant differences among them. Similarly, in the $10\%$ sampling scenario, \textsc{rpt} and \textsc{rppt} occupy the top ranks in terms of performance, followed by other models.

As an overall remark for the ensemble scenario, we recommend \textsc{rpt} as a robust and reliable variant among the tested \textsc{RandomPivotForest} methods across \texttt{tabular}, \texttt{time-series}, and \texttt{text} domains. In the absence of any sampling strategy during training, \textsc{rpt} matches the performance of both \textsc{rdt} and \textsc{rodt} on \texttt{tabular} and \texttt{time-series} data. It further outperforms all alternatives in the \texttt{text} domain, benefitting from training trees in a more interpretable similarity space. The only exception is in the \texttt{images} domain, where \textsc{rppt} and \textsc{roppt} variants significantly outperform all other \textsc{RandomPivotForest} methods.

When sampling strategies are applied, we observe that training efficiency improves considerably, while predictive performance remains stable and competitive. Crucially, the main strengths of top-performing \textsc{RandomPivotForest} variants are also preserved in the sampling scenario.

\smallskip
In comparing ensemble methods to single-model configurations, ensemble-based approaches tend to yield better results than the latter methods paired with \textsc{dt}.

However, they still lag behind \textsc{PivotTree} variants used as selectors with \textsc{knn}, suggesting that even a single trained \textsc{PivotTree} combined with \textsc{knn} can produce reliable and robust performance across data types.
Based on the results, different models may be preferred depending on the desired trade-off between interpretability and performance. 
While simpler models or more interpretable structures may be attractive in settings where transparency and explainability are critical, \textsc{PivotTree} combined with \textsc{knn} consistently demonstrates stronger predictive performance. 
This indicates that the selector role played by \textsc{PivotTree} is effective at identifying informative partitions of the feature space, allowing \textsc{knn} to operate on more relevant local neighborhoods.

Furthermore, the results suggest that the benefits of this hybrid approach are not limited to a specific dataset or data distribution. 
Across multiple experimental settings, the \textsc{PivotTree}-based selectors maintain stable behavior, highlighting their robustness and general applicability. 
In practice, this makes them a compelling choice when performance is prioritized and when moderate model complexity is acceptable. 
At the same time, their tree-based structure still provides a degree of interpretability, offering some insight into how the data is partitioned before the final \textsc{knn}-based prediction is performed.
If interpretability is a primary concern and a similarity-based \textsc{knn} approach is suitable, we recommend \textsc{ppts}$_{Z}$ with \textsc{knn} due to its consistent performance and conceptual simplicity.
For scenarios where faster inference time than \textsc{knn} is required and a more structured set of rules is beneficial, \textsc{rpt} emerges as the preferred choice due to its broad applicability across data types. 
The only exception is in the \texttt{images} domain, where \textsc{rppt} achieves superior performance. 
Finally, if a standalone single interpretable tree structure is desired, both \textsc{ptc}$_{Z}$ and \textsc{pptc}$_{Z}$ are strong and reliable candidates, suitable for most data types.

\smallskip
\textbf{Impact of Distance Measures Evaluation.}
%}
%
%
\begin{table}[t]
\caption{
Average weighted F1-score $\pm$ std. dev. for \textsc{RandomPivotForest} with 100 \textsc{PivotTree} estimators, where each estimator is trained on 10\% of the training set.
Results are reported for different distance measures.
Best results are shown in \textbf{bold}, and second-best results in \textit{italics}.
}
\centering
\setlength{\tabcolsep}{3mm}
\label{ref:tab_dist_comp_ensemble}
\begin{tabular}{c l c c c c c}
\toprule
 & model & \texttt{tabular} & \texttt{images} & \texttt{time-series} & \texttt{text} & \texttt{all} \\
\midrule

\multirow{4}{*}{\rotatebox{90}{\texttt{euclidean}}}
& $\textsc{rpt}$   
& \textbf{.78} $\pm$ .16 
& .86 $\pm$ .10 
& \textbf{.64} $\pm$ .26 
& \textit{.57} $\pm$ .24 
& \textit{.73} $\pm$ .21 \\

& $\textsc{ropt}$  
& \textit{.76} $\pm$ .17 
& .86 $\pm$ .11 
& \textbf{.64} $\pm$ .26 
& .52 $\pm$ .24 
& .72 $\pm$ .21 \\

& $\textsc{rppt}$  
& \textbf{.77} $\pm$ .17 
& \textit{.93} $\pm$ .05 
& \textit{.63} $\pm$ .26 
& .53 $\pm$ .25 
& \textbf{.74} $\pm$ .22 \\

& $\textsc{roppt}$ 
& \textbf{.77} $\pm$ .17 
& \textit{.93} $\pm$ .05 
& \textit{.63} $\pm$ .26 
& .52 $\pm$ .25 
& \textbf{.74} $\pm$ .22 \\

\midrule
\multirow{4}{*}{\rotatebox{90}{\texttt{cosine}}}
& $\textsc{rpt}$   
& \textit{.77} $\pm$ .16 
& .88 $\pm$ .09 
& \textit{.63} $\pm$ .26 
& \textbf{.58} $\pm$ .24 
& \textit{.73} $\pm$ .21 \\

& $\textsc{ropt}$  
& .76 $\pm$ .17 
& .87 $\pm$ .09 
& \textbf{.64} $\pm$ .26 
& .54 $\pm$ .24 
& \textit{.73} $\pm$ .21 \\

& $\textsc{rppt}$  
& .76 $\pm$ .17 
& \textbf{.94} $\pm$ .04 
& .62 $\pm$ .26 
& .56 $\pm$ .24 
& \textbf{.74} $\pm$ .22 \\

& $\textsc{roppt}$ 
& .76 $\pm$ .16 
& \textbf{.94} $\pm$ .05 
& \textbf{.64} $\pm$ .26 
& .54 $\pm$ .24 
& \textbf{.74} $\pm$ .22 \\

\midrule
\multirow{4}{*}{\rotatebox{90}{\texttt{manhattan}}}
& $\textsc{rpt}$   
& \textbf{.78} $\pm$ .16 
& .86 $\pm$ .10 
& \textbf{.64} $\pm$ .26 
& .55 $\pm$ .24 
& \textit{.73} $\pm$ .21 \\

& $\textsc{ropt}$  
& \textit{.77} $\pm$ .17 
& .87 $\pm$ .09 
& \textbf{.64} $\pm$ .26 
& .52 $\pm$ .24 
& .72 $\pm$ .22 \\

& $\textsc{rppt}$  
& \textbf{.78} $\pm$ .17 
& \textit{.93} $\pm$ .05 
& \textit{.63} $\pm$ .26 
& .52 $\pm$ .25 
& \textbf{.74} $\pm$ .23 \\

& $\textsc{roppt}$ 
& \textbf{.78} $\pm$ .16 
& \textit{.93} $\pm$ .05 
& \textit{.63} $\pm$ .26 
& .52 $\pm$ .26 
& \textbf{.74} $\pm$ .23 \\

\bottomrule
\end{tabular}
\end{table}

We complete our quantitative analysis by analyzing the impact of different distance measures on the overall performance of \textsc{PivotTree} and \textsc{RandomPivotForest} variants across different data modalities. As reference distance measures, we compare the standard Euclidean distance used in the main experiments with the Manhattan and cosine distances.
Since our objective in this context is to compare the effect of various distance measures on model performance, we fix to $4$ the parameter $\mathit{maxdepth}$ of the trees, both when used as standalone models and within ensembles, and compare their performance on the same test set for each dataset used in previous experiments.

For ensemble variants, we fix the number of samples used to train each estimator to 10\% of the training set, while for standalone models we work in the unconstrained setting w.r.t. the number of pivots.
Table~\ref{ref:tab_dist_comp_standalone} reports the average weighted F1-score of the different \textsc{PivotTree} variants w.r.t. different distance measures and across multiple data modalities.
Overall, performance appears largely comparable across distance measures, both when \textsc{PivotTree} is employed directly as a classification model and when it is used as a selector in combination with \textsc{dt}.
Interestingly, a different trend emerges for the \texttt{images} modality, where a substantial improvement is observed when cosine distance is adopted.
In particular, \textsc{pptc}$_Z$ achieves the strongest results in this setting.
This behaviour may be influenced by the specific architectures used to extract latent representations from images, suggesting that the choice of distance measure should be carefully considered when applying \textsc{PivotTree} variants to domains that rely on learned feature embeddings.
In our experiments, we also assessed model performance in terms of different distance measures using the Nemenyi post-hoc test. 
The results indicate that there are no statistically significant differences among the different \textsc{PivotTree} variants employing different distance measures at the $95\%$ significance level.
We complete the analysis of the impact of distance measures by reporting in Table~\ref{ref:tab_dist_comp_ensemble} the weighted F1-score achieved by the different \textsc{RandomPivotForest} variants.
In this setting, the results are largely comparable across distance measures, with no single distance measure consistently emerging as the best choice across all data modalities.
Also in this case, the Nemenyi post-hoc test conducted in our experiments revealed that the ensembles of \textsc{PivotTree}s are statistically indistinguishable.

\subsection{Qualitative Results}
\label{subsec:qual_results}\

\begin{figure}[t]
    \centering
    \includegraphics[width=1.0\linewidth]{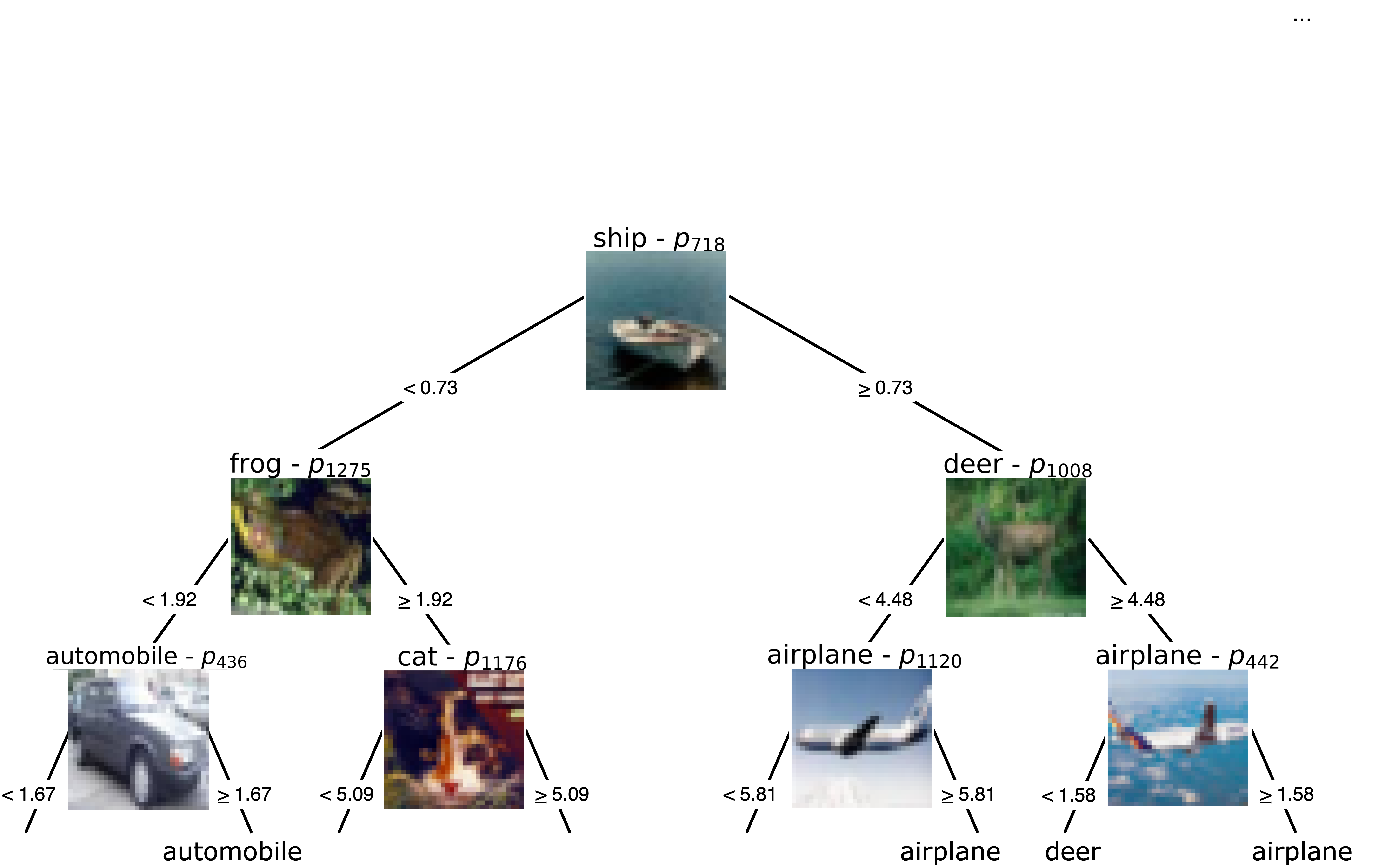}
    \caption{Example of \textsc{ptc}$_{Z}$ with $\mathit{maxdepth} = 4$ on  \texttt{cifar10}. Only partial structure shown for visualization purposes.}
    \label{fig:cifar10_univar}
\end{figure}

In the following, we illustrate some qualitative examples on different data types to show the usability of \textsc{PivotTree} at prediction and explanation time using a decision tree as interpretable model $f$.
In Figure~\ref{fig:cifar10_univar}, we report an example of prediction and explanation on \texttt{cifar10} using \textsc{ptc}$_{Z}$\footnote{For visualization purposes, rules of the form ``$s(x, p_i) \geq \beta$'' are reported as ``$s(x, p_i) \geq \lceil \mu \rceil - \beta$'', where $\mu$ is the maximum threshold value found in the trained tree.}.

The trained \textsc{ptc}$_{Z}$ identifies a single pivot per node and performs splits based on similarity to a threshold value: given a hypothetical instance $x$ to predict, the predictive reasoning employed by the trained model proceeds as follows: $x$ is first compared to $p_{718}$, an instance of an \textit{ship} class. 
If it is similar enough to such instance, meaning $s(x,p_{718})\geq 0.73$, it is then compared to $p_{1008}$, an example of \textit{deer}. 
If $x$ is sufficiently similar to $p_{1008}$ ($s(x,p_{1008})\geq 4.48$) the model performs a further check with another \textit{airplane} instance, $p_{442}$. 
If $x$ is similar enough to $p_{442}$ ($s(x,p_{1008})\geq 1.58$), then it is identified as an \textit{airplane}, otherwise as a \textit{deer}. 
Analogous reasoning applies for remaining branches.

The trained \textsc{ptc}$_{Z}$ identifies a single pivot per node and performs splits based on similarity to a threshold value: given a hypothetical instance $x$ to predict, the predictive reasoning employed by the trained model proceeds as follows: $x$ is first compared to $p_{718}$, an instance of an \textit{ship} class. 
If it is similar enough to such instance, meaning $s(x,p_{718})\geq 0.73$, it is then compared to $p_{1008}$, an example of \textit{deer}. 
If $x$ is sufficiently similar to $p_{1008}$ ($s(x,p_{1008})\geq 4.48$) the model performs a further check with another \textit{airplane} instance, $p_{442}$. 
If $x$ is similar enough to $p_{442}$ ($s(x,p_{1008})\geq 1.58$), then it is identified as an \textit{airplane}, otherwise as a \textit{deer}. 
Analogous reasoning applies for remaining branches.
We notice that, in this process, pivots are selected with a clear variety in their semantic role within the tree. For instance, the initial comparison with a \textit{ship} pivot provides a coarse-grained separation of instances that resemble large structured objects, while subsequent pivots such as the \textit{deer} or \textit{frog} examples perform progressively finer distinctions among remaining candidates.
This qualitative diversity highlights how each pivot contributes a specific semantic cue that guides the decision process along the path of the tree.

\begin{figure}[t]
    \centering
    \includegraphics[width=1.0\linewidth]{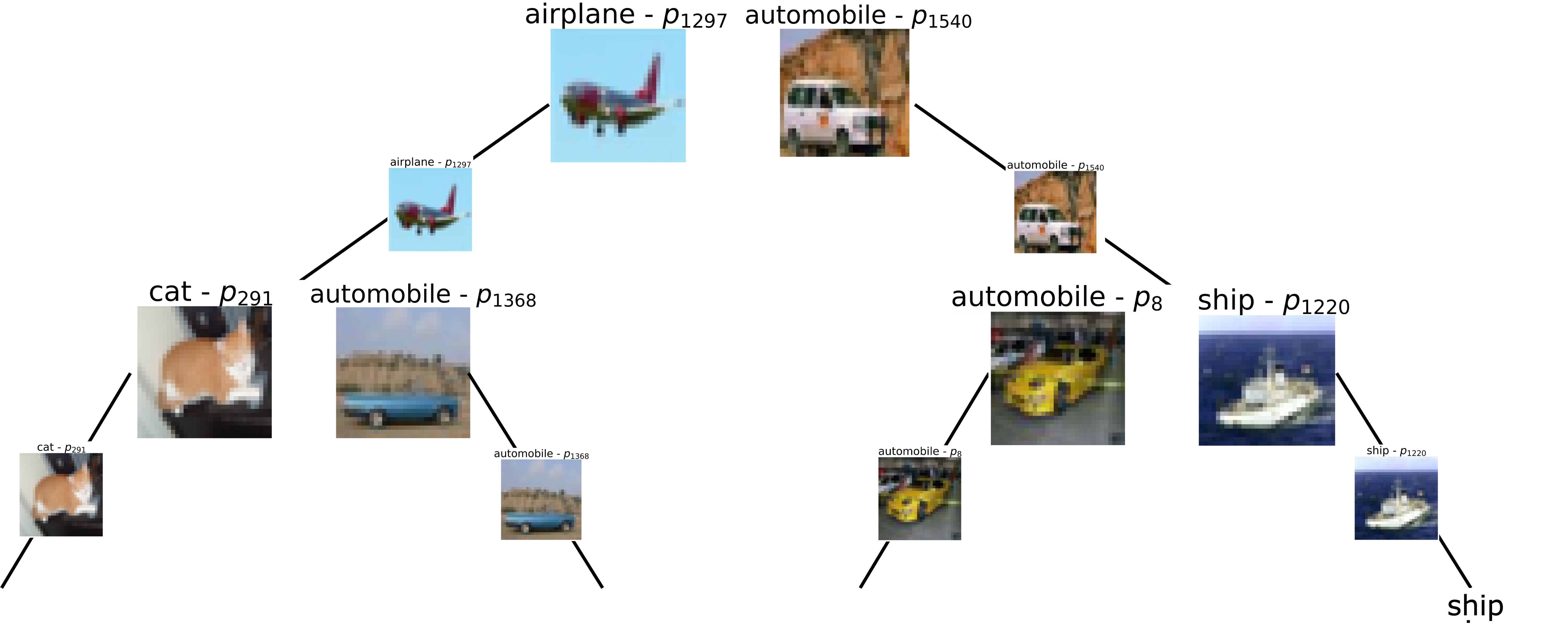}
    \caption{Example of  \textsc{pptc}$_{Z}$ with $\mathit{maxdepth} = 4$ on \texttt{cifar10}. Only partial structure shown for visualization purposes.}
    \label{fig:cifar10}
\end{figure}

Instead, in Figure~\ref{fig:cifar10} we report the proximity-based variant obtained for the same dataset.
In the case of \textsc{pptc}$_{Z}$, the splitting strategy differs, leading also to a different selection of pivots. 
Given a hypothetical instance $x$ to classify, it is first compared to $p_{1297}$ (an \textit{airplane}) and $p_{1540}$ (an \textit{automobile}). 
If $x$ is closer to $p_{1540}$, it follows the right branch.  
In this case, it is then compared to $p_{8}$ (another \textit{automobile}) and to $p_{1220}$, a \textit{ship}, for further validation. 
If $x$ is found to be more similar to $p_{1220}$ than to $p_{8}$, the model classifies it as a \textit{ship}.  
This reasoning can be interpreted in natural language as follows:  
``If $x$ is more similar to \textit{automobile} $p_{1540}$ than to \textit{airplane} $p_{1297}$, but not similar enough to a further \textit{automobile} instance $p_{8}$ and at the same time similar enough to a \textit{ship} $p_{1220}$, then it should be classified as a \textit{ship}.''
Also in this case, pivots are identified with a relevant variety in their semantic function within the tree. Some pivots provide broad reference exemplars that allow the model to distinguish between visually distinct categories, such as \textit{airplane} and \textit{automobile}, while others refine the decision by introducing more specific comparisons with instances that capture subtle similarities across classes. For example, the presence of two \textit{automobile} pivots in different nodes highlights how different representative instances of the same class may play distinct roles: one supports a coarse discrimination at higher levels of the tree, while another helps verify whether the instance truly belongs to that class or should instead be associated with a visually related category such as \textit{ship}.

\begin{figure}[t]
    \centering
    \includegraphics[width=1.0\linewidth]{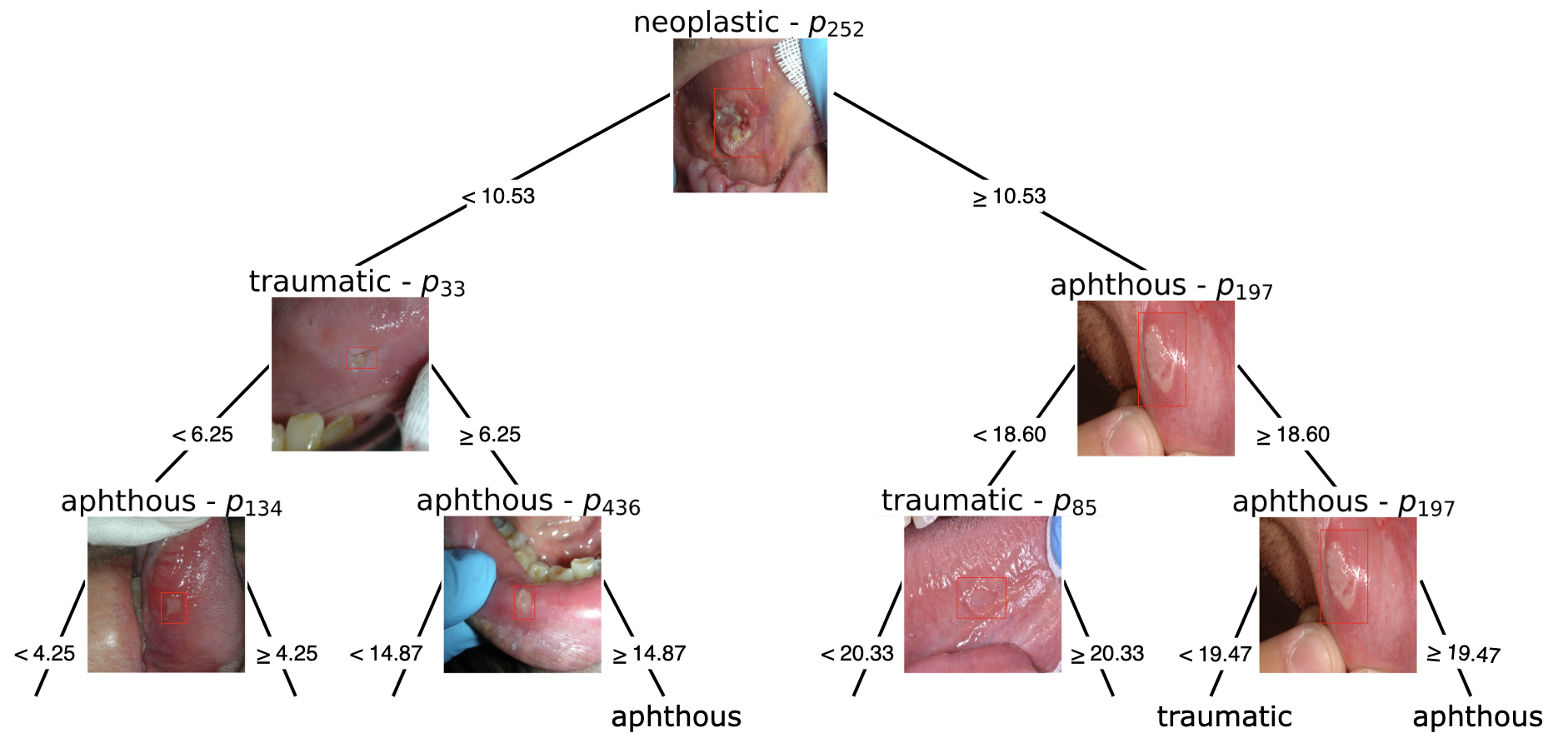}
    \caption{Example of \textsc{ptc}$_{Z}$ with $\mathit{maxdepth} = 4$ on \texttt{oral}. Only partial structure shown for visualization purposes.}
    \label{fig:oral_ptc}
\end{figure}

\begin{figure}[t]
    \centering
    \includegraphics[width=1.0\linewidth]{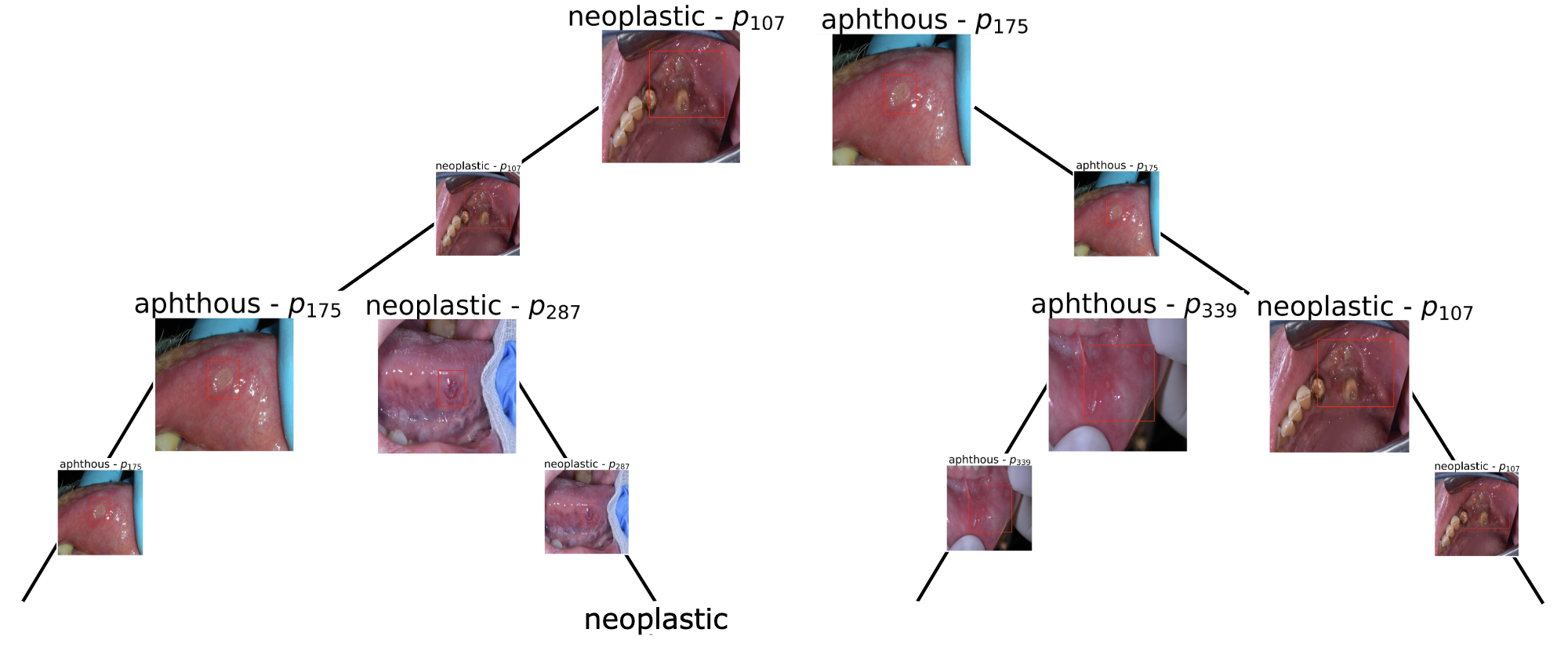}
    \caption{Example of \textsc{pptc}$_{Z}$ with $\mathit{maxdepth} = 4$ on \texttt{oral}. Only partial structure shown for visualization purposes.}
    \label{fig:oral_ppt}
\end{figure}

\smallskip
We also provide a qualitative example of \textsc{PivotTree}s using instances from a real-world healthcare case study~\cite{cascione2024oral}.
In Figure~\ref{fig:oral_ptc}, we present a \textsc{ptc}$_{Z}$ model trained on the \texttt{oral} training set. 
As in previous cases, a hypothetical instance $x$ is evaluated by comparing it with a small number of exemplar instances in order to make a prediction. 
If $x$ does not exhibit a similarity greater than or equal to $10.53$ with the \textit{neoplastic} instance $p_{252}$, it is then compared with the \textit{traumatic} instance $p_{33}$. 
If $x$ is sufficiently similar to $p_{33}$, it is further compared with the \textit{aphthous} instance $p_{435}$.
If it is more similar to $p_{435}$, it is classified as \textit{aphthous}; otherwise, alternative decision paths are followed.

On the other hand, Figure~\ref{fig:oral_ppt} presents a trained \textsc{pptc}$_{Z}$ on the same dataset, offering a different case-based solution to the same problem: an input instance $x$ is first compared with two reference instances: a \textit{neoplastic} case $p_{107}$ and an \textit{aphthous} case $p_{175}$. 
If $x$ is more similar to $p_{107}$, it follows the left branch, where it is further compared with $p_{175}$ again and with another \textit{neoplastic} instance, $p_{287}$. 
If $x$ is found to be closer to $p_{287}$, it is labeled as \textit{neoplastic}; otherwise, additional comparisons are performed. 
Analogous reasoning is applied recursively along other branches of the tree.
We report that, on the \texttt{oral} dataset, $\textsc{ptc}_{Z}$ achieves a weighted F1-score of 0.774 using 6 pivots, while $\textsc{pptc}_{Z}$ achieves 0.772 using 11 pivots. 
In comparison, the \textsc{CNN} used to extract latent representations achieves a weighted F1-score of 0.851. 
This remark that, when interpretability is requested, it might be necessary to sacrifice the performance.

\begin{figure}[t]
    \centering
    \includegraphics[width=1\linewidth]{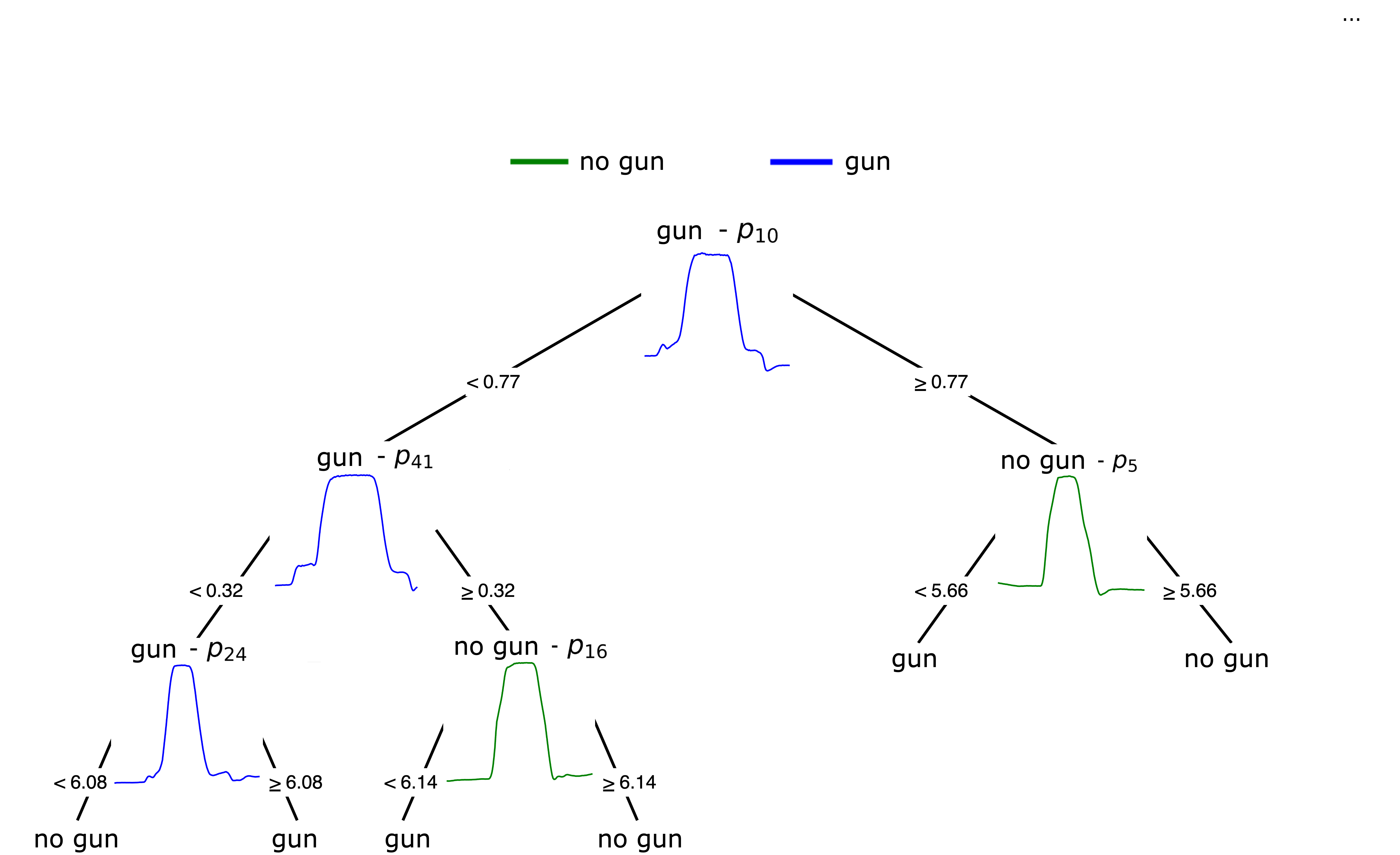}
    \caption{Example of \textsc{ptc}$_{Z}$ with $\mathit{maxdepth} = 4$ on \texttt{gun}. Only partial structure shown for visualization purposes.}
    \label{fig:gun_univar}
\end{figure}

\begin{figure}[t]
    \centering
    \includegraphics[width=1\linewidth]{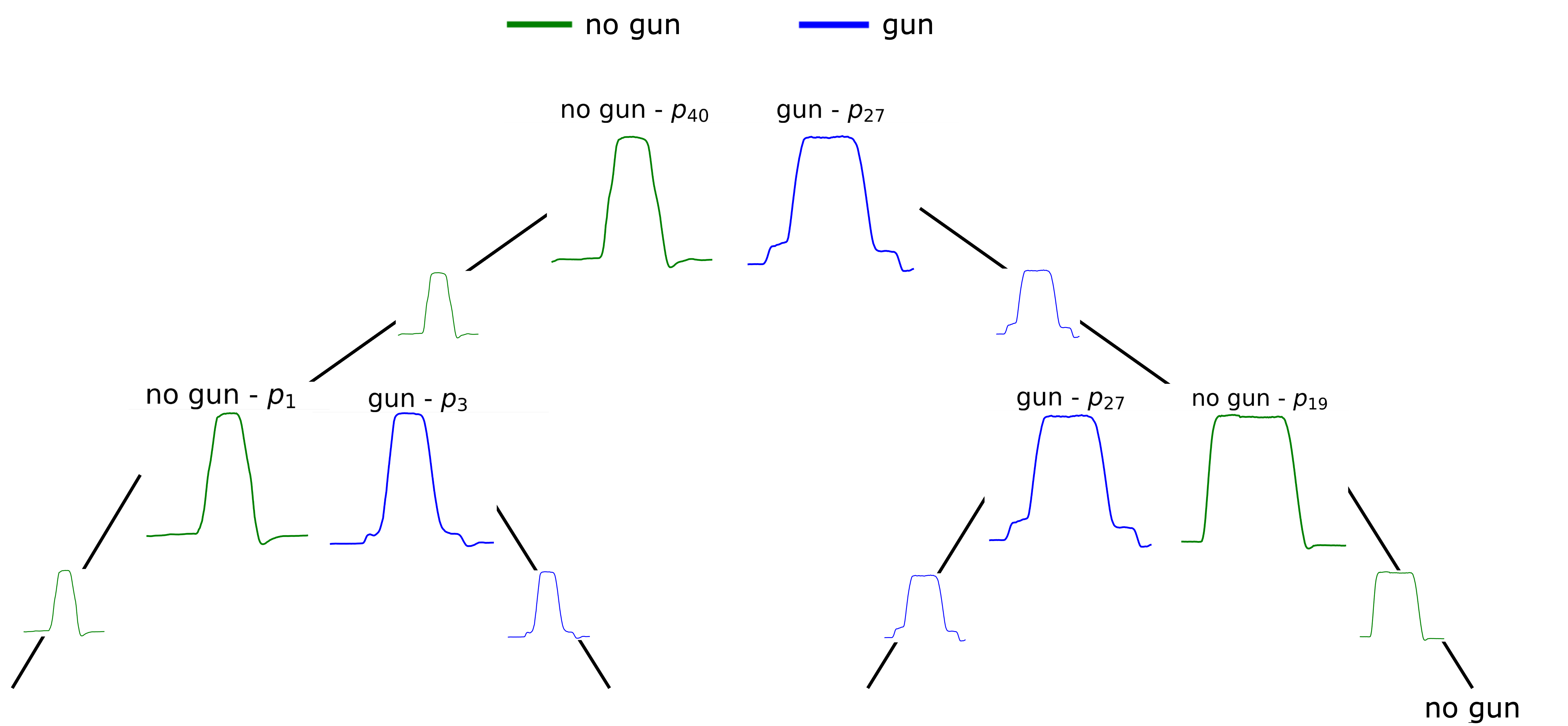}
    \caption{Example of \textsc{pptc}$_{Z}$ with $\mathit{maxdepth} = 4$ on \texttt{gun}. Only partial structure shown for visualization purposes.}
    \label{fig:gun_prox}
\end{figure}

Focusing on a different data type, we provide visualizations of trained \textsc{PivotTrees} on the \texttt{gun} dataset, which consists of univariate \texttt{time-series}. This dataset captures hand motion data from two actors performing two types of gestures: \textit{gun} and \textit{no gun}. In the \textit{gun} class, actors draw a replica gun from a hip-mounted holster, point it at a target for approximately one second, and then return it. In the \textit{no gun}, actors point with their index finger in a similar motion, but without drawing a weapon. 
The time-series represent the X-axis movement of the right-hand centroid during these actions.

Figure~\ref{fig:gun_univar} illustrates a trained \textsc{ptc}$_{Z}$ tree on the \texttt{gun} dataset. The reasoning follows a similar logic to the cases presented above. Given a test instance $x$, the model first compares it with the \textit{gun} pivoy $p_{10}$. This pivot exhibits the characteristic sharp peaks at the beginning and at the end of the time series, a pattern typically associated with gun-draw sequences and corresponding to the rapid arm movement occurring during the action. 
If the similarity score satisfies $s(x, p_{10}) \geq 0.77$, $x$ is then compared with a \textit{no gun} prototype $p_{5}$. In contrast to the previous pivot, $p_{5}$ shows a smoother and more regular temporal profile, lacking the pronounced spikes that characterize gun-related motions. If the similarity with $p_{5}$ exceeds the threshold $5.66$, the instance is classified as \textit{no gun}; otherwise, it is labeled as \textit{gun}. Analogous reasoning is recursively applied along the other branches of the tree.
If instead $x$ is not sufficiently similar to $p_{10}$, the model compares it with another \textit{gun} pivot $p_{41}$. This pivot still presents a clear peak structure, but with a slightly different shape and temporal alignment with respect to $p_{10}$, suggesting that it captures a variant of the gun-draw motion occurring at a different phase or with a different intensity.
Further down the tree, pivot $p_{24}$ represents another \textit{gun} example characterized by a narrower and more localized peak, which helps refine the classification in more ambiguous cases. Conversely, the \textit{no gun} pivot $p_{16}$ displays a broader and smoother oscillatory pattern, reflecting the more gradual arm movements typically observed in the negative class.
Overall, these examples highlight that the pivots selected by the tree are not only limited in number, but also qualitatively heterogeneous. Each pivot captures a distinct and interpretable temporal pattern within the dataset, allowing the decision process to be understood as a sequence of comparisons with representative reference time series.

On the other hand, Figure~\ref{fig:gun_prox} illustrates a trained \textsc{pptc}$_{Z}$ tree on the same \textit{time-series} dataset. Given an input time series $x$, the tree performs a sequence of comparisons against selected pivots. In particular, $x$ is first compared with $p_{40}$ (a \textit{no gun} example) and $p_{27}$ (a \textit{gun} example). The pivot $p_{40}$ exhibits a relatively smooth temporal profile with a broader peak, which is typical of the gradual arm movements characterizing \textit{no gun} sequences. In contrast, $p_{27}$ shows a sharper and more abrupt peak structure, corresponding to the rapid motion associated with gun-draw actions.
If $x$ is more similar to $p_{27}$, it follows the right branch and is subsequently compared with $p_{19}$ (another \textit{no gun} example). Interestingly, although $p_{19}$ belongs to the \textit{no gun} class, it presents a noticeably different temporal structure compared to $p_{40}$. In particular, $p_{19}$ exhibits a wider and smoother peak with more gradual transitions than $p_{40}$. This comparison, therefore, allows the model to further verify whether the input time series truly resembles the rapid motion associated with gun-draw actions or instead corresponds to a smoother motion pattern typical of the \textit{no gun} class.
If $x$ is found to be more similar to $p_{19}$, it is classified as \textit{no gun}; otherwise, additional comparisons further down the tree determine the final label. This step highlights how pivots belonging to the same class may still capture different characteristic patterns within the data, enabling the model to refine the decision through progressively more specific similarity checks.
The left subtree also reveals meaningful variability among the pivots. For example, the \textit{no gun} pivot $p_{1}$ displays a smoother and slightly asymmetric peak, representing a typical negative example with gradual motion dynamics. Conversely, the \textit{gun} pivot $p_{3}$ presents a more pronounced and localized peak, highlighting the sudden movement that distinguishes gun-draw sequences.
The pivots selected by the proximity-based tree again demonstrate qualitative diversity. Rather than relying on a single prototypical pattern, the model selects multiple representative time series capturing different variants of the motion patterns present in the dataset. As a result, the decision path can be interpreted as a sequence of comparisons with meaningful reference examples, where each pivot contributes a specific temporal cue that helps discriminate between \textit{gun} and \textit{no gun} sequences.

\section{Conclusion}
\label{sec:conclusion}
In this paper, we have presented \textsc{PivotTree}, a hierarchical and interpretable case-based model, inspired by decision trees~\cite{breiman1984classification}, that selectively identifies descriptive and discriminative instances to tackle decision-making tasks.
Unlike traditional decision trees, which learn feature-based rules, \textsc{PivotTree} rules are defined in terms of similarity to a set of reference instances named \emph{pivots}.
Rules can be univariate with one pivot per split, oblique with multiple pivots per split, and proximity-based considering relative similarity to a small set of pivots determines the outcome.
Moreover, \textsc{PivotTree} is a data-agnostic model, capable of being applied across various data modalities, simultaneously addressing both pivot selection and prediction tasks. 

We then leverage single \textsc{PivotTree} models to build a forest ensemble, \textsc{RandomPivotForest}, which we optimize for fast tree construction and high performance.
We have compared \textsc{PivotTree} with other case-based models and tree models, both quantitatively and qualitatively, on a large benchmark of 45 datasets of different sizes, dimensionality, and modality.
Quantitatively, as a prediction model, \textsc{PivotTree} shows state-of-the-art performance: it is as accurate as the most accurate models, while being simultaneously more interpretable.
These results are particularly evident on non-relational data, where \textsc{PivotTree} outperforms all competitors.
The same holds when considering pivot selection: interpretable models learned on the pivots selected by \textsc{PivotTree} are both highly interpretable, due to their small size, yet retain high performance.

Our ensemble proposal, \textsc{RandomPivotTree}, while achieving good performances, also displays high variance: in some modalities, e.g., tabular, it performs on par with the competition, while in others, e.g., image, it is markedly better.
Model complexity does not follow the same trend, with \textsc{RandomPivotForest} achieving high variance in complexity, i.e., number of trees.
Our additional analysis of the impact of different distance measures reveals only minor performance variations among the \textsc{PivotTree} variants, with no statistically significant differences overall. Notably, the cosine distance exhibits particularly promising behavior when applied to \texttt{images}, suggesting its potential suitability for this type of data.
Future research could explore hybrid strategies for splitting, combining pivot-based splits and traditional univariate and oblique splits.
Moreover, given the promising results highlighted in this paper by \textsc{RandomPivotForest}, we could further explore the ensemble potential of \textsc{PivotTree} as a weak learner for boosting, rather than bagging, models.
Additionally, future research directions include validating our approach in real-world decision-making scenarios involving human subjects to better assess its practical effectiveness.

\section*{Declarations}

\bmhead{Funding} 
This work has been partially supported by the Italian Project Fondo Italiano per la Scienza FIS00001966 ``MIMOSA'', by the European Community Horizon~2020 programme under the funding schemes ERC-2018-ADG G.A. 834756 ``XAI'', by the European Commission under the NextGeneration EU programme – National Recovery and Resilience Plan (Piano Nazionale di Ripresa e Resilienza, PNRR) Project: ``SoBigData.it – Strengthening the Italian RI for Social Mining and Big Data Analytics'' – Prot. IR0000013 –  Av. n. 3264 del 28/12/2021, M4C2 - Investimento 1.3, Partenariato Esteso PE00000013 - ``FAIR'' - Future Artificial Intelligence Research'' - Spoke 1 ``Human-centered AI'', and ``TANGO'' that has received funding from the European Union's Horizon Europe research and innovation program under G.A. 101120763.

\bmhead{Conflict of interest/Competing interests}
The authors declare that they have no conflict of interests.

\bmhead{Ethics approval and consent to participate}
Not applicable.

\bmhead{Consent for publication}
The authors declare that they all provide consent for publication. 

\bmhead{Data availability} 
The open-source datasets adopted in this work are available at: 
\texttt{tabular:}
\href{https://archive.ics.uci.edu/dataset/52/ionosphere}{\texttt{ion}},
\href{https://archive.ics.uci.edu/dataset/547/algerian+forest+fires+dataset}{\texttt{fire}},
\href{https://archive.ics.uci.edu/dataset/110/yeast}{\texttt{yeast}},
\href{https://archive.ics.uci.edu/dataset/159/magic+gamma+telescope}{\texttt{magic}},
\href{https://archive.ics.uci.edu/dataset/151/connectionist+bench+sonar+mines+vs+rocks}{\texttt{sonar}},
\href{https://github.com/propublica/compas-analysis}{\texttt{compas}},
\href{https://www.openml.org/d/821}{\texttt{house}},
\href{https://archive.ics.uci.edu/dataset/144/statlog+german+credit+data}{\texttt{german}},
\href{https://archive.ics.uci.edu/dataset/94/spambase}{\texttt{spamb}},
\href{https://www.openml.org/search?type=data&id=1507}{\texttt{norm}},
\href{https://archive.ics.uci.edu/dataset/93/low+resolution+spectrometer}{\texttt{lrs}},
\href{https://archive.ics.uci.edu/dataset/212/vertebral+column}{\texttt{vert}},
\href{https://archive.ics.uci.edu/dataset/105/iris}{\texttt{iris}},
\href{https://archive.ics.uci.edu/dataset/186/wine+quality}{\texttt{wine}},
\href{https://archive.ics.uci.edu/dataset/153/diva}{\texttt{diva}} No public release for privacy concerns. Dataset described in~\cite{lusito2024diva},
\href{https://archive.ics.uci.edu/dataset/17/breast+cancer+wisconsin+diagnostic}{\texttt{breast}},
\href{https://archive.ics.uci.edu/dataset/198/steel+plates+faults}{\texttt{steel}},
\href{https://archive.ics.uci.edu/dataset/39/ecoli}{\texttt{ecoli}},
\href{https://www.kaggle.com/datasets/averkiyoliabev/home-equity-line-of-creditheloc}{\texttt{heloc}},
\href{https://archive.ics.uci.edu/dataset/78/page+blocks+classification}{\texttt{page}}. \texttt{images:}
\href{https://mlpi.ing.unipi.it/doctoralai/}{\texttt{oral}},
\href{https://tinyurl.com/torch-mnist}{\texttt{mnist}},
\href{https://tinyurl.com/cifar10t}{\texttt{cifar10}},
\href{https://kaggle.com/competitions/dogs-vs-cats/data}{\texttt{catsdogs}},
\href{https://tinyurl.com/PT-waterbirds}{\texttt{birds}},
\href{https://tinyurl.com/oxfordpets}{\texttt{pets}},
\href{https://zenodo.org/records/6496656/files/organamnist.npz}{\texttt{organa}},
\href{https://zenodo.org/records/6496656/files/bloodmnist.npz}{\texttt{blood}},
\href{https://tinyurl.com/torch-svhn}{\texttt{svhn}}.
\texttt{time-series:}
\href{https://www.timeseriesclassification.com/aeon-toolkit/Yoga.zip}{\texttt{yoga}},
\href{https://www.timeseriesclassification.com/aeon-toolkit/StarLightCurves.zip}{\texttt{star}},
\href{https://timeseriesclassification.com/aeon-toolkit/ChlorineConcentration.zip}{\texttt{chlorine}},
\href{https://timeseriesclassification.com/aeon-toolkit/SmallKitchenAppliances.zip}{\texttt{kitchen}},
\href{https://timeseriesclassification.com/aeon-toolkit/SharePriceIncrease.zip}{\texttt{share}},
\href{https://timeseriesclassification.com/aeon-toolkit/ElectricDevices.zip}{\texttt{devices}},
\href{https://www.timeseriesclassification.com/aeon-toolkit/GunPoint.zip}{\texttt{gun}},
\href{https://timeseriesclassification.com/aeon-toolkit/WormsTwoClass.zip}{\texttt{worms}},
\href{https://timeseriesclassification.com/aeon-toolkit/ECG5000.zip}{\texttt{ecg}},
\href{https://www.timeseriesclassification.com/aeon-toolkit/Wafer.zip}{\texttt{wafer}}.
\texttt{text:}
\href{https://github.com/sebischair/Medical-Abstracts-TC-Corpus}{\texttt{medabs}},
\href{https://huggingface.co/datasets/HANSEN-REPO/HANSEN}{\texttt{vicuna}},
\href{https://huggingface.co/datasets/HANSEN-REPO/HANSEN}{\texttt{pted}},
\href{https://huggingface.co/datasets/HANSEN-REPO/HANSEN}{\texttt{tgpt}},
\href{https://www.cs.cornell.edu/people/pabo/movie-review-data/review_polarity.tar.gz}{\texttt{pol}},
\href{https://github.com/tfs4/liar_dataset}{\texttt{liar}}.

\bmhead{Code availability}  
The code is open and available here: \url{https://github.com/fismimosa/PivotTree}.

\bmhead{Author contribution}
Alessio Cascione: Conceptualization, Methodology, Software, Validation, Investigation, Resources, Writing – Original Draft, Writing – Review \& Editing, Visualization. 
Mattia Setzu: Conceptualization, Methodology, Investigation, Writing-Original Draft, Writing-Review \& Editing, Visualization.
Cristiano Landi: Conceptualization, Methodology, Software, Validation, Resources.
Paolo Maria Mancarella: Conceptualization, Writing – Review \& Editing.
Riccardo Guidotti: Conceptualization, Methodology, Investigation, Resources, Writing – Original Draft, Writing – Review \& Editing, Supervision, Project administration.

\bibliography{biblio}% common bib file

@article{waa2021evaluating,
  author       = {Jasper van der Waa and
                  Elisabeth Nieuwburg and
                  Anita H. M. Cremers and
                  Mark A. Neerincx},
  title        = {Evaluating {XAI:} {A} comparison of rule-based and example-based explanations},
  journal      = {Artificial Intelligence},
  volume       = {291},
}

@incollection{schank2014knowledge,
  title={Knowledge and memory: The real story},
  author={Schank, Roger C and Abelson, Robert P},
  booktitle={Knowledge and memory: The real story},
  pages={1--85},
  year={2014},
  publisher={Psychology Press},
  editor = {Robert S. and Wyer, Jr.},
  address = {East Sussex, UK}
 
}

@article{johnson2010mental,
  title={Mental models and human reasoning},
  author={Johnson-Laird, Philip N},
  journal={Proceedings of the National Academy of Sciences},
  volume={107},
  number={43},
  pages={18243--18250},
  year={2010},
  publisher={National Acad Sciences}
}

@inproceedings{frosst2017distilling,
  author       = {Nicholas Frosst and
                  Geoffrey E. Hinton},
  title        = {Distilling a Neural Network Into a Soft Decision Tree},
  booktitle    = {CEx@AI*IA},
  series       = {{CEUR} Workshop Proceedings},
  volume       = {2071},
  publisher    = {CEUR-WS.org},
  year         = {2017},
  address = {Aachen}
}

@book{breiman1984classification,
  author       = {Leo Breiman and
                  J. H. Friedman and
                  R. A. Olshen and
                  C. J. Stone},
  title        = {Classification and Regression Trees},
  publisher    = {Wadsworth},
  year         = {1984},
address = {{Belmont, CA , USA}}
}

@article{golding1995review,
  author       = {Andrew R. Golding},
  title        = {A Review of Case-Based Reasoning},
  journal      = {{AI} Mag.},
  volume       = {16},
  number       = {2},
  pages        = {85--86},
  year         = {1995}
}

@article{tan2006data,
  title={Data mining introduction},
  author={Tan, Pang-Ning and Steinbach, Michael and Kumar, Vipin},
  journal={People’s Posts and Telecommunications Publishing House, Beijing},
  year={2006}
}

@book{fix1985discriminatory,
  title={Discriminatory analysis: nonparametric discrimination, consistency properties},
  author={Fix, Evelyn},
  volume={1},
  year={1985},
  publisher={USAF school of Aviation Medicine},
  address ={{Fairborn, OH, USA}}
}

@inproceedings{DBLP:conf/hcomp/HaseCLR19,
  author       = {Peter Hase and
                  Chaofan Chen and
                  Oscar Li and
                  Cynthia Rudin},
  title        = {Interpretable Image Recognition with Hierarchical Prototypes},
  booktitle    = {{HCOMP}},
  pages        = {32--40},
  publisher    = {{AAAI} Press},
  year         = {2019},
  address = {{Washington, DC, USA}}
}

@inproceedings{DBLP:conf/cvpr/ZhangYMW19,
  author       = {Quanshi Zhang and
                  Yu Yang and
                  Haotian Ma and
                  Ying Nian Wu},
  title        = {{Interpreting CNNs via Decision Trees}},
  booktitle    = {{CVPR}},
  pages        = {6261--6270},
  publisher    = {Computer Vision Foundation / {IEEE}},
  year         = {2019},
  address = {New York, NY, USA},
}

@inproceedings{Setzu2023tree,
  title = {Correlation and Unintended Biases on Univariate and Multivariate Decision Trees},
  booktitle = {2023 IEEE International Conference on Big Data (BigData)},
  publisher = {IEEE},
  author = {Setzu,  M. and Ruggieri,  S.},
  year = {2023},
  month = dec,
  address = {Piscataway, NJ, USA},

}

@book{DBLP:series/smpai/PekalskaD05,
  author       = {Elzbieta Pekalska and
                  Robert P. W. Duin},
  title        = {The Dissimilarity Representation for Pattern Recognition - Foundations
                  and Applications},
  series       = {Series in Machine Perception and Artificial Intelligence},
  volume       = {64},
  publisher    = {WorldScientific},
  year         = {2005},
  address = {{Singapore}}
}

@inproceedings{DBLP:conf/nips/ChenLTBRS19,
  author       = {Chaofan Chen and
                  Oscar Li and
                  Daniel Tao and
                  Alina Barnett and
                  Cynthia Rudin and
                  Jonathan Su},
  title        = {This Looks Like That: Deep Learning for Interpretable Image Recognition},
  booktitle    = {NeurIPS},
  pages        = {8928--8939},
  year         = {2019}
}

@article{demvsar2006statistical,
  title={Statistical comparisons of classifiers over multiple data sets},
  author={Dem{\v{s}}ar, Janez},
  journal={Journal of Machine learning research},
  volume={7},
  number={Jan},
  pages={1--30},
  year={2006}
}

@inproceedings{nauta2021prototree,
  author       = {Meike Nauta and
                  Ron van Bree and
                  Christin Seifert},
  title        = {Neural Prototype Trees for Interpretable Fine-Grained Image Recognition},
  booktitle    = {{CVPR}},
}

@article{bien2011hierarchical,
  title={Hierarchical clustering with prototypes via minimax linkage},
  author={Bien, Jacob and Tibshirani, Robert},
  journal={Journal of the American Statistical Association},
  volume={106},
  number={495},
  pages={1075--1084},
  year={2011},
  publisher={Taylor \& Francis}
}

@article{bien2011epsilon, 
  title={Prototype selection for interpretable classification},
  author={Bien, Jacob and Tibshirani, Robert},
  journal = {The Annals of Applied Statistics},
  year={2011},
  pages = {2403--2424},
}

@article{DBLP:journals/datamine/LucasSPOZGPW19,
  author       = {Benjamin Lucas and
                  Ahmed Shifaz and
                  Charlotte Pelletier and
                  Lachlan O'Neill and
                  Nayyar Abbas Zaidi and
                  Bart Goethals and
                  Francois Petitjean and
                  Geoffrey I. Webb},
  title        = {Proximity Forest: an effective and scalable distance-based classifier
                  for time series},
  journal      = {Data Min. Knowl. Discov.},
  volume       = {33},
  number       = {3},
  pages        = {607--635},
  year         = {2019}
}

@article{DeFauw2018ex1,
  title={Clinically applicable deep learning for diagnosis and referral in retinal disease},
  author={De Fauw, Jeffrey and others},
  journal={Nature medicine},
  volume={24},
  number={9},
  pages={1342--1350},
  year={2018},
  publisher={Nature Publishing Group}
}

@article{chatzakou2019cyber,
  author       = {Despoina Chatzakou and
                  Ilias Leontiadis and
                  Jeremy Blackburn and
                  Emiliano De Cristofaro and
                  Gianluca Stringhini and
                  Athena Vakali and
                  Nicolas Kourtellis},
  title        = {Detecting Cyberbullying and Cyberaggression in Social Media},
  journal      = {{ACM} Trans. Web},
  volume       = {13},
  number       = {3},
  pages        = {17:1--17:51},
  year         = {2019}
}

@article{Guidotti2019market,
  author       = {Riccardo Guidotti and
                  Giulio Rossetti and
                  Luca Pappalardo and
                  Fosca Giannotti and
                  Dino Pedreschi},
  title        = {Personalized Market Basket Prediction with Temporal Annotated Recurring
                  Sequences},
  journal      = {{IEEE} Trans. Knowl. Data Eng.},
  volume       = {31},
  number       = {11},
  pages        = {2151--2163},
  year         = {2019}
}

@article{yang2022unbox,
  author       = {Guang Yang and
                  Qinghao Ye and
                  Jun Xia},
  title        = {Unbox the black-box for the medical explainable {AI} via multi-modal
                  and multi-centre data fusion: {A} mini-review, two showcases and beyond},
  journal      = {Inf. Fusion},
  volume       = {77},
  pages        = {29--52},
  year         = {2022}
}

@inproceedings{DBLP:conf/aies/KasirzadehC21,
  author       = {Atoosa Kasirzadeh and
                  Damian Clifford},
  editor       = {Marion Fourcade and
                  Benjamin Kuipers and
                  Seth Lazar and
                  Deirdre K. Mulligan},
  title        = {Fairness and Data Protection Impact Assessments},
  booktitle    = {{AIES} '21: {AAAI/ACM} Conference on AI, Ethics, and Society, Virtual
                  Event, USA, May 19-21, 2021},
  pages        = {146--153},
  publisher    = {{ACM}},
  year         = {2021},
 address =    "{New York, NY, USA}"

}

@inproceedings{DBLP:conf/ACISicis/ThaiphanP18,
  author       = {Rattayagon Thaiphan and
                  Thimaporn Phetkaew},
  title        = {Comparative Analysis of Discretization Algorithms on Decision Tree},
  booktitle    = {{ICIS}},
  pages        = {63--67},
  publisher    = {{IEEE} Computer Society},
  year         = {2018},
  address = {{Washington, DC, USA}}
}

@book{spelke2022babies,
  title={What babies know: Core knowledge and composition volume 1},
  author={Spelke, Elizabeth S},
  volume={1},
  year={2022},
  publisher={Oxford University Press},
  address = {Oxford, UK}
}

@article{korenius2007cosine,
  author       = {Tuomo Korenius and
                  Jorma Laurikkala and
                  Martti Juhola},
  title        = {On principal component analysis, cosine and Euclidean measures in
                  information retrieval},
  journal      = {Inf. Sci.},
  volume       = {177},
  number       = {22},
  pages        = {4893--4905},
  year         = {2007}
}

@article{guidotti2019survey,
  author       = {Riccardo Guidotti and
                  Anna Monreale and
                  Salvatore Ruggieri and
                  Franco Turini and
                  Fosca Giannotti and
                  Dino Pedreschi},
  title        = {A Survey of Methods for Explaining Black Box Models},
  journal      = {{ACM} Comput. Surv.},
  volume       = {51},
  number       = {5},
  pages        = {93:1--93:42},
  year         = {2019}
}

@article{chui2022state,
  title={The state of AI in 2022-and a half decade in review},
  author={Chui, Michael and Hall, Bryce and Mayhew, Hellen and Singla, Alex and Sukharevsky, Alex and by McKinsey, AI},
  journal={Mc Kinsey},
  year={2022},
  publisher={McKinsey \& Company}
}

@article{bodria2023benchmarking,
  author       = {Francesco Bodria and
                  Fosca Giannotti and
                  Riccardo Guidotti and
                  Francesca Naretto and
                  Dino Pedreschi and
                  Salvatore Rinzivillo},
  title        = {Benchmarking and survey of explanation methods for black box models},
  journal      = {Data Min. Knowl. Discov.},
  volume       = {37},
  number       = {5},
  pages        = {1719--1778},
  year         = {2023}
}

@article{healthcase2006,
  author       = {Isabelle Bichindaritz and
                  Cindy Marling},
  title        = {Case-based reasoning in the health sciences: What's next?},
  journal      = {Artif. Intell. Medicine},
  volume       = {36},
  number       = {2},
  pages        = {127--135},
  year         = {2006}
}

@article{hong2023protorynet,
  title={ProtoryNet-Interpretable Text Classification Via Prototype Trajectories},
  author={Hong, Dat and Wang, Tong and Baek, Stephen},
  journal={Journal of Machine Learning Research},
  volume={24},
  number={264},
  pages={1--39},
  year={2023}
}

@inproceedings{DBLP:conf/dis/FedeleGP24,
  author       = {Andrea Fedele and
                  Riccardo Guidotti and
                  Dino Pedreschi},
  title        = {This Sounds Like That: Explainable Audio Classification via Prototypical
                  Parts},
  booktitle    = {{DS} {(2)}},
  series       = {Lecture Notes in Computer Science},
  volume       = {15244},
  pages        = {348--363},
  publisher    = {Springer},
  year         = {2024},
  address = {New York, NY, USA},
}

@inproceedings{DBLP:conf/kdd/MingXQR19,
  author       = {Yao Ming and
                  Panpan Xu and
                  Huamin Qu and
                  Liu Ren},
  title        = {Interpretable and Steerable Sequence Learning via Prototypes},
  booktitle    = {{KDD}},
  pages        = {903--913},
  publisher    = {{ACM}},
  year         = {2019},
  address =    "{New York, NY, USA}"
 
}

@article{li2022data,
  title={A data-driven explainable case-based reasoning approach for financial risk detection},
  author={Li, Wei and Paraschiv, Florentina and Sermpinis, Georgios},
  journal={Quantitative Finance},
  volume={22},
  number={12},
  pages={2257--2274},
  year={2022},
  publisher={Taylor \& Francis}
}

@inproceedings{adebayo2018sanity,
  author       = {Julius Adebayo and
                  Justin Gilmer and
                  Michael Muelly and
                  Ian J. Goodfellow and
                  Moritz Hardt and
                  Been Kim},
  title        = {Sanity Checks for Saliency Maps},
  booktitle    = {NeurIPS},
  pages        = {9525--9536},
  year         = {2018}
}

@inproceedings{jeyakumar2020howcan,
  author       = {Jeya Vikranth Jeyakumar and
                  Joseph Noor and
                  Yu{-}Hsi Cheng and
                  Luis Garcia and
                  Mani B. Srivastava},
  title        = {How Can {I} Explain This to You? An Empirical Study of Deep Neural
                  Network Explanation Methods},
  booktitle    = {NeurIPS},
  year         = {2020}
}

@inproceedings{nguyen2021effectiveness,
  author       = {Giang Nguyen and
                  Daeyoung Kim and
                  Anh Nguyen},
  title        = {The effectiveness of feature attribution methods and its correlation
                  with automatic evaluation scores},
  booktitle    = {NeurIPS},
  pages        = {26422--26436},
  year         = {2021}
}

@inproceedings{DBLP:conf/kdd/KairgeldinC24,
  author       = {Rasul Kairgeldin and
                  Miguel {\'{A}}. Carreira{-}Perpi{\~{n}}{\'{a}}n},
  title        = {Bivariate Decision Trees: Smaller, Interpretable, More Accurate},
  booktitle    = {{KDD}},
  pages        = {1336--1347},
  publisher    = {{ACM}},
  year         = {2024},
  address = {New York, NY, USA},
}

@article{breiman2001randomforest,
  author       = {Leo Breiman},
  title        = {Random Forests},
  journal      = {Mach. Learn.},
  volume       = {45},
  number       = {1},
  pages        = {5--32},
  year         = {2001},
}

@inproceedings{alkhouryW2024splitting,
  author       = {Fouad Alkhoury and
                  Pascal Welke},
  title        = {Splitting Stump Forests: Tree Ensemble Compression for Edge Devices},
  booktitle    = {{DS} {(2)}},
  series       = {Lecture Notes in Computer Science},
  volume       = {15244},
  pages        = {3--18},
  publisher    = {Springer},
  year         = {2024},
  editor       = {Springer},
  address = {New York, NY, USA},
}

@article{wickramarachchi2016hhcart,
  author       = {Darshana Chitraka Wickramarachchi and
                  Blair Lennon Robertson and
                  Marco Reale and
                  Christopher John Price and
                  Jennifer Brown},
  title        = {{HHCART:} An oblique decision tree},
  journal      = {Comput. Stat. Data Anal.},
  volume       = {96},
  pages        = {12--23},
  year         = {2016}
}

@article{murthy1994system,
  author       = {Sreerama K. Murthy and
                  Simon Kasif and
                  Steven Salzberg},
  title        = {A System for Induction of Oblique Decision Trees},
  journal      = {J. Artif. Intell. Res.},
  volume       = {2},
  pages        = {1--32},
  year         = {1994}
}

@article{tan2025proximity,
  author       = {Chang Wei Tan and
                  Matthieu Herrmann and
                  Mahsa Salehi and
                  Geoffrey I. Webb},
  title        = {Proximity forest 2.0: a new effective and scalable similarity-based
                  classifier for time series},
  journal      = {Data Min. Knowl. Discov.},
  volume       = {39},
  number       = {2},
  pages        = {14},
  year         = {2025}
}

@inproceedings{zhang2021proximity,
  author       = {Yue Zhang and
                  Zhihai Wang and
                  Jidong Yuan},
  title        = {{A Proximity Forest for Multivariate Time Series Classification}},
  booktitle    = {{PAKDD} {(1)}},
  series       = {Lecture Notes in Computer Science},
  volume       = {12712},
  pages        = {766--778},
  publisher    = {Springer},
  year         = {2021},
  address = {New York, NY, USA},
}

@inproceedings{cascione2024oral,
  author       = {Alessio Cascione and
                  Mattia Setzu and
                  Federico A. Galatolo and
                  Mario G. C. A. Cimino and
                  Riccardo Guidotti},
  title        = {Interpretable Machine Learning for Oral Lesion Diagnosis Through Prototypical
                  Instances Identification},
  booktitle    = {{DS} {(2)}},
  series       = {Lecture Notes in Computer Science},
  volume       = {15244},
  pages        = {316--331},
  publisher    = {Springer},
  year         = {2024},
  address =     "{New York, NY, USA}"
}

@inproceedings{cascione2024pivottree,
  author       = {Alessio Cascione and
                  Mattia Setzu and
                  Riccardo Guidotti},
  title        = {Data-Agnostic Pivotal Instances Selection for Decision-Making Models},
  booktitle    = {{ECML/PKDD} {(1)}},
  series       = {Lecture Notes in Computer Science},
  volume       = {14941},
  pages        = {367--386},
  publisher    = {Springer},
  year         = {2024},
  address = {{New York, NY, USA}}
}

@article{DBLP:journals/siamrev/McCaffrey92,
  author       = {Daniel F. McCaffrey},
  title        = {{Generalized Additive Models {(T.} J. Hastie and R. J. Tibshirani)}},
  journal      = {{SIAM} Rev.},
  volume       = {34},
  number       = {4},
  pages        = {675--678},
  year         = {1992}
}

@article{pugnana2024benchmark,
  author       = {Andrea Pugnana and
                  Lorenzo Perini and
                  Jesse Davis and
                  Salvatore Ruggieri},
  title        = {Deep Neural Network Benchmarks for Selective Classification},
  journal      = {CoRR},
  volume       = {abs/2401.12708},
  year         = {2024}
}

@article{DBLP:journals/corr/abs-2410-04723,
  author       = {Guangzhi Xiong and
                  Sanchit Sinha and
                  Aidong Zhang},
  title        = {ProtoNAM: Prototypical Neural Additive Models for Interpretable Deep
                  Tabular Learning},
  journal      = {CoRR},
  volume       = {abs/2410.04723},
  year         = {2024}
}

@inproceedings{DBLP:conf/acl/0001GKLL22,
  author       = {Anubrata Das and
                  Chitrank Gupta and
                  Venelin Kovatchev and
                  Matthew Lease and
                  Junyi Jessy Li},
  title        = {ProtoTEx: Explaining Model Decisions with Prototype Tensors},
  booktitle    = {{ACL} {(1)}},
  pages        = {2986--2997},
  publisher    = {Association for Computational Linguistics},
  year         = {2022},
  address = {{Stroudsburg, PA, USA}}
}

@article{DBLP:journals/jmlr/HongWB23,
  author       = {Dat Hong and
                  Tong Wang and
                  Stephen Baek},
  title        = {ProtoryNet - Interpretable Text Classification Via Prototype Trajectories},
  journal      = {J. Mach. Learn. Res.},
  volume       = {24},
  pages        = {264:1--264:39},
  year         = {2023},
  address = {{Washington, DC, USA}}
}

@article{DBLP:journals/telo/FilhoLP22,
  author       = {Renato Miranda Filho and
                  An{\'{\i}}sio M. Lacerda and
                  Gisele L. Pappa},
  title        = {Explainable Regression Via Prototypes},
  journal      = {{ACM} Trans. Evol. Learn. Optim.},
  volume       = {2},
  number       = {4},
  pages        = {14:1--14:26},
  year         = {2022}
}

@article{DBLP:journals/datamine/TanHSW25,
  author       = {Chang Wei Tan and
                  Matthieu Herrmann and
                  Mahsa Salehi and
                  Geoffrey I. Webb},
  title        = {Proximity forest 2.0: a new effective and scalable similarity-based
                  classifier for time series},
  journal      = {Data Min. Knowl. Discov.},
  volume       = {39},
  number       = {2},
  pages        = {14},
  year         = {2025}
}

@article{DBLP:journals/datamine/KarlssonPB16,
  author       = {Isak Karlsson and
                  Panagiotis Papapetrou and
                  Henrik Bostr{\"{o}}m},
  title        = {Generalized random shapelet forests},
  journal      = {Data Min. Knowl. Discov.},
  volume       = {30},
  number       = {5},
  pages        = {1053--1085},
  year         = {2016}
}

@inproceedings{DBLP:conf/kdd/SatheA17,
  author       = {Saket Sathe and
                  Charu C. Aggarwal},
  title        = {Similarity Forests},
  booktitle    = {{KDD}},
  pages        = {395--403},
  publisher    = {{ACM}},
  year         = {2017},
  address = {New York, NY, USA},
}

@inproceedings{DBLP:conf/icml/HaghiriGL18,
  author       = {Siavash Haghiri and
                  Damien Garreau and
                  Ulrike von Luxburg},
  title        = {{Comparison-Based Random Forests}},
  booktitle    = {{ICML}},
  series       = {Proceedings of Machine Learning Research},
  volume       = {80},
  pages        = {1866--1875},
  publisher    = {{PMLR}},
  year         = {2018},
  address = {Cambridge, MA, USA}
}

@inproceedings{DBLP:conf/pakdd/ShiWYL18,
  author       = {Mohan Shi and
                  Zhihai Wang and
                  Jidong Yuan and
                  Haiyang Liu},
  title        = {Random Pairwise Shapelets Forest},
  booktitle    = {{PAKDD} {(1)}},
  series       = {Lecture Notes in Computer Science},
  volume       = {10937},
  pages        = {68--80},
  publisher    = {Springer},
  year         = {2018},
  address = {New York, NY, USA},
}

@article{lusito2024diva,
  author       = {Salvatore Lusito and
                  Andrea Pugnana and
                  Riccardo Guidotti},
  title        = {Solving imbalanced learning with outlier detection and features reduction},
  journal      = {Mach. Learn.},
  volume       = {113},
  number       = {8},
  pages        = {5273--5330},
  year         = {2024}
}

@inproceedings{bonsignori2021deriving,
  title={Deriving a single interpretable model by merging tree-based classifiers},
  author={Bonsignori, Valerio and Guidotti, Riccardo and Monreale, Anna},
  booktitle={International Conference on Discovery Science},
  pages={347--357},
  year={2021},
  organization={Springer}
}

@article{pensa2025explaining,
  title={Explaining Random Forest and XGBoost with Shallow Decision Trees by Co-clustering Feature Importance},
  author={Pensa, Ruggero G and Crombach, Anton and Peignier, Sergio and Rigotti, Christophe},
  journal={Machine Learning},
  volume={114},
  number={12},
  pages={287},
  year={2025},
  publisher={Springer}
}

@article{baniecki2025birds,
  title={Birds look like cars: adversarial analysis of intrinsically interpretable deep learning},
  author={Baniecki, Hubert and Biecek, Przemyslaw},
  journal={Machine Learning},
  volume={114},
  number={12},
  pages={284},
  year={2025},
  publisher={Springer}
}

@incollection{furnkranz1994incremental,
title = {Incremental Reduced Error Pruning},
editor = {William W. Cohen and Haym Hirsh},
booktitle = {Machine Learning Proceedings},
publisher = {Morgan Kaufmann},
address = {San Francisco (CA)},
pages = {70-77},
year = {1994},
author = {Johannes F{\"{u}}rnkranz and
                  Gerhard Widmer},
}

@inproceedings{cohen1995fast,
  title={Fast effective rule induction},
  author={Cohen, William W and others},
  booktitle={{ICML}},
  pages={115--123},
  year={1995}
}

@inproceedings{prokhorenkova2018catboost,
author = {Prokhorenkova, Liudmila and Gusev, Gleb and Vorobev, Aleksandr and Dorogush, Anna Veronika and Gulin, Andrey},
title = {CatBoost: unbiased boosting with categorical features},
year = {2018},
publisher = {Curran Associates Inc.},
address = {Red Hook, NY, USA},
booktitle = {Proceedings of the 32nd International Conference on Neural Information Processing Systems},
pages = {6639–6649},
numpages = {11},
location = {Montr\'{e}al, Canada},
series = {NIPS'18}
}

@article{gohiya2018survey,
  title={A Survey of Xgboost system},
  author={Gohiya, H and Lohiya, H and Patidar, K},
  journal={Int. J. Adv. Technol. Eng. Res},
  volume={8},
  pages={25--30},
  year={2018}
}

@article{ke2017lgbm,
  title={Lightgbm: A highly efficient gradient boosting decision tree},
  author={Ke, Guolin and Meng, Qi and Finley, Thomas and Wang, Taifeng and Chen, Wei and Ma, Weidong and Ye, Qiwei and Liu, Tie-Yan},
  journal={Advances in neural information processing systems},
  volume={30},
  year={2017}
}

@article{miller1956magical,
  title={The magical number seven, plus or minus two: Some limits on our capacity for processing information.},
  author={Miller, George A},
  journal={Psychological review},
  volume={63},
  number={2},
  pages={81},
  year={1956},
  publisher={American Psychological Association}
}

@article{kim2016examples,
  title={Examples are not enough, learn to criticize! criticism for interpretability},
  author={Kim, Been and Khanna, Rajiv and Koyejo, Oluwasanmi O},
  journal={Advances in neural information processing systems},
  volume={29},
  year={2016}
}

\newpage

\appendix

\section{Appendix}
\label{sec:appendix}
%\noindent
In this appendix, we present a more fine-grained analysis of the performance of all approaches discussed in the main paper.
For completeness, we also include results in terms of Balanced Accuracy.
Overall, the results for weighted F1-score and Balanced Accuracy are consistent, and the observations and comments reported in the main text similarly apply to this complementary evaluation metric.

Moreover, we report results for each individual trained model in the \textit{unconstrained setting}, i.e., without any limitation on the number of extracted pivots (the tables in the main paper consider at most 20 pivots).

Finally,  we provide detailed performance results, including weighted F1-score, Balanced Accuracy and the number of pivots or estimators for each dataset individually, for both standalone and ensemble models.

\begin{table}[t]
    \caption{Average Balanced Accuracy $\pm$ std. dev. for  \textsc{PivotTree} classifiers and selectors combined with \textsc{dt} (limited to at most 20 pivots), baselines and competitors. Subscripts indicate the average number of pivots $\pm$ std. dev.
    Best results in \textbf{bold}, second best in \textit{italics}.}
    \centering
    \setlength{\tabcolsep}{2.8mm}
    \label{ref:tab_dt_model_max20_balanc_acc}
    \begin{tabular}{cccccc}
    \toprule
{model} & $\texttt{tabular}$ & $\texttt{images}$ & $\texttt{time-series}$ & $\texttt{text}$ & \texttt{all} \\
\midrule
$\textsc{grule}_{X}$ & $.62_{} \pm .28_{}$ & $.42_{} \pm .39_{}$ & $.43_{} \pm .24_{}$ & $.50_{} \pm .25_{}$ & $.52_{} \pm .30_{}$ \\
$\textsc{irep}_{X}$ & $.70_{} \pm .17_{}$ & $\textit{.88}_{} \pm .10_{}$ & $.52_{} \pm .19_{}$ & $.52_{} \pm .21_{}$ & $\textit{.67}_{} \pm .21_{}$ \\
$\textsc{ripper}_{X}$ & $.72_{} \pm .19_{}$ & $\textbf{.90}_{} \pm .09_{}$ & $.56_{} \pm .20_{}$ & $.54_{} \pm .21_{}$ & $\textbf{.70}_{} \pm .22_{}$ \\
\midrule
$\textsc{dt}_{X}$ & $\textit{.74}_{} \pm .18_{}$ & $.67_{} \pm .19_{}$ & $\textit{.59}_{} \pm .21_{}$ & $.57_{} \pm .22_{}$ & $\textit{.67}_{} \pm .20_{}$ \\
$\textsc{odt}_{X}$ & $\textbf{.75}_{} \pm .19_{}$ & $.67_{} \pm .20_{}$ & $\textit{.59}_{} \pm .22_{}$ & $.57_{} \pm .21_{}$ & $\textit{.67}_{} \pm .21_{}$ \\
\midrule
$\textsc{kms}_{Z}$ & $.67_{13} \pm .22_{6}$ & $.66_{10} \pm .20_{5}$ & $.56_{12} \pm .21_{6}$ & $.55_{13} \pm .20_{4}$ & $.63_{12} \pm .21_{6}$ \\
$\textsc{kmd}_{Z}$ & $.70_{14} \pm .19_{4}$ & $.67_{14} \pm .20_{6}$ & $.58_{11} \pm .22_{6}$ & $.56_{15} \pm .20_{3}$ & $.65_{14} \pm .20_{5}$ \\
$\textsc{ebl}_{Z}$ & $.73_{18} \pm .19_{5}$ & $.69_{15} \pm .21_{6}$ & $.57_{17} \pm .20_{6}$ & $\textit{.59}_{19} \pm .22_{2}$ & $\textit{.67}_{17} \pm .21_{5}$ \\
\midrule
$\textsc{ptc}_{Z}$ & $.72_{\textbf{10}} \pm .20_{4}$ & $.67_{\textbf{6}} \pm .20_{4}$ & $\textit{.59}_{\textbf{8}} \pm .21_{4}$ & $\textbf{.60}_{\textbf{10}} \pm .21_{4}$ & $\textit{.67}_{\textbf{9}} \pm .21_{4}$ \\
$\textsc{pptc}_{Z}$ & $.72_{12} \pm .18_{6}$ & $.70_{{10}} \pm .17_{6}$ & $.56_{12} \pm .20_{6}$ & $\textit{.59}_{\textit{11}} \pm .20_{4}$ & $.66_{12} \pm .20_{6}$ \\
$\textsc{optc}_{Z}$ & $.68_{15} \pm .20_{7}$ & $.65_{\textit{9}} \pm .20_{7}$ & $.57_{11} \pm .21_{4}$ & $.54_{14} \pm .19_{5}$ & $.63_{13} \pm .20_{7}$ \\
$\textsc{opptc}_{Z}$ & $.73_{12} \pm .18_{5}$ & $.71_{11} \pm .19_{6}$ & $.54_{\textit{10}} \pm .20_{6}$ & $.55_{13} \pm .20_{6}$ & $.66_{\textit{11}} \pm .20_{6}$ \\
\midrule
$\textsc{pts}_{Z}$ & $.72_{\textit{11}} \pm .18_{5}$ & $.67_{{10}} \pm .21_{5}$ & $.58_{\textbf{8}} \pm .21_{4}$ & $\textit{.59}_{14} \pm .21_{3}$ & $.66_{\textit{11}} \pm .20_{5}$ \\
$\textsc{ppts}_{Z}$ & $.72_{14} \pm .20_{6}$ & $.67_{11} \pm .20_{7}$ & $\textit{.59}_{15} \pm .21_{6}$ & $\textbf{.60}_{15} \pm .21_{8}$ & $.66_{14} \pm .20_{6}$ \\
$\textsc{opts}_{Z}$ & $.69_{13} \pm .19_{7}$ & $.65_{13} \pm .20_{5}$ & $.57_{\textit{10}} \pm .20_{5}$ & $.54_{{12}} \pm .20_{6}$ & $.64_{{12}} \pm .20_{6}$ \\
$\textsc{oppts}_{Z}$ & $.69_{14} \pm .21_{7}$ & $.64_{11} \pm .21_{7}$ & $.56_{{11}} \pm .20_{6}$ & $.54_{\textit{11}} \pm .20_{6}$ & $.63_{13} \pm .21_{6}$ \\
\bottomrule
\end{tabular}
\end{table}

\begin{table}[t]
    \caption{Average Balanced Accuracy $\pm$ std. dev. for \textsc{PivotTree} selectors combined with \textsc{knn} (limited to at most 20 pivots), baselines and competitors. 
    Subscripts indicate the average number of pivots $\pm$ std. dev.
    Best results in \textbf{bold}, second best in \textit{italics}.}
    \centering
    \setlength{\tabcolsep}{2.8mm}
    \label{ref:tab_knn_model_max20_balanc_acc}
    \begin{tabular}{cccccc}
    \toprule
    {model} & $\texttt{tabular}$ & $\texttt{images}$ & $\texttt{time-series}$ & $\texttt{text}$ & \texttt{all} \\
\midrule    
$\textsc{knn}_{X}$ &
$\textbf{.78}_{} \pm .17_{}$ &
$\textbf{.93}_{} \pm .06_{}$ &
$\textbf{.63}_{} \pm .27_{}$ &
$.61_{} \pm .22_{}$ &
$\textbf{.76}_{} \pm .21_{}$ \\
\midrule

$\textsc{kms}_{Z}$ &
$.76_{14} \pm .17_{6}$ &
$\textit{.92}_{14} \pm .07_{5}$ &
$\textit{.62}_{15} \pm .23_{5}$ &
$.58_{15} \pm .22_{3}$ &
$.74_{14} \pm .21_{5}$ \\

$\textsc{kmd}_{Z}$ &
$\textit{.77}_{16} \pm .17_{5}$ &
$\textit{.92}_{14} \pm .06_{6}$ &
$\textit{.62}_{\textbf{11}} \pm .22_{5}$ &
$.59_{19} \pm .22_{3}$ &
$.74_{15} \pm .21_{6}$ \\

$\textsc{ebl}_{Z}$ &
$\textbf{.78}_{19} \pm .17_{4}$ &
$\textit{.92}_{18} \pm .07_{4}$ &
$\textit{.62}_{17} \pm .23_{6}$ &
$\textit{.62}_{20} \pm .23_{2}$ &
$\textit{.75}_{19} \pm .20_{5}$ \\
\midrule

$\textsc{pts}_{Z}$ &
$\textit{.77}_{\textit{12}} \pm .17_{5}$ &
$\textit{.92}_{14} \pm .06_{5}$ &
$\textit{.62}_{\textbf{11}} \pm .24_{6}$ &
$\textit{.62}_{\textit{12}} \pm .24_{2}$ &
$\textit{.75}_{\textbf{12}} \pm .21_{5}$ \\

$\textsc{ppts}_{Z}$ &
$\textit{.77}_{13} \pm .18_{6}$ &
$\textit{.92}_{14} \pm .07_{8}$ &
$\textbf{.63}_{15} \pm .24_{5}$ &
$.61_{16} \pm .22_{7}$ &
$.74_{14} \pm .21_{6}$ \\

$\textsc{opts}_{Z}$ &
$.76_{14} \pm .18_{5}$ &
$\textit{.92}_{15} \pm .07_{6}$ &
$.61_{13} \pm .24_{5}$ &
$.60_{17} \pm .22_{8}$ &
$.74_{14} \pm .21_{6}$ \\

$\textsc{oppts}_{Z}$ &
$.76_{14} \pm .18_{5}$ &
$.91_{13} \pm .08_{6}$ &
$.61_{\textit{12}} \pm .23_{7}$ &
$.58_{14} \pm .22_{4}$ &
$.73_{\textit{13}} \pm .21_{5}$ \\
\midrule

$\textsc{kms}_{P}$ &
$.68_{16} \pm .20_{4}$ &
$.91_{15} \pm .09_{5}$ &
$.54_{14} \pm .23_{5}$ &
$.52_{19} \pm .20_{2}$ &
$.67_{16} \pm .23_{5}$ \\

$\textsc{kmd}_{P}$ &
$.68_{16} \pm .20_{5}$ &
$.90_{13} \pm .11_{7}$ &
$.57_{17} \pm .23_{5}$ &
$.51_{13} \pm .21_{7}$ &
$.68_{15} \pm .23_{6}$ \\

$\textsc{ebl}_{P}$ &
$.74_{20} \pm .18_{3}$ &
$\textit{.92}_{18} \pm .07_{5}$ &
$.59_{20} \pm .24_{3}$ &
$.58_{20} \pm .23_{0}$ &
$.72_{19} \pm .22_{3}$ \\
\midrule

$\textsc{pts}_{P}$ &
$.67_{\textit{12}} \pm .18_{4}$ &
$.78_{\textit{11}} \pm .22_{6}$ &
$.54_{14} \pm .22_{4}$ &
$.54_{13} \pm .20_{6}$ &
$.65_{\textbf{12}} \pm .22_{5}$ \\

$\textsc{ppts}_{P}$ &
$.69_{\textbf{11}} \pm .20_{7}$ &
$.70_{14} \pm .25_{8}$ &
$.55_{13} \pm .26_{6}$ &
$.55_{14} \pm .21_{8}$ &
$.64_{\textbf{12}} \pm .23_{7}$ \\

$\textsc{opts}_{P}$ &
$.69_{13} \pm .17_{7}$ &
$.78_{15} \pm .21_{6}$ &
$.52_{14} \pm .23_{6}$ &
$.53_{15} \pm .22_{9}$ &
$.65_{14} \pm .22_{7}$ \\

$\textsc{oppts}_{P}$ &
$.71_{13} \pm .20_{9}$ &
$.66_{\textbf{10}} \pm .28_{6}$ &
$.55_{13} \pm .23_{6}$ &
$.52_{\textbf{10}} \pm .21_{6}$ &
$.64_{\textbf{12}} \pm .23_{7}$ \\
\bottomrule
\end{tabular}
\end{table}

\begin{figure}[t]
    \includegraphics[width=0.5\linewidth]{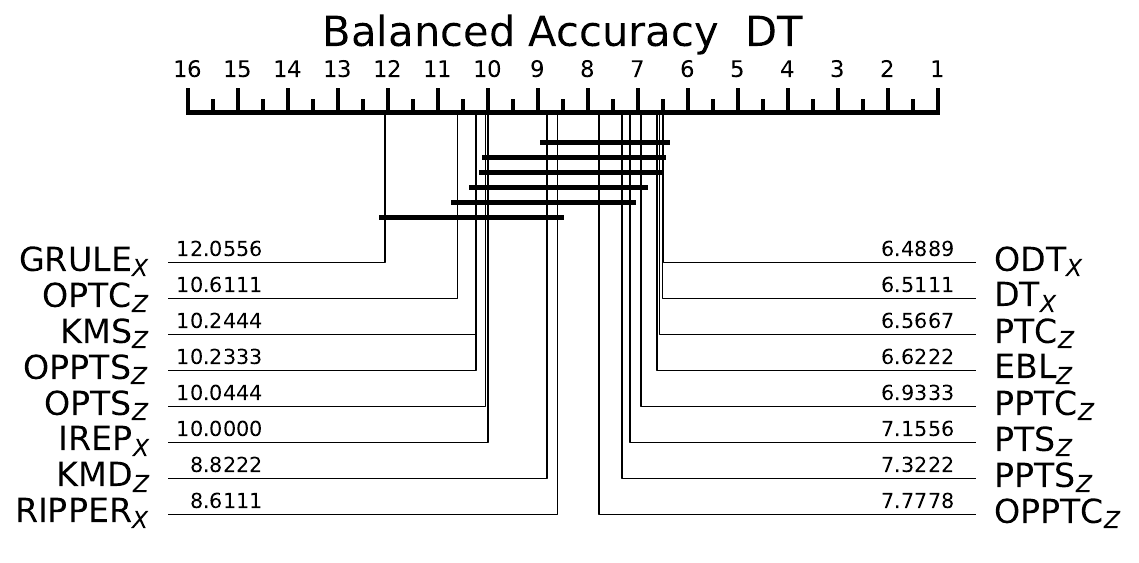}
    \includegraphics[width=0.5\linewidth]{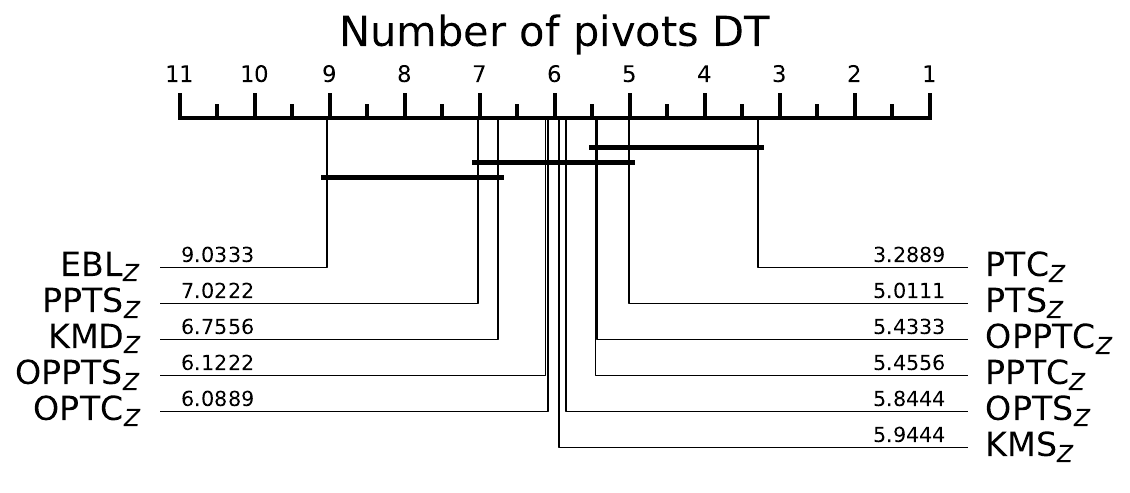}
    \includegraphics[width=0.5\linewidth]{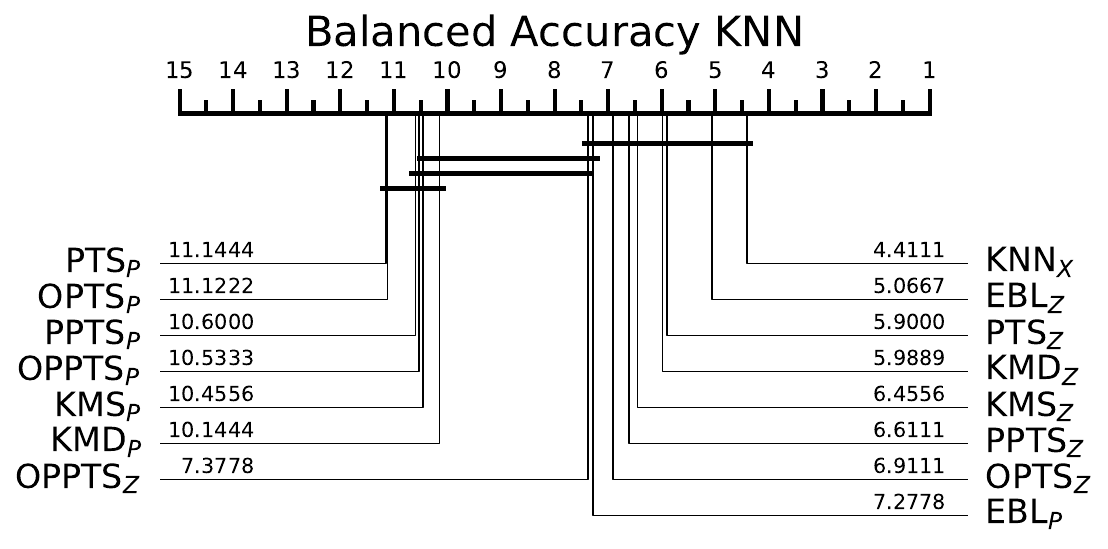}
    \includegraphics[width=0.5\linewidth]{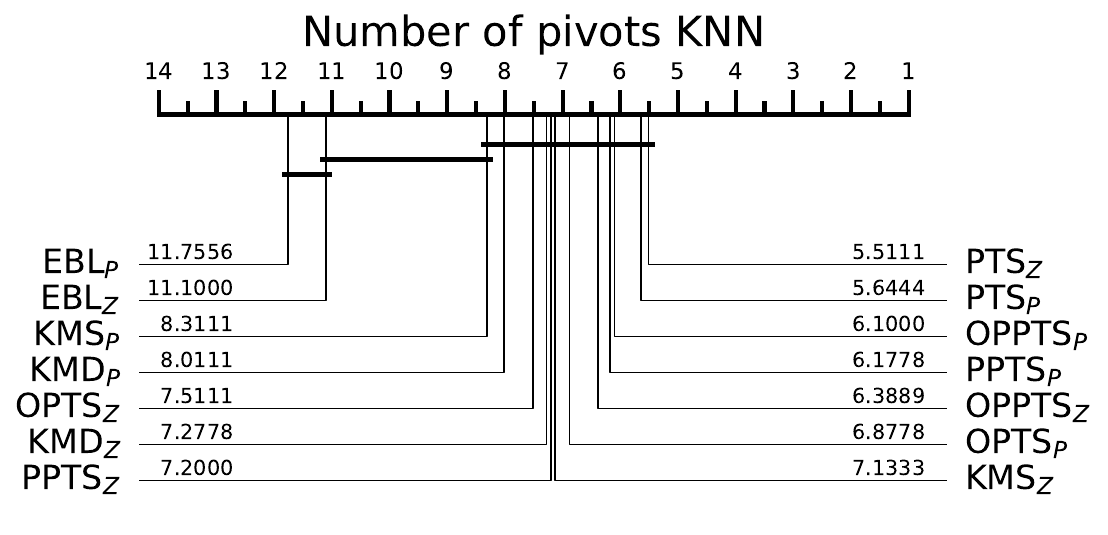}
    \caption{Critical difference plot of model's rank in terms of Balanced Accuracy and number of pivots against each other with Nemenyi test w.r.t all datasets, considering at most 20 pivots for each configuration. Models that are not significantly different at $95$\% significance level are connected. Best models on the right.}
    \label{fig:cd_plots_bal_accuracy}
\end{figure}

\begin{table}[h]
    \caption{Average weighted F1-score $\pm$ std. dev. for  \textsc{PivotTree} classifiers and selectors combined with \textsc{dt} in the unconstrained setting, baselines and competitors. Subscripts indicate the average number of pivots $\pm$ std. dev.
    Best results in \textbf{bold}, second best in \textit{italics}.}
    \centering
    \footnotesize
    \setlength{\tabcolsep}{2.8mm}
    \label{tab:model_comparison_unconstrained_dt_f1_score}
\begin{tabular}{cccccc}
    \toprule
    model & $\texttt{tabular}$ & $\texttt{images}$ & $\texttt{time-series}$ & $\texttt{text}$ & \texttt{all} \\
\midrule
$\textsc{grule}_{X}$ & $.65_{} \pm .29_{}$ & $.38_{} \pm .42_{}$ & $.40_{} \pm .32_{}$ & $.45_{} \pm .32_{}$ & $.51_{} \pm .34_{}$ \\

$\textsc{irep}_{X}$ & $.76_{} \pm .14_{}$ & $\textit{.90}_{} \pm .09_{}$ & $.55_{} \pm .27_{}$ & $.52_{} \pm .25_{}$ & $\textit{.71}_{} \pm .23_{}$ \\

$\textsc{ripper}_{X}$ & $.76_{} \pm .15_{}$ & $\textbf{.91}_{} \pm .08_{}$ & $.61_{} \pm .25_{}$ & $.52_{} \pm .25_{}$ & $.72_{} \pm .22_{}$ \\
\midrule
$\textsc{dt}_{X}$ & $\textbf{.81}_{} \pm .14_{}$ & $.70_{} \pm .19_{}$ & $\textbf{.64}_{} \pm .26_{}$ & $.58_{} \pm .22_{}$ & $\textbf{.72}_{} \pm .21_{}$ \\

$\textsc{odt}_{X}$ & $\textbf{.81}_{} \pm .13_{}$ & $.70_{} \pm .19_{}$ & $\textbf{.64}_{} \pm .27_{}$ & $.58_{} \pm .22_{}$ & $\textbf{.72}_{} \pm .21_{}$ \\
\midrule
$\textsc{kms}_{Z}$ & $.76_{30} \pm .15_{16}$ & $.69_{19} \pm .20_{17}$ & $.61_{26} \pm .26_{17}$ & $.56_{41} \pm .19_{8}$ & $.68_{29} \pm .20_{17}$ \\

$\textsc{kmd}_{Z}$ & $.77_{33} \pm .15_{16}$ & $.70_{26} \pm .20_{16}$ & $\textit{.63}_{28} \pm .27_{15}$ & $.56_{33} \pm .20_{15}$ & $.70_{31} \pm .20_{15}$ \\
\bottomrule
\end{tabular}
\end{table}

\begin{table}[h]
    \caption{Average weighted F1-score $\pm$ std. dev. for \textsc{PivotTree} classifiers and selectors combined with \textsc{knn} in the unconstrained setting, baselines and competitors. Subscripts indicate the average number of pivots $\pm$ std. dev.
    Best results in \textbf{bold}, second best in \textit{italics}.}
    \centering
    \footnotesize
    \setlength{\tabcolsep}{2.8mm}
    \label{tab:model_comparison_clean_knn_unc}
    \begin{tabular}{cccccc}
\toprule
model & $\texttt{tabular}$ & $\texttt{images}$ & $\texttt{time-series}$ & $\texttt{text}$ & \texttt{all} \\
\midrule
$\textsc{knn}_{X}$ & $\textbf{.83}_{} \pm .13_{}$ & $\textbf{.94}_{} \pm .05_{}$ & $\textit{.68}_{} \pm .27_{}$ & $.61_{} \pm .22_{}$ & $\textbf{.79}_{} \pm .20_{}$ \\
\midrule

$\textsc{kms}_{Z}$ & $.80_{13} \pm .14_{6}$ & $\textit{.93}_{13} \pm .06_{5}$ & $\textit{.68}_{14} \pm .24_{6}$ & $.58_{15} \pm .22_{4}$ & $.77_{14} \pm .20_{5}$ \\

$\textsc{kmd}_{Z}$ & $.81_{15} \pm .14_{5}$ & $\textit{.93}_{14} \pm .06_{6}$ & $\textit{.68}_{\textbf{12}} \pm .24_{5}$ & $.60_{19} \pm .22_{3}$ & $\textit{.78}_{15} \pm .19_{5}$ \\

$\textsc{ebl}_{Z}$ & $\textbf{.83}_{207} \pm .13_{281}$ & $\textit{.93}_{53} \pm .05_{57}$ & $\textit{.68}_{322} \pm .24_{638}$ & $\textbf{.63}_{150} \pm .21_{178}$ & 
$\textbf{.79}_{194} \pm .19_{360}$\\

\midrule

$\textsc{pts}_{Z}$ & $\textit{.82}_{\textit{12}} \pm .14_{5}$ & $\textit{.93}_{15} \pm .05_{3}$ & $\textbf{.69}_{\textbf{12}} \pm .25_{7}$ & $\textit{.62}_{\textit{13}} \pm .23_{4}$ & $\textbf{.79}_{\textit{13}} \pm .19_{5}$ \\

$\textsc{ppts}_{Z}$ & $\textit{.82}_{\textit{12}} \pm .14_{6}$ & $\textit{.93}_{16} \pm .05_{7}$ & $\textbf{.69}_{15} \pm .25_{5}$ & $.61_{16} \pm .22_{7}$ & $\textit{.78}_{14} \pm .20_{6}$ \\

$\textsc{opts}_{Z}$ & $.81_{13} \pm .14_{5}$ & $\textit{.93}_{15} \pm .07_{7}$ & $.67_{\textit{13}} \pm .25_{5}$ & $.60_{17} \pm .22_{8}$ & $\textit{.78}_{14} \pm .20_{6}$ \\
\bottomrule
\end{tabular}
\end{table}

\begin{table}[t]
    \caption{Average Balanced Accuracy $\pm$ std. dev. for  \textsc{PivotTree} classifiers and selectors combined with \textsc{dt} in the unconstrained setting, baselines and competitors. Subscripts indicate the average number of pivots $\pm$ std. dev.
    Best results in \textbf{bold}, second best in \textit{italics}.}
    \centering
    \setlength{\tabcolsep}{2.8mm}
    \label{ref:tab_decisiontr_model_unconstrained_balanc_acc}
    \begin{tabular}{cccccc}
    \toprule
    {model} & $\texttt{tabular}$ & $\texttt{images}$ & $\texttt{time-series}$ & $\texttt{text}$ & \texttt{all} \\
\midrule
$\textsc{grule}_{X}$ & $.62_{} \pm .28_{}$ & $.42_{} \pm .39_{}$ & $.43_{} \pm .24_{}$ & $.50_{} \pm .25_{}$ & $.52_{} \pm .30_{}$ \\

$\textsc{irep}_{X}$ & $.70_{} \pm .17_{}$ & $\textit{.88}_{} \pm .10_{}$ & $.52_{} \pm .19_{}$ & $.52_{} \pm .21_{}$ & $\textit{.67}_{} \pm .21_{}$ \\

$\textsc{ripper}_{X}$ & $.72_{} \pm .19_{}$ & $\textbf{.90}_{} \pm .09_{}$ & $.56_{} \pm .20_{}$ & $.54_{} \pm .21_{}$ & $\textbf{.70}_{} \pm .22_{}$ \\
\midrule
$\textsc{dt}_{X}$ & $\textit{.74}_{} \pm .18_{}$ & $.67_{} \pm .19_{}$ & $\textit{.59}_{} \pm .21_{}$ & $.57_{} \pm .22_{}$ & $\textit{.67}_{} \pm .20_{}$ \\

$\textsc{odt}_{X}$ & $\textbf{.75}_{} \pm .19_{}$ & $.67_{} \pm .20_{}$ & $\textit{.59}_{} \pm .22_{}$ & $.57_{} \pm .21_{}$ & $\textit{.67}_{} \pm .21_{}$ \\
\midrule
$\textsc{kms}_{Z}$ & $.70_{26} \pm .20_{16}$ & $.67_{17} \pm .20_{17}$ & $.56_{25} \pm .20_{16}$ & $.56_{39} \pm .20_{11}$ & $.64_{26} \pm .21_{17}$ \\

$\textsc{kmd}_{Z}$ & $.71_{33} \pm .19_{15}$ & $.68_{27} \pm .20_{16}$ & $.58_{31} \pm .21_{17}$ & $.57_{26} \pm .20_{16}$ & $.66_{30} \pm .20_{16}$ \\
\bottomrule
\end{tabular}
\end{table}

\begin{table}[t]
    \caption{Average Balanced Accuracy $\pm$ std. dev. for  \textsc{PivotTree} selectors combined with \textsc{knn} in the unconstrained setting, baselines and competitors. Subscripts indicate the average number of pivots $\pm$ std. dev.
    Best results in \textbf{bold}, second best in \textit{italics}.}
    \centering
    \setlength{\tabcolsep}{2.8mm}
    \label{ref:tab_knn_model_unconstrained_balanc_acc}
    \begin{tabular}{cccccc}
    \toprule
    {model} & $\texttt{tabular}$ & $\texttt{images}$ & $\texttt{time-series}$ & $\texttt{text}$ & \texttt{all} \\
    \midrule

$\textsc{knn}_{X}$ & $\textbf{.78}_{} \pm .17_{}$ & $\textbf{.93}_{} \pm .06_{}$ & $\textbf{.63}_{} \pm .27_{}$ & $.61_{} \pm .22_{}$ & $\textbf{.76}_{} \pm .21_{}$ \\
\midrule
$\textsc{kms}_{Z}$ & $.76_{14} \pm .17_{6}$ & $\textit{.92}_{14} \pm .07_{5}$ & $\textit{.62}_{15} \pm .23_{5}$ & $.58_{15} \pm .22_{3}$ & $.74_{14} \pm .21_{5}$ \\

$\textsc{kmd}_{Z}$ & $\textit{.77}_{16} \pm .17_{5}$ & $\textit{.92}_{14} \pm .06_{6}$ & $\textit{.62}_{\textbf{11}} \pm .22_{5}$ & $.59_{19} \pm .22_{3}$ & $.74_{15} \pm .21_{6}$ \\

$\textsc{ebl}_{Z}$ & $\textbf{.78}_{204} \pm .17_{282}$ & $\textbf{.93}_{50} \pm .06_{58}$ & $\textit{.62}_{354} \pm .23_{628}$ & $\textbf{.63}_{160} \pm .21_{173}$ & $\textbf{.76}_{201} \pm .21_{360}$ \\
\midrule
$\textsc{pts}_{Z}$ & $\textit{.77}_{\textit{12}} \pm .17_{5}$ & $\textit{.92}_{14} \pm .06_{5}$ & $\textit{.62}_{\textbf{11}} \pm .24_{6}$ & $\textit{.62}_{\textit{12}} \pm .24_{2}$ & $\textit{.75}_{\textbf{12}} \pm .21_{5}$ \\

$\textsc{ppts}_{Z}$ & $\textit{.77}_{13} \pm .18_{6}$ & $\textit{.92}_{14} \pm .07_{8}$ & $\textbf{.63}_{15} \pm .24_{5}$ & $.61_{16} \pm .22_{7}$ & $.74_{14} \pm .20_{6}$ \\

$\textsc{opts}_{Z}$ & $.76_{14} \pm .18_{5}$ & $\textit{.92}_{15} \pm .07_{6}$ & $.61_{13} \pm .24_{5}$ & $.60_{17} \pm .22_{8}$ & $.74_{14} \pm .21_{6}$ \\
\bottomrule
\end{tabular}
\end{table}

\begin{sidewaystable}
\caption{Average weighted F1-score for \textsc{PivotTree} classifiers and selectors combined with \textsc{dt}, baselines and competitors. Subscripts indicate the average number of pivots $\pm$ std. dev. Best results in \textbf{bold}, second best in \textit{italics}}
\label{tab:dt_single_dataset_unconstrained}
\centering
\footnotesize
\setlength{\tabcolsep}{1.4mm} % Reduce column 
\begin{tabular}{l|lll|ll|lll|llll|llll}
\toprule
\texttt{dataset} & $\textsc{grule}_{X}$ & $\textsc{irep}_{X}$ & $\textsc{ripper}_{X}$ & $\textsc{dt}_{X}$ & $\textsc{odt}_{X}$ & $\textsc{kms}_{Z}$ & $\textsc{kmd}_{Z}$ & $\textsc{ebl}_{Z}$ & $\textsc{ptc}_{Z}$ & $\textsc{pptc}_{Z}$ & $\textsc{optc}_{Z}$ & $\textsc{opptc}_{Z}$ & $\textsc{pts}_{Z}$ & $\textsc{ppts}_{Z}$ & $\textsc{opts}_{Z}$ & $\textsc{oppts}_{Z}$ \\
\midrule
\texttt{ion} & $.87_{}$ & $.71_{}$ & $.73_{}$ & $.92_{}$ & $.92_{}$ & $.91_{31}$ & $\textit{.97}_{11}$ & $.96_{54}$ & $.94_{8}$ & $.87_{12}$ & $.88_{13}$ & $.81_{12}$ & $.93_{17}$ & $.94_{19}$ & $.91_{28}$ & $.96_{22}$ \\
\texttt{fire} & $\textbf{1.00}_{}$ & $.93_{}$ & $.97_{}$ & $\textbf{1.00}_{}$ & $\textit{.99}_{}$ & $.85_{26}$ & $.89_{33}$ & $.88_{10}$ & $.88_{5}$ & $.89_{13}$ & $.85_{4}$ & $.85_{7}$ & $.86_{4}$ & $.82_{7}$ & $.88_{4}$ & $.90_{30}$ \\
\texttt{yeast} & $.19_{}$ & $.50_{}$ & $.47_{}$ & $.52_{}$ & $\textbf{.56}_{}$ & $.53_{40}$ & $.49_{48}$ & $.49_{230}$ & $.52_{14}$ & $.53_{16}$ & $.46_{24}$ & $\textit{.55}_{19}$ & $.52_{96}$ & $.54_{78}$ & $\textit{.55}_{123}$ & $.47_{66}$ \\
\texttt{magic} & $.76_{}$ & $.79_{}$ & $.75_{}$ & $\textbf{.81}_{}$ & $\textit{.80}_{}$ & $.77_{34}$ & $.76_{49}$ & $\textit{.80}_{1139}$ & $\textit{.80}_{11}$ & $.79_{14}$ & $.74_{16}$ & $.78_{13}$ & $\textit{.80}_{42}$ & $.77_{29}$ & $.78_{25}$ & $.77_{44}$ \\
\texttt{sonar} & $.68_{}$ & $.66_{}$ & $.62_{}$ & $.71_{}$ & $.73_{}$ & $.70_{5}$ & $.75_{10}$ & $.74_{12}$ & $\textit{.79}_{8}$ & $.76_{13}$ & $.70_{16}$ & $\textbf{.81}_{17}$ & $\textit{.79}_{10}$ & $.75_{9}$ & $.75_{36}$ & $.78_{30}$ \\
\texttt{compas} & $.52_{}$ & $.47_{}$ & $.47_{}$ & $\textit{.61}_{}$ & $.59_{}$ & $.60_{26}$ & $.60_{46}$ & $\textit{.61}_{156}$ & $.58_{10}$ & $.59_{6}$ & $.57_{6}$ & $\textbf{.62}_{8}$ & $.59_{10}$ & $.56_{5}$ & $.58_{13}$ & $.55_{18}$ \\
\bottomrule
\end{tabular}
\end{sidewaystable}

\begin{table}[t]
    \centering
    \caption{Average weighted F1-score for \textsc{PivotTree} classifiers and selectors combined with \textsc{knn}, baselines and competitors. Subscripts indicate the average number of pivots $\pm$ std. dev. Best results in \textbf{bold}, second best in \textit{italics}}
    \label{tab:knn_fulltable_indvividual_f1}
    \centering
    \footnotesize
    \setlength{\tabcolsep}{0.2mm} % Reduce column spacing
    \begin{tabular}{l|l|lll|llll|lll|llll}
        \toprule
        \texttt{dataset} & $\textsc{knn}_{X}$ & $\textsc{kms}_{Z}$ & $\textsc{kmd}_{Z}$ & $\textsc{ebl}_{Z}$ & $\textsc{pts}_{Z}$ & $\textsc{ppts}_{Z}$ & $\textsc{opts}_{Z}$ & $\textsc{oppts}_{Z}$ & $\textsc{kms}_{P}$ & $\textsc{kmd}_{P}$ & $\textsc{ebl}_{P}$  & $\textsc{pts}_{P}$ & $\textsc{ppts}_{P}$ & $\textsc{opts}_{P}$ & $\textsc{oppts}_{P}$ \\
        \midrule
        \texttt{ion} & $.89_{}$ & $\textit{.96}_{12}$ & $\textbf{.99}_{14}$ & $\textit{.96}_{157}$ & $\textit{.96}_{8}$ & $.95_{12}$ & $.95_{13}$ & $\textit{.96}_{22}$ & $.88_{20}$ & $.87_{20}$ & $\textit{.96}_{80}$ & $.89_{16}$ & $.79_{15}$ & $.83_{11}$ & $.85_{20}$ \\
        \texttt{fire} & $.89_{}$ & $.89_{5}$ & $.91_{15}$ & $\textit{.92}_{11}$ & $.84_{6}$ & $\textit{.92}_{7}$ & $\textbf{.93}_{9}$ & $.88_{14}$ & $.90_{9}$ & $.86_{11}$ & $\textit{.92}_{38}$ & $.72_{14}$ & $.81_{4}$ & $.89_{9}$ & $.80_{2}$ \\
        \texttt{yeast} & $\textit{.57}_{}$ & $.52_{11}$ & $\textit{.57}_{15}$ & $\textbf{.58}_{153}$ & $.50_{14}$ & $.50_{9}$ & $.53_{11}$ & $.55_{19}$ & $.43_{15}$ & $.48_{20}$ & $.56_{74}$ & $.35_{14}$ & $.25_{16}$ & $.23_{24}$ & $.27_{19}$ \\
        \texttt{magic} & $\textbf{.83}_{}$ & $.80_{19}$ & $.80_{17}$ & $.79_{902}$ & $\textit{.81}_{11}$ & $.80_{9}$ & $.80_{11}$ & $.80_{16}$ & $.72_{18}$ & $.65_{8}$ & $\textbf{.83}_{1542}$ & $.75_{3}$ & $.76_{3}$ & $.72_{6}$ & $.77_{11}$ \\
        \texttt{sonar} & $\textbf{.94}_{}$ & $.84_{10}$ & $.79_{9}$ & $\textit{.92}_{50}$ & $.86_{16}$ & $.73_{9}$ & $.86_{18}$ & $.81_{14}$ & $.70_{20}$ & $.75_{18}$ & $.87_{96}$ & $.55_{14}$ & $.71_{13}$ & $.67_{15}$ & $.78_{13}$ \\
        \texttt{compas} & $\textbf{.59}_{}$ & $\textit{.58}_{14}$ & $.57_{7}$ & $\textbf{.59}_{724}$ & $\textbf{.59}_{3}$ & $\textbf{.59}_{6}$ & $.57_{12}$ & $\textbf{.59}_{11}$ & $.55_{10}$ & $.47_{8}$ & $.56_{467}$ & $.56_{15}$ & $\textit{.58}_{11}$ & $\textbf{.59}_{13}$ & $\textit{.58}_{15}$ \\
\bottomrule
    \end{tabular}
\end{table}

\begin{sidewaystable}
\caption{Average weighted F1-score for \textsc{PivotTree} classifiers and selectors combined with \textsc{dt} (max 20 pivots), baselines and competitors. Subscripts indicate the average number of pivots $\pm$ std. dev. Best results in \textbf{bold}, second best in \textit{italics}}
\label{tab:dt_single_dataset_max20}
\centering
\footnotesize
\setlength{\tabcolsep}{1.4mm} % Reduce column spacing
\begin{tabular}{l|lll|ll|lll|llll|llll}
\toprule
\texttt{dataset} & $\textsc{grule}_{X}$ & $\textsc{irep}_{X}$ & $\textsc{ripper}_{X}$ & $\textsc{dt}_{X}$ & $\textsc{odt}_{X}$ & $\textsc{kms}_{Z}$ & $\textsc{kmd}_{Z}$ & $\textsc{ebl}_{Z}$ & $\textsc{ptc}_{Z}$ & $\textsc{pptc}_{Z}$ & $\textsc{optc}_{Z}$ & $\textsc{opptc}_{Z}$ & $\textsc{pts}_{Z}$ & $\textsc{ppts}_{Z}$ & $\textsc{opts}_{Z}$ & $\textsc{oppts}_{Z}$ \\
\midrule
\texttt{ion} & $.87_{}$ & $.71_{}$ & $.73_{}$ & $.92_{}$ & $.92_{}$ & $.89_{9}$ & $\textit{.97}_{11}$ & $\textbf{.98}_{20}$ & $.94_{8}$ & $.87_{12}$ & $.88_{13}$ & $.81_{12}$ & $.93_{17}$ & $.94_{19}$ & $.88_{11}$ & $.96_{22}$ \\
\texttt{fire} & $\textbf{1.00}_{}$ & $.93_{}$ & $.97_{}$ & $\textbf{1.00}_{}$ & $\textit{.99}_{}$ & $.92_{10}$ & $.86_{17}$ & $.88_{10}$ & $.88_{5}$ & $.89_{13}$ & $.85_{4}$ & $.85_{7}$ & $.86_{4}$ & $.82_{7}$ & $.88_{4}$ & $.90_{21}$ \\
\texttt{yeast} & $.19_{}$ & $.50_{}$ & $.47_{}$ & $.52_{}$ & $\textbf{.56}_{}$ & $.51_{13}$ & $.50_{7}$ & $.47_{20}$ & $.52_{14}$ & $.53_{16}$ & $.46_{24}$ & $\textit{.55}_{19}$ & $.51_{14}$ & $.50_{16}$ & $.50_{11}$ & $.44_{4}$ \\
\texttt{magic} & $.76_{}$ & $.79_{}$ & $.75_{}$ & $\textbf{.81}_{}$ & $\textit{.80}_{}$ & $.76_{19}$ & $.76_{17}$ & $.77_{20}$ & $\textit{.80}_{11}$ & $.79_{14}$ & $.74_{16}$ & $.78_{13}$ & $\textit{.80}_{11}$ & $.77_{15}$ & $.74_{11}$ & $.76_{16}$ \\
\texttt{sonar} & $.68_{}$ & $.66_{}$ & $.62_{}$ & $.71_{}$ & $.73_{}$ & $.70_{5}$ & $.75_{10}$ & $.74_{12}$ & $\textit{.79}_{8}$ & $.76_{13}$ & $.70_{16}$ & $\textbf{.81}_{17}$ & $\textit{.79}_{10}$ & $.75_{9}$ & $.68_{18}$ & $.70_{13}$ \\
\texttt{compas} & $.52_{}$ & $.47_{}$ & $.47_{}$ & $\textit{.61}_{}$ & $.59_{}$ & $.57_{11}$ & $.60_{17}$ & $.60_{20}$ & $.58_{10}$ & $.59_{6}$ & $.57_{6}$ & $\textbf{.62}_{8}$ & $.59_{10}$ & $.56_{5}$ & $.58_{13}$ & $.55_{18}$ \\
\bottomrule
\end{tabular}
\end{sidewaystable}

\begin{table}[t]
        \centering
    \caption{Average weighted F1-score for \textsc{PivotTree} classifiers and selectors combined with \textsc{knn} (max 20 pivots), baselines and competitors. Subscripts indicate the average number of pivots $\pm$ std. dev. Best results in \textbf{bold}, second best in \textit{italics}}
    \label{tab:knn_fulltable_indvividual_max_20_piv}
\centering
\footnotesize
\setlength{\tabcolsep}{0.2mm} % Reduce column spacing
\begin{tabular}{l|l|lll|llll|lll|llll}
\toprule
\texttt{dataset} & $\textsc{knn}_{X}$ & $\textsc{kms}_{Z}$ & $\textsc{kmd}_{Z}$ & $\textsc{ebl}_{Z}$ & $\textsc{pts}_{Z}$ & $\textsc{ppts}_{Z}$ & $\textsc{opts}_{Z}$ & $\textsc{oppts}_{Z}$ & $\textsc{kms}_{P}$ & $\textsc{kmd}_{P}$ & $\textsc{ebl}_{P}$ & $\textsc{pts}_{P}$ & $\textsc{ppts}_{P}$ & $\textsc{opts}_{P}$ & $\textsc{oppts}_{P}$ \\
\midrule
\texttt{ion} & $.89_{}$ & $\textit{.96}_{12}$ & $\textbf{.99}_{14}$ & $.95_{20}$ & $\textit{.96}_{8}$ & $.95_{12}$ & $.95_{13}$ & $\textit{.96}_{22}$ & $.88_{20}$ & $.87_{20}$ & $.89_{20}$ & $.89_{16}$ & $.79_{15}$ & $.83_{11}$ & $.85_{20}$ \\
\texttt{fire} & $.89_{}$ & $.89_{5}$ & $.91_{15}$ & $\textit{.92}_{11}$ & $.84_{6}$ & $\textit{.92}_{7}$ & $\textbf{.93}_{9}$ & $.88_{14}$ & $.90_{9}$ & $.86_{11}$ & $\textbf{.93}_{20}$ & $.72_{14}$ & $.81_{4}$ & $.89_{9}$ & $.80_{2}$ \\
\texttt{yeast} & $\textit{.57}_{}$ & $.52_{11}$ & $\textit{.57}_{15}$ & $.55_{20}$ & $.50_{14}$ & $.50_{9}$ & $.53_{11}$ & $.55_{19}$ & $.43_{15}$ & $.48_{20}$ & $\textbf{.58}_{20}$ & $.35_{14}$ & $.25_{16}$ & $.23_{24}$ & $.27_{19}$ \\
\texttt{magic} & $\textbf{.83}_{}$ & $.80_{19}$ & $.80_{17}$ & $\textit{.81}_{20}$ & $\textit{.81}_{11}$ & $.80_{9}$ & $.80_{11}$ & $.80_{16}$ & $.72_{18}$ & $.65_{8}$ & $.77_{20}$ & $.75_{3}$ & $.76_{3}$ & $.72_{6}$ & $.77_{11}$ \\
\texttt{sonar} & $\textbf{.94}_{}$ & $.84_{10}$ & $.79_{9}$ & $.76_{20}$ & $.86_{16}$ & $.73_{9}$ & $.86_{18}$ & $.81_{14}$ & $.70_{20}$ & $.75_{18}$ & $.78_{12}$ & $.55_{14}$ & $.71_{13}$ & $.67_{15}$ & $.78_{13}$ \\
\texttt{compas} & $\textbf{.59}_{}$ & $\textit{.58}_{14}$ & $.57_{7}$ & $\textbf{.59}_{20}$ & $\textbf{.59}_{3}$ & $\textbf{.59}_{6}$ & $.57_{12}$ & $\textbf{.59}_{11}$ & $.55_{10}$ & $.47_{8}$ & $.54_{20}$ & $.56_{15}$ & $\textit{.58}_{11}$ & $\textbf{.59}_{13}$ & $\textit{.58}_{15}$ \\
\bottomrule
\end{tabular}
\end{table}

\begin{figure}[t]
    \includegraphics[width=0.48\linewidth]{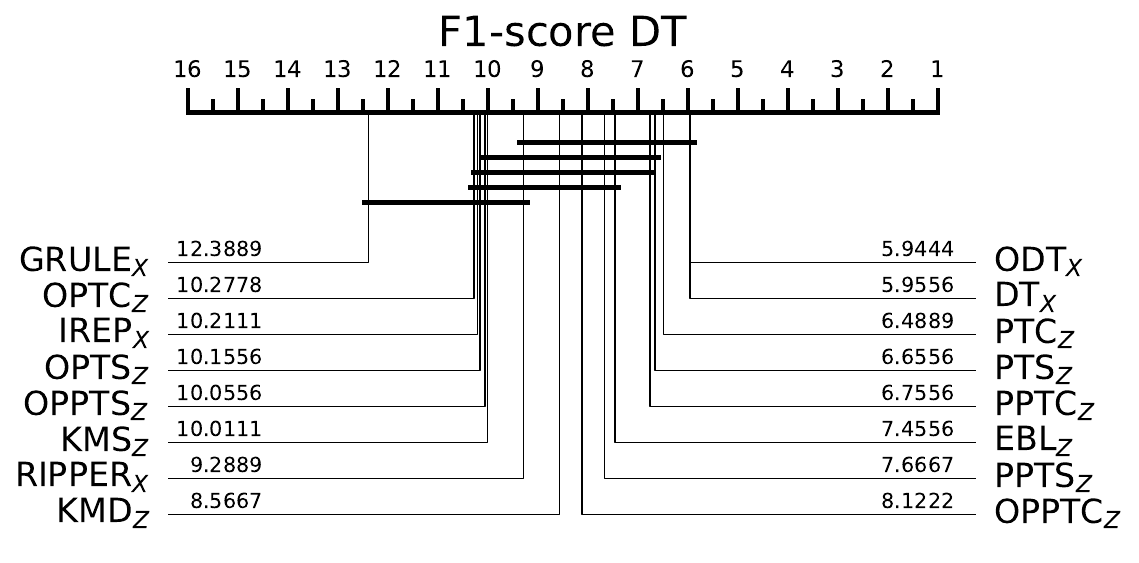}
    \includegraphics[width=0.48\linewidth]{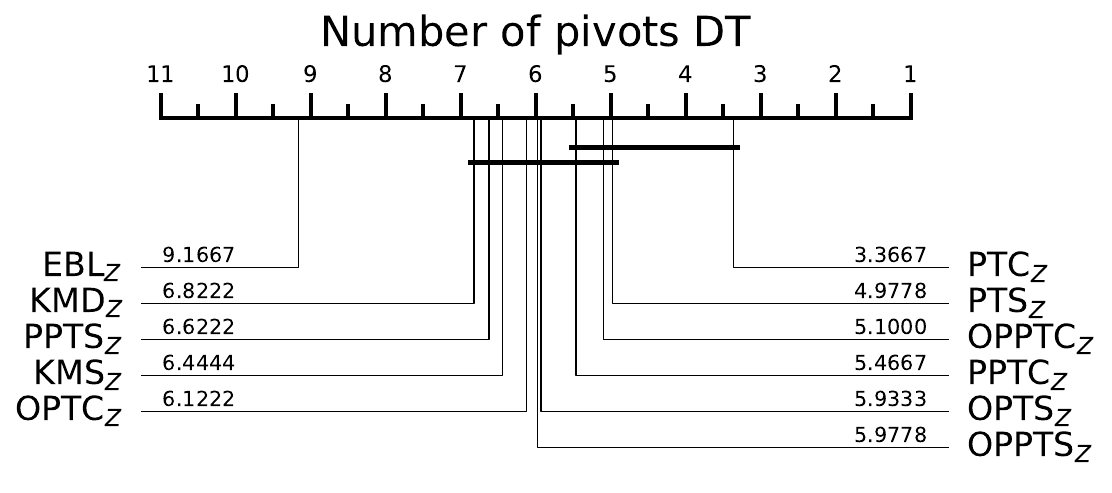}
    \includegraphics[width=0.5\linewidth]{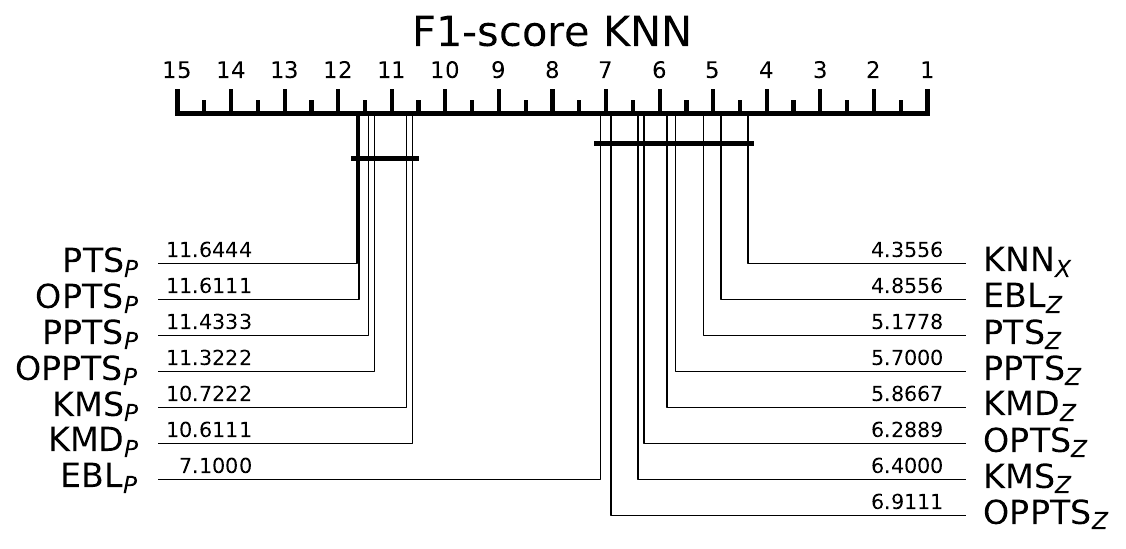}
    \includegraphics[width=0.5\linewidth]{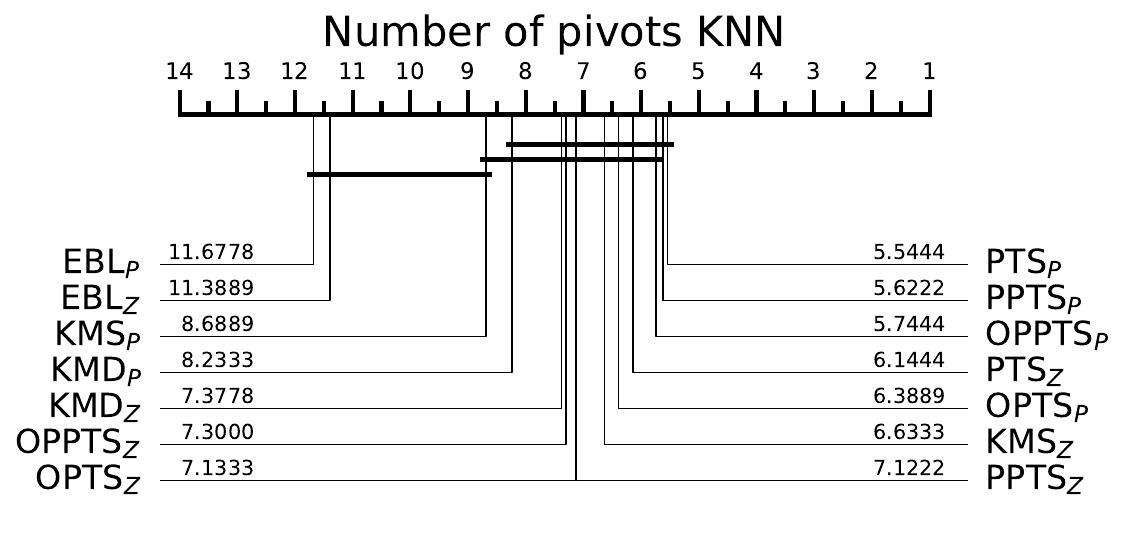}
    \caption{Critical difference plot of model's rank in terms of weighted F1-score and number of pivots against each other with Nemenyi test w.r.t all datasets in the constrained setting. Models that are not significantly different at $95$\% significance level are connected. Best models on the right.}
    \label{fig:cd_plots_f1_full}
\end{figure}

\begin{sidewaystable}
\caption{Average Balanced Accuracy for \textsc{PivotTree} classifiers and selectors combined with \textsc{dt}, baselines and competitors. Subscripts indicate the average number of pivots $\pm$ std. dev. Best results in \textbf{bold}, second best in \textit{italics}}
\label{tab:dt_single_dataset_unconstrained_balance_acc}
\centering
\footnotesize
\setlength{\tabcolsep}{1.4mm} % Reduce column 
\begin{tabular}{l|lll|ll|lll|llll|llll}
\toprule
\texttt{dataset} & $\textsc{grule}_{X}$ & $\textsc{irep}_{X}$ & $\textsc{ripper}_{X}$ & $\textsc{dt}_{X}$ & $\textsc{odt}_{X}$ & $\textsc{kms}_{Z}$ & $\textsc{kmd}_{Z}$ & $\textsc{ebl}_{Z}$ & $\textsc{ptc}_{Z}$ & $\textsc{pptc}_{Z}$ & $\textsc{optc}_{Z}$ & $\textsc{opptc}_{Z}$ & $\textsc{pts}_{Z}$ & $\textsc{ppts}_{Z}$ & $\textsc{opts}_{Z}$ & $\textsc{oppts}_{Z}$ \\
\midrule
\texttt{ion} & $.83_{}$ & $.66_{}$ & $.73_{}$ & $.92_{}$ & $.92_{}$ & $.92_{31}$ & $.97_{13}$ & $.96_{54}$ & $.93_{8}$ & $.83_{12}$ & $.86_{13}$ & $.76_{12}$ & $.85_{16}$ & $.96_{19}$ & $.90_{28}$ & $.96_{22}$ \\
\texttt{fire} & $\textbf{1.00}_{}$ & $.92_{}$ & $\textit{.98}_{}$ & $\textbf{1.00}_{}$ & $\textit{.98}_{}$ & $.85_{26}$ & $.90_{26}$ & $.87_{10}$ & $.87_{5}$ & $.87_{13}$ & $.85_{4}$ & $.83_{7}$ & $.86_{4}$ & $.83_{7}$ & $.86_{4}$ & $.90_{30}$ \\
\texttt{yeast} & $.15_{}$ & $.40_{}$ & $.37_{}$ & $.42_{}$ & $\textit{.44}_{}$ & $.19_{3}$ & $.31_{39}$ & $.33_{230}$ & $.36_{14}$ & $.43_{16}$ & $.29_{24}$ & $\textbf{.49}_{19}$ & $.36_{67}$ & $.33_{16}$ & $.36_{123}$ & $.33_{87}$ \\
\texttt{magic} & $.71_{}$ & $.74_{}$ & $.74_{}$ & $\textbf{.77}_{}$ & $\textit{.76}_{}$ & $.67_{32}$ & $.71_{49}$ & $\textbf{.77}_{1139}$ & $\textbf{.77}_{11}$ & $.75_{14}$ & $.69_{16}$ & $.73_{5}$ & $\textbf{.77}_{42}$ & $.74_{29}$ & $.73_{38}$ & $.74_{44}$ \\
\texttt{sonar} & $.69_{}$ & $.66_{}$ & $.62_{}$ & $.72_{}$ & $.74_{}$ & $.70_{5}$ & $.75_{10}$ & $.78_{12}$ & $\textit{.80}_{8}$ & $.75_{13}$ & $.70_{16}$ & $\textbf{.81}_{17}$ & $\textit{.80}_{10}$ & $.75_{9}$ & $.76_{36}$ & $.78_{30}$ \\
\texttt{compas} & $.47_{}$ & $.42_{}$ & $.42_{}$ & $.53_{}$ & $.51_{}$ & $.48_{11}$ & $\textbf{.55}_{41}$ & $\textit{.54}_{156}$ & $.52_{3}$ & $.50_{6}$ & $.48_{6}$ & $\textit{.54}_{8}$ & $.53_{10}$ & $.49_{5}$ & $.51_{13}$ & $.47_{68}$ \\
\bottomrule
\end{tabular}
\end{sidewaystable}

\begin{sidewaystable}[t]
\caption{Average Balanced Accuracy for \textsc{PivotTree} classifiers and selectors combined with \textsc{dt} (max 20 pivots), baselines and competitors. Subscripts indicate the average number of pivots $\pm$ std. dev. Best results in \textbf{bold}, second best in \textit{italics}}
\label{tab:dt_single_dataset_max_20}
\centering
\footnotesize
\setlength{\tabcolsep}{1.4mm} % Reduce column 
\begin{tabular}{l|lll|ll|lll|llll|llll}
\toprule
\texttt{dataset} & $\textsc{grule}_{X}$ & $\textsc{irep}_{X}$ & $\textsc{ripper}_{X}$ & $\textsc{dt}_{X}$ & $\textsc{odt}_{X}$ & $\textsc{kms}_{Z}$ & $\textsc{kmd}_{Z}$ & $\textsc{ebl}_{Z}$ & $\textsc{ptc}_{Z}$ & $\textsc{pptc}_{Z}$ & $\textsc{optc}_{Z}$ & $\textsc{opptc}_{Z}$ & $\textsc{pts}_{Z}$ & $\textsc{ppts}_{Z}$ & $\textsc{opts}_{Z}$ & $\textsc{oppts}_{Z}$ \\
\midrule
\texttt{ion} & $.83_{}$ & $.66_{}$ & $.73_{}$ & $.92_{}$ & $.92_{}$ & $.88_{9}$ & $.97_{13}$ & $\textbf{.99}_{20}$ & $.93_{\textbf{8}}$ & $.83_{12}$ & $.86_{13}$ & $.76_{12}$ & $.85_{16}$ & $.96_{19}$ & $.87_{11}$ & $.96_{22}$ \\
\texttt{fire} & $\textbf{1.00}_{}$ & $.92_{}$ & $\textit{.98}_{}$ & $\textbf{1.00}_{}$ & $\textit{.98}_{}$ & $.92_{10}$ & $.85_{17}$ & $.87_{10}$ & $.87_{5}$ & $.87_{13}$ & $.85_{4}$ & $.83_{7}$ & $.86_{4}$ & $.83_{7}$ & $.86_{4}$ & $.90_{21}$ \\
\texttt{yeast} & $.15_{}$ & $.40_{}$ & $.37_{}$ & $.42_{}$ & $\textit{.44}_{}$ & $.19_{3}$ & $.31_{11}$ & $.33_{20}$ & $.36_{14}$ & $.43_{16}$ & $.29_{24}$ & $\textbf{.49}_{19}$ & $.36_{14}$ & $.33_{16}$ & $.27_{4}$ & $.35_{4}$ \\
\texttt{magic} & $.71_{}$ & $.74_{}$ & $.74_{}$ & $\textbf{.77}_{}$ & $\textit{.76}_{}$ & $.67_{20}$ & $.73_{10}$ & $\textit{.76}_{20}$ & $\textbf{.77}_{11}$ & $.75_{14}$ & $.69_{16}$ & $.73_{5}$ & $\textbf{.77}_{11}$ & $.73_{15}$ & $.70_{11}$ & $.73_{16}$ \\
\texttt{sonar} & $.69_{}$ & $.66_{}$ & $.62_{}$ & $.72_{}$ & $.74_{}$ & $.66_{5}$ & $.75_{10}$ & $.78_{12}$ & $\textit{.80}_{8}$ & $.75_{13}$ & $.70_{16}$ & $\textbf{.81}_{17}$ & $\textit{.80}_{10}$ & $.75_{9}$ & $.70_{18}$ & $.57_{14}$ \\
\texttt{compas} & $.47_{}$ & $.42_{}$ & $.42_{}$ & $.53_{}$ & $.51_{}$ & $.48_{11}$ & $\textit{.54}_{17}$ & $.50_{20}$ & $.52_{3}$ & $.50_{6}$ & $.48_{6}$ & $\textit{.54}_{8}$ & $.53_{10}$ & $.49_{5}$ & $.51_{13}$ & $.50_{18}$ \\
\bottomrule
\end{tabular}
\end{sidewaystable}

\begin{table}[t]
        \centering
    \caption{Average Balanced Accuracy for \textsc{PivotTree} classifiers and selectors combined with \textsc{knn}, baselines and competitors. Subscripts indicate the average number of pivots $\pm$ std. dev. Best results in \textbf{bold}, second best in \textit{italics}}
    \label{tab:knn_fulltable_indvividual_acc}
\centering
\footnotesize
\setlength{\tabcolsep}{0.2mm} % Reduce column spacing
\begin{tabular}{l|l|lll|llll|lll|llll}
\toprule
\texttt{dataset} & $\textsc{knn}_{X}$ & $\textsc{kms}_{Z}$ & $\textsc{kmd}_{Z}$ & $\textsc{ebl}_{Z}$ & $\textsc{pts}_{Z}$ & $\textsc{ppts}_{Z}$ & $\textsc{opts}_{Z}$ & $\textsc{oppts}_{Z}$ & $\textsc{kms}_{P}$ & $\textsc{kmd}_{P}$ & $\textsc{ebl}_{P}$ & $\textsc{pts}_{P}$ & $\textsc{ppts}_{P}$ & $\textsc{opts}_{P}$ & $\textsc{oppts}_{P}$ \\
\midrule
\texttt{ion} & $.86_{}$ & $.96_{20}$ & $\textbf{.99}_{14}$ & $.96_{157}$ & $.95_{8}$ & $\textit{.97}_{14}$ & $.96_{13}$ & $.96_{22}$ & $.86_{20}$ & $.84_{5}$ & $.95_{80}$ & $.88_{16}$ & $.80_{15}$ & $.81_{11}$ & $.84_{20}$ \\
\texttt{fire} & $.88_{}$ & $.89_{5}$ & $.91_{19}$ & $\textit{.92}_{32}$ & $.84_{6}$ & $.90_{7}$ & $\textbf{.93}_{9}$ & $.88_{14}$ & $.91_{9}$ & $.90_{15}$ & $.91_{38}$ & $.72_{14}$ & $.79_{4}$ & $.89_{9}$ & $.81_{2}$ \\
\texttt{yeast} & $.47_{}$ & $.50_{12}$ & $\textit{.52}_{12}$ & $.47_{60}$ & $.48_{14}$ & $.48_{16}$ & $.45_{11}$ & $.47_{9}$ & $.39_{19}$ & $.45_{20}$ & $\textit{.52}_{74}$ & $.23_{14}$ & $.20_{16}$ & $.32_{24}$ & $.26_{19}$ \\
\texttt{magic} & $\textit{.79}_{}$ & $.76_{10}$ & $.77_{17}$ & $.75_{902}$ & $.78_{11}$ & $.77_{9}$ & $.77_{11}$ & $.77_{11}$ & $.68_{18}$ & $.59_{8}$ & $\textbf{.80}_{1542}$ & $.72_{13}$ & $.73_{14}$ & $.73_{6}$ & $.73_{13}$ \\
\texttt{sonar} & $\textbf{.94}_{}$ & $.84_{10}$ & $.79_{9}$ & $\textit{.92}_{50}$ & $.86_{16}$ & $.74_{9}$ & $.86_{18}$ & $.81_{14}$ & $.69_{17}$ & $.77_{15}$ & $.87_{96}$ & $.54_{14}$ & $.73_{13}$ & $.67_{15}$ & $.79_{17}$ \\
\texttt{compas} & $.51_{}$ & $.49_{13}$ & $.49_{7}$ & $.51_{724}$ & $.51_{3}$ & $.51_{6}$ & $.48_{12}$ & $.49_{8}$ & $.47_{10}$ & $.40_{8}$ & $.50_{467}$ & $\textit{.52}_{12}$ & $\textbf{.53}_{11}$ & $\textit{.52}_{13}$ & $.51_{17}$ \\
\texttt{house} & $\textbf{.83}_{}$ & $.79_{20}$ & $.78_{19}$ & $\textit{.82}_{238}$ & $.80_{14}$ & $.80_{19}$ & $.76_{16}$ & $.77_{16}$ & $.51_{13}$ & $.70_{18}$ & $.81_{2753}$ & $.76_{10}$ & $.75_{6}$ & $.75_{20}$ & $.76_{16}$ \\
\bottomrule
\end{tabular}
\end{table}

\begin{table}[t]
        \centering
    \caption{Average Balanced Accuracy for \textsc{PivotTree} classifiers and selectors combined with \textsc{knn} (max 20 pivots), baselines and competitors. Subscripts indicate the average number of pivots $\pm$ std. dev. Best results in \textbf{bold}, second best in \textit{italics}}
    \label{tab:knn_fulltable_indvividual_acc_max_20}
\centering
\footnotesize
\setlength{\tabcolsep}{0.2mm} % Reduce column spacing
\begin{tabular}{l|l|lll|llll|lll|llll}
\toprule
\texttt{dataset} & $\textsc{knn}_{X}$ & $\textsc{kms}_{Z}$ & $\textsc{kmd}_{Z}$ & $\textsc{ebl}_{Z}$ & $\textsc{pts}_{Z}$ & $\textsc{ppts}_{Z}$ & $\textsc{opts}_{Z}$ & $\textsc{oppts}_{Z}$ & $\textsc{kms}_{P}$ & $\textsc{kmd}_{P}$ & $\textsc{ebl}_{P}$ & $\textsc{pts}_{P}$ & $\textsc{ppts}_{P}$ & $\textsc{opts}_{P}$ & $\textsc{oppts}_{P}$ \\
\midrule
\texttt{ion} & $.86_{}$ & $.96_{20}$ & $\textbf{.99}_{14}$ & $.95_{20}$ & $.95_{8}$ & $\textit{.97}_{14}$ & $.96_{13}$ & $.96_{22}$ & $.86_{20}$ & $.84_{5}$ & $.85_{20}$ & $.88_{16}$ & $.80_{15}$ & $.81_{11}$ & $.84_{20}$ \\
\texttt{fire} & $.88_{}$ & $.89_{5}$ & $.91_{19}$ & $.91_{11}$ & $.84_{6}$ & $.90_{7}$ & $\textbf{.93}_{9}$ & $.88_{14}$ & $.91_{9}$ & $.90_{15}$ & $\textbf{.93}_{20}$ & $.72_{14}$ & $.79_{4}$ & $.89_{9}$ & $.81_{2}$ \\
\texttt{yeast} & $.47_{}$ & $.50_{12}$ & $\textit{.52}_{12}$ & $\textbf{.54}_{20}$ & $.48_{14}$ & $.48_{16}$ & $.45_{11}$ & $.47_{9}$ & $.39_{19}$ & $.45_{20}$ & $\textit{.52}_{20}$ & $.23_{14}$ & $.20_{16}$ & $.32_{24}$ & $.26_{19}$ \\
\texttt{magic} & $\textit{.79}_{}$ & $.76_{10}$ & $.77_{17}$ & $.78_{20}$ & $.78_{11}$ & $.77_{9}$ & $.77_{11}$ & $.77_{11}$ & $.68_{18}$ & $.59_{8}$ & $.73_{20}$ & $.72_{13}$ & $.73_{14}$ & $.73_{6}$ & $.73_{13}$ \\
\texttt{sonar} & $\textbf{.94}_{}$ & $.84_{10}$ & $.79_{9}$ & $.78_{20}$ & $.86_{16}$ & $.74_{9}$ & $.86_{18}$ & $.81_{14}$ & $.69_{17}$ & $.77_{15}$ & $.79_{12}$ & $.54_{14}$ & $.73_{13}$ & $.67_{15}$ & $.79_{17}$ \\
\texttt{compas} & $.51_{}$ & $.49_{13}$ & $.49_{7}$ & $.51_{20}$ & $.51_{3}$ & $.51_{6}$ & $.48_{12}$ & $.49_{8}$ & $.47_{10}$ & $.40_{8}$ & $.51_{20}$ & $\textit{.52}_{12}$ & $\textbf{.53}_{11}$ & $\textit{.52}_{13}$ & $.51_{17}$ \\
\texttt{house} & $\textbf{.83}_{}$ & $.79_{20}$ & $.78_{19}$ & $\textit{.82}_{20}$ & $.80_{14}$ & $.80_{19}$ & $.76_{16}$ & $.77_{16}$ & $.51_{13}$ & $.70_{18}$ & $.77_{20}$ & $.76_{10}$ & $.75_{6}$ & $.75_{20}$ & $.76_{16}$ \\
\bottomrule
\end{tabular}
\end{table}

\begin{table}[t]
    \caption{Average Balanced Accuracy $\pm$ std. dev. for \textsc{RandomPivotForest} with $100$ \textsc{PivotTree} \textsc{classifiers} as estimators, and no sampling, against baselines. 
    When the splitting stump forest is adopted, subscripts indicate the average number of stumps $\pm$ std. dev.
    Best results in \textbf{bold}, second best in \textit{italics}.}
    \label{tab:ensemble_full_bal_acc_full}
    \setlength{\tabcolsep}{1.4mm}
    \centering
    \begin{tabular}{cccccc}
    \toprule

{model} & $\texttt{tabular}$ & $\texttt{images}$ & $\texttt{time-series}$ & $\texttt{text}$ & \texttt{all} \\
\midrule
$\textsc{xgb}$ & $\textbf{.84}_{} \pm .14_{}$ & $\textbf{.94}_{} \pm .05_{}$ & $\textbf{.67}_{} \pm .25_{}$ & $\textbf{.70}_{} \pm .25_{}$ & $\textbf{.80}_{} \pm .20_{}$ \\
$\textsc{lgbm}$ & $\textit{.83}_{} \pm .15_{}$ & $\textbf{.94}_{} \pm .05_{}$ & $.64_{} \pm .24_{}$ & $\textbf{.70}_{} \pm .25_{}$ & $\textit{.79}_{} \pm .20_{}$ \\
$\textsc{catb}$ & $\textit{.83}_{} \pm .15_{}$ & $\textit{.93}_{} \pm .06_{}$ & $\textit{.66}_{} \pm .25_{}$ & $\textit{.68}_{} \pm .25_{}$ & $\textit{.79}_{} \pm .20_{}$ \\
\midrule
$\textsc{rdt}$ & $.72_{} \pm .21_{}$ & $\textit{.93}_{} \pm .05_{}$ & $.62_{} \pm .23_{}$ & $.56_{} \pm .25_{}$ & $.72_{} \pm .23_{}$ \\
$\textsc{rodt}$ & $.72_{} \pm .21_{}$ & $.92_{} \pm .06_{}$ & $.61_{} \pm .23_{}$ & $.55_{} \pm .24_{}$ & $.72_{} \pm .23_{}$ \\
\midrule
$\textsc{rpt}$ & $.72_{} \pm .22_{}$ & $.73_{} \pm .17_{}$ & $.63_{} \pm .23_{}$ & $.57_{} \pm .25_{}$ & $.69_{} \pm .22_{}$ \\
$\textsc{ropt}$ & $.70_{} \pm .21_{}$ & $.74_{} \pm .16_{}$ & $.62_{} \pm .23_{}$ & $.53_{} \pm .24_{}$ & $.67_{} \pm .21_{}$ \\
\bottomrule
\end{tabular}
\end{table}

\begin{table}[t]
    \caption{Average Balanced Accuracy $\pm$ std. dev. for \textsc{RandomPivotForest} with $100$ \textsc{PivotTree} \textsc{classifiers} as estimators, and 10\% of the training set is used to train each estimator, against baselines. 
    When the splitting stump forest is adopted, subscripts indicate the average number of stumps $\pm$ std. dev.
    Best results in \textbf{bold}, second best in \textit{italics}.}
    \label{tab:ensemble_subsample_bal_acc}
    \centering
    \setlength{\tabcolsep}{1.4mm}
    \begin{tabular}{cccccc}
\toprule
{model} & $\texttt{tabular}$ & $\texttt{images}$ & $\texttt{time-series}$ & $\texttt{text}$ & \texttt{all} \\
\midrule
$\textsc{xgb}$ & $.80_{} \pm .17_{}$ & $\textit{.93}_{} \pm .05_{}$ & $.61_{} \pm .21_{}$ & $\textit{.68}_{} \pm .25_{}$ & $\textit{.77}_{} \pm .21_{}$ \\
$\textsc{lgbm}$ & $\textbf{.83}_{} \pm .15_{}$ & $\textbf{.94}_{} \pm .05_{}$ & $\textit{.64}_{} \pm .24_{}$ & $\textbf{.70}_{} \pm .25_{}$ & $\textbf{.79}_{} \pm .20_{}$ \\
$\textsc{catb}$ & $\textit{.82}_{} \pm .16_{}$ & $\textbf{.94}_{} \pm .05_{}$ & $\textbf{.66}_{} \pm .25_{}$ & $\textit{.68}_{} \pm .25_{}$ & $\textbf{.79}_{} \pm .20_{}$ \\
\midrule
$\textsc{rdt}$ & $.69_{} \pm .23_{}$ & $.92_{} \pm .06_{}$ & $.55_{} \pm .17_{}$ & $.53_{} \pm .25_{}$ & $.68_{} \pm .23_{}$ \\
$\textsc{rodt}$ & $.68_{} \pm .23_{}$ & $.92_{} \pm .06_{}$ & $.56_{} \pm .17_{}$ & $.52_{} \pm .25_{}$ & $.68_{} \pm .23_{}$ \\
\midrule
$\textsc{rpt}$ & $.69_{} \pm .23_{}$ & $.83_{} \pm .14_{}$ & $.58_{} \pm .20_{}$ & $.56_{} \pm .24_{}$ & $.68_{} \pm .22_{}$ \\
$\textsc{ropt}$ & $.67_{} \pm .23_{}$ & $.82_{} \pm .15_{}$ & $.58_{} \pm .19_{}$ & $.53_{} \pm .24_{}$ & $.66_{} \pm .22_{}$ \\
\bottomrule
\end{tabular}
\end{table}

\begin{figure}[t]
    \centering
    \includegraphics[width=0.48\linewidth]{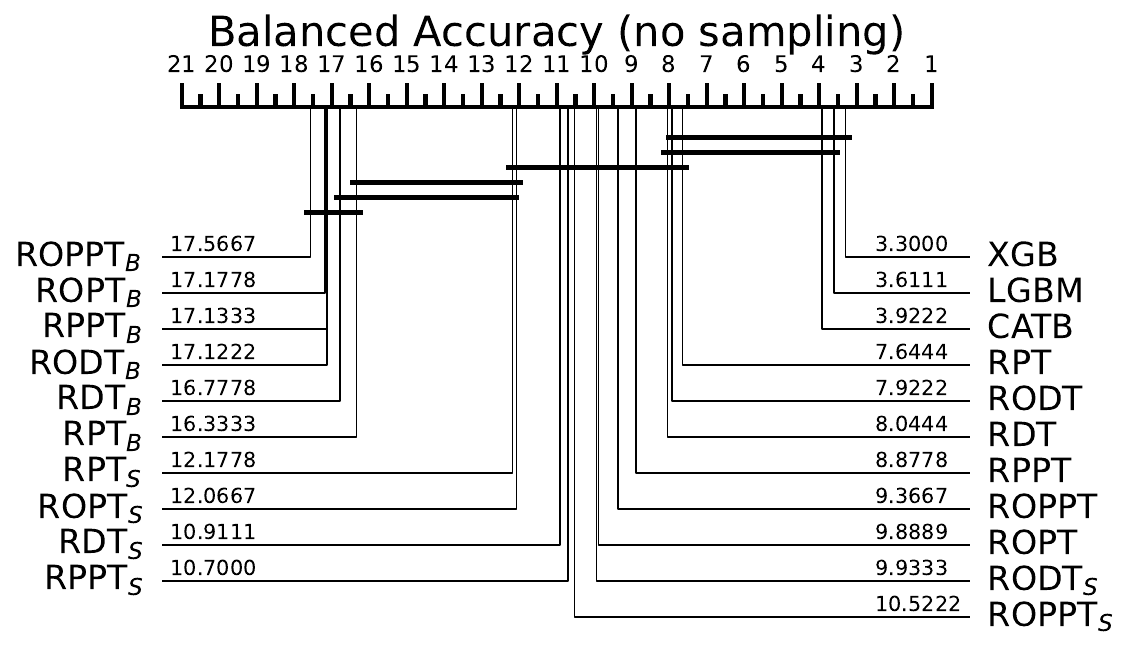}
    \includegraphics[width=0.48\linewidth]{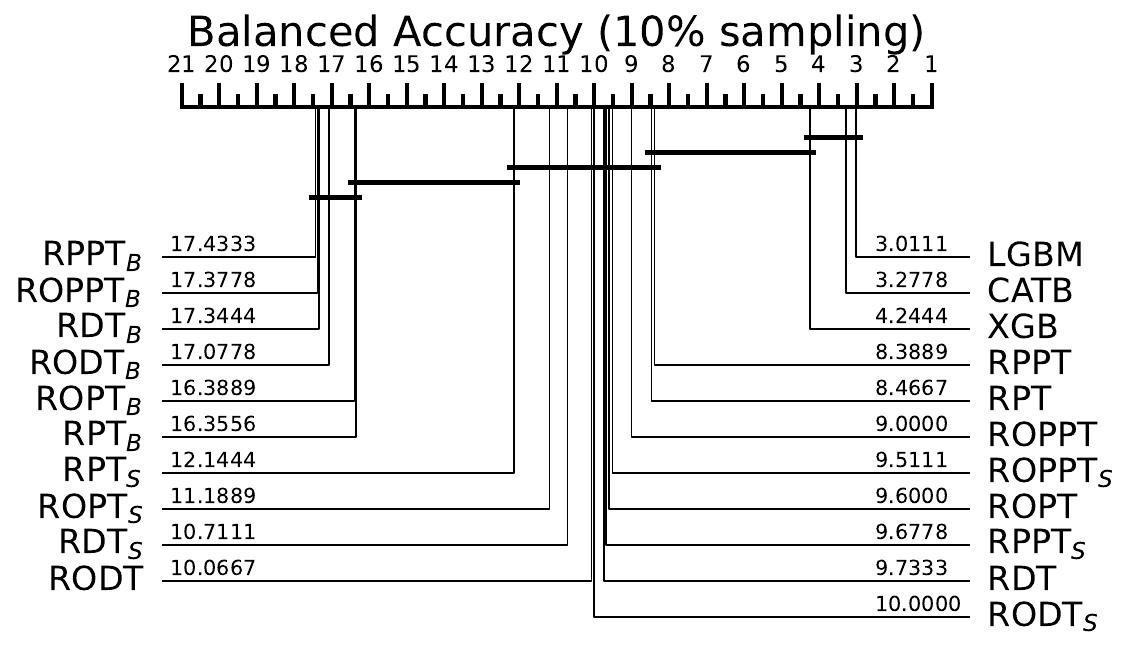}
    \caption{Critical difference plots of ensemble models' rankings in terms of Balanced Accuracy, evaluated using the Nemenyi test across all datasets. Models that are not significantly different at the $95\%$ significance level are connected. The best models are positioned on the right.}
    \label{fig:cd_plots_ensemble_balanced_acc}
\end{figure}

\begin{sidewaystable}[t]
\caption{Average Weighted F1-Score for \textsc{RandomPivotForest} with $100$ \textsc{PivotTree} \textsc{c}lassifiers as estimators, and no sampling, against baselines, for each \texttt{dataset}.  When the splitting stump forest is adopted, subscripts indicate the average number of stumps $\pm$ std. dev. Best results in \textbf{bold}, second best in \textit{italics}.}
    \label{tab:ensemble_sample_TOTAL}
\centering
\footnotesize
\setlength{\tabcolsep}{0.1mm} % Reduce column spacing
\begin{tabular}{l|lll|ll|llll|ll|llll|ll|llll}
\toprule
\texttt{dataset} & $\textsc{xgb}$ & $\textsc{lgbm}$ & $\textsc{catb}$ & $\textsc{rdt}$ & $\textsc{rodt}$ & $\textsc{rpt}$ & $\textsc{ropt}$ & $\textsc{rppt}$ & $\textsc{roppt}$ & $\textsc{rdt}_{\textsc{s}}$ & $\textsc{rodt}_{\textsc{s}}$ & $\textsc{rpt}_{\textsc{s}}$ & $\textsc{ropt}_{\textsc{s}}$ & $\textsc{rppt}_{\textsc{s}}$ & $\textsc{roppt}_{\textsc{s}}$ & $\textsc{rdt}_{\textsc{b}}$ & $\textsc{rodt}_{\textsc{b}}$ & $\textsc{rpt}_{\textsc{b}}$ & $\textsc{ropt}_{\textsc{b}}$ & $\textsc{rppt}_{\textsc{b}}$ & $\textsc{roppt}_{\textsc{b}}$ \\
\midrule
\texttt{ion} & $\textit{.94}_{}$ & $\textbf{.95}_{}$ & $\textit{.94}_{}$ & $.91_{}$ & $\textbf{.95}_{}$ & $.93_{}$ & $.93_{}$ & $.93_{}$ & $\textit{.94}_{}$ & $.87_{30}$ & $.86_{48}$ & $.88_{57}$ & $.90_{157}$ & $.86_{176}$ & $.91_{85}$ & $.80_{11}$ & $.80_{96}$ & $.79_{16}$ & $.68_{107}$ & $.64_{101}$ & $.67_{69}$ \\
\texttt{fire} & $\textbf{.99}_{}$ & $\textbf{.99}_{}$ & $\textbf{.99}_{}$ & $\textbf{.99}_{}$ & $\textit{.97}_{}$ & $.95_{}$ & $.93_{}$ & $.95_{}$ & $.95_{}$ & $.62_{20}$ & $.73_{101}$ & $.66_{82}$ & $.73_{73}$ & $.68_{222}$ & $.67_{69}$ & $.85_{24}$ & $.85_{222}$ & $.89_{55}$ & $.73_{88}$ & $.81_{73}$ & $.71_{36}$ \\
\texttt{yeast} & $.57_{}$ & $\textit{.60}_{}$ & $\textbf{.61}_{}$ & $.52_{}$ & $.51_{}$ & $.51_{}$ & $.53_{}$ & $.52_{}$ & $.53_{}$ & $.23_{1}$ & $.31_{21}$ & $.33_{13}$ & $.35_{15}$ & $.34_{9}$ & $.32_{7}$ & $.23_{1}$ & $.20_{3}$ & $.22_{1}$ & $.20_{2}$ & $.21_{1}$ & $.27_{7}$ \\
\texttt{magic} & $\textbf{.88}_{}$ & $\textbf{.88}_{}$ & $\textit{.87}_{}$ & $.70_{}$ & $.70_{}$ & $.70_{}$ & $.69_{}$ & $.69_{}$ & $.67_{}$ & $.75_{33}$ & $.75_{53}$ & $.76_{44}$ & $.76_{66}$ & $.78_{180}$ & $.76_{78}$ & $.48_{14}$ & $.55_{39}$ & $.65_{14}$ & $.63_{22}$ & $.51_{51}$ & $.60_{28}$ \\
\texttt{sonar} & $\textit{.87}_{}$ & $\textit{.87}_{}$ & $.84_{}$ & $\textit{.87}_{}$ & $.84_{}$ & $\textbf{.90}_{}$ & $.86_{}$ & $.86_{}$ & $.86_{}$ & $.62_{128}$ & $.63_{21}$ & $.59_{38}$ & $.57_{28}$ & $.60_{437}$ & $.54_{97}$ & $.60_{153}$ & $.67_{194}$ & $.67_{126}$ & $.74_{73}$ & $.68_{106}$ & $.67_{97}$ \\
\texttt{compas} & $\textit{.63}_{}$ & $\textit{.63}_{}$ & $\textbf{.64}_{}$ & $.50_{}$ & $.50_{}$ & $.50_{}$ & $.49_{}$ & $.47_{}$ & $.48_{}$ & $.47_{7}$ & $.50_{45}$ & $.51_{38}$ & $.50_{29}$ & $.50_{59}$ & $.51_{48}$ & $.38_{22}$ & $.38_{69}$ & $.38_{95}$ & $.38_{89}$ & $.38_{53}$ & $.38_{47}$ \\
\texttt{house} & $\textit{.89}_{}$ & $\textbf{.90}_{}$ & $\textbf{.90}_{}$ & $.81_{}$ & $.82_{}$ & $.81_{}$ & $.74_{}$ & $.76_{}$ & $.75_{}$ & $.80_{59}$ & $.80_{179}$ & $.79_{187}$ & $.79_{262}$ & $.77_{406}$ & $.77_{392}$ & $.58_{1}$ & $.58_{3}$ & $.58_{5}$ & $.58_{116}$ & $.58_{114}$ & $.58_{126}$ \\
\texttt{german} & $.75_{}$ & $\textbf{.77}_{}$ & $\textit{.76}_{}$ & $.57_{}$ & $.57_{}$ & $.57_{}$ & $.57_{}$ & $.57_{}$ & $.57_{}$ & $.64_{7}$ & $.70_{20}$ & $.70_{13}$ & $.70_{24}$ & $.69_{9}$ & $.70_{24}$ & $.57_{16}$ & $.57_{28}$ & $.57_{45}$ & $.57_{24}$ & $.57_{10}$ & $.57_{9}$ \\
\texttt{spamb} & $\textbf{.96}_{}$ & $\textit{.95}_{}$ & $\textit{.95}_{}$ & $.88_{}$ & $.88_{}$ & $.89_{}$ & $.87_{}$ & $.86_{}$ & $.85_{}$ & $.79_{40}$ & $.81_{120}$ & $.80_{221}$ & $.80_{228}$ & $.81_{281}$ & $.81_{214}$ & $.79_{50}$ & $.73_{96}$ & $.74_{103}$ & $.67_{99}$ & $.68_{119}$ & $.68_{216}$ \\
\texttt{norm} & $\textit{.97}_{}$ & $\textbf{.98}_{}$ & $\textbf{.98}_{}$ & $.96_{}$ & $\textit{.97}_{}$ & $\textbf{.98}_{}$ & $\textbf{.98}_{}$ & $\textbf{.98}_{}$ & $\textbf{.98}_{}$ & $.80_{181}$ & $.81_{128}$ & $.81_{90}$ & $.82_{105}$ & $.81_{101}$ & $.81_{101}$ & $.94_{188}$ & $\textit{.97}_{104}$ & $\textit{.97}_{105}$ & $\textit{.97}_{105}$ & $\textbf{.98}_{99}$ & $\textit{.97}_{106}$ \\
\texttt{lrs} & $\textbf{.85}_{}$ & $\textbf{.85}_{}$ & $\textit{.83}_{}$ & $.80_{}$ & $.80_{}$ & $.75_{}$ & $.77_{}$ & $.79_{}$ & $.80_{}$ & $.60_{357}$ & $.57_{297}$ & $.54_{154}$ & $.54_{188}$ & $.57_{174}$ & $.54_{179}$ & $.37_{187}$ & $.41_{285}$ & $.42_{72}$ & $.44_{116}$ & $.51_{81}$ & $.38_{87}$ \\
\texttt{vert} & $\textit{.83}_{}$ & $\textbf{.85}_{}$ & $.82_{}$ & $.75_{}$ & $.78_{}$ & $.75_{}$ & $.75_{}$ & $.78_{}$ & $.79_{}$ & $.68_{14}$ & $.67_{80}$ & $.65_{18}$ & $.68_{15}$ & $.64_{47}$ & $.67_{31}$ & $.71_{14}$ & $.62_{22}$ & $.70_{9}$ & $.67_{24}$ & $.61_{25}$ & $.61_{21}$ \\
\texttt{iris} & $\textbf{1.00}_{}$ & $\textbf{1.00}_{}$ & $\textbf{1.00}_{}$ & $\textit{.98}_{}$ & $\textit{.98}_{}$ & $\textbf{1.00}_{}$ & $\textit{.98}_{}$ & $\textbf{1.00}_{}$ & $.96_{}$ & $.81_{11}$ & $.75_{13}$ & $.73_{13}$ & $.72_{15}$ & $.76_{12}$ & $.73_{11}$ & $.28_{14}$ & $.31_{30}$ & $.70_{11}$ & $.64_{13}$ & $.34_{12}$ & $.83_{2}$ \\
\texttt{wine} & $\textbf{1.00}_{}$ & $\textbf{1.00}_{}$ & $\textit{.99}_{}$ & $.98_{}$ & $.98_{}$ & $\textit{.99}_{}$ & $.98_{}$ & $.98_{}$ & $.98_{}$ & $.91_{81}$ & $.91_{190}$ & $.91_{101}$ & $.90_{129}$ & $.91_{151}$ & $.92_{92}$ & $.69_{52}$ & $.65_{342}$ & $.63_{48}$ & $.65_{55}$ & $.63_{106}$ & $.65_{111}$ \\
\texttt{diva} & $\textit{.92}_{}$ & $\textit{.92}_{}$ & $\textbf{.93}_{}$ & $.76_{}$ & $.76_{}$ & $.71_{}$ & $.67_{}$ & $.65_{}$ & $.64_{}$ & $.79_{34}$ & $.79_{46}$ & $.79_{125}$ & $.78_{128}$ & $.80_{98}$ & $.80_{175}$ & $.63_{41}$ & $.63_{42}$ & $.63_{52}$ & $.63_{45}$ & $.63_{41}$ & $.63_{49}$ \\
\texttt{breast} & $\textit{.97}_{}$ & $\textit{.97}_{}$ & $\textbf{.98}_{}$ & $.96_{}$ & $\textit{.97}_{}$ & $\textit{.97}_{}$ & $.95_{}$ & $.96_{}$ & $\textit{.97}_{}$ & $.96_{113}$ & $.96_{53}$ & $.95_{103}$ & $.93_{50}$ & $.94_{106}$ & $.95_{28}$ & $.89_{114}$ & $.94_{348}$ & $.95_{103}$ & $.91_{199}$ & $.93_{60}$ & $.92_{88}$ \\
\texttt{steel} & $.79_{}$ & $\textbf{.81}_{}$ & $\textit{.80}_{}$ & $.63_{}$ & $.64_{}$ & $.62_{}$ & $.57_{}$ & $.60_{}$ & $.62_{}$ & $.52_{127}$ & $.54_{142}$ & $.52_{44}$ & $.53_{49}$ & $.55_{206}$ & $.54_{97}$ & $.23_{49}$ & $.26_{25}$ & $.26_{36}$ & $.29_{3}$ & $.20_{21}$ & $.33_{8}$ \\
\texttt{ecoli} & $\textbf{.85}_{}$ & $\textit{.83}_{}$ & $\textit{.83}_{}$ & $.77_{}$ & $.73_{}$ & $.73_{}$ & $.73_{}$ & $.72_{}$ & $.72_{}$ & $.61_{44}$ & $.66_{29}$ & $.64_{29}$ & $.58_{33}$ & $.61_{57}$ & $.62_{23}$ & $.49_{18}$ & $.49_{21}$ & $.47_{27}$ & $.50_{28}$ & $.46_{19}$ & $.50_{21}$ \\
\texttt{heloc} & $\textbf{.71}_{}$ & $\textbf{.71}_{}$ & $\textbf{.71}_{}$ & $\textit{.70}_{}$ & $\textit{.70}_{}$ & $\textbf{.71}_{}$ & $\textit{.70}_{}$ & $\textit{.70}_{}$ & $\textit{.70}_{}$ & $\textit{.70}_{40}$ & $\textit{.70}_{58}$ & $.68_{133}$ & $.69_{124}$ & $\textit{.70}_{88}$ & $.69_{92}$ & $.51_{33}$ & $.55_{55}$ & $.56_{106}$ & $.50_{27}$ & $.51_{142}$ & $.45_{104}$ \\
\texttt{page} & $\textbf{.97}_{}$ & $\textbf{.97}_{}$ & $\textbf{.97}_{}$ & $.95_{}$ & $.95_{}$ & $.95_{}$ & $.93_{}$ & $.93_{}$ & $.93_{}$ & $.92_{112}$ & $.95_{153}$ & $.93_{93}$ & $.95_{92}$ & $\textit{.96}_{143}$ & $\textit{.96}_{149}$ & $.85_{41}$ & $.85_{13}$ & $.86_{74}$ & $.85_{53}$ & $.84_{7}$ & $.84_{0}$ \\
\texttt{yoga} & $\textbf{.79}_{}$ & $\textit{.77}_{}$ & $\textbf{.79}_{}$ & $.73_{}$ & $.72_{}$ & $.75_{}$ & $.74_{}$ & $.73_{}$ & $\textit{.77}_{}$ & $.60_{126}$ & $.58_{68}$ & $.61_{21}$ & $.59_{27}$ & $.70_{110}$ & $.67_{83}$ & $.50_{66}$ & $.45_{4}$ & $.52_{10}$ & $.45_{8}$ & $.44_{300}$ & $.38_{242}$ \\
\texttt{star} & $\textbf{.95}_{}$ & $\textbf{.95}_{}$ & $\textit{.94}_{}$ & $.90_{}$ & $.90_{}$ & $.91_{}$ & $.89_{}$ & $.83_{}$ & $.79_{}$ & $.79_{444}$ & $.80_{99}$ & $.80_{147}$ & $.81_{218}$ & $.80_{139}$ & $.79_{113}$ & $.41_{127}$ & $.42_{121}$ & $.49_{104}$ & $.41_{104}$ & $.70_{38}$ & $.47_{39}$ \\
\texttt{chlorine} & $.06_{}$ & $.08_{}$ & $.09_{}$ & $.08_{}$ & $.08_{}$ & $.08_{}$ & $.07_{}$ & $.08_{}$ & $.08_{}$ & $.12_{20}$ & $.17_{38}$ & $.17_{38}$ & $.10_{19}$ & $\textit{.21}_{645}$ & $\textbf{.22}_{384}$ & $.13_{34}$ & $.13_{42}$ & $.08_{47}$ & $.08_{32}$ & $.09_{29}$ & $.09_{37}$ \\
\texttt{kitchen} & $\textit{.70}_{}$ & $.68_{}$ & $.69_{}$ & $\textit{.70}_{}$ & $\textit{.70}_{}$ & $\textbf{.73}_{}$ & $\textbf{.73}_{}$ & $.61_{}$ & $.61_{}$ & $.63_{166}$ & $.69_{372}$ & $.69_{121}$ & $.67_{125}$ & $.61_{265}$ & $.62_{219}$ & $.41_{417}$ & $.40_{181}$ & $.40_{79}$ & $.33_{64}$ & $.53_{66}$ & $.46_{48}$ \\
\texttt{share} & $\textbf{.64}_{}$ & $\textit{.63}_{}$ & $\textit{.63}_{}$ & $.56_{}$ & $.56_{}$ & $.56_{}$ & $.56_{}$ & $.56_{}$ & $.56_{}$ & $.61_{136}$ & $.59_{196}$ & $.61_{157}$ & $.59_{221}$ & $.59_{53}$ & $.60_{85}$ & $.58_{140}$ & $.56_{15}$ & $.56_{367}$ & $.56_{47}$ & $.56_{11}$ & $.56_{24}$ \\
\texttt{devices} & $\textbf{.67}_{}$ & $\textit{.66}_{}$ & $.63_{}$ & $.43_{}$ & $.43_{}$ & $.42_{}$ & $.43_{}$ & $.40_{}$ & $.40_{}$ & $.45_{689}$ & $.44_{744}$ & $.41_{815}$ & $.44_{789}$ & $.44_{846}$ & $.44_{803}$ & $.27_{172}$ & $.26_{198}$ & $.23_{102}$ & $.22_{170}$ & $.20_{418}$ & $.18_{248}$ \\
\texttt{gun} & $.87_{}$ & $.63_{}$ & $\textit{.91}_{}$ & $\textbf{.93}_{}$ & $\textit{.91}_{}$ & $.85_{}$ & $.87_{}$ & $.86_{}$ & $.85_{}$ & $.69_{97}$ & $.67_{131}$ & $.69_{33}$ & $.67_{46}$ & $.67_{58}$ & $.67_{45}$ & $.70_{85}$ & $.57_{92}$ & $.77_{61}$ & $.76_{19}$ & $.53_{43}$ & $.45_{53}$ \\
\texttt{worms} & $\textit{.60}_{}$ & $\textbf{.62}_{}$ & $\textbf{.62}_{}$ & $\textit{.60}_{}$ & $.53_{}$ & $\textbf{.62}_{}$ & $.56_{}$ & $\textit{.60}_{}$ & $.59_{}$ & $.49_{65}$ & $.51_{64}$ & $.56_{116}$ & $.50_{34}$ & $.56_{44}$ & $.55_{316}$ & $.56_{8}$ & $.45_{9}$ & $.58_{24}$ & $.46_{34}$ & $.48_{65}$ & $.49_{109}$ \\
\texttt{ecg} & $\textbf{.93}_{}$ & $\textbf{.93}_{}$ & $\textbf{.93}_{}$ & $.91_{}$ & $.91_{}$ & $\textit{.92}_{}$ & $.91_{}$ & $.91_{}$ & $.91_{}$ & $.88_{181}$ & $.89_{184}$ & $.87_{132}$ & $.88_{123}$ & $.89_{84}$ & $.88_{62}$ & $.67_{103}$ & $.60_{125}$ & $.47_{16}$ & $.64_{16}$ & $.68_{9}$ & $.64_{14}$ \\
\texttt{wafer} & $\textbf{.99}_{}$ & $\textbf{.99}_{}$ & $\textbf{.99}_{}$ & $\textit{.97}_{}$ & $\textit{.97}_{}$ & $\textbf{.99}_{}$ & $\textbf{.99}_{}$ & $\textbf{.99}_{}$ & $\textbf{.99}_{}$ & $.94_{133}$ & $.96_{168}$ & $.92_{44}$ & $.94_{41}$ & $.96_{516}$ & $\textit{.97}_{562}$ & $.84_{206}$ & $.84_{220}$ & $.84_{4}$ & $.84_{131}$ & $.84_{126}$ & $.84_{85}$ \\
\texttt{oral} & $\textit{.88}_{}$ & $\textit{.88}_{}$ & $.85_{}$ & $.86_{}$ & $.84_{}$ & $\textbf{.89}_{}$ & $.85_{}$ & $\textit{.88}_{}$ & $\textbf{.89}_{}$ & $.65_{237}$ & $.71_{350}$ & $.70_{394}$ & $.64_{280}$ & $.78_{403}$ & $.70_{289}$ & $.29_{122}$ & $.20_{32}$ & $.33_{102}$ & $.45_{36}$ & $.18_{403}$ & $.32_{183}$ \\
\texttt{MNIST} & $\textbf{.99}_{}$ & $\textbf{.99}_{}$ & $\textbf{.99}_{}$ & $\textbf{.99}_{}$ & $\textbf{.99}_{}$ & $.55_{}$ & $.55_{}$ & $.88_{}$ & $\textit{.89}_{}$ & $\textbf{.99}_{58}$ & $\textbf{.99}_{71}$ & $\textbf{.99}_{185}$ & $\textbf{.99}_{169}$ & $\textbf{.99}_{217}$ & $\textbf{.99}_{73}$ & $.05_{466}$ & $.10_{96}$ & $.02_{410}$ & $.02_{390}$ & $.11_{205}$ & $.03_{68}$ \\
\texttt{cifar10} & $\textbf{.95}_{}$ & $\textbf{.95}_{}$ & $\textbf{.95}_{}$ & $\textbf{.95}_{}$ & $\textbf{.95}_{}$ & $.62_{}$ & $.72_{}$ & $\textbf{.95}_{}$ & $\textbf{.95}_{}$ & $\textit{.94}_{145}$ & $\textbf{.95}_{51}$ & $.14_{3}$ & $.15_{2}$ & $\textit{.94}_{92}$ & $\textit{.94}_{67}$ & $.11_{28}$ & $.14_{82}$ & $.06_{3}$ & $.19_{17}$ & $.11_{15}$ & $.07_{1}$ \\
\texttt{catsdogs} & $\textbf{.95}_{}$ & $\textbf{.95}_{}$ & $\textbf{.95}_{}$ & $\textbf{.95}_{}$ & $\textbf{.95}_{}$ & $\textbf{.95}_{}$ & $\textbf{.95}_{}$ & $\textbf{.95}_{}$ & $\textbf{.95}_{}$ & $\textbf{.95}_{51}$ & $\textbf{.95}_{55}$ & $\textbf{.95}_{100}$ & $\textbf{.95}_{45}$ & $\textbf{.95}_{15}$ & $\textbf{.95}_{19}$ & $\textbf{.95}_{97}$ & $\textbf{.95}_{100}$ & $\textbf{.95}_{100}$ & $\textbf{.95}_{100}$ & $.85_{17}$ & $\textit{.93}_{46}$ \\
\texttt{birds} & $.88_{}$ & $\textbf{.90}_{}$ & $\textbf{.90}_{}$ & $\textit{.89}_{}$ & $\textbf{.90}_{}$ & $\textit{.89}_{}$ & $.88_{}$ & $\textit{.89}_{}$ & $\textit{.89}_{}$ & $\textit{.89}_{24}$ & $\textit{.89}_{22}$ & $.87_{118}$ & $.86_{28}$ & $.88_{119}$ & $.88_{55}$ & $.88_{115}$ & $.86_{119}$ & $.84_{105}$ & $.76_{28}$ & $.86_{42}$ & $.86_{28}$ \\
\texttt{pets} & $\textit{.94}_{}$ & $\textit{.94}_{}$ & $\textit{.94}_{}$ & $\textit{.94}_{}$ & $\textit{.94}_{}$ & $\textbf{.95}_{}$ & $\textit{.94}_{}$ & $\textit{.94}_{}$ & $\textit{.94}_{}$ & $\textit{.94}_{235}$ & $\textit{.94}_{225}$ & $\textit{.94}_{167}$ & $\textit{.94}_{91}$ & $\textit{.94}_{47}$ & $\textit{.94}_{271}$ & $.55_{152}$ & $.55_{2}$ & $.64_{4}$ & $.64_{18}$ & $.93_{1}$ & $.55_{125}$ \\
\texttt{organa} & $\textbf{1.00}_{}$ & $\textbf{1.00}_{}$ & $\textbf{1.00}_{}$ & $.98_{}$ & $.98_{}$ & $.75_{}$ & $.76_{}$ & $.98_{}$ & $.97_{}$ & $\textit{.99}_{590}$ & $\textit{.99}_{985}$ & $.89_{203}$ & $.27_{4}$ & $.97_{83}$ & $.98_{105}$ & $.07_{1096}$ & $.08_{12}$ & $.20_{105}$ & $.17_{88}$ & $.09_{1}$ & $.06_{10}$ \\
\texttt{blood} & $\textbf{.95}_{}$ & $\textbf{.95}_{}$ & $\textit{.94}_{}$ & $.92_{}$ & $.92_{}$ & $.69_{}$ & $.68_{}$ & $.93_{}$ & $.90_{}$ & $.93_{1160}$ & $.93_{634}$ & $.89_{377}$ & $.89_{341}$ & $.93_{453}$ & $.93_{600}$ & $.14_{1176}$ & $.14_{1257}$ & $.05_{273}$ & $.05_{224}$ & $.07_{79}$ & $.09_{43}$ \\
\texttt{SVHN} & $\textit{.95}_{}$ & $\textit{.95}_{}$ & $\textbf{.96}_{}$ & $\textit{.95}_{}$ & $\textit{.95}_{}$ & $.57_{}$ & $.57_{}$ & $\textit{.95}_{}$ & $\textit{.95}_{}$ & $\textit{.95}_{29}$ & $\textit{.95}_{123}$ & $\textit{.95}_{151}$ & $.93_{131}$ & $\textit{.95}_{80}$ & $\textit{.95}_{87}$ & $.06_{55}$ & $.15_{2}$ & $.23_{101}$ & $.13_{103}$ & $.08_{34}$ & $.07_{24}$ \\
\texttt{medabs} & $\textbf{.61}_{}$ & $\textbf{.61}_{}$ & $\textbf{.61}_{}$ & $.47_{}$ & $.47_{}$ & $.49_{}$ & $.47_{}$ & $.49_{}$ & $.48_{}$ & $.53_{240}$ & $.53_{274}$ & $.52_{185}$ & $.53_{180}$ & $\textit{.55}_{170}$ & $\textit{.55}_{169}$ & $.17_{240}$ & $.17_{252}$ & $.20_{4}$ & $.17_{9}$ & $.17_{27}$ & $.17_{19}$ \\
\texttt{vicuna} & $\textbf{.87}_{}$ & $\textbf{.87}_{}$ & $\textit{.86}_{}$ & $.54_{}$ & $.54_{}$ & $.63_{}$ & $.43_{}$ & $.44_{}$ & $.43_{}$ & $.65_{286}$ & $.66_{308}$ & $.66_{362}$ & $.66_{160}$ & $.67_{149}$ & $.67_{302}$ & $.42_{12}$ & $.42_{12}$ & $.42_{139}$ & $.42_{70}$ & $.42_{14}$ & $.42_{80}$ \\
\texttt{pTED} & $\textbf{.79}_{}$ & $.77_{}$ & $\textit{.78}_{}$ & $.49_{}$ & $.49_{}$ & $.53_{}$ & $.46_{}$ & $.47_{}$ & $.46_{}$ & $.64_{282}$ & $.63_{282}$ & $.63_{182}$ & $.62_{254}$ & $.64_{130}$ & $.61_{112}$ & $.46_{151}$ & $.46_{153}$ & $.46_{174}$ & $.46_{64}$ & $.46_{60}$ & $.46_{49}$ \\
\texttt{tgpt} & $\textbf{.97}_{}$ & $\textbf{.97}_{}$ & $\textit{.96}_{}$ & $.92_{}$ & $.91_{}$ & $.92_{}$ & $.89_{}$ & $.93_{}$ & $.91_{}$ & $.78_{126}$ & $.77_{155}$ & $.78_{123}$ & $.81_{332}$ & $.80_{239}$ & $.78_{132}$ & $.57_{91}$ & $.75_{103}$ & $.57_{94}$ & $.44_{51}$ & $.85_{99}$ & $.45_{93}$ \\
\texttt{pol} & $\textbf{.73}_{}$ & $\textbf{.73}_{}$ & $.70_{}$ & $.71_{}$ & $.71_{}$ & $\textit{.72}_{}$ & $.70_{}$ & $.71_{}$ & $.71_{}$ & $.60_{72}$ & $.55_{92}$ & $.59_{63}$ & $.57_{51}$ & $.56_{91}$ & $.58_{66}$ & $.54_{72}$ & $.49_{97}$ & $.57_{106}$ & $.58_{110}$ & $.50_{39}$ & $.42_{66}$ \\
\texttt{liar} & $\textbf{.26}_{}$ & $\textit{.25}_{}$ & $\textit{.25}_{}$ & $.18_{}$ & $.17_{}$ & $.18_{}$ & $.17_{}$ & $.18_{}$ & $.17_{}$ & $.20_{100}$ & $.21_{86}$ & $.20_{131}$ & $.21_{195}$ & $.18_{102}$ & $.19_{109}$ & $.10_{16}$ & $.10_{25}$ & $.07_{422}$ & $.11_{274}$ & $.08_{31}$ & $.07_{554}$ \\
\midrule
\texttt{Avg.} & $\textbf{.83}_{}$ & $\textit{.82}_{}$ & $\textbf{.83}_{}$ & $.77_{}$ & $.76_{}$ & $.74_{}$ & $.72_{}$ & $.75_{}$ & $.75_{}$ & $.71_{158.47}$ & $.72_{168.31}$ & $.69_{132.53}$ & $.68_{127.02}$ & $.72_{180.27}$ & $.72_{158.22}$ & $.50_{139.78}$ & $.49_{119.73}$ & $.52_{90.20}$ & $.50_{\textit{77.67}}$ & $.50_{\textbf{75.84}}$ & $.48_{78.22}$ \\
\texttt{Std.} & $.19_{}$ & $.19_{}$ & $.19_{}$ & $.22_{}$ & $.22_{}$ & $.21_{}$ & $.22_{}$ & $.22_{}$ & $.22_{}$ & $.21_{212.61}$ & $.20_{195.44}$ & $.21_{138.94}$ & $.22_{136.65}$ & $.20_{175.86}$ & $.20_{167.23}$ & $\textit{.26}_{238.70}$ & ${.26}_{196.27}$ & ${.26}_{99.75}$ & $.25_{{76.54}}$ & ${.27}_{{92.89}}$ & ${.26}_{94.75}$ \\
\texttt{Rank} & $3.7_{10.1}$ & $3.6_{10.1}$ & $3.5_{10.1}$ & $7.9_{10.1}$ & $8.1_{10.1}$ & $7.7_{10.1}$ & $10.1_{10.1}$ & $8.9_{10.1}$ & $9.3_{10.1}$ & $10.9_{10.9}$ & $10.0_{9.1}$ & $11.3_{9.5}$ & $11.7_{10.5}$ & $10.3_{8.1}$ & $10.3_{10.4}$ & $16.8_{12.6}$ & $17.3_{12.0}$ & $16.6_{12.7}$ & $17.3_{13.9}$ & $17.6_{15.2}$ & $18.1_{15.1}$\\
\bottomrule
\end{tabular}
\end{sidewaystable}

\begin{sidewaystable}[t]
\caption{Average Weighted F1-Score for \textsc{RandomPivotForest} with $100$ \textsc{PivotTree} \textsc{c}lassifiers as estimators, and 10\% of the training set is used to train each estimator, against baselines, for each \texttt{dataset}.  When the splitting stump forest is adopted, subscripts indicate the average number of stumps $\pm$ std. dev. Best results in \textbf{bold}, second best in \textit{italics}.}
    \label{tab:ensemble_sample_subset_f1_score}
\centering
\footnotesize
\setlength{\tabcolsep}{0.3mm} % Reduce column spacing
\begin{tabular}{l|lll|ll|llll|ll|llll|ll|llll}
\toprule
\texttt{dataset} & $\textsc{xgb}$ & $\textsc{lgbm}$ & $\textsc{catb}$ & $\textsc{rdt}$ & $\textsc{rodt}$ & $\textsc{rpt}$ & $\textsc{ropt}$ & $\textsc{rppt}$ & $\textsc{roppt}$ & $\textsc{rdt}_{\textsc{s}}$ & $\textsc{rodt}_{\textsc{s}}$ & $\textsc{rpt}_{\textsc{s}}$ & $\textsc{ropt}_{\textsc{s}}$ & $\textsc{rppt}_{\textsc{s}}$ & $\textsc{roppt}_{\textsc{s}}$ & $\textsc{rdt}_{\textsc{b}}$ & $\textsc{rodt}_{\textsc{b}}$ & $\textsc{rpt}_{\textsc{b}}$ & $\textsc{ropt}_{\textsc{b}}$ & $\textsc{rppt}_{\textsc{b}}$ & $\textsc{roppt}_{\textsc{b}}$ \\
\midrule
\texttt{ion} & $\textit{.93}_{}$ & $\textbf{.95}_{}$ & $\textbf{.95}_{}$ & $.90_{}$ & $.91_{}$ & $.89_{}$ & $.91_{}$ & $.86_{}$ & $.86_{}$ & $.86_{46}$ & $.86_{41}$ & $.89_{37}$ & $.88_{36}$ & $.91_{72}$ & $.91_{103}$ & $.50_{\textbf{17}}$ & $.50_{74}$ & $.73_{107}$ & $.71_{99}$ & $.69_{\textit{26}}$ & $.62_{40}$ \\
\texttt{fire} & $\textbf{.99}_{}$ & $\textbf{.99}_{}$ & $\textbf{.99}_{}$ & $\textit{.96}_{}$ & $\textit{.96}_{}$ & $.93_{}$ & $.92_{}$ & $.93_{}$ & $.93_{}$ & $.58_{\textbf{14}}$ & $.69_{\textit{53}}$ & $.69_{80}$ & $.71_{95}$ & $.74_{83}$ & $.71_{109}$ & $.89_{98}$ & $.90_{101}$ & $.73_{107}$ & $.82_{99}$ & $.78_{109}$ & $.72_{73}$ \\
\texttt{yeast} & $.56_{}$ & $\textit{.57}_{}$ & $\textbf{.58}_{}$ & $.52_{}$ & $.48_{}$ & $.53_{}$ & $.49_{}$ & $.52_{}$ & $.52_{}$ & $.34_{12}$ & $.34_{51}$ & $.33_{81}$ & $.35_{46}$ & $.33_{59}$ & $.35_{62}$ & $.25_{21}$ & $.18_{\textit{3}}$ & $.20_{11}$ & $.27_{36}$ & $.14_{\textbf{2}}$ & $.19_{91}$ \\
\texttt{magic} & $\textit{.87}_{}$ & $\textbf{.88}_{}$ & $\textit{.87}_{}$ & $.70_{}$ & $.70_{}$ & $.70_{}$ & $.69_{}$ & $.69_{}$ & $.68_{}$ & $.73_{53}$ & $.75_{57}$ & $.75_{44}$ & $.75_{67}$ & $.78_{267}$ & $.75_{68}$ & $.56_{34}$ & $.59_{57}$ & $.67_{\textbf{10}}$ & $.65_{\textit{25}}$ & $.51_{91}$ & $.53_{30}$ \\
\texttt{sonar} & $.78_{}$ & $\textbf{.87}_{}$ & $\textit{.84}_{}$ & $.77_{}$ & $.76_{}$ & $.79_{}$ & $.65_{}$ & $\textit{.84}_{}$ & $.81_{}$ & $.62_{102}$ & $.59_{123}$ & $.65_{53}$ & $.65_{\textit{10}}$ & $.59_{\textbf{8}}$ & $.59_{14}$ & $.74_{100}$ & $.61_{76}$ & $.63_{83}$ & $.62_{\textit{10}}$ & $.60_{151}$ & $.55_{14}$ \\
\texttt{compas} & $\textbf{.64}_{}$ & $\textit{.63}_{}$ & $\textit{.63}_{}$ & $.48_{}$ & $.48_{}$ & $.51_{}$ & $.50_{}$ & $.49_{}$ & $.49_{}$ & $.47_{\textbf{11}}$ & $.53_{49}$ & $.48_{40}$ & $.50_{57}$ & $.49_{58}$ & $.52_{56}$ & $.38_{\textit{36}}$ & $.38_{70}$ & $.38_{96}$ & $.38_{89}$ & $.38_{71}$ & $.38_{85}$ \\
\texttt{house} & $\textit{.89}_{}$ & $\textbf{.90}_{}$ & $\textit{.89}_{}$ & $.81_{}$ & $.81_{}$ & $.80_{}$ & $.75_{}$ & $.75_{}$ & $.75_{}$ & $.80_{179}$ & $.80_{223}$ & $.78_{187}$ & $.77_{209}$ & $.77_{415}$ & $.78_{455}$ & $.58_{158}$ & $.58_{155}$ & $.59_{\textbf{9}}$ & $.58_{\textit{109}}$ & $.58_{115}$ & $.58_{123}$ \\
\texttt{german} & $.73_{}$ & $\textbf{.77}_{}$ & $\textit{.76}_{}$ & $.57_{}$ & $.57_{}$ & $.57_{}$ & $.57_{}$ & $.57_{}$ & $.57_{}$ & $.64_{\textit{7}}$ & $.70_{15}$ & $.70_{42}$ & $.70_{39}$ & $.70_{37}$ & $.70_{21}$ & $.57_{10}$ & $.57_{10}$ & $.57_{8}$ & $.57_{11}$ & $.57_{\textbf{2}}$ & $.57_{41}$ \\
\texttt{spamb} & $\textbf{.95}_{}$ & $\textbf{.95}_{}$ & $\textbf{.95}_{}$ & $.87_{}$ & $.87_{}$ & $\textit{.88}_{}$ & $.87_{}$ & $.86_{}$ & $.85_{}$ & $.79_{\textbf{56}}$ & $.81_{99}$ & $.79_{116}$ & $.78_{87}$ & $.81_{108}$ & $.82_{175}$ & $.75_{122}$ & $.80_{98}$ & $.73_{\textit{86}}$ & $.54_{116}$ & $.71_{123}$ & $.71_{99}$ \\
\texttt{norm} & $\textit{.97}_{}$ & $\textbf{.98}_{}$ & $\textit{.97}_{}$ & $\textit{.97}_{}$ & $\textit{.97}_{}$ & $\textbf{.98}_{}$ & $\textit{.97}_{}$ & $\textbf{.98}_{}$ & $\textbf{.98}_{}$ & $.81_{222}$ & $.81_{\textit{49}}$ & $.81_{71}$ & $.82_{\textbf{45}}$ & $.81_{63}$ & $.81_{68}$ & $.95_{226}$ & $\textit{.97}_{118}$ & $\textit{.97}_{100}$ & $\textbf{.98}_{87}$ & $\textbf{.98}_{95}$ & $\textbf{.98}_{106}$ \\
\texttt{lrs} & $\textit{.81}_{}$ & $\textbf{.88}_{}$ & $\textit{.81}_{}$ & $.80_{}$ & $.79_{}$ & $.75_{}$ & $.75_{}$ & $.76_{}$ & $.79_{}$ & $.57_{252}$ & $.57_{237}$ & $.55_{149}$ & $.53_{137}$ & $.56_{\textit{80}}$ & $.55_{\textbf{57}}$ & $.38_{280}$ & $.40_{186}$ & $.45_{134}$ & $.42_{165}$ & $.35_{93}$ & $.35_{172}$ \\
\texttt{vert} & $\textbf{.86}_{}$ & $\textit{.85}_{}$ & $.84_{}$ & $.76_{}$ & $.76_{}$ & $.75_{}$ & $.73_{}$ & $.78_{}$ & $.78_{}$ & $.66_{\textit{21}}$ & $.62_{66}$ & $.62_{68}$ & $.62_{\textbf{18}}$ & $.65_{151}$ & $.62_{46}$ & $.63_{75}$ & $.63_{86}$ & $.60_{106}$ & $.63_{104}$ & $.65_{152}$ & $.71_{111}$ \\
\texttt{iris} & $\textit{.98}_{}$ & $\textbf{1.00}_{}$ & $\textbf{1.00}_{}$ & $\textbf{1.00}_{}$ & $\textit{.98}_{}$ & $.96_{}$ & $\textbf{1.00}_{}$ & $\textbf{1.00}_{}$ & $\textbf{1.00}_{}$ & $.82_{\textbf{18}}$ & $.71_{113}$ & $.71_{38}$ & $.70_{35}$ & $.75_{59}$ & $.73_{\textit{19}}$ & $.69_{123}$ & $.45_{53}$ & $.87_{38}$ & $.89_{36}$ & $.65_{20}$ & $.89_{\textit{19}}$ \\
\texttt{wine} & $\textit{.99}_{}$ & $\textbf{1.00}_{}$ & $\textit{.99}_{}$ & $.98_{}$ & $.98_{}$ & $.98_{}$ & $.98_{}$ & $.98_{}$ & $.98_{}$ & $.91_{89}$ & $.91_{88}$ & $.91_{129}$ & $.91_{102}$ & $.91_{211}$ & $.91_{136}$ & $.69_{\textit{82}}$ & $.65_{94}$ & $.69_{\textbf{11}}$ & $.64_{133}$ & $.63_{84}$ & $.63_{102}$ \\
\texttt{diva} & $\textit{.91}_{}$ & $\textbf{.92}_{}$ & $\textit{.91}_{}$ & $.74_{}$ & $.74_{}$ & $.70_{}$ & $.67_{}$ & $.65_{}$ & $.64_{}$ & $.80_{\textbf{34}}$ & $.80_{53}$ & $.78_{74}$ & $.79_{69}$ & $.80_{133}$ & $.80_{117}$ & $.63_{\textbf{34}}$ & $.63_{\textbf{34}}$ & $.63_{75}$ & $.63_{54}$ & $.63_{\textit{40}}$ & $.63_{46}$ \\
\texttt{breast} & $\textbf{.99}_{}$ & $.95_{}$ & $\textit{.98}_{}$ & $.96_{}$ & $.96_{}$ & $.95_{}$ & $.94_{}$ & $.95_{}$ & $.95_{}$ & $.95_{53}$ & $.95_{64}$ & $.94_{92}$ & $.95_{24}$ & $.94_{67}$ & $.94_{\textit{19}}$ & $.93_{62}$ & $.88_{\textbf{12}}$ & $.91_{115}$ & $.92_{89}$ & $.91_{100}$ & $.90_{95}$ \\
\texttt{steel} & $.74_{}$ & $\textbf{.80}_{}$ & $\textit{.77}_{}$ & $.50_{}$ & $.49_{}$ & $.57_{}$ & $.56_{}$ & $.57_{}$ & $.59_{}$ & $.53_{88}$ & $.54_{133}$ & $.53_{62}$ & $.53_{75}$ & $.53_{270}$ & $.57_{205}$ & $.28_{68}$ & $.20_{11}$ & $.21_{\textit{6}}$ & $.31_{\textbf{2}}$ & $.20_{11}$ & $.37_{8}$ \\
\texttt{ecoli} & $.82_{}$ & $\textit{.83}_{}$ & $\textbf{.88}_{}$ & $.65_{}$ & $.64_{}$ & $.64_{}$ & $.65_{}$ & $.67_{}$ & $.67_{}$ & $.59_{20}$ & $.56_{105}$ & $.56_{158}$ & $.55_{140}$ & $.55_{141}$ & $.53_{135}$ & $.43_{\textit{19}}$ & $.49_{37}$ & $.51_{133}$ & $.48_{22}$ & $.45_{\textbf{13}}$ & $.41_{88}$ \\
\texttt{heloc} & $\textbf{.71}_{}$ & $\textbf{.71}_{}$ & $\textbf{.71}_{}$ & $\textit{.70}_{}$ & $\textit{.70}_{}$ & $\textit{.70}_{}$ & $\textit{.70}_{}$ & $\textit{.70}_{}$ & $\textit{.70}_{}$ & $\textit{.70}_{50}$ & $\textit{.70}_{\textbf{31}}$ & $\textit{.70}_{91}$ & $.69_{133}$ & $.69_{113}$ & $.69_{92}$ & $.50_{75}$ & $.61_{134}$ & $.66_{102}$ & $.46_{\textit{34}}$ & $.39_{127}$ & $.47_{40}$ \\
\texttt{page} & $\textit{.96}_{}$ & $\textbf{.97}_{}$ & $\textbf{.97}_{}$ & $.93_{}$ & $.93_{}$ & $.94_{}$ & $.93_{}$ & $.93_{}$ & $.93_{}$ & $.92_{134}$ & $.94_{130}$ & $.93_{93}$ & $.95_{64}$ & $\textit{.96}_{143}$ & $\textit{.96}_{149}$ & $.84_{\textbf{0}}$ & $.84_{\textbf{0}}$ & $.86_{101}$ & $.85_{53}$ & $.84_{19}$ & $.84_{\textit{8}}$ \\
\texttt{yoga} & $.68_{}$ & $\textit{.77}_{}$ & $\textbf{.79}_{}$ & $.59_{}$ & $.58_{}$ & $.63_{}$ & $.65_{}$ & $.67_{}$ & $.69_{}$ & $.64_{144}$ & $.63_{155}$ & $.62_{99}$ & $.66_{69}$ & $.67_{123}$ & $.68_{139}$ & $.37_{\textbf{13}}$ & $.37_{\textit{14}}$ & $.45_{15}$ & $.49_{\textit{14}}$ & $.55_{125}$ & $.46_{23}$ \\
\texttt{star} & $.92_{}$ & $\textbf{.95}_{}$ & $\textit{.94}_{}$ & $.79_{}$ & $.79_{}$ & $.89_{}$ & $.90_{}$ & $.80_{}$ & $.80_{}$ & $.79_{235}$ & $.80_{224}$ & $.81_{114}$ & $.80_{104}$ & $.82_{284}$ & $.81_{217}$ & $.79_{100}$ & $.79_{100}$ & $.41_{121}$ & $.42_{111}$ & $.42_{\textbf{4}}$ & $.42_{\textit{81}}$ \\
\texttt{chlorine} & $.12_{}$ & $.08_{}$ & $.10_{}$ & $.08_{}$ & $.08_{}$ & $.09_{}$ & $.08_{}$ & $.08_{}$ & $.08_{}$ & $\textit{.17}_{98}$ & $\textbf{.21}_{172}$ & $.16_{43}$ & $.16_{38}$ & $\textbf{.21}_{645}$ & $\textbf{.21}_{279}$ & $.09_{\textit{13}}$ & $.09_{20}$ & $.09_{21}$ & $.08_{\textbf{7}}$ & $.09_{150}$ & $.09_{347}$ \\
\texttt{kitchen} & $.66_{}$ & $\textit{.68}_{}$ & $\textbf{.70}_{}$ & $.59_{}$ & $.59_{}$ & $.67_{}$ & $.66_{}$ & $.59_{}$ & $.53_{}$ & $.63_{134}$ & $.62_{86}$ & $.66_{109}$ & $.63_{101}$ & $.63_{145}$ & $.60_{145}$ & $.50_{96}$ & $.64_{166}$ & $.44_{\textbf{18}}$ & $.43_{\textit{37}}$ & $.46_{117}$ & $.35_{113}$ \\
\texttt{share} & $.59_{}$ & $\textbf{.63}_{}$ & $\textit{.62}_{}$ & $.56_{}$ & $.56_{}$ & $.56_{}$ & $.56_{}$ & $.56_{}$ & $.56_{}$ & $.60_{176}$ & $.60_{70}$ & $.55_{257}$ & $.59_{159}$ & $.58_{29}$ & $.59_{76}$ & $.56_{\textbf{5}}$ & $.56_{\textit{6}}$ & $.56_{43}$ & $.60_{\textit{6}}$ & $.56_{52}$ & $.56_{65}$ \\
\texttt{devices} & $.61_{}$ & $\textbf{.66}_{}$ & $\textit{.63}_{}$ & $.43_{}$ & $.43_{}$ & $.42_{}$ & $.43_{}$ & $.41_{}$ & $.40_{}$ & $.42_{701}$ & $.43_{718}$ & $.41_{246}$ & $.41_{320}$ & $.43_{236}$ & $.42_{181}$ & $.26_{214}$ & $.27_{221}$ & $.27_{\textit{81}}$ & $.27_{145}$ & $.16_{92}$ & $.16_{\textbf{5}}$ \\
\texttt{gun} & $.65_{}$ & $.63_{}$ & $\textbf{.94}_{}$ & $\textit{.81}_{}$ & $.79_{}$ & $.70_{}$ & $.67_{}$ & $.74_{}$ & $.76_{}$ & $.66_{60}$ & $.65_{61}$ & $.68_{58}$ & $.63_{53}$ & $.63_{\textit{16}}$ & $.66_{\textbf{15}}$ & $.42_{49}$ & $.43_{49}$ & $.72_{66}$ & $.71_{70}$ & $.36_{19}$ & $.44_{19}$ \\
\texttt{worms} & $.57_{}$ & $\textit{.62}_{}$ & $.56_{}$ & $.45_{}$ & $.55_{}$ & $.56_{}$ & $.59_{}$ & $.59_{}$ & $.58_{}$ & $.53_{94}$ & $.55_{133}$ & $.53_{19}$ & $.53_{89}$ & $.44_{112}$ & $.54_{168}$ & $.55_{\textit{14}}$ & $\textbf{.63}_{\textbf{12}}$ & $.49_{107}$ & $.59_{\textbf{12}}$ & $.55_{42}$ & $.49_{171}$ \\
\texttt{ecg} & $\textit{.91}_{}$ & $\textbf{.93}_{}$ & $\textbf{.93}_{}$ & $.88_{}$ & $.88_{}$ & $.89_{}$ & $.88_{}$ & $.89_{}$ & $.89_{}$ & $.89_{215}$ & $.89_{171}$ & $.87_{124}$ & $.87_{93}$ & $.89_{128}$ & $.89_{145}$ & $.78_{97}$ & $.49_{166}$ & $.57_{\textit{88}}$ & $.56_{\textbf{51}}$ & $.75_{104}$ & $.86_{105}$ \\
\texttt{wafer} & $\textit{.98}_{}$ & $\textbf{.99}_{}$ & $\textbf{.99}_{}$ & $.93_{}$ & $.94_{}$ & $.97_{}$ & $.97_{}$ & $\textit{.98}_{}$ & $\textbf{.99}_{}$ & $.94_{46}$ & $.94_{93}$ & $.92_{44}$ & $.94_{41}$ & $.96_{516}$ & $.97_{562}$ & $.84_{\textbf{4}}$ & $.83_{\textit{5}}$ & $.84_{113}$ & $.84_{131}$ & $.84_{105}$ & $.84_{99}$ \\
\texttt{oral} & $\textbf{.88}_{}$ & $\textbf{.88}_{}$ & $\textbf{.88}_{}$ & $\textit{.85}_{}$ & $.84_{}$ & $.82_{}$ & $.78_{}$ & $.82_{}$ & $.84_{}$ & $.70_{\textbf{54}}$ & $.69_{264}$ & $.64_{177}$ & $.79_{158}$ & $.74_{251}$ & $.76_{175}$ & $.27_{105}$ & $.21_{128}$ & $.45_{112}$ & $.35_{\textit{85}}$ & $.23_{88}$ & $.26_{109}$ \\
\texttt{MNIST} & $\textbf{.99}_{}$ & $\textbf{.99}_{}$ & $\textbf{.99}_{}$ & $\textbf{.99}_{}$ & $\textbf{.99}_{}$ & $\textbf{.99}_{}$ & $\textbf{.99}_{}$ & $\textbf{.99}_{}$ & $\textbf{.99}_{}$ & $\textbf{.99}_{124}$ & $\textbf{.99}_{130}$ & $.44_{35}$ & $.44_{34}$ & $\textbf{.99}_{42}$ & $\textit{.74}_{\textbf{11}}$ & $.08_{\textit{29}}$ & $.03_{489}$ & $.02_{200}$ & $.02_{204}$ & $.14_{201}$ & $.07_{234}$ \\
\texttt{cifar10} & $\textbf{.95}_{}$ & $\textbf{.95}_{}$ & $\textbf{.95}_{}$ & $\textbf{.95}_{}$ & $\textbf{.95}_{}$ & $\textbf{.95}_{}$ & $\textbf{.95}_{}$ & $\textbf{.95}_{}$ & $\textbf{.95}_{}$ & $\textit{.94}_{74}$ & $\textit{.94}_{72}$ & $.60_{62}$ & $.63_{53}$ & $\textit{.94}_{36}$ & $\textit{.94}_{49}$ & $.07_{95}$ & $.06_{51}$ & $.11_{21}$ & $.19_{\textit{17}}$ & $.06_{44}$ & $.11_{\textbf{7}}$ \\
\texttt{catsdogs} & $\textbf{.95}_{}$ & $\textbf{.95}_{}$ & $\textbf{.95}_{}$ & $\textbf{.95}_{}$ & $\textbf{.95}_{}$ & $\textbf{.95}_{}$ & $\textbf{.95}_{}$ & $\textbf{.95}_{}$ & $\textbf{.95}_{}$ & $\textbf{.95}_{126}$ & $\textbf{.95}_{133}$ & $\textbf{.95}_{\textit{91}}$ & $\textbf{.95}_{\textbf{16}}$ & $\textbf{.95}_{304}$ & $\textbf{.95}_{107}$ & $.33_{142}$ & $.33_{107}$ & $\textbf{.95}_{100}$ & $\textbf{.95}_{100}$ & $\textit{.94}_{109}$ & $.80_{489}$ \\
\texttt{birds} & $\textit{.89}_{}$ & $\textbf{.90}_{}$ & $\textbf{.90}_{}$ & $\textit{.89}_{}$ & $\textit{.89}_{}$ & $\textit{.89}_{}$ & $.88_{}$ & $\textit{.89}_{}$ & $\textit{.89}_{}$ & $\textit{.89}_{92}$ & $\textbf{.90}_{58}$ & $.87_{81}$ & $.88_{57}$ & $.88_{55}$ & $.88_{163}$ & $.87_{118}$ & $.87_{68}$ & $.83_{\textit{19}}$ & $.73_{\textbf{2}}$ & $.87_{29}$ & $.86_{30}$ \\
\texttt{pets} & $\textit{.94}_{}$ & $\textit{.94}_{}$ & $\textit{.94}_{}$ & $\textbf{.95}_{}$ & $\textit{.94}_{}$ & $\textit{.94}_{}$ & $\textit{.94}_{}$ & $\textit{.94}_{}$ & $\textit{.94}_{}$ & $\textit{.94}_{37}$ & $\textit{.94}_{47}$ & $.93_{90}$ & $.93_{56}$ & $\textit{.94}_{89}$ & $\textit{.94}_{254}$ & $.55_{96}$ & $.16_{\textbf{0}}$ & $.62_{\textit{1}}$ & $.48_{\textit{1}}$ & $.58_{6}$ & $.55_{7}$ \\
\texttt{organa} & $\textbf{1.00}_{}$ & $\textbf{1.00}_{}$ & $\textbf{1.00}_{}$ & $.97_{}$ & $.98_{}$ & $.76_{}$ & $.76_{}$ & $.98_{}$ & $.98_{}$ & $\textit{.99}_{849}$ & $\textit{.99}_{496}$ & $.40_{10}$ & $.45_{\textit{8}}$ & $.97_{85}$ & $.98_{111}$ & $.05_{21}$ & $.05_{286}$ & $.05_{10}$ & $.09_{\textbf{7}}$ & $.05_{198}$ & $.07_{21}$ \\
\texttt{blood} & $\textit{.94}_{}$ & $\textbf{.95}_{}$ & $\textbf{.95}_{}$ & $.91_{}$ & $.91_{}$ & $.75_{}$ & $.72_{}$ & $.92_{}$ & $.92_{}$ & $.93_{739}$ & $.93_{509}$ & $.61_{59}$ & $.77_{51}$ & $.93_{310}$ & $.93_{367}$ & $.07_{128}$ & $.08_{145}$ & $.04_{\textit{7}}$ & $.07_{\textbf{4}}$ & $.12_{111}$ & $.20_{148}$ \\
\texttt{SVHN} & $\textit{.95}_{}$ & $\textit{.95}_{}$ & $\textit{.95}_{}$ & $\textit{.95}_{}$ & $\textbf{.96}_{}$ & $.72_{}$ & $.73_{}$ & $\textit{.95}_{}$ & $.94_{}$ & $\textit{.95}_{29}$ & $\textit{.95}_{77}$ & $.44_{14}$ & $.26_{\textbf{7}}$ & $\textit{.95}_{68}$ & $\textit{.95}_{64}$ & $.19_{530}$ & $.18_{521}$ & $.19_{13}$ & $.08_{\textit{9}}$ & $.06_{35}$ & $.06_{64}$ \\
\texttt{medabs} & $\textbf{.61}_{}$ & $\textbf{.61}_{}$ & $\textit{.60}_{}$ & $.46_{}$ & $.46_{}$ & $.48_{}$ & $.47_{}$ & $.49_{}$ & $.49_{}$ & $.54_{230}$ & $.54_{228}$ & $.53_{154}$ & $.54_{91}$ & $.54_{167}$ & $.54_{174}$ & $.17_{199}$ & $.17_{200}$ & $.08_{\textbf{0}}$ & $.18_{\textit{4}}$ & $.17_{19}$ & $.17_{26}$ \\
\texttt{vicuna} & $.84_{}$ & $\textbf{.87}_{}$ & $\textit{.86}_{}$ & $.43_{}$ & $.43_{}$ & $.60_{}$ & $.46_{}$ & $.44_{}$ & $.42_{}$ & $.66_{169}$ & $.66_{166}$ & $.66_{190}$ & $.66_{137}$ & $.67_{231}$ & $.69_{269}$ & $.42_{82}$ & $.42_{82}$ & $.44_{\textit{18}}$ & $.42_{\textbf{13}}$ & $.42_{22}$ & $.42_{21}$ \\
\texttt{pTED} & $\textbf{.77}_{}$ & $\textbf{.77}_{}$ & $\textit{.75}_{}$ & $.46_{}$ & $.46_{}$ & $.53_{}$ & $.46_{}$ & $.47_{}$ & $.46_{}$ & $.62_{155}$ & $.61_{227}$ & $.61_{154}$ & $.61_{195}$ & $.63_{232}$ & $.61_{138}$ & $.46_{63}$ & $.46_{66}$ & $.46_{280}$ & $.46_{\textbf{12}}$ & $.46_{\textit{62}}$ & $.46_{68}$ \\
\texttt{tgpt} & $\textit{.95}_{}$ & $\textbf{.97}_{}$ & $\textit{.95}_{}$ & $.90_{}$ & $.91_{}$ & $.91_{}$ & $.90_{}$ & $.91_{}$ & $.90_{}$ & $.78_{152}$ & $.76_{366}$ & $.80_{269}$ & $.77_{236}$ & $.78_{236}$ & $.79_{251}$ & $.65_{\textit{100}}$ & $.56_{\textit{100}}$ & $.49_{132}$ & $.43_{111}$ & $.38_{128}$ & $.58_{\textbf{32}}$ \\
\texttt{pol} & $.69_{}$ & $\textbf{.73}_{}$ & $\textit{.70}_{}$ & $\textit{.70}_{}$ & $.67_{}$ & $\textit{.70}_{}$ & $.66_{}$ & $.69_{}$ & $.69_{}$ & $.54_{38}$ & $.55_{\textit{36}}$ & $.60_{52}$ & $.59_{48}$ & $.59_{89}$ & $.59_{41}$ & $.49_{115}$ & $.49_{58}$ & $.54_{89}$ & $.58_{86}$ & $.50_{\textbf{35}}$ & $.50_{41}$ \\
\texttt{liar} & $\textit{.24}_{}$ & $\textbf{.25}_{}$ & $\textit{.24}_{}$ & $.11_{}$ & $.11_{}$ & $.18_{}$ & $.18_{}$ & $.18_{}$ & $.17_{}$ & $.21_{151}$ & $.21_{159}$ & $.19_{124}$ & $.21_{156}$ & $.20_{55}$ & $.20_{99}$ & $.07_{27}$ & $.11_{\textbf{7}}$ & $.13_{\textit{12}}$ & $.11_{17}$ & $.15_{\textit{12}}$ & $.15_{54}$ \\
\midrule
\texttt{Avg.} & $\textit{.81}_{}$ & $\textbf{.82}_{}$ & $\textbf{.82}_{}$ & $.74_{}$ & $.74_{}$ & $.73_{}$ & $.72_{}$ & $.74_{}$ & $.74_{}$ & $.71_{141.84}$ & $.71_{147.36}$ & $.66_{97.33}$ & $.66_{85.73}$ & $.72_{156.04}$ & $.71_{140.13}$ & $.49_{91.00}$ & $.47_{99.47}$ & $.51_{\textit{69.44}}$ & $.50_{\textbf{58.42}}$ & $.48_{74.51}$ & $.49_{83.78}$ \\
\texttt{Std.} & $.19_{{}}$ & $.19_{\textbf{}}$ & $.19_{\textbf{}}$ & $.23_{\textbf{}}$ & $.23_{\textbf{}}$ & $.21_{\textbf{}}$ & $.21_{\textbf{}}$ & $.22_{\textbf{}}$ & $.22_{\textbf{}}$ & $.20_{181.93}$ & $.20_{139.48}$ & $.19_{63.14}$ & $.20_{65.55}$ & $.20_{132.55}$ & $.20_{114.28}$ & ${.26}_{92.47}$ & ${.26}_{110.43}$ & ${.27}_{59.53}$ & ${.26}_{{52.38}}$ & ${.26}_{54.11}$ & ${.26}_{90.24}$ \\
%\texttt{Rank} & $4.4$ & $2.9$ & $2.9$ & $9.5$ & $9.5$ & $8.3$ & $9.7$ & $8.6$ & $9.0$ & $10.3$ & $9.7$ & $12.2$ & $11.5$ & $10.0$ & $9.8$ & $17.3$ & $17.3$ & $16.6$ & $16.2$ & $17.7$ & $17.6$ \\
\texttt{Rank} & $4.4_{5.0}$ & $2.9_{5.0}$ & $2.9_{5.0}$ & $9.5_{5.0}$ & $9.5_{5.0}$ & $8.3_{5.0}$ & $9.7_{5.0}$ & $8.6_{5.0}$ & $9.0_{5.0}$ & $10.3_{15.0}$ & $9.7_{14.0}$ & $12.2_{14.9}$ & $11.5_{16.1}$ & $10.0_{13.6}$ & $9.8_{13.9}$ & $17.3_{16.3}$ & $17.3_{16.0}$ & $16.6_{16.4}$ & $16.2_{17.4}$ & $17.7_{16.4}$ & $17.6_{16.0}$\\
\bottomrule

\end{tabular}

\end{sidewaystable}

\begin{figure}[t]
    \centering
    \includegraphics[width=0.48\linewidth]{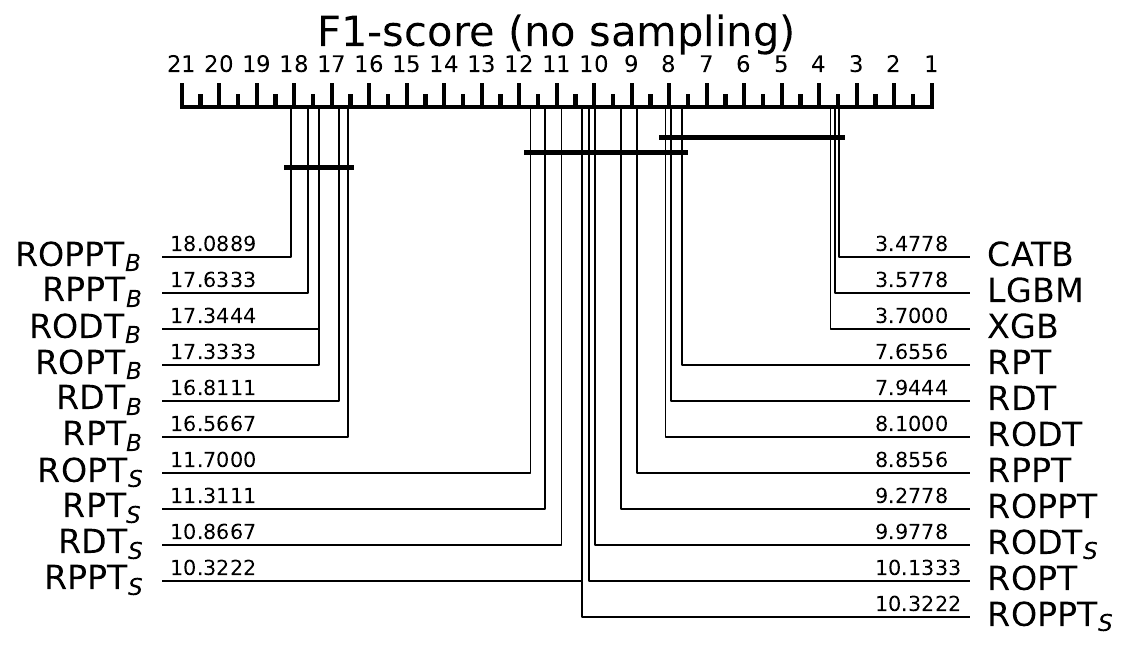}
    \includegraphics[width=0.48\linewidth]{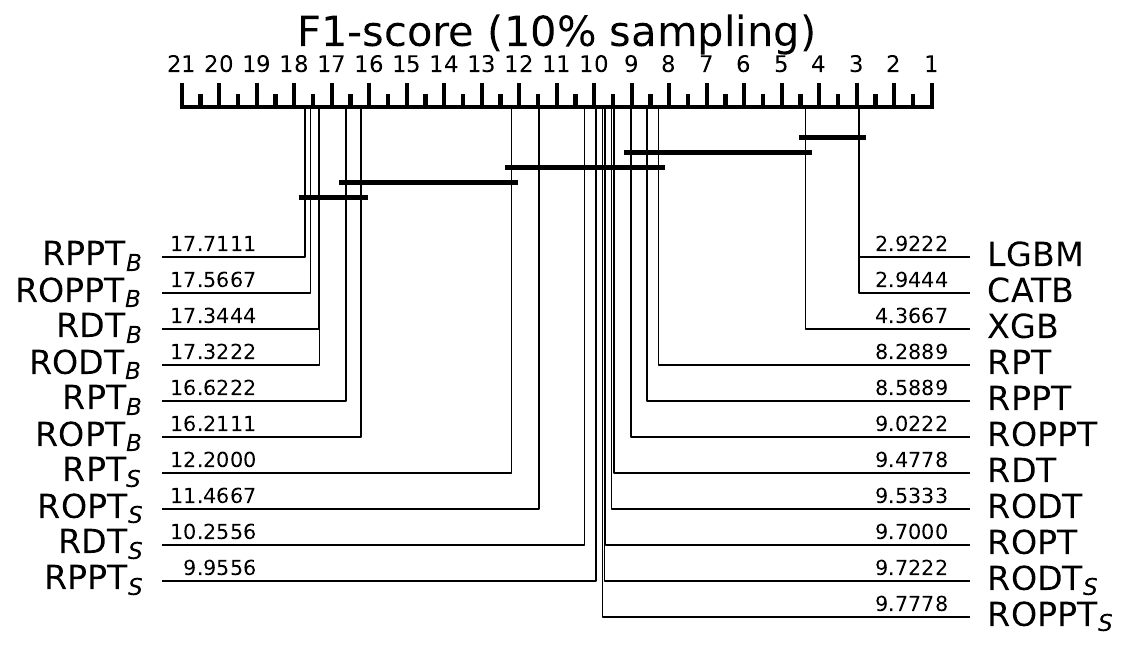}
    \caption{Critical difference plots of ensemble models' rankings in terms of weighted F1-score, evaluated using the Nemenyi test across all datasets. Models that are not significantly different at the $95\%$ significance level are connected. The best models are positioned on the right.}
    \label{fig:cd_plots_ensemble_full}
\end{figure}

\begin{sidewaystable}[t]
\caption{Average Balanced Accuracy for \textsc{RandomPivotForest} with $100$ \textsc{PivotTree} \textsc{c}lassifiers as estimators, and no sampling, against baselines, for each \texttt{dataset}.  When the splitting stump forest is adopted, subscripts indicate the average number of stumps $\pm$ std. dev. Best results in \textbf{bold}, second best in \textit{italics}.}
    \label{tab:ensemble_sample_TOTAL_bal_accc}
\centering
\footnotesize
\setlength{\tabcolsep}{0.3mm} % Reduce column spacing
\begin{tabular}{l|lll|ll|llll|ll|llll|ll|llll}
\toprule
\texttt{dataset} & $\textsc{xgb}$ & $\textsc{lgbm}$ & $\textsc{catb}$ & $\textsc{rdt}$ & $\textsc{rodt}$ & $\textsc{rpt}$ & $\textsc{ropt}$ & $\textsc{rppt}$ & $\textsc{roppt}$ & $\textsc{rdt}_{\textsc{s}}$ & $\textsc{rodt}_{\textsc{s}}$ & $\textsc{rpt}_{\textsc{s}}$ & $\textsc{ropt}_{\textsc{s}}$ & $\textsc{rppt}_{\textsc{s}}$ & $\textsc{roppt}_{\textsc{s}}$ & $\textsc{rdt}_{\textsc{b}}$ & $\textsc{rodt}_{\textsc{b}}$ & $\textsc{rpt}_{\textsc{b}}$ & $\textsc{ropt}_{\textsc{b}}$ & $\textsc{rppt}_{\textsc{b}}$ & $\textsc{roppt}_{\textsc{b}}$ \\
\midrule
\texttt{ion} & $\textit{.93}_{}$ & $\textbf{.94}_{}$ & $\textit{.93}_{}$ & $.89_{}$ & $\textbf{.94}_{}$ & $.92_{}$ & $.92_{}$ & $.91_{}$ & $\textit{.93}_{}$ & $.85_{30}$ & $.83_{48}$ & $.87_{83}$ & $.89_{157}$ & $.84_{176}$ & $.89_{85}$ & $.63_{17}$ & $.63_{126}$ & $.78_{16}$ & $.67_{107}$ & $.60_{101}$ & $.62_{69}$ \\
\texttt{fire} & $\textbf{.99}_{}$ & $\textit{.98}_{}$ & $\textbf{.99}_{}$ & $\textbf{.99}_{}$ & $.97_{}$ & $.94_{}$ & $.93_{}$ & $.94_{}$ & $.94_{}$ & $.63_{17}$ & $.74_{101}$ & $.65_{82}$ & $.71_{73}$ & $.67_{222}$ & $.67_{69}$ & $.85_{24}$ & $.85_{222}$ & $.88_{55}$ & $.70_{88}$ & $.80_{73}$ & $.68_{36}$ \\
\texttt{yeast} & $\textbf{.59}_{}$ & $.49_{}$ & $\textit{.56}_{}$ & $.29_{}$ & $.30_{}$ & $.36_{}$ & $.39_{}$ & $.30_{}$ & $.31_{}$ & $.15_{1}$ & $.25_{26}$ & $.25_{13}$ & $.30_{15}$ & $.27_{9}$ & $.28_{7}$ & $.12_{1}$ & $.11_{0}$ & $.11_{1}$ & $.10_{2}$ & $.11_{0}$ & $.11_{0}$ \\
\texttt{magic} & $\textit{.85}_{}$ & $\textbf{.86}_{}$ & $\textit{.85}_{}$ & $.65_{}$ & $.64_{}$ & $.65_{}$ & $.63_{}$ & $.63_{}$ & $.62_{}$ & $.70_{33}$ & $.70_{53}$ & $.71_{44}$ & $.71_{66}$ & $.73_{180}$ & $.72_{78}$ & $.55_{33}$ & $.49_{39}$ & $.59_{14}$ & $.58_{22}$ & $.50_{51}$ & $.55_{28}$ \\
\texttt{sonar} & $\textit{.88}_{}$ & $.87_{}$ & $.84_{}$ & $.87_{}$ & $.85_{}$ & $\textbf{.90}_{}$ & $.85_{}$ & $.85_{}$ & $.85_{}$ & $.62_{128}$ & $.63_{21}$ & $.58_{38}$ & $.57_{28}$ & $.61_{437}$ & $.55_{97}$ & $.60_{153}$ & $.68_{194}$ & $.67_{126}$ & $.73_{73}$ & $.67_{106}$ & $.67_{97}$ \\
\texttt{compas} & $\textit{.55}_{}$ & $\textit{.55}_{}$ & $\textbf{.56}_{}$ & $.44_{}$ & $.44_{}$ & $.44_{}$ & $.44_{}$ & $.42_{}$ & $.42_{}$ & $.42_{6}$ & $.43_{45}$ & $.43_{38}$ & $.43_{28}$ & $.44_{46}$ & $.43_{48}$ & $.33_{22}$ & $.33_{69}$ & $.33_{95}$ & $.33_{89}$ & $.33_{53}$ & $.33_{47}$ \\
\texttt{house} & $\textbf{.87}_{}$ & $\textbf{.87}_{}$ & $\textbf{.87}_{}$ & $.74_{}$ & $.74_{}$ & $.73_{}$ & $.64_{}$ & $.66_{}$ & $.66_{}$ & $.74_{98}$ & $\textit{.75}_{312}$ & $.72_{246}$ & $.73_{262}$ & $.69_{406}$ & $.70_{392}$ & $.50_{1}$ & $.50_{3}$ & $.50_{107}$ & $.50_{116}$ & $.50_{114}$ & $.50_{126}$ \\
\texttt{german} & $\textit{.68}_{}$ & $\textbf{.70}_{}$ & $.65_{}$ & $.50_{}$ & $.50_{}$ & $.50_{}$ & $.50_{}$ & $.50_{}$ & $.50_{}$ & $.55_{7}$ & $.63_{20}$ & $.63_{13}$ & $.63_{21}$ & $.62_{9}$ & $.63_{24}$ & $.50_{16}$ & $.50_{28}$ & $.50_{45}$ & $.50_{24}$ & $.50_{10}$ & $.50_{9}$ \\
\texttt{spamb} & $\textbf{.96}_{}$ & $\textit{.95}_{}$ & $\textit{.95}_{}$ & $.86_{}$ & $.87_{}$ & $.87_{}$ & $.85_{}$ & $.84_{}$ & $.83_{}$ & $.77_{68}$ & $.79_{120}$ & $.78_{221}$ & $.78_{228}$ & $.79_{281}$ & $.79_{214}$ & $.77_{50}$ & $.71_{96}$ & $.72_{103}$ & $.66_{99}$ & $.66_{119}$ & $.67_{216}$ \\
\texttt{norm} & $\textit{.97}_{}$ & $\textbf{.98}_{}$ & $\textbf{.98}_{}$ & $.96_{}$ & $\textit{.97}_{}$ & $\textbf{.98}_{}$ & $\textbf{.98}_{}$ & $\textbf{.98}_{}$ & $\textbf{.98}_{}$ & $.80_{181}$ & $.81_{128}$ & $.81_{90}$ & $.82_{105}$ & $.81_{95}$ & $.81_{101}$ & $.94_{188}$ & $\textit{.97}_{104}$ & $\textit{.97}_{105}$ & $\textit{.97}_{105}$ & $\textbf{.98}_{99}$ & $\textit{.97}_{106}$ \\
\texttt{lrs} & $\textit{.62}_{}$ & $\textbf{.63}_{}$ & $.61_{}$ & $.41_{}$ & $.48_{}$ & $.41_{}$ & $.40_{}$ & $.45_{}$ & $.47_{}$ & $.26_{357}$ & $.24_{297}$ & $.20_{106}$ & $.21_{188}$ & $.21_{133}$ & $.22_{110}$ & $.18_{99}$ & $.14_{285}$ & $.15_{107}$ & $.15_{97}$ & $.15_{77}$ & $.13_{87}$ \\
\texttt{vert} & $\textbf{.80}_{}$ & $\textbf{.80}_{}$ & $\textit{.77}_{}$ & $.65_{}$ & $.69_{}$ & $.65_{}$ & $.66_{}$ & $.67_{}$ & $.70_{}$ & $.60_{14}$ & $.54_{80}$ & $.52_{18}$ & $.56_{15}$ & $.51_{47}$ & $.57_{31}$ & $.65_{14}$ & $.48_{22}$ & $.59_{9}$ & $.53_{24}$ & $.50_{36}$ & $.47_{21}$ \\
\texttt{iris} & $\textbf{1.00}_{}$ & $\textbf{1.00}_{}$ & $\textbf{1.00}_{}$ & $\textit{.97}_{}$ & $\textit{.97}_{}$ & $\textbf{1.00}_{}$ & $\textit{.97}_{}$ & $\textbf{1.00}_{}$ & $.92_{}$ & $.79_{11}$ & $.72_{13}$ & $.69_{13}$ & $.67_{15}$ & $.72_{12}$ & $.72_{29}$ & $.44_{14}$ & $.46_{5}$ & $.70_{11}$ & $.68_{13}$ & $.51_{12}$ & $.82_{2}$ \\
\texttt{wine} & $\textbf{.99}_{}$ & $\textbf{.99}_{}$ & $\textbf{.99}_{}$ & $.96_{}$ & $.96_{}$ & $\textit{.98}_{}$ & $.96_{}$ & $.97_{}$ & $.96_{}$ & $.87_{81}$ & $.87_{190}$ & $.87_{48}$ & $.86_{129}$ & $.87_{12}$ & $.89_{92}$ & $.55_{52}$ & $.52_{342}$ & $.50_{48}$ & $.52_{55}$ & $.50_{106}$ & $.51_{111}$ \\
\texttt{diva} & $\textbf{.90}_{}$ & $\textbf{.90}_{}$ & $\textbf{.90}_{}$ & $.62_{}$ & $.62_{}$ & $.57_{}$ & $.53_{}$ & $.52_{}$ & $.50_{}$ & $.70_{34}$ & $.70_{46}$ & $.70_{125}$ & $.69_{128}$ & $.71_{98}$ & $\textit{.72}_{175}$ & $.50_{41}$ & $.50_{42}$ & $.50_{52}$ & $.50_{45}$ & $.50_{41}$ & $.50_{49}$ \\
\texttt{breast} & $\textbf{.97}_{}$ & $.95_{}$ & $\textbf{.97}_{}$ & $\textit{.96}_{}$ & $\textit{.96}_{}$ & $\textbf{.97}_{}$ & $.94_{}$ & $.95_{}$ & $\textit{.96}_{}$ & $\textit{.96}_{113}$ & $\textit{.96}_{53}$ & $.95_{103}$ & $.93_{50}$ & $.93_{33}$ & $\textit{.96}_{28}$ & $.88_{114}$ & $.91_{57}$ & $.93_{115}$ & $.91_{199}$ & $.93_{60}$ & $.91_{88}$ \\
\texttt{steel} & $\textit{.81}_{}$ & $\textbf{.82}_{}$ & $.79_{}$ & $.51_{}$ & $.51_{}$ & $.51_{}$ & $.46_{}$ & $.47_{}$ & $.49_{}$ & $.41_{127}$ & $.46_{142}$ & $.45_{44}$ & $.43_{59}$ & $.49_{206}$ & $.48_{181}$ & $.15_{49}$ & $.17_{25}$ & $.16_{36}$ & $.18_{3}$ & $.14_{74}$ & $.20_{8}$ \\
\texttt{ecoli} & $\textbf{.72}_{}$ & $\textit{.70}_{}$ & $.68_{}$ & $.61_{}$ & $.58_{}$ & $.57_{}$ & $.58_{}$ & $.57_{}$ & $.57_{}$ & $.35_{6}$ & $.42_{28}$ & $.39_{35}$ & $.37_{33}$ & $.33_{8}$ & $.42_{7}$ & $.25_{18}$ & $.26_{21}$ & $.24_{27}$ & $.26_{28}$ & $.23_{19}$ & $.26_{21}$ \\
\texttt{heloc} & $\textbf{.71}_{}$ & $\textbf{.71}_{}$ & $\textbf{.71}_{}$ & $\textit{.70}_{}$ & $\textit{.70}_{}$ & $\textit{.70}_{}$ & $\textit{.70}_{}$ & $\textit{.70}_{}$ & $\textit{.70}_{}$ & $.69_{40}$ & $.69_{58}$ & $.68_{133}$ & $.69_{124}$ & $.69_{88}$ & $.69_{92}$ & $.51_{33}$ & $.55_{55}$ & $.58_{106}$ & $.52_{27}$ & $.54_{142}$ & $.47_{104}$ \\
\texttt{page} & $\textit{.92}_{}$ & $\textbf{.93}_{}$ & $.91_{}$ & $.79_{}$ & $.80_{}$ & $.82_{}$ & $.73_{}$ & $.74_{}$ & $.74_{}$ & $.75_{112}$ & $.87_{153}$ & $.78_{93}$ & $.84_{92}$ & $.90_{143}$ & $.89_{149}$ & $.51_{41}$ & $.53_{13}$ & $.53_{74}$ & $.51_{53}$ & $.50_{7}$ & $.50_{0}$ \\
\texttt{yoga} & $\textbf{.79}_{}$ & $.77_{}$ & $\textit{.78}_{}$ & $.73_{}$ & $.72_{}$ & $.75_{}$ & $.74_{}$ & $.73_{}$ & $.77_{}$ & $.60_{126}$ & $.58_{68}$ & $.62_{21}$ & $.59_{27}$ & $.69_{110}$ & $.66_{83}$ & $.57_{127}$ & $.58_{24}$ & $.54_{10}$ & $.56_{23}$ & $.50_{22}$ & $.50_{242}$ \\
\texttt{star} & $\textbf{.93}_{}$ & $\textbf{.93}_{}$ & $\textit{.91}_{}$ & $.82_{}$ & $.82_{}$ & $.83_{}$ & $.80_{}$ & $.71_{}$ & $.67_{}$ & $.69_{444}$ & $.70_{409}$ & $.71_{147}$ & $.70_{218}$ & $.69_{139}$ & $.68_{113}$ & $.31_{127}$ & $.33_{121}$ & $.37_{104}$ & $.32_{104}$ & $.55_{38}$ & $.32_{1}$ \\
\texttt{chlorine} & $.12_{}$ & $.12_{}$ & $.13_{}$ & $.26_{}$ & $.26_{}$ & $.25_{}$ & $.27_{}$ & $.28_{}$ & $.29_{}$ & $.31_{20}$ & $.30_{38}$ & $.30_{30}$ & $.29_{19}$ & $.26_{645}$ & $.27_{384}$ & $\textbf{.33}_{34}$ & $\textit{.32}_{42}$ & $.30_{47}$ & $.29_{54}$ & $\textbf{.33}_{29}$ & $\textbf{.33}_{37}$ \\
\texttt{kitchen} & $.71_{}$ & $.70_{}$ & $.70_{}$ & $.71_{}$ & $\textit{.72}_{}$ & $\textbf{.74}_{}$ & $\textbf{.74}_{}$ & $.65_{}$ & $.64_{}$ & $.65_{183}$ & $.70_{372}$ & $.70_{121}$ & $.68_{84}$ & $.64_{265}$ & $.59_{48}$ & $.49_{417}$ & $.49_{181}$ & $.49_{79}$ & $.43_{64}$ & $.55_{49}$ & $.56_{48}$ \\
\texttt{share} & $\textbf{.56}_{}$ & $\textit{.54}_{}$ & $\textit{.54}_{}$ & $.50_{}$ & $.50_{}$ & $.50_{}$ & $.50_{}$ & $.50_{}$ & $.50_{}$ & $.50_{27}$ & $.51_{196}$ & $.49_{367}$ & $.51_{221}$ & $.51_{53}$ & $.51_{85}$ & $.51_{140}$ & $.50_{204}$ & $.50_{367}$ & $.50_{47}$ & $.50_{11}$ & $.50_{24}$ \\
\texttt{devices} & $\textbf{.59}_{}$ & $\textit{.58}_{}$ & $.55_{}$ & $.35_{}$ & $.36_{}$ & $.36_{}$ & $.35_{}$ & $.34_{}$ & $.34_{}$ & $.36_{689}$ & $.35_{744}$ & $.32_{815}$ & $.34_{789}$ & $.34_{846}$ & $.34_{803}$ & $.22_{172}$ & $.22_{225}$ & $.21_{95}$ & $.20_{77}$ & $.16_{226}$ & $.16_{248}$ \\
\texttt{gun} & $.87_{}$ & $.63_{}$ & $\textit{.91}_{}$ & $\textbf{.93}_{}$ & $\textit{.91}_{}$ & $.85_{}$ & $.87_{}$ & $.86_{}$ & $.85_{}$ & $.69_{97}$ & $.67_{131}$ & $.69_{33}$ & $.67_{46}$ & $.67_{58}$ & $.67_{45}$ & $.70_{85}$ & $.57_{92}$ & $.78_{126}$ & $.75_{7}$ & $.53_{43}$ & $.55_{53}$ \\
\texttt{worms} & $\textit{.59}_{}$ & $\textbf{.61}_{}$ & $\textbf{.61}_{}$ & $\textit{.59}_{}$ & $.52_{}$ & $\textbf{.61}_{}$ & $.56_{}$ & $\textit{.59}_{}$ & $.58_{}$ & $.48_{65}$ & $.50_{64}$ & $.55_{116}$ & $.49_{34}$ & $.55_{44}$ & $.54_{316}$ & $.56_{8}$ & $.45_{9}$ & $\textit{.59}_{24}$ & $.48_{34}$ & $.55_{163}$ & $.53_{109}$ \\
\texttt{ecg} & $\textit{.52}_{}$ & $\textbf{.53}_{}$ & $.51_{}$ & $.45_{}$ & $.44_{}$ & $.49_{}$ & $.47_{}$ & $.45_{}$ & $.45_{}$ & $.48_{181}$ & $.48_{308}$ & $.42_{118}$ & $.44_{123}$ & $.48_{72}$ & $.45_{62}$ & $.22_{337}$ & $.25_{125}$ & $.19_{16}$ & $.27_{16}$ & $.29_{9}$ & $.27_{14}$ \\
\texttt{wafer} & $\textbf{.98}_{}$ & $\textbf{.98}_{}$ & $\textit{.97}_{}$ & $.88_{}$ & $.88_{}$ & $.95_{}$ & $.95_{}$ & $\textit{.97}_{}$ & $\textbf{.98}_{}$ & $.80_{133}$ & $.82_{168}$ & $.71_{44}$ & $.78_{41}$ & $.86_{516}$ & $.89_{562}$ & $.50_{206}$ & $.50_{0}$ & $.50_{4}$ & $.50_{7}$ & $.50_{126}$ & $.50_{85}$ \\
\texttt{oral} & $\textit{.88}_{}$ & $\textit{.88}_{}$ & $.84_{}$ & $.87_{}$ & $.84_{}$ & $\textbf{.89}_{}$ & $.85_{}$ & $\textit{.88}_{}$ & $\textbf{.89}_{}$ & $.65_{237}$ & $.71_{350}$ & $.70_{394}$ & $.64_{280}$ & $.78_{403}$ & $.70_{289}$ & $.33_{188}$ & $.35_{236}$ & $.39_{102}$ & $.40_{239}$ & $.26_{403}$ & $.40_{183}$ \\
\texttt{MNIST} & $\textbf{.99}_{}$ & $\textbf{.99}_{}$ & $\textbf{.99}_{}$ & $\textbf{.99}_{}$ & $\textbf{.99}_{}$ & $.60_{}$ & $.60_{}$ & $\textit{.90}_{}$ & $\textit{.90}_{}$ & $\textbf{.99}_{58}$ & $\textbf{.99}_{57}$ & $\textbf{.99}_{185}$ & $\textbf{.99}_{169}$ & $\textbf{.99}_{217}$ & $\textbf{.99}_{73}$ & $.10_{466}$ & $.20_{96}$ & $.10_{410}$ & $.10_{0}$ & $.20_{205}$ & $.10_{68}$ \\
\texttt{cifar10} & $\textbf{.95}_{}$ & $\textbf{.95}_{}$ & $\textbf{.95}_{}$ & $\textbf{.95}_{}$ & $\textbf{.95}_{}$ & $.67_{}$ & $.77_{}$ & $\textbf{.95}_{}$ & $\textbf{.95}_{}$ & $\textit{.94}_{145}$ & $\textbf{.95}_{51}$ & $.22_{3}$ & $.29_{2}$ & $\textit{.94}_{92}$ & $\textit{.94}_{67}$ & $.04_{154}$ & $.06_{137}$ & $.13_{3}$ & $.10_{0}$ & $.04_{14}$ & $.11_{19}$ \\
\texttt{catsdogs} & $\textbf{.95}_{}$ & $\textbf{.95}_{}$ & $\textbf{.95}_{}$ & $\textbf{.95}_{}$ & $\textbf{.95}_{}$ & $\textbf{.95}_{}$ & $\textbf{.95}_{}$ & $\textbf{.95}_{}$ & $\textbf{.95}_{}$ & $\textbf{.95}_{51}$ & $\textbf{.95}_{55}$ & $\textbf{.95}_{100}$ & $\textbf{.95}_{45}$ & $\textbf{.95}_{15}$ & $\textbf{.95}_{19}$ & $\textbf{.95}_{97}$ & $\textbf{.95}_{100}$ & $\textbf{.95}_{100}$ & $\textbf{.95}_{100}$ & $.85_{17}$ & $\textit{.93}_{46}$ \\
\texttt{birds} & $\textbf{.84}_{}$ & $\textbf{.84}_{}$ & $\textbf{.84}_{}$ & $\textit{.83}_{}$ & $\textit{.83}_{}$ & $.82_{}$ & $.80_{}$ & $.82_{}$ & $.80_{}$ & $.82_{24}$ & $.82_{22}$ & $.80_{118}$ & $.77_{28}$ & $.81_{119}$ & $.81_{86}$ & $.78_{115}$ & $.74_{119}$ & $.72_{105}$ & $.60_{28}$ & $\textit{.83}_{6}$ & $.79_{28}$ \\
\texttt{pets} & $\textbf{.94}_{}$ & $\textbf{.94}_{}$ & $\textbf{.94}_{}$ & $\textit{.93}_{}$ & $\textbf{.94}_{}$ & $\textbf{.94}_{}$ & $\textbf{.94}_{}$ & $\textbf{.94}_{}$ & $\textbf{.94}_{}$ & $\textbf{.94}_{235}$ & $\textbf{.94}_{225}$ & $\textit{.93}_{167}$ & $\textit{.93}_{91}$ & $\textbf{.94}_{47}$ & $\textbf{.94}_{271}$ & $.50_{152}$ & $.50_{2}$ & $.56_{4}$ & $.87_{7}$ & $\textit{.93}_{1}$ & $.50_{0}$ \\
\texttt{organa} & $\textit{.99}_{}$ & $\textbf{1.00}_{}$ & $\textbf{1.00}_{}$ & $.97_{}$ & $.97_{}$ & $.63_{}$ & $.63_{}$ & $.95_{}$ & $.94_{}$ & $\textit{.99}_{590}$ & $\textit{.99}_{985}$ & $.87_{203}$ & $.29_{4}$ & $.97_{83}$ & $.97_{105}$ & $.09_{1096}$ & $.09_{12}$ & $.09_{0}$ & $.09_{0}$ & $.09_{1}$ & $.09_{35}$ \\
\texttt{blood} & $\textit{.93}_{}$ & $\textbf{.94}_{}$ & $\textbf{.94}_{}$ & $.89_{}$ & $.89_{}$ & $.61_{}$ & $.60_{}$ & $.90_{}$ & $.85_{}$ & $.92_{1160}$ & $.92_{634}$ & $.86_{377}$ & $.87_{341}$ & $.92_{453}$ & $.92_{600}$ & $.15_{102}$ & $.23_{237}$ & $.12_{0}$ & $.12_{224}$ & $.13_{79}$ & $.16_{89}$ \\
\texttt{SVHN} & $\textbf{.95}_{}$ & $\textbf{.95}_{}$ & $\textbf{.95}_{}$ & $\textbf{.95}_{}$ & $\textbf{.95}_{}$ & $.48_{}$ & $.49_{}$ & $\textbf{.95}_{}$ & $\textbf{.95}_{}$ & $\textbf{.95}_{29}$ & $\textbf{.95}_{23}$ & $\textit{.94}_{151}$ & $.93_{131}$ & $\textbf{.95}_{80}$ & $\textbf{.95}_{64}$ & $.10_{55}$ & $.19_{2}$ & $.10_{0}$ & $.10_{0}$ & $.10_{80}$ & $.10_{24}$ \\
\texttt{medabs} & $\textit{.61}_{}$ & $\textbf{.62}_{}$ & $.60_{}$ & $.41_{}$ & $.41_{}$ & $.43_{}$ & $.41_{}$ & $.44_{}$ & $.43_{}$ & $.50_{270}$ & $.51_{274}$ & $.51_{228}$ & $.51_{180}$ & $.53_{170}$ & $.53_{247}$ & $.20_{240}$ & $.20_{252}$ & $.20_{20}$ & $.20_{47}$ & $.20_{59}$ & $.20_{21}$ \\
\texttt{vicuna} & $\textit{.86}_{}$ & $\textbf{.87}_{}$ & $.85_{}$ & $.57_{}$ & $.56_{}$ & $.63_{}$ & $.50_{}$ & $.51_{}$ & $.51_{}$ & $.64_{286}$ & $.65_{308}$ & $.65_{362}$ & $.65_{160}$ & $.66_{149}$ & $.66_{302}$ & $.50_{12}$ & $.50_{12}$ & $.50_{124}$ & $.50_{70}$ & $.50_{20}$ & $.50_{80}$ \\
\texttt{pTED} & $\textbf{.77}_{}$ & $\textit{.75}_{}$ & $\textit{.75}_{}$ & $.52_{}$ & $.52_{}$ & $.54_{}$ & $.50_{}$ & $.50_{}$ & $.50_{}$ & $.61_{282}$ & $.60_{282}$ & $.60_{182}$ & $.58_{359}$ & $.61_{130}$ & $.58_{112}$ & $.50_{151}$ & $.50_{153}$ & $.50_{174}$ & $.50_{64}$ & $.50_{60}$ & $.50_{49}$ \\
\texttt{tgpt} & $\textbf{.97}_{}$ & $\textbf{.97}_{}$ & $\textit{.96}_{}$ & $.92_{}$ & $.91_{}$ & $.92_{}$ & $.89_{}$ & $.93_{}$ & $.91_{}$ & $.78_{126}$ & $.77_{155}$ & $.78_{123}$ & $.81_{332}$ & $.80_{239}$ & $.78_{132}$ & $.58_{91}$ & $.75_{103}$ & $.59_{94}$ & $.48_{51}$ & $.85_{99}$ & $.51_{93}$ \\
\texttt{pol} & $\textit{.73}_{}$ & $\textbf{.74}_{}$ & $.70_{}$ & $.71_{}$ & $.71_{}$ & $.72_{}$ & $.70_{}$ & $.71_{}$ & $.71_{}$ & $.60_{72}$ & $.55_{92}$ & $.60_{63}$ & $.57_{51}$ & $.57_{91}$ & $.56_{111}$ & $.54_{72}$ & $.53_{97}$ & $.57_{106}$ & $.58_{110}$ & $.51_{39}$ & $.51_{66}$ \\
\texttt{liar} & $\textbf{.24}_{}$ & $\textbf{.24}_{}$ & $\textit{.23}_{}$ & $.20_{}$ & $.20_{}$ & $.20_{}$ & $.19_{}$ & $.20_{}$ & $.19_{}$ & $.19_{100}$ & $.20_{86}$ & $.19_{131}$ & $.19_{195}$ & $.17_{102}$ & $.18_{109}$ & $.17_{13}$ & $.17_{19}$ & $.17_{34}$ & $.17_{60}$ & $.16_{53}$ & $.17_{401}$ \\
\midrule
\texttt{Avg.} & $\textbf{.80}_{}$ & $\textit{.79}_{}$ & $\textit{.79}_{}$ & $.72_{}$ & $.72_{}$ & $.69_{}$ & $.67_{}$ & $.70_{}$ & $.70_{}$ & $.66_{157.64}$ & $.67_{178.47}$ & $.64_{137.44}$ & $.63_{128.58}$ & $.67_{172.87}$ & $.67_{157.71}$ & $.45_{125.27}$ & $.45_{96.62}$ & $.47_{75.56}$ & $.46_{\textbf{60.04}}$ & $.46_{\textit{70.04}}$ & $.45_{73.07}$ \\
\texttt{Std.} & $.20_{}$ & $.20_{}$ & $.20_{}$ & $.23_{}$ & $.23_{}$ & $.22_{}$ & $.21_{}$ & $.23_{}$ & $.23_{}$ & $.22_{213.80}$ & $.21_{201.69}$ & $.22_{144.81}$ & $.22_{139.83}$ & $.22_{179.99}$ & $.22_{167.68}$ & $\textit{.24}_{181.00}$ & $.23_{89.89}$ & ${.25}_{82.83}$ & ${.24}_{{56.47}}$ & ${.25}_{{73.60}}$ & $.23_{78.98}$ \\
\texttt{Rank} & $3.3_{9.9}$ & $3.6_{9.9}$ & $3.9_{9.9}$ & $8.0_{9.9}$ & $7.9_{9.9}$ & $7.6_{9.9}$ & $9.9_{9.9}$ & $8.9_{9.9}$ & $9.4_{9.9}$ & $10.9_{11.2}$ & $9.9_{9.1}$ & $12.2_{9.7}$ & $12.1_{10.5}$ & $10.7_{9.2}$ & $10.5_{9.8}$ & $16.8_{12.2}$ & $17.1_{12.0}$ & $16.3_{13.3}$ & $17.2_{15.2}$ & $17.1_{14.9}$ & $17.6_{15.2}$\\
\bottomrule
\end{tabular}
\end{sidewaystable}

\begin{sidewaystable}[t]
\caption{Average Balanced Accuracy for \textsc{RandomPivotForest} with $100$ \textsc{PivotTree} \textsc{c}lassifiers as estimators, and 10\% of the training set is used to train each estimator, against baselines, for each \texttt{dataset}.  When the splitting stump forest is adopted, subscripts indicate the average number of stumps $\pm$ std. dev. Best results in \textbf{bold}, second best in \textit{italics}.}
    \label{tab:ensemble_sample_10perc_bal_acc}
\centering
\footnotesize
\setlength{\tabcolsep}{0.3mm} % Reduce column spacing
\begin{tabular}{l|lll|ll|llll|ll|llll|ll|llll}
\toprule
\texttt{dataset} & $\textsc{xgb}$ & $\textsc{lgbm}$ & $\textsc{catb}$ & $\textsc{rdt}$ & $\textsc{rodt}$ & $\textsc{rpt}$ & $\textsc{ropt}$ & $\textsc{rppt}$ & $\textsc{roppt}$ & $\textsc{rdt}_{\textsc{s}}$ & $\textsc{rodt}_{\textsc{s}}$ & $\textsc{rpt}_{\textsc{s}}$ & $\textsc{ropt}_{\textsc{s}}$ & $\textsc{rppt}_{\textsc{s}}$ & $\textsc{roppt}_{\textsc{s}}$ & $\textsc{rdt}_{\textsc{b}}$ & $\textsc{rodt}_{\textsc{b}}$ & $\textsc{rpt}_{\textsc{b}}$ & $\textsc{ropt}_{\textsc{b}}$ & $\textsc{rppt}_{\textsc{b}}$ & $\textsc{roppt}_{\textsc{b}}$ \\
\midrule
\texttt{ion} & $\textit{.91}_{}$ & $\textbf{.94}_{}$ & $\textbf{.94}_{}$ & $.87_{}$ & $.88_{}$ & $.87_{}$ & $.89_{}$ & $.82_{}$ & $.82_{}$ & $.83_{46}$ & $.83_{41}$ & $.87_{37}$ & $.86_{36}$ & $.90_{72}$ & $.89_{103}$ & $.50_{17}$ & $.50_{74}$ & $.70_{54}$ & $.78_{23}$ & $.64_{26}$ & $.58_{40}$ \\
\texttt{fire} & $\textit{.98}_{}$ & $\textit{.98}_{}$ & $\textbf{.99}_{}$ & $.95_{}$ & $.95_{}$ & $.93_{}$ & $.91_{}$ & $.93_{}$ & $.93_{}$ & $.56_{14}$ & $.68_{53}$ & $.69_{80}$ & $.72_{83}$ & $.74_{83}$ & $.70_{109}$ & $.87_{98}$ & $.89_{101}$ & $.70_{107}$ & $.79_{99}$ & $.76_{109}$ & $.70_{73}$ \\
\texttt{yeast} & $.46_{}$ & $\textit{.49}_{}$ & $\textbf{.52}_{}$ & $.30_{}$ & $.25_{}$ & $.30_{}$ & $.26_{}$ & $.29_{}$ & $.28_{}$ & $.25_{29}$ & $.30_{51}$ & $.27_{40}$ & $.33_{45}$ & $.26_{59}$ & $.28_{62}$ & $.12_{21}$ & $.10_{4}$ & $.10_{11}$ & $.13_{36}$ & $.11_{5}$ & $.11_{91}$ \\
\texttt{magic} & $\textit{.84}_{}$ & $\textbf{.86}_{}$ & $\textit{.84}_{}$ & $.64_{}$ & $.64_{}$ & $.64_{}$ & $.63_{}$ & $.63_{}$ & $.63_{}$ & $.68_{53}$ & $.70_{57}$ & $.70_{44}$ & $.70_{67}$ & $.73_{267}$ & $.70_{68}$ & $.52_{34}$ & $.53_{28}$ & $.62_{10}$ & $.61_{25}$ & $.50_{91}$ & $.50_{30}$ \\
\texttt{sonar} & $.78_{}$ & $\textbf{.87}_{}$ & $\textit{.84}_{}$ & $.76_{}$ & $.75_{}$ & $.79_{}$ & $.65_{}$ & $.83_{}$ & $.80_{}$ & $.62_{102}$ & $.60_{123}$ & $.65_{53}$ & $.64_{10}$ & $.59_{8}$ & $.60_{14}$ & $.73_{100}$ & $.62_{76}$ & $.62_{83}$ & $.61_{10}$ & $.60_{151}$ & $.57_{14}$ \\
\texttt{compas} & $\textbf{.56}_{}$ & $\textit{.55}_{}$ & $\textit{.55}_{}$ & $.43_{}$ & $.43_{}$ & $.44_{}$ & $.44_{}$ & $.43_{}$ & $.43_{}$ & $.42_{7}$ & $.45_{33}$ & $.42_{40}$ & $.43_{57}$ & $.43_{58}$ & $.44_{56}$ & $.33_{36}$ & $.33_{70}$ & $.33_{96}$ & $.33_{89}$ & $.33_{71}$ & $.33_{85}$ \\
\texttt{house} & $\textbf{.87}_{}$ & $\textbf{.87}_{}$ & $\textbf{.87}_{}$ & $.73_{}$ & $.73_{}$ & $.72_{}$ & $.65_{}$ & $.66_{}$ & $.66_{}$ & $\textit{.75}_{179}$ & $\textit{.75}_{223}$ & $.71_{187}$ & $.70_{209}$ & $.69_{415}$ & $.71_{455}$ & $.50_{158}$ & $.50_{155}$ & $.51_{9}$ & $.50_{109}$ & $.50_{115}$ & $.50_{123}$ \\
\texttt{german} & $.66_{}$ & $\textbf{.70}_{}$ & $\textit{.68}_{}$ & $.50_{}$ & $.50_{}$ & $.50_{}$ & $.50_{}$ & $.50_{}$ & $.50_{}$ & $.55_{7}$ & $.63_{15}$ & $.63_{42}$ & $.62_{39}$ & $.62_{37}$ & $.63_{21}$ & $.50_{10}$ & $.50_{10}$ & $.50_{8}$ & $.50_{11}$ & $.50_{2}$ & $.50_{41}$ \\
\texttt{spamb} & $\textit{.94}_{}$ & $\textbf{.95}_{}$ & $\textbf{.95}_{}$ & $.86_{}$ & $.86_{}$ & $.87_{}$ & $.85_{}$ & $.84_{}$ & $.83_{}$ & $.78_{56}$ & $.79_{99}$ & $.77_{116}$ & $.77_{181}$ & $.79_{108}$ & $.80_{175}$ & $.67_{73}$ & $.74_{98}$ & $.72_{86}$ & $.55_{116}$ & $.69_{123}$ & $.69_{99}$ \\
\texttt{norm} & $\textit{.97}_{}$ & $\textbf{.98}_{}$ & $\textit{.97}_{}$ & $\textit{.97}_{}$ & $\textit{.97}_{}$ & $\textbf{.98}_{}$ & $\textit{.97}_{}$ & $\textbf{.98}_{}$ & $\textbf{.98}_{}$ & $.81_{222}$ & $.81_{49}$ & $.81_{71}$ & $.82_{45}$ & $.81_{63}$ & $.81_{68}$ & $.95_{226}$ & $\textit{.97}_{118}$ & $\textit{.97}_{100}$ & $\textbf{.98}_{87}$ & $\textbf{.98}_{95}$ & $\textbf{.98}_{106}$ \\
\texttt{lrs} & $.43_{}$ & $\textbf{.63}_{}$ & $\textit{.50}_{}$ & $.42_{}$ & $.41_{}$ & $.40_{}$ & $.39_{}$ & $.41_{}$ & $.42_{}$ & $.25_{252}$ & $.25_{237}$ & $.21_{149}$ & $.21_{137}$ & $.22_{116}$ & $.22_{56}$ & $.13_{280}$ & $.14_{186}$ & $.15_{158}$ & $.15_{165}$ & $.12_{93}$ & $.12_{172}$ \\
\texttt{vert} & $\textbf{.82}_{}$ & $.80_{}$ & $\textit{.81}_{}$ & $.64_{}$ & $.63_{}$ & $.66_{}$ & $.63_{}$ & $.72_{}$ & $.72_{}$ & $.58_{21}$ & $.50_{66}$ & $.50_{68}$ & $.54_{67}$ & $.54_{151}$ & $.50_{46}$ & $.50_{75}$ & $.50_{86}$ & $.46_{106}$ & $.49_{102}$ & $.51_{152}$ & $.60_{111}$ \\
\texttt{iris} & $\textit{.97}_{}$ & $\textbf{1.00}_{}$ & $\textbf{1.00}_{}$ & $\textbf{1.00}_{}$ & $\textit{.97}_{}$ & $.95_{}$ & $\textbf{1.00}_{}$ & $\textbf{1.00}_{}$ & $\textbf{1.00}_{}$ & $.77_{21}$ & $.67_{113}$ & $.67_{38}$ & $.67_{35}$ & $.72_{59}$ & $.70_{19}$ & $.69_{123}$ & $.54_{53}$ & $.85_{38}$ & $.87_{36}$ & $.68_{20}$ & $.87_{19}$ \\
\texttt{wine} & $\textbf{.99}_{}$ & $\textbf{.99}_{}$ & $\textbf{.99}_{}$ & $.95_{}$ & $.96_{}$ & $\textit{.97}_{}$ & $.96_{}$ & $\textit{.97}_{}$ & $.96_{}$ & $.87_{89}$ & $.88_{88}$ & $.88_{129}$ & $.87_{133}$ & $.87_{48}$ & $.89_{47}$ & $.55_{82}$ & $.52_{94}$ & $.53_{3}$ & $.50_{133}$ & $.50_{84}$ & $.50_{1}$ \\
\texttt{diva} & $\textit{.88}_{}$ & $\textbf{.90}_{}$ & $.87_{}$ & $.60_{}$ & $.60_{}$ & $.56_{}$ & $.53_{}$ & $.52_{}$ & $.50_{}$ & $.71_{34}$ & $.71_{53}$ & $.69_{74}$ & $.69_{69}$ & $.71_{133}$ & $.72_{162}$ & $.50_{34}$ & $.50_{34}$ & $.50_{75}$ & $.50_{54}$ & $.50_{40}$ & $.50_{46}$ \\
\texttt{breast} & $\textbf{.98}_{}$ & $.95_{}$ & $\textbf{.98}_{}$ & $.95_{}$ & $.94_{}$ & $.94_{}$ & $.93_{}$ & $.94_{}$ & $.94_{}$ & $.94_{53}$ & $.95_{64}$ & $.93_{92}$ & $.94_{102}$ & $\textit{.96}_{22}$ & $.94_{19}$ & $.92_{62}$ & $.86_{12}$ & $.91_{72}$ & $.92_{89}$ & $.89_{100}$ & $.89_{95}$ \\
\texttt{steel} & $.74_{}$ & $\textbf{.82}_{}$ & $\textit{.77}_{}$ & $.41_{}$ & $.40_{}$ & $.39_{}$ & $.38_{}$ & $.43_{}$ & $.47_{}$ & $.43_{141}$ & $.46_{133}$ & $.46_{70}$ & $.45_{75}$ & $.48_{270}$ & $.51_{205}$ & $.18_{68}$ & $.14_{11}$ & $.18_{6}$ & $.18_{2}$ & $.12_{11}$ & $.25_{8}$ \\
\texttt{ecoli} & $.69_{}$ & $\textit{.70}_{}$ & $\textbf{.76}_{}$ & $.36_{}$ & $.36_{}$ & $.36_{}$ & $.37_{}$ & $.38_{}$ & $.38_{}$ & $.36_{20}$ & $.35_{105}$ & $.34_{158}$ & $.31_{140}$ & $.33_{141}$ & $.37_{12}$ & $.21_{19}$ & $.26_{37}$ & $.26_{133}$ & $.25_{22}$ & $.23_{13}$ & $.20_{88}$ \\
\texttt{heloc} & $\textbf{.71}_{}$ & $\textbf{.71}_{}$ & $\textbf{.71}_{}$ & $.69_{}$ & $.69_{}$ & $\textit{.70}_{}$ & $\textit{.70}_{}$ & $\textit{.70}_{}$ & $\textit{.70}_{}$ & $.69_{50}$ & $\textit{.70}_{31}$ & $.69_{91}$ & $.69_{133}$ & $.69_{113}$ & $.69_{92}$ & $.52_{75}$ & $.61_{134}$ & $.66_{102}$ & $.53_{34}$ & $.49_{127}$ & $.51_{92}$ \\
\texttt{page} & $.88_{}$ & $\textbf{.93}_{}$ & $.89_{}$ & $.74_{}$ & $.71_{}$ & $.78_{}$ & $.74_{}$ & $.73_{}$ & $.73_{}$ & $.74_{134}$ & $.82_{130}$ & $.79_{101}$ & $.84_{92}$ & $\textit{.90}_{143}$ & $.89_{149}$ & $.50_{0}$ & $.50_{0}$ & $.53_{101}$ & $.51_{53}$ & $.50_{19}$ & $.50_{8}$ \\
\texttt{yoga} & $.68_{}$ & $\textit{.77}_{}$ & $\textbf{.79}_{}$ & $.61_{}$ & $.61_{}$ & $.64_{}$ & $.65_{}$ & $.67_{}$ & $.68_{}$ & $.63_{144}$ & $.62_{155}$ & $.62_{99}$ & $.66_{69}$ & $.67_{123}$ & $.67_{139}$ & $.50_{13}$ & $.50_{51}$ & $.52_{92}$ & $.51_{14}$ & $.50_{70}$ & $.51_{23}$ \\
\texttt{star} & $.89_{}$ & $\textbf{.93}_{}$ & $\textit{.90}_{}$ & $.66_{}$ & $.66_{}$ & $.81_{}$ & $.81_{}$ & $.68_{}$ & $.67_{}$ & $.68_{235}$ & $.70_{224}$ & $.70_{114}$ & $.69_{104}$ & $.73_{284}$ & $.70_{217}$ & $.66_{100}$ & $.66_{100}$ & $.33_{0}$ & $.33_{111}$ & $.33_{4}$ & $.33_{0}$ \\
\texttt{chlorine} & $.18_{}$ & $.12_{}$ & $.17_{}$ & $.29_{}$ & $.30_{}$ & $.28_{}$ & $.30_{}$ & $\textit{.31}_{}$ & $.30_{}$ & $.29_{98}$ & $.28_{172}$ & $.30_{43}$ & $.30_{38}$ & $.26_{645}$ & $.27_{279}$ & $\textbf{.33}_{13}$ & $\textbf{.33}_{20}$ & $\textbf{.33}_{21}$ & $.29_{7}$ & $\textbf{.33}_{150}$ & $\textbf{.33}_{125}$ \\
\texttt{kitchen} & $.67_{}$ & $\textit{.70}_{}$ & $\textbf{.71}_{}$ & $.64_{}$ & $.64_{}$ & $.69_{}$ & $.67_{}$ & $.63_{}$ & $.61_{}$ & $.65_{134}$ & $.65_{86}$ & $.66_{109}$ & $.64_{101}$ & $.65_{145}$ & $.62_{145}$ & $.52_{96}$ & $.64_{166}$ & $.47_{35}$ & $.47_{37}$ & $.59_{99}$ & $.44_{113}$ \\
\texttt{share} & $\textbf{.54}_{}$ & $\textbf{.54}_{}$ & $\textbf{.54}_{}$ & $.50_{}$ & $.50_{}$ & $.50_{}$ & $.50_{}$ & $.50_{}$ & $.50_{}$ & $.50_{6}$ & $\textit{.52}_{70}$ & $.47_{257}$ & $.51_{159}$ & $.51_{29}$ & $.51_{76}$ & $.50_{5}$ & $.50_{6}$ & $.50_{43}$ & $\textit{.52}_{6}$ & $.50_{52}$ & $.50_{65}$ \\
\texttt{devices} & $.52_{}$ & $\textbf{.58}_{}$ & $\textit{.54}_{}$ & $.36_{}$ & $.36_{}$ & $.35_{}$ & $.36_{}$ & $.35_{}$ & $.34_{}$ & $.33_{701}$ & $.33_{718}$ & $.32_{246}$ & $.33_{320}$ & $.34_{236}$ & $.33_{181}$ & $.21_{214}$ & $.21_{221}$ & $.21_{77}$ & $.19_{78}$ & $.13_{92}$ & $.15_{5}$ \\
\texttt{gun} & $.68_{}$ & $.63_{}$ & $\textbf{.94}_{}$ & $\textit{.82}_{}$ & $.80_{}$ & $.72_{}$ & $.70_{}$ & $.74_{}$ & $.76_{}$ & $.66_{60}$ & $.65_{61}$ & $.68_{58}$ & $.63_{53}$ & $.63_{16}$ & $.67_{15}$ & $.51_{49}$ & $.48_{77}$ & $.74_{66}$ & $.72_{70}$ & $.51_{19}$ & $.51_{19}$ \\
\texttt{worms} & $.56_{}$ & $\textit{.61}_{}$ & $.55_{}$ & $.48_{}$ & $.55_{}$ & $.55_{}$ & $.59_{}$ & $.58_{}$ & $.57_{}$ & $.52_{94}$ & $.46_{12}$ & $.53_{19}$ & $.52_{89}$ & $.43_{112}$ & $.52_{168}$ & $.49_{35}$ & $\textbf{.62}_{12}$ & $.53_{107}$ & $.58_{12}$ & $.55_{42}$ & $.50_{171}$ \\
\texttt{ecg} & $.48_{}$ & $\textbf{.53}_{}$ & $\textit{.51}_{}$ & $.39_{}$ & $.39_{}$ & $.40_{}$ & $.39_{}$ & $.39_{}$ & $.39_{}$ & $.48_{215}$ & $.48_{186}$ & $.41_{124}$ & $.44_{93}$ & $.45_{95}$ & $.46_{113}$ & $.33_{97}$ & $.20_{166}$ & $.24_{88}$ & $.26_{93}$ & $.30_{108}$ & $.37_{105}$ \\
\texttt{wafer} & $.91_{}$ & $\textbf{.98}_{}$ & $\textit{.97}_{}$ & $.74_{}$ & $.75_{}$ & $.88_{}$ & $.87_{}$ & $.95_{}$ & $.96_{}$ & $.80_{46}$ & $.81_{93}$ & $.71_{44}$ & $.78_{41}$ & $.86_{516}$ & $.89_{562}$ & $.50_{4}$ & $.49_{5}$ & $.50_{0}$ & $.55_{16}$ & $.50_{105}$ & $.50_{99}$ \\
\texttt{oral} & $\textbf{.88}_{}$ & $\textbf{.88}_{}$ & $\textbf{.88}_{}$ & $\textit{.85}_{}$ & $.84_{}$ & $.82_{}$ & $.79_{}$ & $.82_{}$ & $.84_{}$ & $.70_{54}$ & $.69_{264}$ & $.64_{177}$ & $.79_{158}$ & $.74_{251}$ & $.75_{175}$ & $.28_{105}$ & $.26_{128}$ & $.46_{112}$ & $.42_{93}$ & $.27_{88}$ & $.36_{109}$ \\
\texttt{MNIST} & $\textbf{.99}_{}$ & $\textbf{.99}_{}$ & $\textbf{.99}_{}$ & $\textbf{.99}_{}$ & $\textbf{.99}_{}$ & $\textbf{.99}_{}$ & $\textbf{.99}_{}$ & $\textbf{.99}_{}$ & $\textbf{.99}_{}$ & $\textbf{.99}_{124}$ & $\textbf{.99}_{130}$ & $.50_{35}$ & $.50_{34}$ & $\textbf{.99}_{42}$ & $\textit{.79}_{11}$ & $.10_{32}$ & $.20_{130}$ & $.10_{0}$ & $.10_{0}$ & $.12_{18}$ & $.10_{0}$ \\
\texttt{cifar10} & $\textbf{.95}_{}$ & $\textbf{.95}_{}$ & $\textbf{.95}_{}$ & $\textbf{.95}_{}$ & $\textbf{.95}_{}$ & $\textbf{.95}_{}$ & $\textbf{.95}_{}$ & $\textbf{.95}_{}$ & $\textbf{.95}_{}$ & $\textit{.94}_{74}$ & $\textit{.94}_{72}$ & $.60_{62}$ & $.64_{53}$ & $\textit{.94}_{36}$ & $\textit{.94}_{49}$ & $.07_{95}$ & $.11_{51}$ & $.10_{0}$ & $.10_{0}$ & $.09_{44}$ & $.15_{61}$ \\
\texttt{catsdogs} & $\textbf{.95}_{}$ & $\textbf{.95}_{}$ & $\textbf{.95}_{}$ & $\textbf{.95}_{}$ & $\textbf{.95}_{}$ & $\textbf{.95}_{}$ & $\textbf{.95}_{}$ & $\textbf{.95}_{}$ & $\textbf{.95}_{}$ & $\textbf{.95}_{89}$ & $\textbf{.95}_{133}$ & $\textbf{.95}_{91}$ & $\textbf{.95}_{32}$ & $\textbf{.95}_{304}$ & $\textbf{.95}_{107}$ & $.50_{142}$ & $.50_{107}$ & $\textbf{.95}_{100}$ & $\textbf{.95}_{100}$ & $\textit{.94}_{109}$ & $.80_{489}$ \\
\texttt{birds} & $\textit{.83}_{}$ & $\textbf{.84}_{}$ & $\textbf{.84}_{}$ & $.81_{}$ & $.81_{}$ & $.81_{}$ & $.79_{}$ & $.82_{}$ & $.82_{}$ & $.81_{28}$ & $.81_{21}$ & $.79_{81}$ & $.81_{55}$ & $.82_{38}$ & $.82_{163}$ & $.77_{118}$ & $.76_{68}$ & $.74_{39}$ & $.57_{2}$ & $.79_{29}$ & $.78_{30}$ \\
\texttt{pets} & $\textbf{.94}_{}$ & $\textbf{.94}_{}$ & $\textbf{.94}_{}$ & $\textbf{.94}_{}$ & $\textbf{.94}_{}$ & $\textbf{.94}_{}$ & $\textbf{.94}_{}$ & $\textbf{.94}_{}$ & $\textbf{.94}_{}$ & $\textbf{.94}_{6}$ & $\textbf{.94}_{47}$ & $\textit{.93}_{90}$ & $\textit{.93}_{56}$ & $\textbf{.94}_{89}$ & $\textbf{.94}_{254}$ & $.50_{0}$ & $.50_{0}$ & $.62_{1}$ & $.41_{1}$ & $.52_{6}$ & $.50_{7}$ \\
\texttt{organa} & $\textit{.99}_{}$ & $\textbf{1.00}_{}$ & $\textbf{1.00}_{}$ & $.94_{}$ & $.95_{}$ & $.63_{}$ & $.63_{}$ & $.96_{}$ & $.96_{}$ & $\textit{.99}_{849}$ & $\textit{.99}_{496}$ & $.36_{10}$ & $.42_{8}$ & $.97_{85}$ & $.98_{111}$ & $.09_{21}$ & $.09_{913}$ & $.09_{0}$ & $.09_{0}$ & $.09_{8}$ & $.08_{21}$ \\
\texttt{blood} & $\textit{.93}_{}$ & $\textbf{.94}_{}$ & $\textbf{.94}_{}$ & $.87_{}$ & $.87_{}$ & $.69_{}$ & $.63_{}$ & $.90_{}$ & $.89_{}$ & $.92_{739}$ & $.92_{509}$ & $.55_{59}$ & $.72_{51}$ & $.92_{310}$ & $.92_{367}$ & $.12_{134}$ & $.11_{145}$ & $.08_{7}$ & $.08_{4}$ & $.16_{111}$ & $.19_{148}$ \\
\texttt{SVHN} & $\textbf{.95}_{}$ & $\textbf{.95}_{}$ & $\textbf{.95}_{}$ & $\textbf{.95}_{}$ & $\textbf{.95}_{}$ & $.67_{}$ & $.68_{}$ & $\textbf{.95}_{}$ & $\textit{.94}_{}$ & $\textbf{.95}_{29}$ & $\textbf{.95}_{28}$ & $.39_{14}$ & $.23_{7}$ & $\textbf{.95}_{68}$ & $\textbf{.95}_{64}$ & $.19_{530}$ & $.19_{521}$ & $.19_{13}$ & $.11_{9}$ & $.10_{35}$ & $.16_{5}$ \\
\texttt{medabs} & $\textit{.61}_{}$ & $\textbf{.62}_{}$ & $.60_{}$ & $.40_{}$ & $.40_{}$ & $.42_{}$ & $.41_{}$ & $.44_{}$ & $.43_{}$ & $.51_{230}$ & $.52_{228}$ & $.50_{154}$ & $.52_{182}$ & $.52_{167}$ & $.54_{255}$ & $.20_{199}$ & $.20_{200}$ & $.20_{0}$ & $.20_{23}$ & $.20_{19}$ & $.20_{26}$ \\
\texttt{vicuna} & $.84_{}$ & $\textbf{.87}_{}$ & $\textit{.85}_{}$ & $.50_{}$ & $.50_{}$ & $.61_{}$ & $.52_{}$ & $.51_{}$ & $.50_{}$ & $.65_{169}$ & $.65_{166}$ & $.65_{190}$ & $.64_{137}$ & $.66_{231}$ & $.68_{269}$ & $.50_{82}$ & $.50_{82}$ & $.50_{256}$ & $.50_{70}$ & $.50_{22}$ & $.50_{21}$ \\
\texttt{pTED} & $\textbf{.75}_{}$ & $\textbf{.75}_{}$ & $\textit{.73}_{}$ & $.50_{}$ & $.50_{}$ & $.54_{}$ & $.50_{}$ & $.51_{}$ & $.50_{}$ & $.59_{155}$ & $.58_{227}$ & $.58_{154}$ & $.58_{195}$ & $.61_{232}$ & $.58_{138}$ & $.50_{63}$ & $.50_{66}$ & $.50_{202}$ & $.50_{19}$ & $.50_{62}$ & $.50_{68}$ \\
\texttt{tgpt} & $\textit{.96}_{}$ & $\textbf{.97}_{}$ & $\textit{.96}_{}$ & $.90_{}$ & $.90_{}$ & $.91_{}$ & $.90_{}$ & $.91_{}$ & $.90_{}$ & $.78_{152}$ & $.76_{366}$ & $.80_{269}$ & $.77_{236}$ & $.78_{236}$ & $.79_{251}$ & $.66_{100}$ & $.60_{100}$ & $.51_{132}$ & $.51_{111}$ & $.44_{128}$ & $.42_{251}$ \\
\texttt{pol} & $.69_{}$ & $\textbf{.74}_{}$ & $\textit{.70}_{}$ & $\textit{.70}_{}$ & $.68_{}$ & $\textit{.70}_{}$ & $.66_{}$ & $.69_{}$ & $.69_{}$ & $.54_{38}$ & $.55_{36}$ & $.60_{52}$ & $.59_{48}$ & $.59_{89}$ & $.58_{41}$ & $.50_{115}$ & $.50_{58}$ & $.54_{89}$ & $.53_{71}$ & $.52_{111}$ & $.52_{41}$ \\
\texttt{liar} & $\textit{.23}_{}$ & $\textbf{.24}_{}$ & $\textit{.23}_{}$ & $.17_{}$ & $.17_{}$ & $.20_{}$ & $.20_{}$ & $.19_{}$ & $.19_{}$ & $.19_{151}$ & $.19_{159}$ & $.18_{68}$ & $.20_{156}$ & $.19_{55}$ & $.19_{99}$ & $.17_{27}$ & $.17_{7}$ & $.17_{87}$ & $.17_{66}$ & $.17_{60}$ & $.18_{54}$ \\
\midrule
\texttt{Avg.} & $\textit{.77}_{}$ & $\textbf{.79}_{}$ & $\textbf{.79}_{}$ & $.68_{}$ & $.68_{}$ & $.68_{}$ & $.66_{}$ & $.69_{}$ & $.69_{}$ & $.66_{136.67}$ & $.66_{142.73}$ & $.60_{95.53}$ & $.61_{94.00}$ & $.67_{151.11}$ & $.67_{137.49}$ & $.45_{88.44}$ & $.45_{106.24}$ & $.46_{\textit{65.00}}$ & $.45_{\textbf{51.29}}$ & $.44_{67.51}$ & $.44_{75.51}$ \\
\texttt{Std.} & $.21_{}$ & $.20_{}$ & $.20_{}$ & $\textit{.23}_{}$ & $\textit{.23}_{}$ & $.22_{}$ & $.22_{}$ & ${.23}_{}$ & ${.23}_{}$ & $.22_{183.79}$ & $.21_{142.60}$ & $.20_{63.59}$ & $.20_{66.77}$ & $.22_{134.26}$ & $.21_{117.96}$ & ${.23}_{93.04}$ & $.22_{151.27}$ & ${.24}_{58.37}$ & ${.24}_{{44.49}}$ & ${.24}_{46.50}$ & $.22_{83.87}$ \\
\texttt{Rank} & $4.2_{5.0}$ & $3.0_{5.0}$ & $3.3_{5.0}$ & $9.7_{5.0}$ & $10.1_{5.0}$ & $8.5_{5.0}$ & $9.6_{5.0}$ & $8.4_{5.0}$ & $9.0_{5.0}$ & $10.7_{15.2}$ & $10.0_{14.2}$ & $12.1_{14.7}$ & $11.2_{15.2}$ & $9.7_{13.8}$ & $9.5_{14.1}$ & $17.3_{16.5}$ & $17.1_{15.7}$ & $16.4_{16.6}$ & $16.4_{17.6}$ & $17.4_{16.2}$ & $17.4_{16.1}$\\
\bottomrule
\end{tabular}
\end{sidewaystable}

\begin{table}[t]
\caption{Average Balanced Accuracy $\pm$ std. dev. for \textsc{PivotTree} classifiers and selectors combined with \textsc{dt} in the unconstrained setting, alongside baselines and competing methods, with respect to different distance measures.
Subscripts denote the average number of pivots $\pm$ standard deviation (rounded up to nearest integer).
Best results are shown in \textbf{bold}, and second-best results in \textit{italics}.}
\centering
\setlength{\tabcolsep}{2.2mm}
\label{ref:tab_dist_comp_standalone_bal_cc}
\begin{tabular}{c c c c c c c}
\toprule
 & model & $\texttt{tabular}$ & $\texttt{images}$ & $\texttt{time-series}$ & $\texttt{text}$ & \texttt{all} \\
\midrule
\multirow{8}{*}{\rotatebox{90}{\texttt{euclidean}}} 

& $\textsc{ptc}_{Z}$ 
& $.73_{\textbf{11}} \pm .20_{3}$ 
& $.66_{\textit{11}} \pm .20_{4}$ 
& $.59_{\textbf{9}} \pm .21_{3}$ 
& $.59_{13} \pm .21_{2}$ 
& $.67_{\textbf{11}} \pm .20_{3}$ \\

& $\textsc{pptc}_{Z}$ 
& $.72_{14} \pm .18_{4}$ 
& $.70_{\textit{11}} \pm .18_{5}$ 
& $.57_{12} \pm .21_{6}$ 
& $.59_{\textit{12}} \pm .20_{3}$ 
& $.67_{13} \pm .19_{5}$ \\

& $\textsc{optc}_{Z}$ 
& $.68_{17} \pm .19_{6}$ 
& $.65_{15} \pm .19_{6}$ 
& $.57_{12} \pm .21_{6}$ 
& $.56_{23} \pm .20_{4}$ 
& $.63_{16} \pm .20_{6}$ \\

& $\textsc{opptc}_{Z}$ 
& $.73_{13} \pm .18_{5}$ 
& $.71_{12} \pm .19_{6}$ 
& $.55_{11} \pm .20_{7}$ 
& $.55_{14} \pm .20_{4}$ 
& $.66_{\textit{12}} \pm .20_{5}$ \\

\cmidrule(lr){2-7}

& $\textsc{pts}_{Z}$ 
& $.73_{39} \pm .18_{19}$ 
& $.66_{50} \pm .20_{21}$ 
& $.59_{35} \pm .21_{31}$ 
& $.59_{61} \pm .21_{32}$ 
& $.67_{43} \pm .20_{25}$ \\

& $\textsc{ppts}_{Z}$ 
& $.71_{37} \pm .18_{24}$ 
& $.67_{55} \pm .21_{39}$ 
& $.59_{37} \pm .20_{36}$ 
& $\textit{.60}_{51} \pm .21_{45}$ 
& $.66_{43} \pm .20_{33}$ \\

& $\textsc{opts}_{Z}$ 
& $.71_{50} \pm .19_{30}$ 
& $.65_{66} \pm .20_{36}$ 
& $.58_{47} \pm .19_{48}$ 
& $.58_{96} \pm .21_{68}$ 
& $.65_{59} \pm .20_{43}$ \\

& $\textsc{oppts}_{Z}$ 
& $.72_{46} \pm .20_{31}$ 
& $.66_{69} \pm .20_{49}$ 
& $.55_{40} \pm .20_{34}$ 
& $.57_{76} \pm .19_{60}$ 
& $.65_{53} \pm .20_{41}$ \\

\midrule
\multirow{8}{*}{\rotatebox{90}{\texttt{cosine}}} 

& $\textsc{ptc}_{Z}$ 
& $\textbf{.75}_{\textbf{11}} \pm .18_{3}$ 
& $.66_{\textit{11}} \pm .20_{4}$ 
& $\textbf{.61}_{\textit{10}} \pm .19_{4}$ 
& $\textit{.60}_{\textit{12}} \pm .21_{2}$ 
& $.68_{\textbf{11}} \pm .20_{3}$ \\

& $\textsc{pptc}_{Z}$ 
& $.\textbf{75}_{\textit{12}} \pm .17_{5}$ 
& $\textbf{.85}_{15} \pm .12_{4}$ 
& $\textit{.60}_{17} \pm .20_{7}$ 
& $.59_{14} \pm .22_{5}$ 
& $\textbf{.71}_{14} \pm .19_{6}$ \\

& $\textsc{optc}_{Z}$ 
& $.71_{17} \pm .20_{5}$ 
& $.66_{16} \pm .20_{5}$ 
& $\textbf{.61}_{13} \pm .20_{5}$ 
& $.58_{20} \pm .21_{5}$ 
& $.66_{16} \pm .20_{5}$ \\

& $\textsc{opptc}_{Z}$ 
& $\textit{.74}_{13} \pm .18_{5}$ 
& $\textit{.79}_{15} \pm .13_{5}$ 
& $.57_{15} \pm .21_{8}$ 
& $.59_{\textbf{11}} \pm .21_{4}$ 
& $\textit{.69}_{14} \pm .19_{6}$ \\

\cmidrule(lr){2-7}

& $\textsc{pts}_{Z}$ 
& $\textbf{.75}_{42} \pm .18_{23}$ 
& $.66_{51} \pm .20_{24}$ 
& $\textbf{.61}_{36} \pm .20_{31}$ 
& $\textbf{.61}_{62} \pm .21_{34}$ 
& $.68_{45} \pm .20_{27}$ \\

& $\textsc{ppts}_{Z}$ 
& $.73_{34} \pm .20_{31}$ 
& $.66_{77} \pm .20_{55}$ 
& $\textit{.60}_{46} \pm .19_{36}$ 
& $\textit{.60}_{54} \pm .22_{45}$ 
& $.67_{48} \pm .20_{42}$ \\

& $\textsc{opts}_{Z}$ 
& $.74_{50} \pm .19_{26}$ 
& $.66_{69} \pm .20_{36}$ 
& $.59_{43} \pm .19_{33}$ 
& $\textit{.60}_{87} \pm .21_{63}$ 
& $.67_{57} \pm .20_{38}$ \\

& $\textsc{oppts}_{Z}$ 
& $.72_{51} \pm .18_{42}$ 
& $.66_{111} \pm .20_{72}$ 
& $.59_{61} \pm .19_{56}$ 
& $.59_{67} \pm .21_{52}$ 
& $.66_{67} \pm .19_{56}$ \\

\midrule
\multirow{8}{*}{\rotatebox{90}{\texttt{manhattan}}} 

& $\textsc{ptc}_{Z}$ 
& $.72_{\textit{12}} \pm .19_{3}$ 
& $.66_{\textbf{10}} \pm .20_{3}$ 
& $.58_{\textbf{9}} \pm .21_{4}$ 
& $.58_{13} \pm .20_{2}$ 
& $.66_{\textbf{11}} \pm .20_{3}$ \\

& $\textsc{pptc}_{Z}$ 
& $\textit{.74}_{13} \pm .18_{5}$ 
& $.71_{12} \pm .19_{6}$ 
& $\textit{.60}_{14} \pm .21_{6}$ 
& $.58_{\textit{12}} \pm .21_{3}$ 
& $.68_{13} \pm .20_{5}$ \\

& $\textsc{optc}_{Z}$ 
& $.70_{16} \pm .20_{5}$ 
& $.66_{16} \pm .20_{6}$ 
& $.58_{15} \pm .20_{6}$ 
& $.55_{22} \pm .19_{4}$ 
& $.64_{17} \pm .20_{6}$ \\

& $\textsc{opptc}_{Z}$ 
& $\textit{.74}_{14} \pm .17_{4}$ 
& $.70_{12} \pm .19_{4}$ 
& $\textit{.60}_{12} \pm .21_{6}$ 
& $.57_{13} \pm .20_{2}$ 
& $.68_{13} \pm .20_{4}$ \\

\cmidrule(lr){2-7}

& $\textsc{pts}_{Z}$ 
& $.73_{41} \pm .19_{20}$ 
& $.66_{48} \pm .20_{25}$ 
& $.58_{36} \pm .21_{31}$ 
& $.59_{63} \pm .20_{38}$ 
& $.67_{44} \pm .20_{27}$ \\

& $\textsc{ppts}_{Z}$ 
& $.73_{35} \pm .20_{24}$ 
& $.67_{54} \pm .21_{38}$ 
& $\textit{.60}_{37} \pm .19_{27}$ 
& $.59_{50} \pm .21_{35}$ 
& $.67_{42} \pm .20_{29}$ \\

& $\textsc{opts}_{Z}$ 
& $.73_{49} \pm .19_{21}$ 
& $.67_{70} \pm .20_{37}$ 
& $\textit{.60}_{48} \pm .20_{37}$ 
& $.56_{88} \pm .19_{66}$ 
& $.67_{58} \pm .20_{38}$ \\

& $\textsc{oppts}_{Z}$ 
& $.72_{51} \pm .18_{31}$ 
& $.67_{74} \pm .21_{39}$ 
& $\textit{.60}_{46} \pm .18_{36}$ 
& $.56_{66} \pm .20_{43}$ 
& $.66_{56} \pm .19_{36}$ \\

\bottomrule
\end{tabular}
\end{table}

\begin{table}[t]
\caption{
Average Balanced Accuracy $\pm$ std. dev. for \textsc{RandomPivotForest} with 100 \textsc{PivotTree} estimators, where each estimator is trained on 10\% of the training set.
Results are reported for different distance measures.
Best results are shown in \textbf{bold}, and second-best results in \textit{italics}.
}
\centering
\setlength{\tabcolsep}{2.2mm}
\label{ref:tab_dist_comp_ensemble_bal_acc_small}
\begin{tabular}{c c c c c c c}
\toprule
 & model & \texttt{tabular} & \texttt{images} & \texttt{time-series} & \texttt{text} & \texttt{all} \\
\midrule

\multirow{4}{*}{\rotatebox{90}{\texttt{euclidean}}}
& $\textsc{rpt}$   
& \textit{.68} $\pm$ .23
& .83 $\pm$ .14
& \textit{.58} $\pm$ .20
& \textit{.56} $\pm$ .24
& .67 $\pm$ .22 \\

& $\textsc{rppt}$  
& \textbf{.69} $\pm$ .23
& \textit{.92} $\pm$ .06
& \textbf{.59} $\pm$ .20
& .54 $\pm$ .24
& \textbf{.69} $\pm$ .23 \\

& $\textsc{ropt}$  
& .67 $\pm$ .24
& .82 $\pm$ .15
& \textbf{.59} $\pm$ .19
& .53 $\pm$ .24
& .66 $\pm$ .23 \\

& $\textsc{roppt}$ 
& \textit{.68} $\pm$ .23
& \textit{.92} $\pm$ .06
& \textbf{.59} $\pm$ .21
& .54 $\pm$ .24
& \textbf{.69} $\pm$ .24 \\

\midrule
\multirow{4}{*}{\rotatebox{90}{\texttt{cosine}}}
& $\textsc{rpt}$   
& .67 $\pm$ .23
& .84 $\pm$ .13
& \textit{.58} $\pm$ .20
& \textbf{.58} $\pm$ .24
& .67 $\pm$ .22 \\

& $\textsc{rppt}$  
& .67 $\pm$ .23
& \textbf{.93} $\pm$ .05
& .57 $\pm$ .20
& \textit{.56} $\pm$ .24
& \textbf{.69} $\pm$ .24 \\

& $\textsc{ropt}$  
& .66 $\pm$ .23
& .84 $\pm$ .13
& \textbf{.59} $\pm$ .20
& .55 $\pm$ .24
& .66 $\pm$ .23 \\

& $\textsc{roppt}$ 
& .66 $\pm$ .23
& \textbf{.93} $\pm$ .06
& \textit{.58} $\pm$ .21
& .55 $\pm$ .24
& \textit{.68} $\pm$ .24 \\

\midrule
\multirow{4}{*}{\rotatebox{90}{\texttt{manhattan}}}
& $\textsc{rpt}$   
& \textbf{.69} $\pm$ .23
& .83 $\pm$ .14
& \textit{.58} $\pm$ .20
& .55 $\pm$ .24
& \textit{.68} $\pm$ .22 \\

& $\textsc{rppt}$  
& \textbf{.69} $\pm$ .23
& .91 $\pm$ .06
& \textit{.58} $\pm$ .20
& .54 $\pm$ .24
& \textbf{.69} $\pm$ .23 \\

& $\textsc{ropt}$  
& .67 $\pm$ .24
& .83 $\pm$ .13
& \textbf{.59} $\pm$ .20
& .53 $\pm$ .24
& .66 $\pm$ .23 \\

& $\textsc{roppt}$ 
& \textbf{.69} $\pm$ .23
& .91 $\pm$ .06
& \textit{.58} $\pm$ .20
& .53 $\pm$ .25
& \textbf{.69} $\pm$ .23 \\

\bottomrule
\end{tabular}
\end{table}

\end{document}